\documentclass[a4paper]{cas-dc}

\usepackage[authoryear,longnamesfirst]{natbib}

\usepackage{graphicx}%
\usepackage{multirow}%
\usepackage{amsmath,amssymb,amsfonts}%
\usepackage{amsthm}%
\usepackage{mathrsfs}%

\usepackage{textcomp}%
\usepackage{manyfoot}%
\usepackage{booktabs}%
\usepackage{algorithm}%
\usepackage{algorithmicx}%
\usepackage{algpseudocode}%
\usepackage{listings}%
\usepackage{pifont}
\usepackage{tabularx}
\usepackage{array}
\usepackage{adjustbox}
\usepackage{enumitem}
\usepackage{rotating}
\usepackage{hyperref}

\newcommand{\tabincell}[2]{\begin{tabular}{@{}#1@{}}#2\end{tabular}}
\newcommand{\cmark}{\ding{51}}%
\newcommand{\xmark}{\ding{55}}%

\newcommand{\ie}{\textit{i}.\textit{e}.}
\newcommand{\eg}{\textit{e}.\textit{g}.}
\newcommand{\etc}{etc\@ifnextchar.{}{.\@}}

\begin{document}
\let\WriteBookmarks\relax
\def\floatpagepagefraction{1}
\def\textpagefraction{.001}

\shorttitle{}

\shortauthors{Yifan Jiao et~al.}

\title [mode = title]{PlanarTrack: A High-quality and Challenging Benchmark  for Large-scale Planar Object Tracking}                      

% \tnotetext[1]{This document is the results of the research
%    project funded by the National Science Foundation.}

% \tnotetext[2]{The second title footnote which is a longer text matter
%    to fill through the whole text width and overflow into
%    another line in the footnotes area of the first page.}


\author[1,2]{Yifan Jiao}[style=chinese]
\author[1,2]{Xinran Liu}[style=chinese]
\author[3]{Xiaoqiong Liu}[style=chinese]
\author[3]{Xiaohui Yuan}[style=chinese]
\author[3]{Heng Fan}[style=chinese]
\author[1,2]{Libo Zhang}[style=chinese,orcid=0000-0001-8450-0958]
\ead{libo@iscas.ac.cn}

\cormark[1]


% \credit{Conceptualization of this study, Methodology, Software}

\affiliation[1]{organization={Institute of Software, Chinese Academy of Sciences},
    city={Beijing},
    % citysep={}, % Uncomment if no comma needed between city and postcode
    country={China}}
\affiliation[2]{organization={University of Chinese Academy of Sciences},
    city={Beijing},
    % citysep={}, % Uncomment if no comma needed between city and postcode
    country={China}}

\affiliation[3]{organization={Dept. of Computer Science \& Engineering, University of North Texas},
    city={Denton},
    % citysep={}, % Uncomment if no comma needed between city and postcode
    country={United States of America}}

\cortext[cor1]{Corresponding author}

\begin{abstract}
Planar tracking has drawn increasing interest owing to its key roles in robotics and augmented reality. Despite recent great advancement, further development of planar tracking, particularly in the deep learning era, is largely limited compared to generic tracking due to the lack of large-scale platforms. To mitigate this, we propose \textbf{PlanarTrack}, a large-scale high-quality and challenging benchmark for planar tracking. Specifically, PlanarTrack consists of 1,150 sequences with over 733K frames, including 1,000 short-term and 150 new long-term videos, which enables comprehensive evaluation of short- and long-term tracking performance. All videos in PlanarTrack are recorded in unconstrained conditions from the wild, which makes PlanarTrack challenging but more realistic for real-world applications. To ensure high-quality annotations, each video frame is manually annotated by four corner points with multi-round meticulous inspection and refinement. To enhance target diversity of PlanarTrack, we only capture a unique target in one sequence, which is different from existing benchmarks. To our best knowledge, PlanarTrack is by far the largest and most diverse and challenging dataset dedicated to planar tracking. To understand performance of existing methods on PlanarTrack and to provide a comparison for future research, we evaluate 10 representative planar trackers with extensive comparison and in-depth analysis. Our evaluation reveals that, unsurprisingly, the top planar trackers heavily degrade on the challenging PlanarTrack, which indicates more efforts are required for improving planar tracking. Moreover, we derive a variant named \textbf{PlanarTrack}\textsubscript{BB} from PlanarTrack for generic tracking. Evaluation with 15 generic trackers shows that, surprisingly, our PlanarTrack\textsubscript{BB} is even more challenging than several popular generic tracking benchmarks, and more attention should be paid to dealing with planar targets, though they are rigid. Our data and results will be released at \url{https://github.com/HengLan/PlanarTrack}

\end{abstract}

% \includegraphics{figs/grabs.pdf}

% \begin{highlights}
% \item We present \textbf{PlanarTrack}, a high-quality large-scale benchmark that dedicated for planar object tracking. With more than 700K maunal-annotated frames and 1,150 different targets, there is a huge increase in diversity and scale in the proposed PlanarTrack, compared to the existing benchmarks.
% \item Unlike existing benchmarks that collected in indoor laboratory environments with simple background, we establish our PlanarTrack in the wild, which makes it more challenging.
% \item The proposed PlanarTrack consists of 150 long sequences with an average length of 1,622 and 4 ultra-long sequences longer than 3,000 frames, which makes it possible for evaluation of long-term tracking.
% \item Extensive experiments show that all the 10 planar trackers significantly decline on our challenging PlanarTrack, which indicates the usefulness and effectiveness of our benchmark in future research of planar object tracking.
% \item We also develop \textbf{PlanarTrack\textsubscript{BB}}, a by-product of proposed PlanarTrack to observe the performance of generic trackers. Experiment results of 15 top-performance generic trackers show the limitation of generic trackers in localizing planar-like targets.
% \end{highlights}

% Keywords

\maketitle

\section{Introduction}\label{sec1}

Planar object tracking is a fundamental problem in computer vision. Different from generic object tracking which aims at localizing the target with axis-aligned rectangle bounding boxes~\citep{wu2013online,huang2019got,fan2019lasot}, the goal of planar object tracking is to predict the 2D transformations (\eg, the homograph) of a target (\eg, surface or plane of the object) and locate it with four corner points (see Fig.~\ref{fig1}). Because of its important applications in augmented reality (AR) (\eg,~\citep{comport2003real,wagner2009real,matveichev2021mobile}) and robotics (\eg,~\citep{mondragon20103d,corso2003direct}), planar object tracking has attracted increasing interest in recent years. Particularly, with the introduction of several benchmarks (\eg,~\citep{liang2018planar,liang2021planar,roy2015tracking}), great progress has been seen in planar object tracking (\eg,~\citep{zhan2022homography,zhang2022hvc,vserych2023planar,li2023centroid}. Despite this, these datasets are largely limited in further facilitating the development of planar object tracking, due to the following reasons:
\begin{figure*}[!tb]
    \centering
    \includegraphics[width=0.95\linewidth]{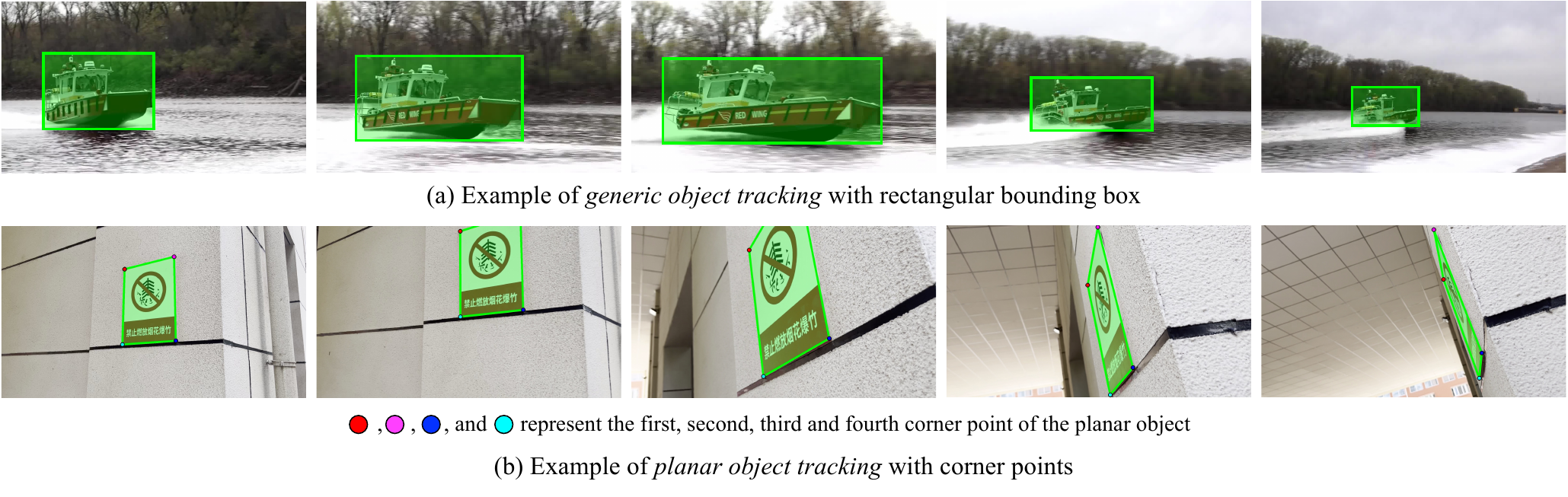}
    \caption{Comparison between generic object tracking (a) and planar object tracking (b). The former estimates axis-aligned rectangular bounding boxes for the target object, while the latter (our focus in this work) calculates 2D transformations of the target object to obtain the corresponding corner points for localization. \emph{All figures throughout this paper are best viewed in color and by zooming in.}}
    \label{fig1}
\end{figure*}

\begin{figure}[!t]
    \centering
    \includegraphics[width=0.95\linewidth]{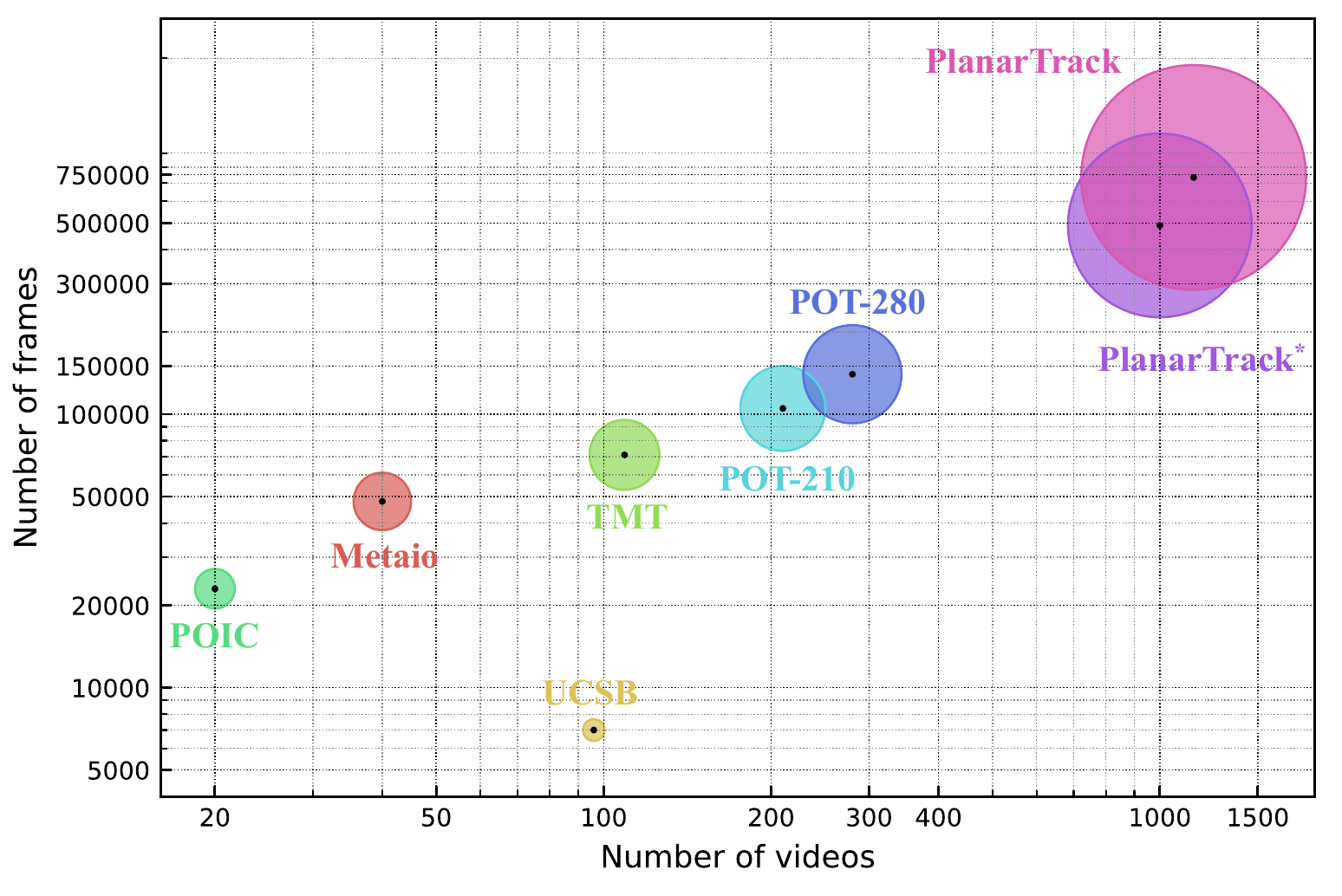}
    \caption{Summary of planar object tracking datasets, containing POT-280 \citep{liang2021planar}, POT-210 \citep{liang2018planar}, TMT \citep{roy2015tracking}, UCSB \citep{gauglitz2011evaluation}, Metiao \citep{lieberknecht2009dataset}, POIC \citep{chen2017illumination}, our PlanarTrack and PlanarTrack$^{*}$ from conference version~\citep{liu2023planartrack}. The circle diameter is in proportion to the number of frames of a dataset. Our PlanarTrack is the \emph{largest} benchmark.}
    \label{fig2}
\end{figure}

\vspace{0.3em}
\noindent\textbf{Small-scale}. One major issue with existing benchmarks is their relatively small scales. Especially, in the deep learning era, in order to unleash the potential of deep planar tracking, a large-scale platform with a great number of video sequences is highly desired for training. As demonstrated in Fig.~\ref{fig2}, however, all existing datasets comprise \emph{less than} 300 video sequences, which is far from being sufficient for training deep planar trackers. As a result, researchers in the community have to utilize synthetic data generated from images (\eg, \citep{lin2014microsoft}) or videos from the generic bounding box-based tracking benchmark (\eg,~\citep{huang2019got}) for deep planar tracking, which may result in suboptimal performance because of  domain gap among different tasks. In addition to the training of deep planar trackers, a large-scale platform is necessary for reliable evaluation and comparison of different algorithms.

\vspace{0.3em}
\noindent\textbf{Less challenging scenario}. Real-world scenarios are often challenging and complicated. Nevertheless, early planar tracking datasets (\eg,~\citep{lieberknecht2009dataset,roy2015tracking,gauglitz2011evaluation,chen2017illumination}) are developed from indoor laboratory environments with simple background, which cannot fully reflect the complicated and diverse scenarios in real applications while evaluating. To handle this, recent datasets (\eg,~\citep{liang2018planar,liang2021planar}) directly collect videos in the wild. However, most sequences in these benchmarks are mainly involved with one challenge factor (or \emph{attribute} in generic tracking), and very few (\eg, 30 videos in~\citep{liang2018planar} and 40 videos in~\citep{liang2021planar}) contain multiple challenges (\ie, the unconstrained condition). This may weaken the difficulties of planar tracking in the wild where arbitrary challenges could occur simultaneously, and thus restricts their usage in evaluating the generalization of planar tracking systems in the real world.

\vspace{0.3em}
\noindent\textbf{Limited diversity}. The diversity of target objects is crucial for a tracking benchmark. In existing planar tracking datasets, the sample planar target is often utilized in multiple sequences, which largely reduces the diversity in target appearance and may lead to bias in performance assessment. For example, for the current largest planar tracking benchmark~\citep{liang2021planar} (one target used in 7 videos), the number of planar targets does not exceed 40 (see Tab.~\ref{tab1}). Such lack of diversity makes it difficult to use the current benchmarks for faithful  evaluation of planar trackers in practice.

\vspace{0.3em}
\noindent\textbf{Lack of long-term tracking}. The task of long-term tracking is more challenging and holds greater practical significance compared to short-term tracking. This is because long-term tracking requires algorithms capable of continuously capturing the target object over extended durations, while effectively handling scenarios wherein the target frequently disappears and reappears. This complexity makes long-term tracking tasks more reflective of real-world applications. In order to be deployed in real applications, a planar tracker is expected to perform well in not only short-term scenarios but also in long-term videos. Yet, existing benchmarks either contain only short-term videos (\eg,~\citep{gauglitz2011evaluation,liang2018planar,liang2021planar}) with an average length of less than 1,000 frames or just a few long-term videos (\eg,~\citep{roy2015tracking,chen2017illumination}). We note that the benchmark of~\citep{lieberknecht2009dataset} could serve as a testbed for long-term planar tracking by containing 40 long sequences with an average length of 1,200 frames. However, its diversity (with 5 targets) and scale (40 sequences in total) are significantly limited in further facilitating the development of planar tracking.

We notice that there exist several large-scale benchmarks (\eg,~\citep{muller2018trackingnet,fan2019lasot,huang2019got,peng2024vasttrack}) for generic tracking. However, planar tracking differs fundamentally from generic tracking: instead of predicting bounding boxes, it requires estimating 2D homography via four corner points, which is crucial for applications such as augmented reality and robotics. Such geometric precision cannot be reliably achieved by post-processing generic trackers, as bounding boxes provide insufficient information and small errors are easily amplified. Owing to these different goals and settings (see Fig.~\ref{fig1}), existing generic datasets are \emph{not} suitable for planar tracking. In addition, a recent benchmark named MPOT-3K~\citep{zhang2023multiple} with 356 videos has been introduced for multi-planar tracking, which differs from the goal of single-planar tracking and is therefore not directly applicable. To further facilitate research on deep planar tracking, a dedicated large-scale benchmark is desired, which motivates our work.

Unlike generic tracking that only predicts bounding boxes, planar tracking estimates 2D homography via four corner points, which is essential for applications such as augmented reality and robotics. Approximating this task by post-processing generic trackers is unreliable, as bounding boxes lack sufficient geometric information and small errors are easily amplified.

\subsection{Contribution}\label{sec1_1}

In this paper, we propose to develop a novel large-scale benchmark, named PlanarTrack, dedicated to planar object tracking. The contributions of PlanarTrack are summarized as follows:

\begin{enumerate}[label=(\arabic*), leftmargin=*, noitemsep]
\setlength{\leftmargin}{0pt}
    \item We present a dedicated large-scale benchmark, \textbf{PlanarTrack}, for planar object tracking. PlanarTrack contains 1,150 sequences with more than 733K frames. All these videos are directly recorded in complicated \emph{unconstrained} conditions from the wild scenarios. Compared to existing datasets (\eg, \citep{chen2017illumination, gauglitz2011evaluation, liang2021planar, liang2018planar, lieberknecht2009dataset, roy2015tracking}), our PlanarTrack is much more challenging yet realistic in real applications. For each frame in PlanarTrack, we carefully inspected and manually annotated the coordinates of four corner points. To ensure annotation quality, each annotation is double-verified and corrected if necessary. As far as we know, PlanarTrack is so far the \emph{largest} (in terms of the number of sequences and frames) and \emph{most challenging} planar tracking dataset with high-quality dense annotations. By developing PlanarTrack, we aim to provide a dedicated large-scale platform for promoting the development and evaluations of deep-learning-based planar trackers. 
    \item There is a huge increase in diversity of targets in PlanarTrack, compared to existing datasets. There are 1,150 different targets while other datasets only contain 40 targets at most. The diversity of PlanarTrack makes a contribution to a more effective training and more equitable evaluations.
    \item PlanarTrack gives an opportunity for evaluation of long-term tracking. 150 out of 1,150 sequences are produced as long sequences with an average length of 1,622 frames. Further more, there are 4 \emph{ultra-long} sequences longer than 3,000 frames, enabling assessment of long-term trackers. Experiments on long-term and short-term sequences show that all planar trackers struggle to maintain target capture over extended periods, indicating the need for further research into long-term tracking.
    \item We offer more challenging information in PlanarTrack. Almost all sequences have multiple challenging factors (\ie, \emph{unconstrained conditions}) which are closer to the realistic scenarios, while existing benchmarks contain no or little unconstrained videos. Researchers can further understand planar trackers by carrying out experiments on different challenging factors.
    \item To analyze PlanarTrack and provide comparisons for future research, we evaluate 10 recent planar object tracking algorithms. Evaluation results show that all the trackers significantly decline on our more challenging PlanarTrack, which indicates that more efforts should be made for improvements. We further conduct an overall analysis of different challenging factors and long-term tracking with discussion to provide a guidance for future research. Besides, our re-training experiments show the usefulness and effectiveness of our benchmark in performance enhancement.
    \item To observe the performance of generic trackers in localizing planar-like targets, we develop \textbf{PlanarTrack}\textsubscript{BB}, a by-product of PlanarTrack which is suitable for generic box tracking. We aim at \emph{large-scale} learning and evaluation of generic trackers on tracking \emph{rigid} targets, which is rarely investigated before. To this end, we select 15 top-performance transformer-based generic trackers for evaluation on PlanarTrack\textsubscript{BB}. Results show that all trackers reveal heavy performance degeneration on PlanarTrack\textsubscript{BB} compared with existing large-scale generic tracking benchmarks (\eg, LaSOT \citep{fan2019lasot} and TrackingNet \citep{muller2018trackingnet}). More efforts should be made to handle planar objects though they are rigid.    
    
\end{enumerate}

This paper extends an early conference version in \citep{liu2023planartrack}. The main new contributions are as follows. \textbf{(i)} We expand the scale of PlanarTrack to be about 1.5 times larger in term of number of frames by introducing 243,326 new images with precise annotations. \textbf{(ii)} For long-term tracking, we introduce 150 long sequences with an average length of 1,622 frames, among which 4 ultra-long sequences longer than 3,000 frames are contained. Additional experiments have been conducted to highlight the significance of long-term planar object tracking. \textbf{(iii)} More details of PlanarTrack construction are provided.  \textbf{(iv)} More thorough experiments and in-depth analysis are conducted on PlanarTrack for planar object tracking and PlanarTrack\textsubscript{BB} for generic tracking relatively, in order to show the advantages and necessity of dedicated large-scale benchmark.

The rest of this paper is organized as follows. Section \ref{sec2} briefly introduces related tracking algorithms and benchmarks. In Section \ref{sec3}, we describe the construction of our PlanarTrack in detail with a comprehensive analysis of benchmark attributes. Experimental evaluation results and in-depth analysis are conducted in Section \ref{sec4} for better understanding. Section \ref{sec5} reports the construction of PlanarTrack\textsubscript{BB} and generic tracking experiments, followed by a conclusion in Section \ref{sec6}.

\begin{table*}[!tb]\small
  \centering
  \caption{Detailed comparison of the proposed PlanarTrack with other existing planar object tracking benchmarks. PlanarTrack$^*$ denotes for the conference version of PlanarTrack.}
  \setlength{\extrarowheight}{1pt}
  \setlength{\tabcolsep}{1.1mm}{
    \begin{tabular}{lcccccccccc}
    \specialrule{.1em}{.05em}{.05em} 
    Benchmark & Year  & Targets & Videos & \tabincell{c}{Min\\frames} & \tabincell{c}{Mean\\frames} & \tabincell{c}{Max\\frames} & \tabincell{c}{Total\\frames} & 
    \tabincell{c}{Annotated\\frames} &\tabincell{c}{Unconstrain-\\ed Videos} & \tabincell{c}{In the\\wild} \\
    \hline
    Metaio~\citep{lieberknecht2009dataset} & 2009  & 8     & 40    & 1,200  & 1,200  & 1,200  & 48K &48K  & n/a       & \xmark \\
    UCSB~\citep{gauglitz2011evaluation}  & 2011  & 6     & 96    & 13    & 72    & 500   & 7K &7K   & n/a     & \xmark \\
    TMT~\citep{roy2015tracking}   & 2015  & 12    & 109   & 191   & 648   & 2,518  & 71K &71K  & n/a      & \xmark \\
    POIC~\citep{chen2017illumination}  & 2017  & 20    & 20    & 283   & 1,149  & 2,666  & 23K &23K  & n/a    & \xmark \\
    POT-210~\citep{liang2018planar} & 2018  & 30    & 210   & 501   & 501   & 501   & 105K &53K  & 30       & \cmark \\
    POT-280~\citep{liang2021planar} & 2021  & 40    & 280   & 501   & 501   & 501   & 140K &70K & 40      & \cmark \\
    % MPOT-3K~\citep{zhang2023multiple} & 2023 & 356 & 3717 & - & - & - & 687K & 687K & 356 & \cmark \\
    \hline
    \rowcolor{gray!15} \textbf{PlanarTrack$^*$~\citep{liu2023planartrack}} & \textbf{2023}  & \textbf{1,000}  & \textbf{1,000}  & \textbf{317}   & \textbf{490}   & \textbf{549}   & \textbf{490K} & \textbf{490K}  & \textbf{1,000}    & \cmark \\
    \rowcolor{gray!15} \textbf{PlanarTrack} & \textbf{2024}  & \textbf{1,150}  & \textbf{1,150}  & \textbf{317}   & \textbf{638}   & \textbf{3,352}   & \textbf{733K} & \textbf{733K}  & \textbf{1,150}    & \cmark \\
    \specialrule{.1em}{.05em}{.05em} 
    \end{tabular}%
  }
  \label{tab1}%
\end{table*}%

% The Introduction section, of referenced text \citep{bib1} expands on the background of the work (some overlap with the Abstract is acceptable). The introduction should not include subheadings.

% Springer Nature does not impose a strict layout as standard however authors are advised to check the individual requirements for the journal they are planning to submit to as there may be journal-level preferences. When preparing your text please also be aware that some stylistic choices are not supported in full text XML (publication version), including coloured font. These will not be replicated in the typeset article if it is accepted. 

\section{Related Work}\label{sec2}

\subsection{Planar Tracking Algorithms}\label{sec2_1}

Planar object tracking is a fundamental computer vision task, which aims at recovering the homography from the template to the current frame. Here we briefly review three mainstream trends including keypoint-based methods, region-based methods and deep-learning-based methods.

\vspace{0.3em}
\noindent\textbf{Keypoint-based methods} Keypoint-based algorithms (\citep{dick2013realtime, ozuysal2009fast, wang2017gracker, hare2012efficient, zhao2015metric}) typically represent an object with a set of points and their descriptors. Their tracking process is divided into two steps. Firstly, trackers detect the keypoints of objects (\eg, SIFT~\citep{lowe2004distinctive}, SURF~\citep{bay2008speeded} and FAST~\citep{rosten2008faster}). A pair of correspondences between object and image keypoints is established through descriptor matching. Then, a robust homography is estimated with geometric estimation algorithms (\eg, RANSAC~\citep{fischler1981random} and its variants~\citep{torr2000mlesac, chum2005matching}). To deal with the huge per-frame motions, an approximate nearest neighbour search to estimate per-frame state updates is introduced in~\citep{dick2013realtime}. Authors in~\citep{ozuysal2009fast} propose to detect objects by leveraging hundreds of binary features and models class posterior probabilities in a naive Bayesian classification framework, making it perform remarkably on datasets containing very significant perspective changes with less computational costs. A graph is applied in~\citep{wang2017gracker} to model a planar object and represent its structure, instead of a simple collection of keypoints. 

\vspace{0.3em}
\noindent\textbf{Region-based methods} Region-based methods (\eg,~\citep{benhimane2004real, richa2011visual, chen2017illumination, tan2014multi}) are sometimes called \emph{direct methods}. These methods formulate the planar tracking task as an image registration problem. They directly estimate the homography by optimizing the alignment of the current frame with the object of the initial frame. The work of~\citep{benhimane2004real} presents a tracking algorithm based on minimizing the sum-of-squared-difference between a given template and the current image. The proposed minimization method is a second-order one, making it unnecessary to compute the Hessian and achieve the high convergence rate. To reduce the impact of non-linear illumination variations, the authors in~\citep{richa2011visual} introduced a direct tracking method based on an image similarity measure called the sum of conditional variance (SCV). The SCV requires less iterations to converge and has a significantly larger convergence radius, and achieves excellent performance under challenging illumination conditions and rapid motions. The work of \citep{chen2017illumination} also measures the similarity between two images through a second-order minimization method for planar object tracking. They suggested a denoising method based on the Perona-Malik function and a mask image to improve the robustness against image noise and low texture. 

\vspace{0.3em}
\noindent\textbf{Deep-learning-based methods} In addition to the above two types, another popular trend is to regress the homography with the deep neural networks \citep{zhan2022homography, zhang2022hvc, li2023centroid, erlik2017homography, wang2018deep, liu2019gift, sarlin2020superglue, vserych2023planar}. A hierarchy of twin convolutional regression networks is introduced in \citep{erlik2017homography} to estimate the homography between a pair of images. The framework achieves high performance with simple hierarchical arrangement of simple models due to the iterative nature. In \citep{zhan2022homography}, a novel homography decomposition approach is proposed to reduce and stabilize the condition number by decomposing the homography transformation into two groups and is trained in a semi-supervised fashion. Dense optical flow with weight is introduced in \citep{vserych2023planar} to estimate a homography by weighted least squares in a fully differentiable manner. HDN~\citep{zhan2022homography} further improves robustness by introducing a homography decomposition network with semi-supervised learning, enabling stable estimation under challenging conditions. More recently, WOFT~\citep{vserych2023planar} formulates planar tracking as weighted optical flow estimation, where homography is obtained via differentiable weighted least squares, achieving strong performance on multiple benchmarks. The above deep-learning-based planar trackers can not only avoid complicated keypoint feature extraction and be trained end to end, but also achieve outstanding performance. Thus, the deep regression-based methods have attracted increasing attention in planar tracking.

\subsection{Planar Tracking Benchmarks}\label{sec2_2}

Datasets have played an important role in facilitating the development of planar object tracking. In recent years, there have been several planar tracking benchmarks, including Metaio \citep{lieberknecht2009dataset}, UCSB \citep{gauglitz2011evaluation}, TMT \citep{roy2015tracking}, POIC \citep{chen2017illumination}, POT (POT-210\citep{liang2018planar}, POT-280 \citep{liang2021planar}) and MPOT-3K \citep{zhang2023multiple}. Table \ref{tab1} provides a detailed comparison between these benchmarks.

\vspace{0.3em}
\noindent \textbf{Metaio} Metaio \citep{lieberknecht2009dataset} is one of the earliest datasets for planar tracking. It consists of 40 videos with eight different textures using a camera mounted on the robotic measurement arm. The ratio of successfully tracked images is used for measuring the performance of the planar trackers.

\vspace{0.3em}
\noindent \textbf{UCSB} UCSB \citep{gauglitz2011evaluation} has 96 sequences, containing six planar textures with 16 motion patterns each. The ground truth is semi-automatically annotated using four red markers fixed on a glass frame. 

\vspace{0.3em}
\noindent \textbf{TMT} TMT \citep{roy2015tracking} comprises 109 sequences and each one is labeled with a challenging factor. Three trackers are used for ground truth annotations. The coordinates of four corners are determined when all three trackers are agree within a certain range. The goal of TMT is to evaluate different planar tracking algorithms for human and robot manipulation tasks.

\vspace{0.3em}
\noindent \textbf{POIC} POIC \citep{chen2017illumination} contains 10 sequences with total of 6663 frames. Objects with varying texture and lambertian/specular materials are provided to evaluate the performance of planar trackers in challenging complicated illumination environments.

\vspace{0.3em}
\noindent \textbf{POT} Different from the above dataset collected from a simple laboratory environment, POT-210 \citep{liang2018planar} is the first one providing a dataset for planar object tracking in the wild, which contains 210 sequences of 30 planar objects. It is further extended to POT-280 in \citep{liang2021planar} by introducing 70 more sequences of another 10 objects. Each planar object in POT \citep{liang2018planar, liang2021planar} is captured in seven videos. However, six of these form one challenge, and only one contains multiple challenges in unconstrained conditions.

% \noindent \textbf{MPOT-3K}

Previous algorithms have primarily relied on the POIC and POT datasets for experimentation and analysis. However, both datasets have significant limitations. On the one hand, POIC is small in scale and lacks sufficient category diversity, making it inadequate for fairly evaluating deep-based planar trackers, while deep-based algorithms are the current mainstream in this field. On the other hand, POT contains only seven sequences, six of which contains a single challenge factor, with only one sequence presenting multiple challenges under unconstrained conditions. This renders POT less representative of real-world scenarios. As a result, the field currently lacks a benchmark that addresses these shortcomings and provides a comprehensive evaluation framework for planar object tracking. To this end, we proposed PlanarTrack, the \emph{largest} and most \emph{challenging} and \emph{diverse} benchmark with \emph{high-quality} annotations for \emph{long-term} planar object tracking. Table \ref{tab1} displays a detailed comparison of our PlanarTrack with existing planar tracking benchmarks.

\subsection{Large-scale Generic Tracking Benchmarks}\label{sec2_3}

Large-scale benchmarks make it possible for efficient training and reliable evaluation, which have greatly facilitated the development of tracking in recent years. Examples of large-scale benchmarks include GOT-10k \citep{huang2019got}, LaSOT \citep{fan2019lasot, fan2021lasot}, TrackingNet \citep{muller2018trackingnet}, OxUvA \citep{valmadre2018long}, TNL2K \citep{wang2021towards}, and VastTrack~\citep{peng2024vasttrack}.

\vspace{0.3em}
\noindent \textbf{GOT-10k} GOT-10k \citep{huang2019got} consists of 10K videos, aiming to provide rich motion trajectories for short-term tracking. It is the first one to propose a novel one-shot evaluation for assessing tracking performance.

\vspace{0.3em}
\noindent \textbf{LaSOT} LaSOT \citep{fan2019lasot} is a high-quality large-scale benchmark for single object tracking with 1400 sequences and more than 3.5M frames. The average sequence length is more than 2500 frames and each sequence has various challenges deriving from the wild. It is later extended in \citep{fan2021lasot} by providing 150 extra sequences.

\vspace{0.3em}
\noindent \textbf{TrackingNet} TrackingNet \citep{muller2018trackingnet} is the first large-scale dataset and benchmark for object tracking in the wild, which contains more than 30K videos with more than 14 million dense annotations. The goal of TrackingNet is to further improve and generalize deep trackers.

\vspace{0.3em}
\noindent \textbf{OxUvA} OxUvA \citep{valmadre2018long} consists of 366 sequences spanning 14 hours, which is designed for long-term tracking. It is more challenging due to the frequent target disappearance. 

\vspace{0.3em}
\noindent \textbf{TNL2K} TNL2K \citep{wang2021towards} comprises 2K sequences with 124K frames and 663 words, aiming to evaluate trackers specifically for vision-language tracking.

\vspace{0.3em}
\noindent \textbf{VastTrack} VastTrack \citep{peng2024vasttrack} is a recently proposed large-scale generic tracking benchmark. It comprises over 50K video sequences with more than 2K categories, aiming to facilitate the exploration of more general and universal tracking.

Different from the aforementioned benchmarks, PlanarTrack is specifically designed for planar object tracking. Rather than using axis-aligned rectangular bounding boxes for targets, PlanarTrack utilizes corner point annotations for improved precision.

\section{The Proposed PlanarTrack Benchmark}\label{sec3}
\subsection{Design Principle}\label{sec3_1}

Our goal is to establish a dedicated benchmark, PlanarTrack, for training and evaluating planar object trackers. To this end, we follow five principles in establishing PlanarTrack, aiming at addressing all the issues of existing planar tracking benchmarks mentioned in previous sections:

\vspace{0.3em}
\noindent \textbf{Dedicated large-scale benchmark} An important motivation for our work is to train and fairly evaluate the deep-learning-based planar trackers by providing a large-scale benchmark. For this purpose, we capture 1,150 sequences with over 733K frames in the proposed benchmark, which is four times larger than the scale of POT-280 \citep{liang2021planar}.

\vspace{0.3em}
\noindent \textbf{Challenging realistic objects in the wild} To preserve tracking challenges in complicated realistic scenarios and faithfully reflect the performance of planar trackers in practice, videos of PlanarTrack are collected from natural scenarios with multiple challenge factors (\ie \ unconstrained condition).

\vspace{0.3em}
\noindent \textbf{Long-term tracking sequences} Frequent disappear and reenter is a common situation in long-term tracking. As a result, some long sequences should be included in the benchmark for evaluating long-term tracking algorithms. 

\begin{figure*}[!t]
    \centering
    \hspace*{-3mm} % 调整这里的值以减少左边的间隔
    \begin{tabular}{c@{\hspace{1mm}}c}
    \includegraphics[width=0.49\linewidth]{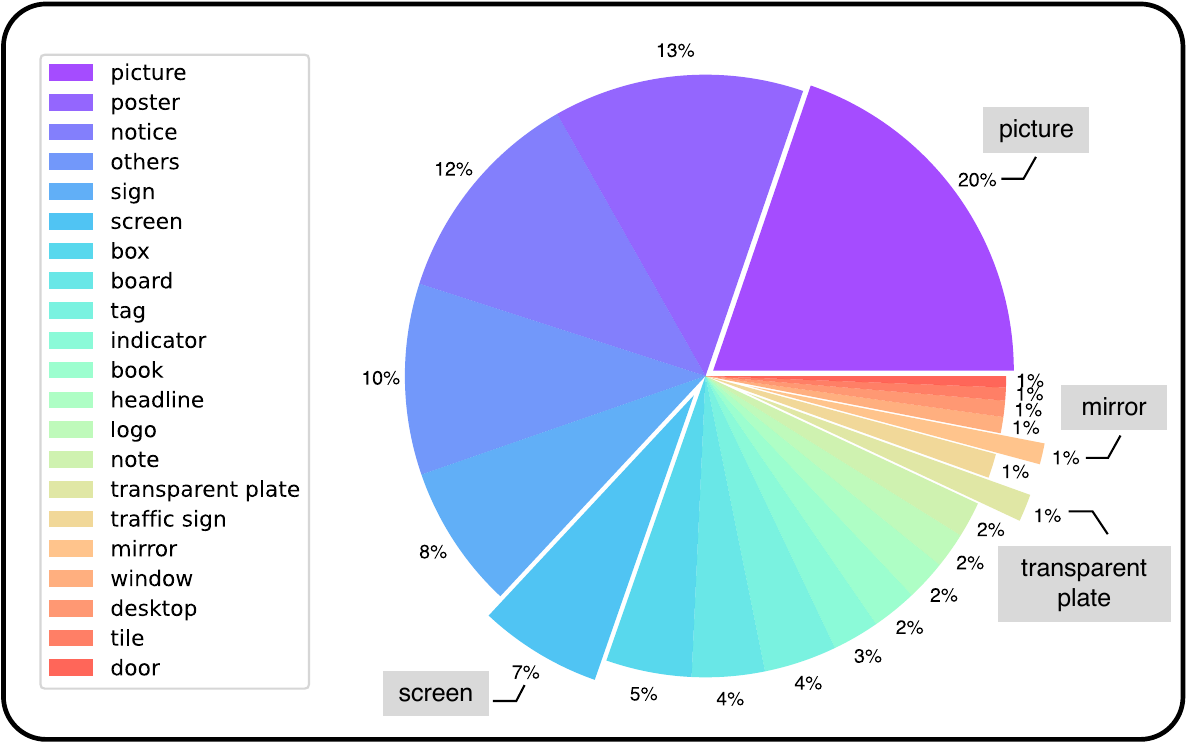} &
    \includegraphics[width=0.49\linewidth]{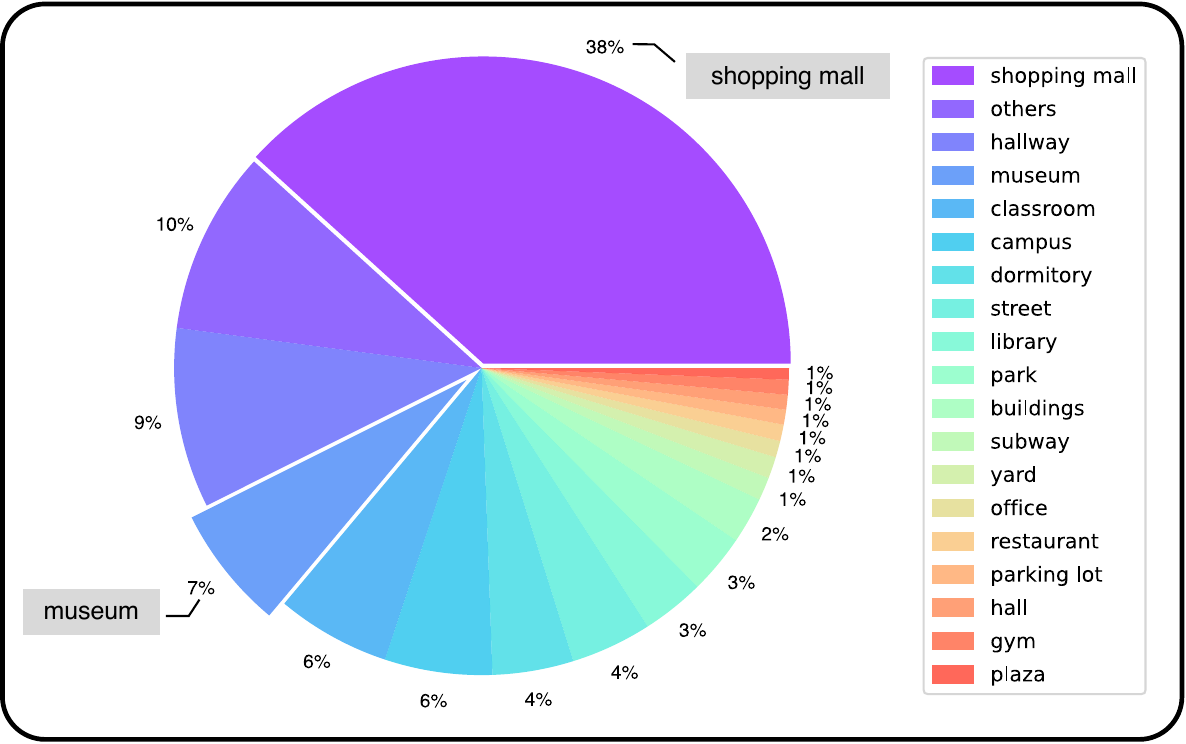} \\
    {\small (a) Distribution of classes in all sequences } & 
    {\small (b) Distribution of scenarios in all sequences } \\
    \end{tabular}
    \caption{Distribution of classes and scenarios in all sequences. (a): Planar targets can be divided into 21 classes. Four representative classes are highlighted. (b): Videos are all collected in these 19 scenarios. }
    \label{classes_scenarios_pie}
\end{figure*}

\vspace{0.3em}
\noindent \textbf{Diverse planar objects} The diversity of objects is crucial for the generalization of planar trackers. Considering this, the planar target in each sequence of our PlanarTrack should be unique, which is different from the existing benchmarks (\eg, POT-210/280 \citep{liang2018planar, liang2021planar}).

\vspace{0.3em}
\noindent \textbf{High-quality dense annotations} Accurate annotations are indispensable for effective training and fair evaluation. Therefore, each frame in PlanarTrack is manually labeled with careful refinement by well-trained annotators, in order to ensure the high-quality annotations.

\subsection{Data Collection}\label{sec3_2}

Different from existing generic object tracking benchmarks \citep{fan2019lasot, huang2019got, muller2018trackingnet,peng2024vasttrack} that source videos from YouTube (\url{https://www.youtube.com/}), we construct our PlanarTrack by recording videos from reality. We record sequences from natural scenarios using mobile phone because we find that there are few videos focused on planar objects on YouTube. Specifically, we invite many volunteers who are familiar with planar tracking to capture videos using various phones with different resolutions, in order to diversify the video sources. Following the principles mentioned above, we select various categories of planar objects, including \textit{box}, \textit{poster}, \textit{tag}, \textit{picture}, \textit{mirror}, \textit{screen}, \textit{traffic sign}, \textit{tile}, \textit{board}, \textit{transparent plate} and so on. Each sequence has a unique target and is captured in unconstrained conditions from various natural scenes (\eg \ \textit{shopping mall}, \textit{restaurant}, \textit{library}, \textit{dormitory}, \textit{museum} for indoor scenarios, \textit{campus}, \textit{street}, \textit{playground}, \textit{park}, \textit{plaza} for outdoor scenarios). We demonstrate the distribution of scenarios and classes in Fig. \ref{classes_scenarios_pie}. From \ref{classes_scenarios_pie} we can see that, our PlanarTrack is highly diverse in both scenarios and classes. All sequences are collected in 19 scenarios, while the shopping mall occupies the highest percentage. For the diversity of objects, all planar targets are divided into 21 classes, in which the picture has the greatest number. We purposely capture some targets with unconventional appearance changes (\eg, \emph{screen}, \emph{transparent plate} and \emph{mirror}) to enhance the challenge of our dataset.

In total, PlanarTrack is divided into two parts. The first part (\textit{part-1} for short) contains 1,000 sequences with an average length of 490 frames. Initially, we collected over 2,500 videos for \textit{part-1}. After a careful inspection, we choose 1,000 sequences which best meet the principles mentioned above. For these 1,000 videos, we further verify their contents and remove inappropriate parts to ensure that they are suitable for planar tracking. Although the sequence length of \textit{part-1} can reach the level of the existing benchmark, \textit{part-1} does not address the issue of long-term tracking. To this end, we introduce another part (\textit{part-2} for short), which comprises 150 \textit{long} sequences with an average length of 1,622 frames, which contains 4 \textit{ultra-long} sequences of more than 3,000 frames. We at first recorded more than 300 sequences in other places different from \textit{part-1}. In these long sequences, we capture objects that frequently enter and leave the view to reflect the real-world scenarios. After carrying through the same selecting and preprocessing flow, we provide 150 sequences with the best quality in \textit{part-2}. Eventually, we compile our PlanarTrack, a large-scale challenging benchmark dedicated to planar tracking by including 1,150 unconstrained sequences with more than 733K frames from 1,150 unique planar objects. Table \ref{tab1} provides a detailed summary of PlanarTrack and its comparison with existing planar tracking benchmarks.

% \begin{figure*}[!t]
%     \centering
%     \includegraphics[width=\linewidth]{classes_scenarios_pie.pdf}
%     \caption{Distribution of the scenarios and  }
%     \label{classes_scenarios_pie}
% \end{figure*}

\subsection{Annotation}\label{sec3_3}

\begin{figure*}[!t]
		\centering
		\begin{tabular}{c@{\hspace{0.4mm}}c}
			%\begin{tabular}{@{\hspace{.0mm}}c@{\hspace{1.75mm}} @{\hspace{.0mm}}c@{\hspace{.0mm}} @{\hspace{.0mm}}c@{\hspace{.0mm}}}
\includegraphics[width=0.158\linewidth]{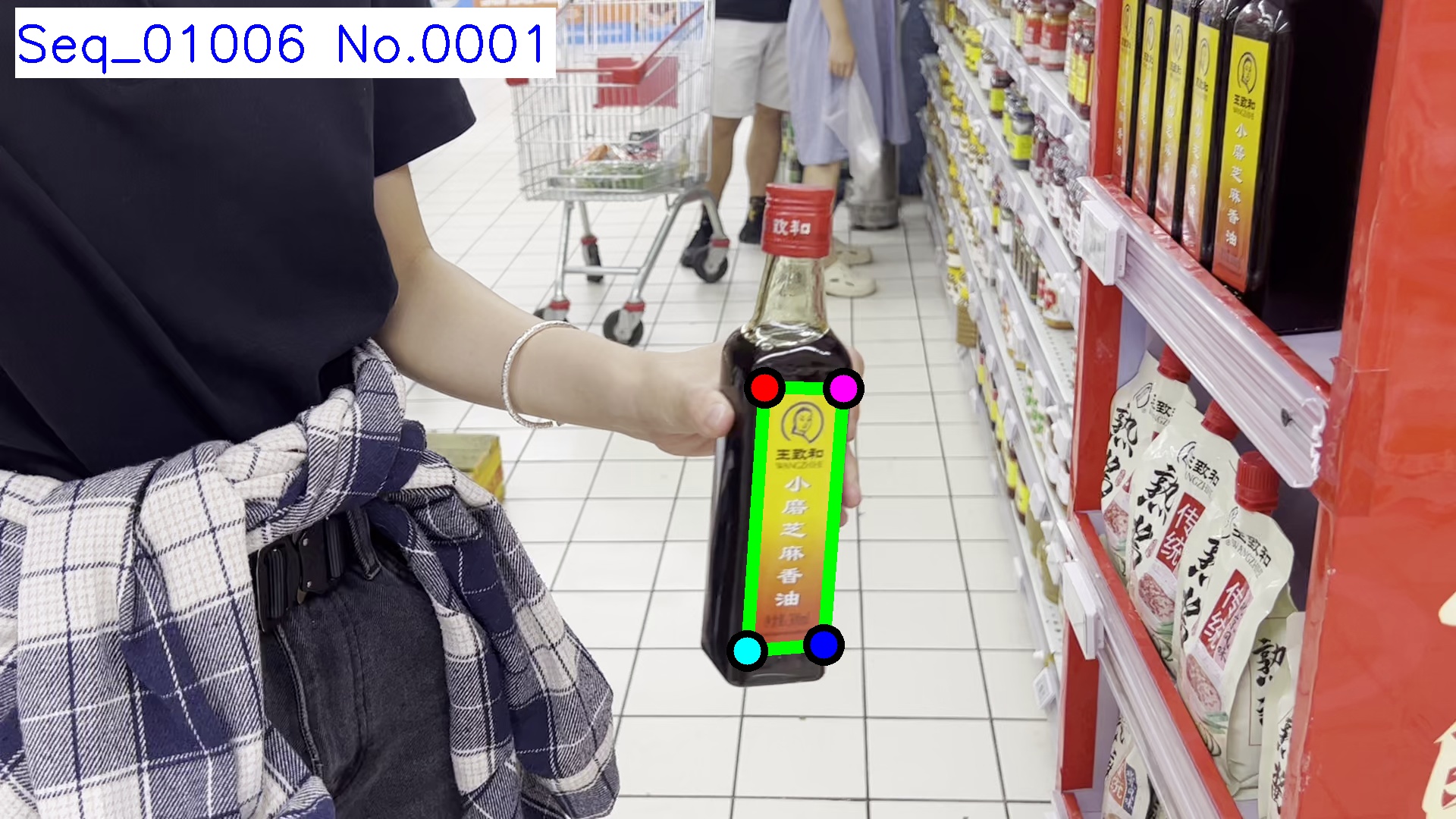} \includegraphics[width=0.158\linewidth]{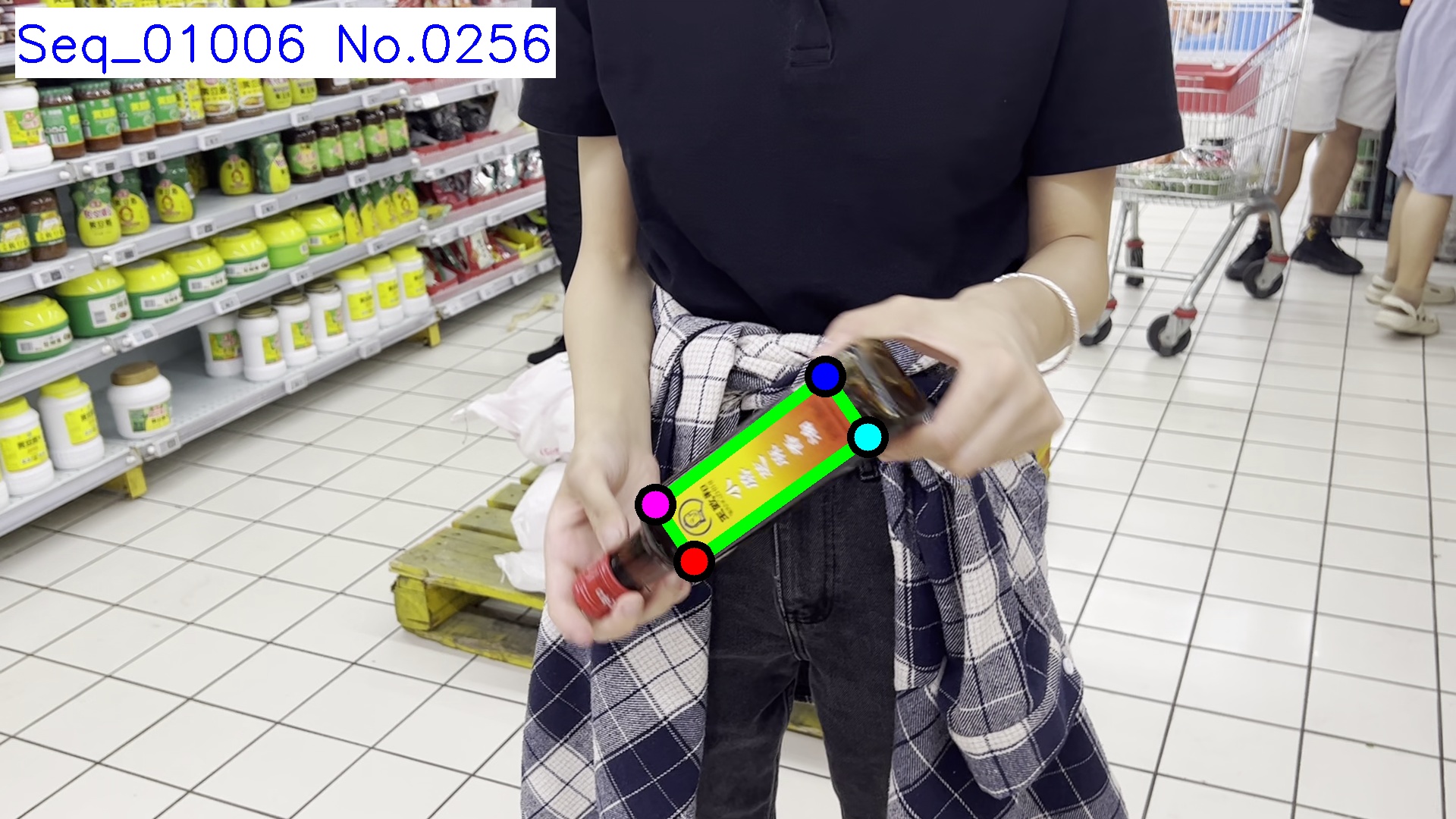} \includegraphics[width=0.158\linewidth]{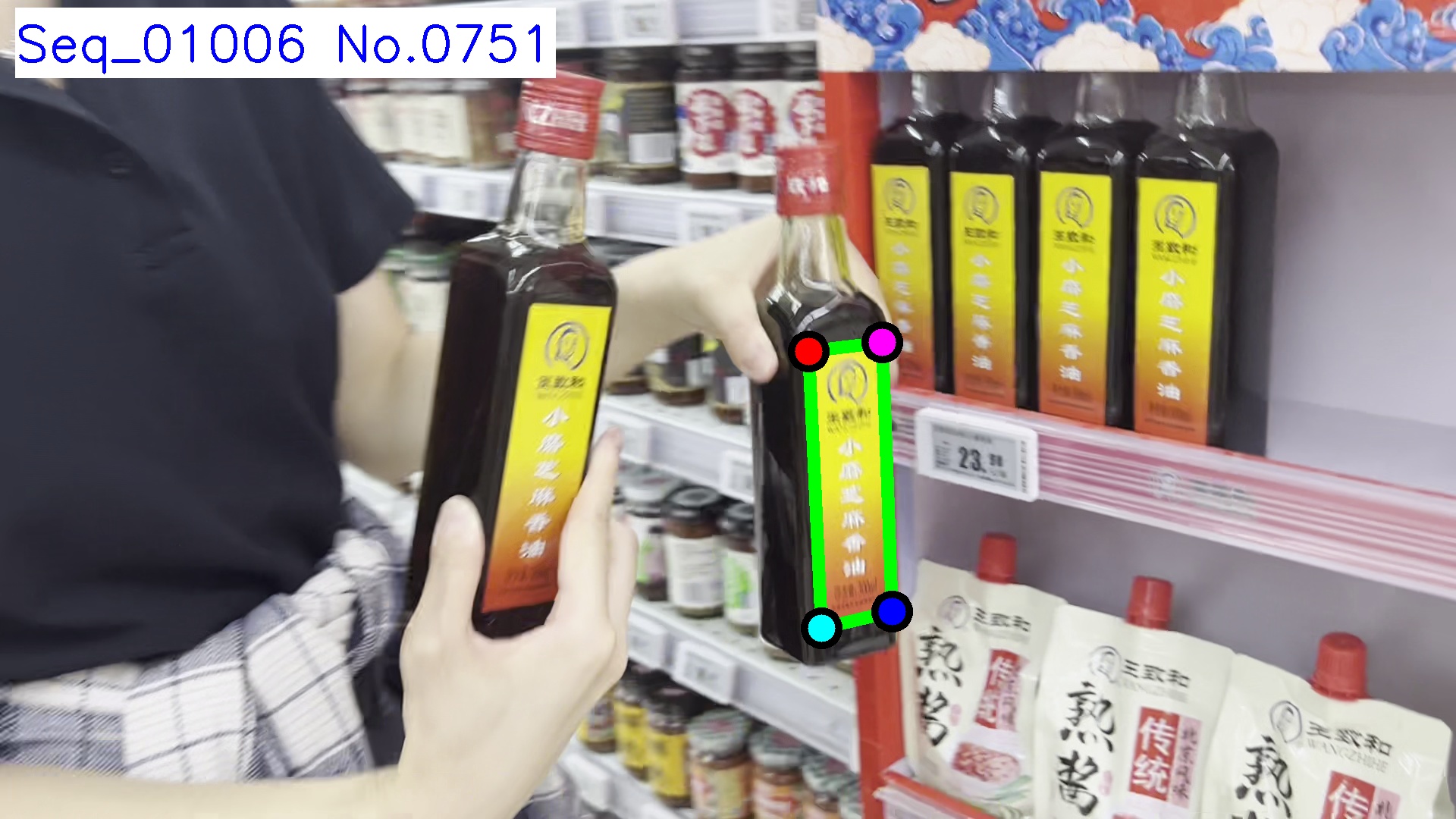} \includegraphics[width=0.158\linewidth]{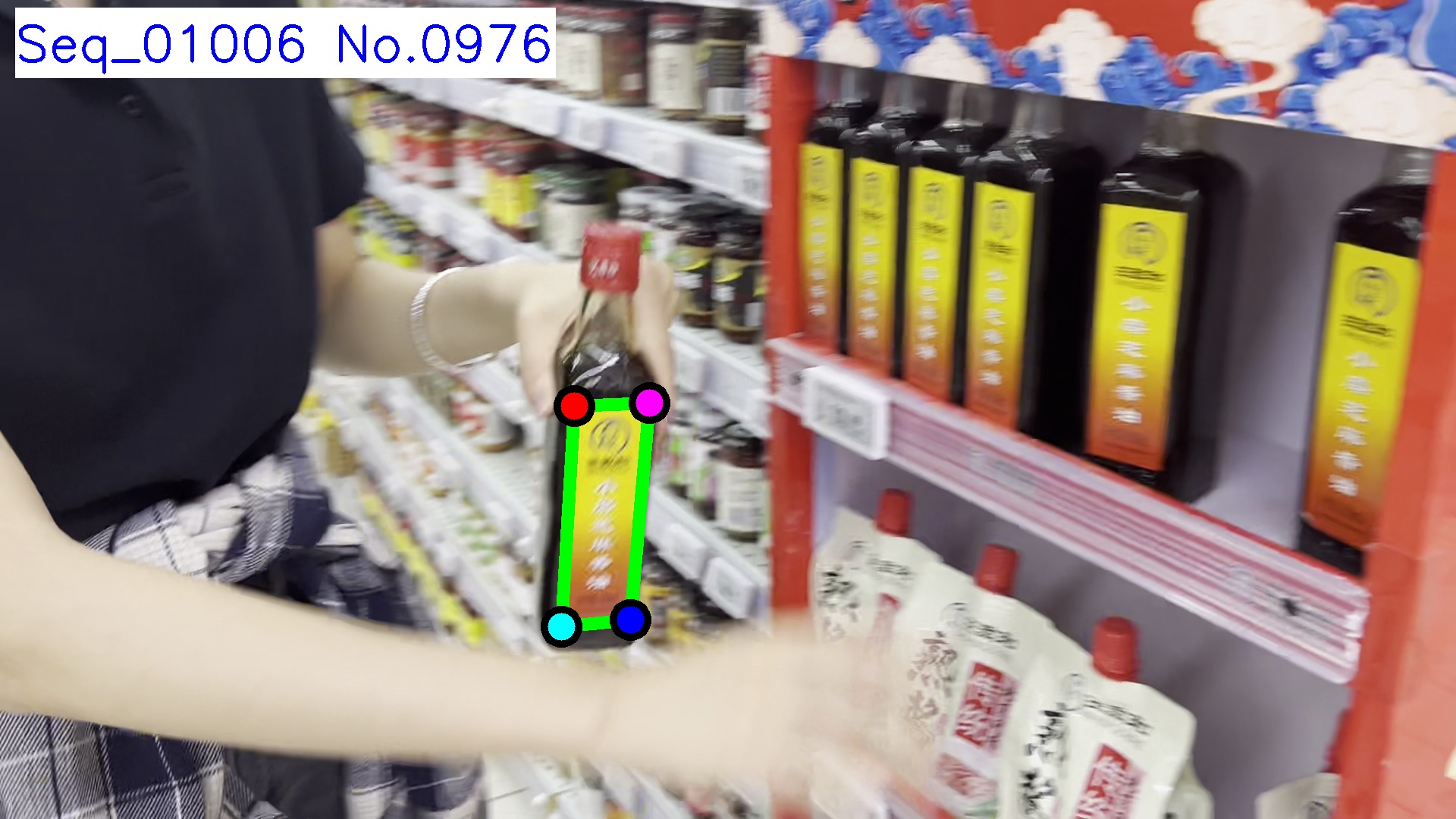}  \includegraphics[width=0.158\linewidth]{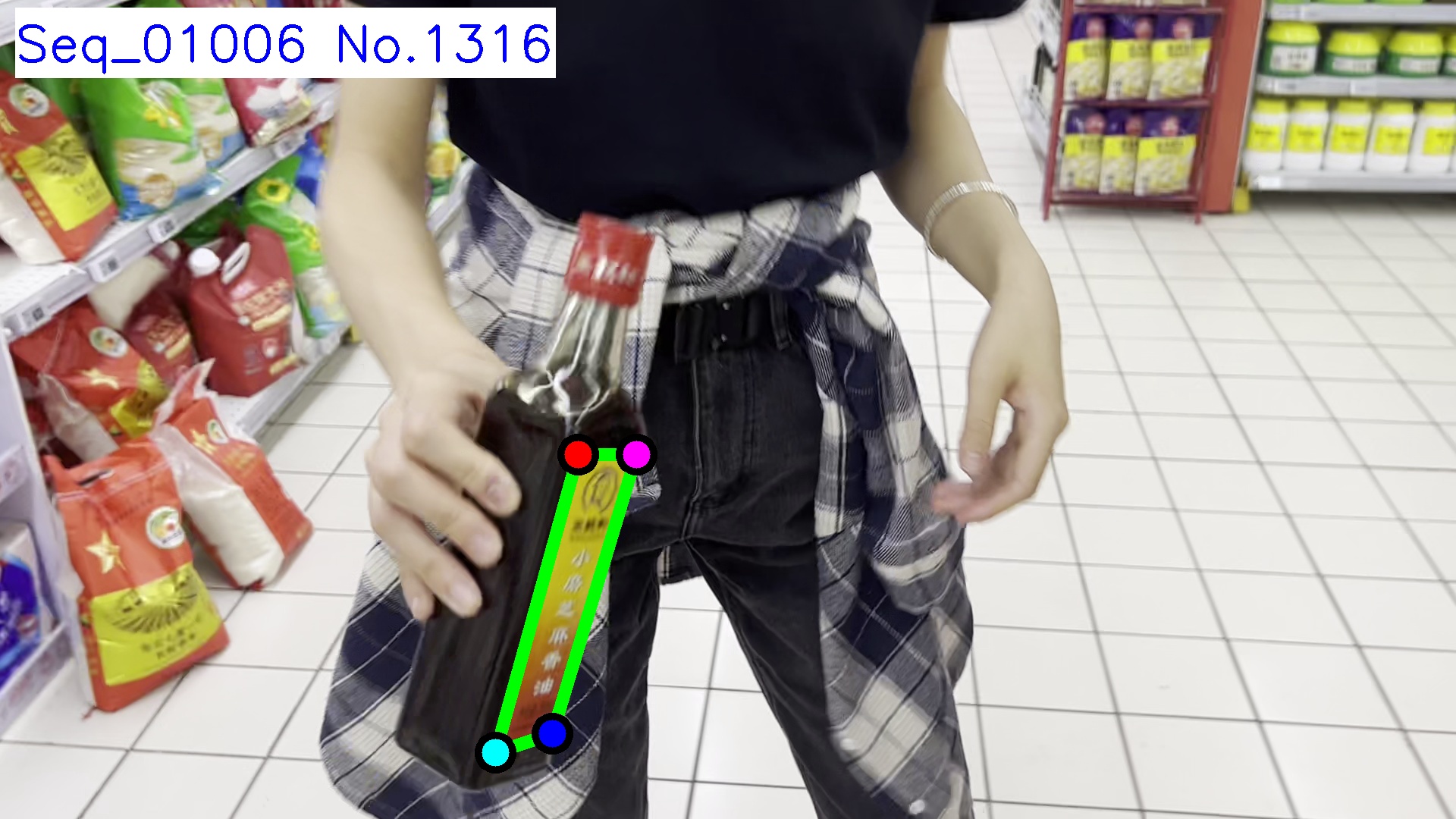} \includegraphics[width=0.158\linewidth]{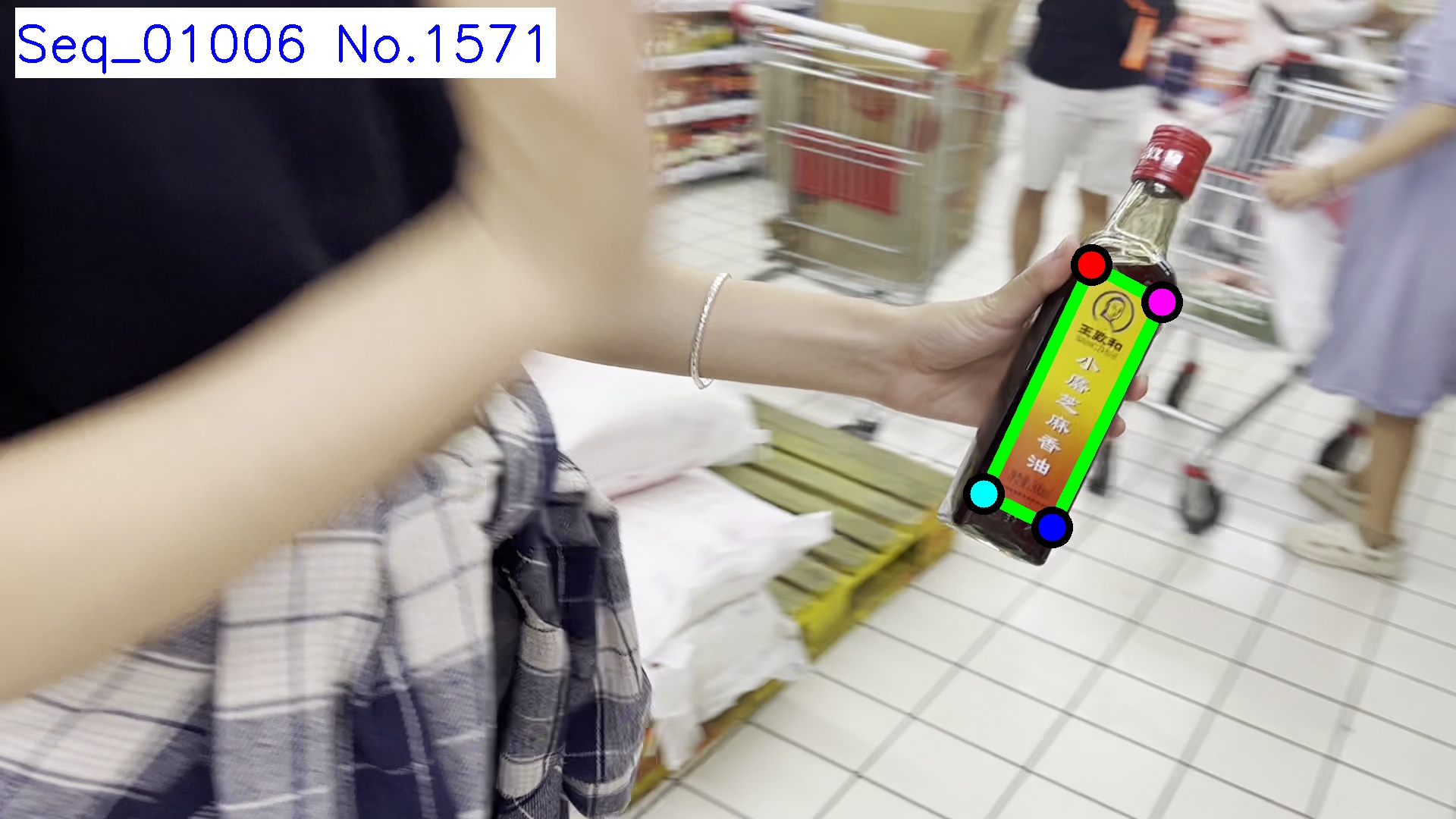} \\
{\small (a) Sequence with similar planar targets} \\
\includegraphics[width=0.158\linewidth]{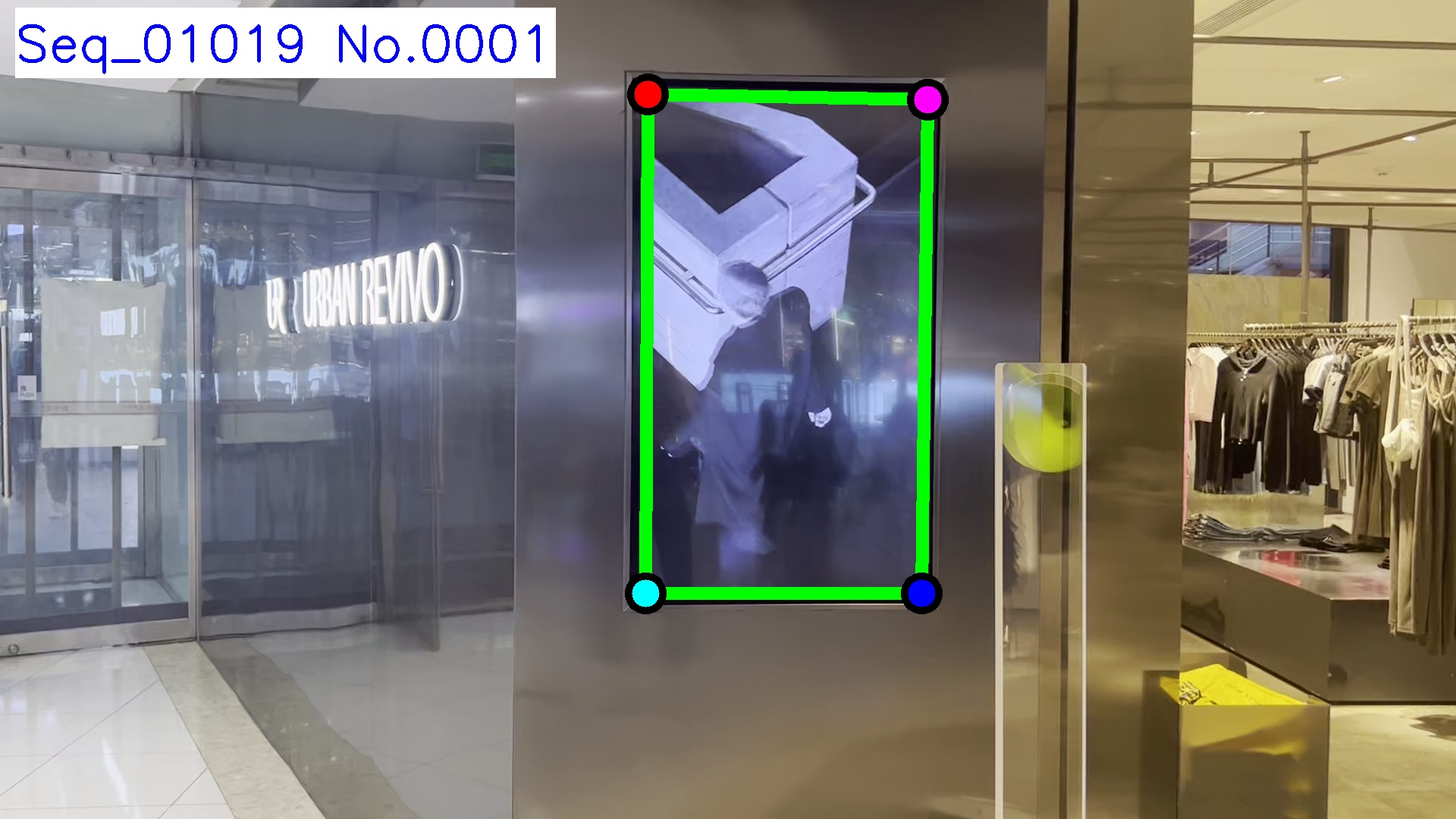} \includegraphics[width=0.158\linewidth]{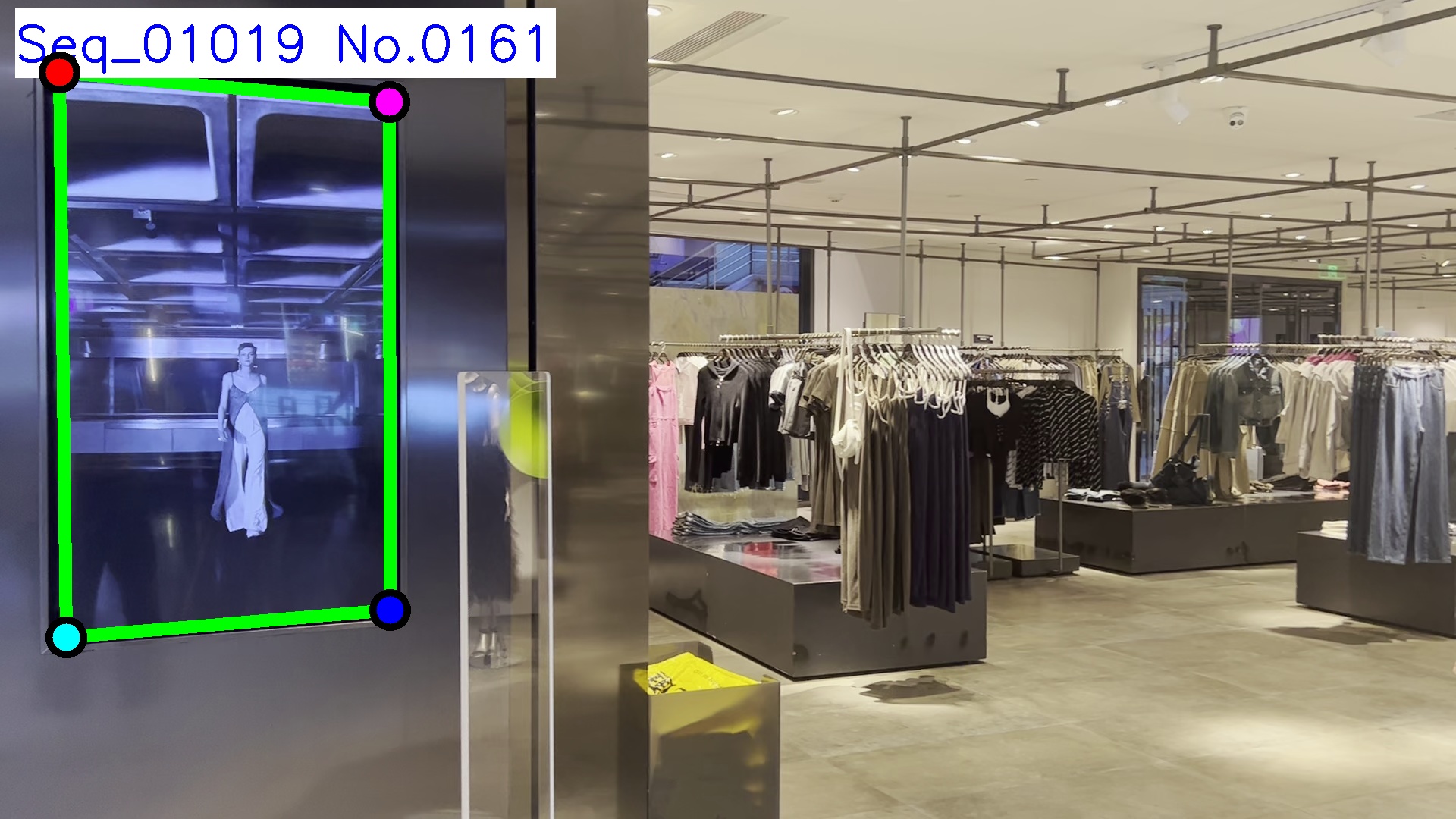} \includegraphics[width=0.158\linewidth]{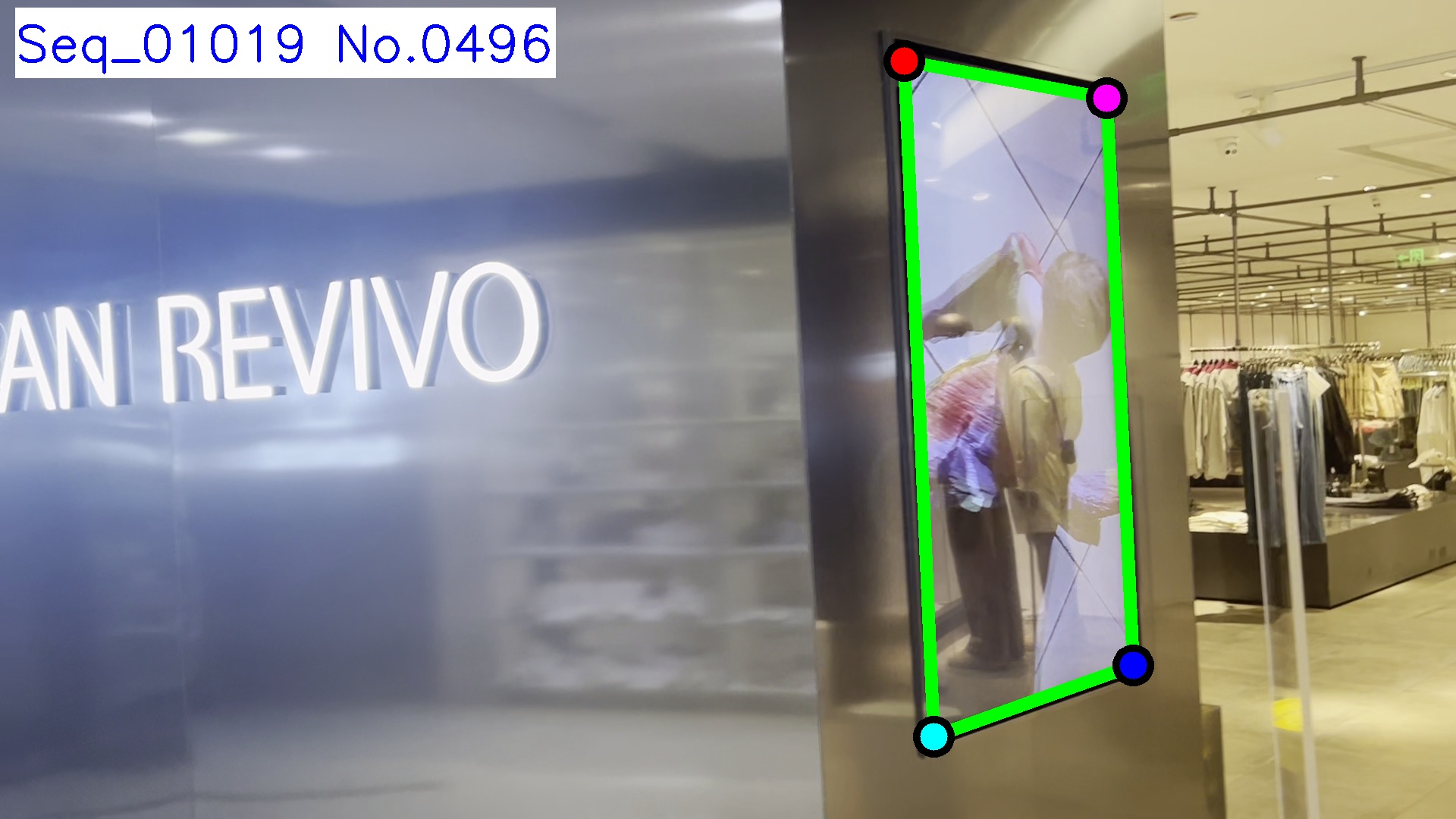} \includegraphics[width=0.158\linewidth]{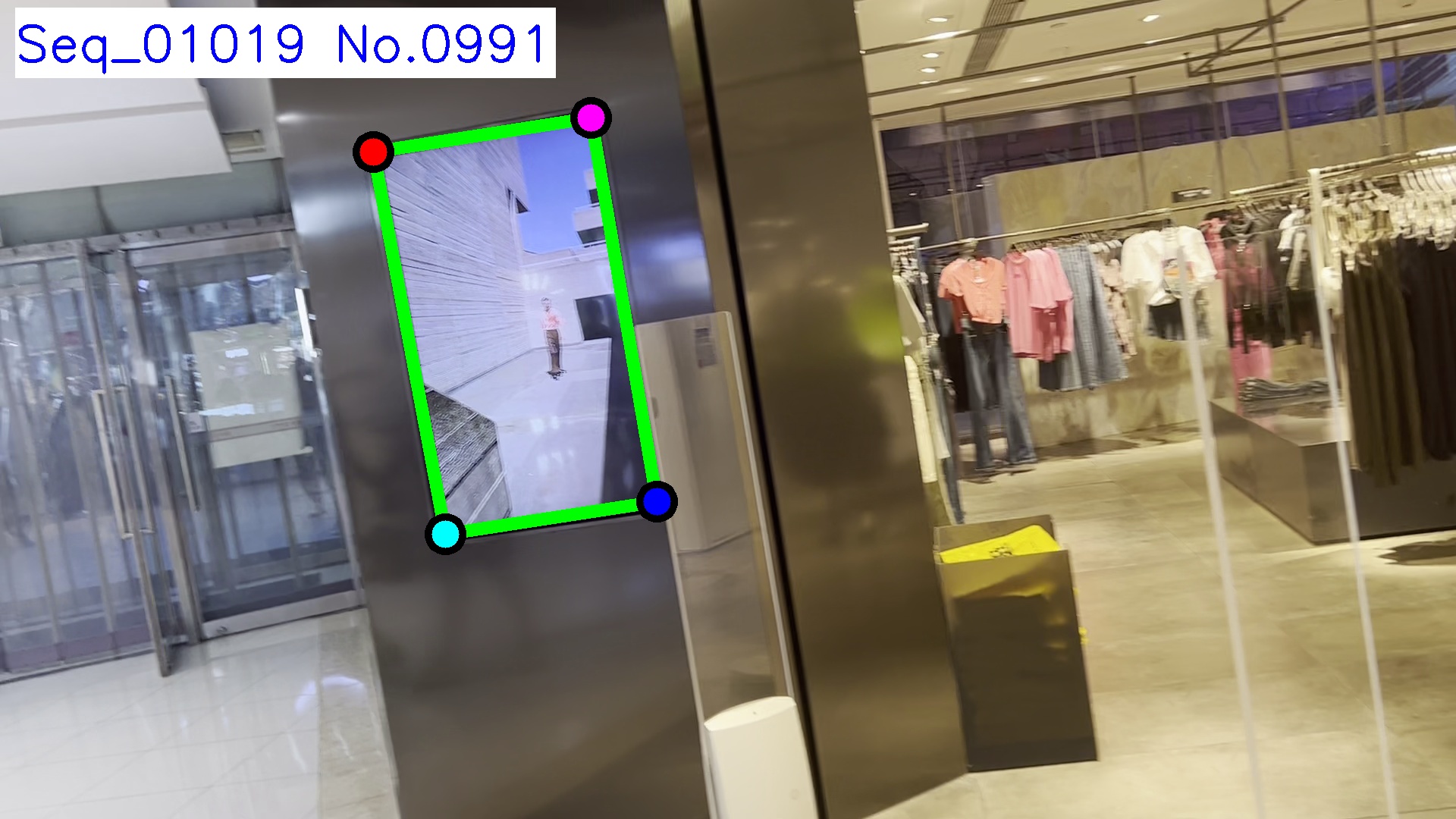}  \includegraphics[width=0.158\linewidth]{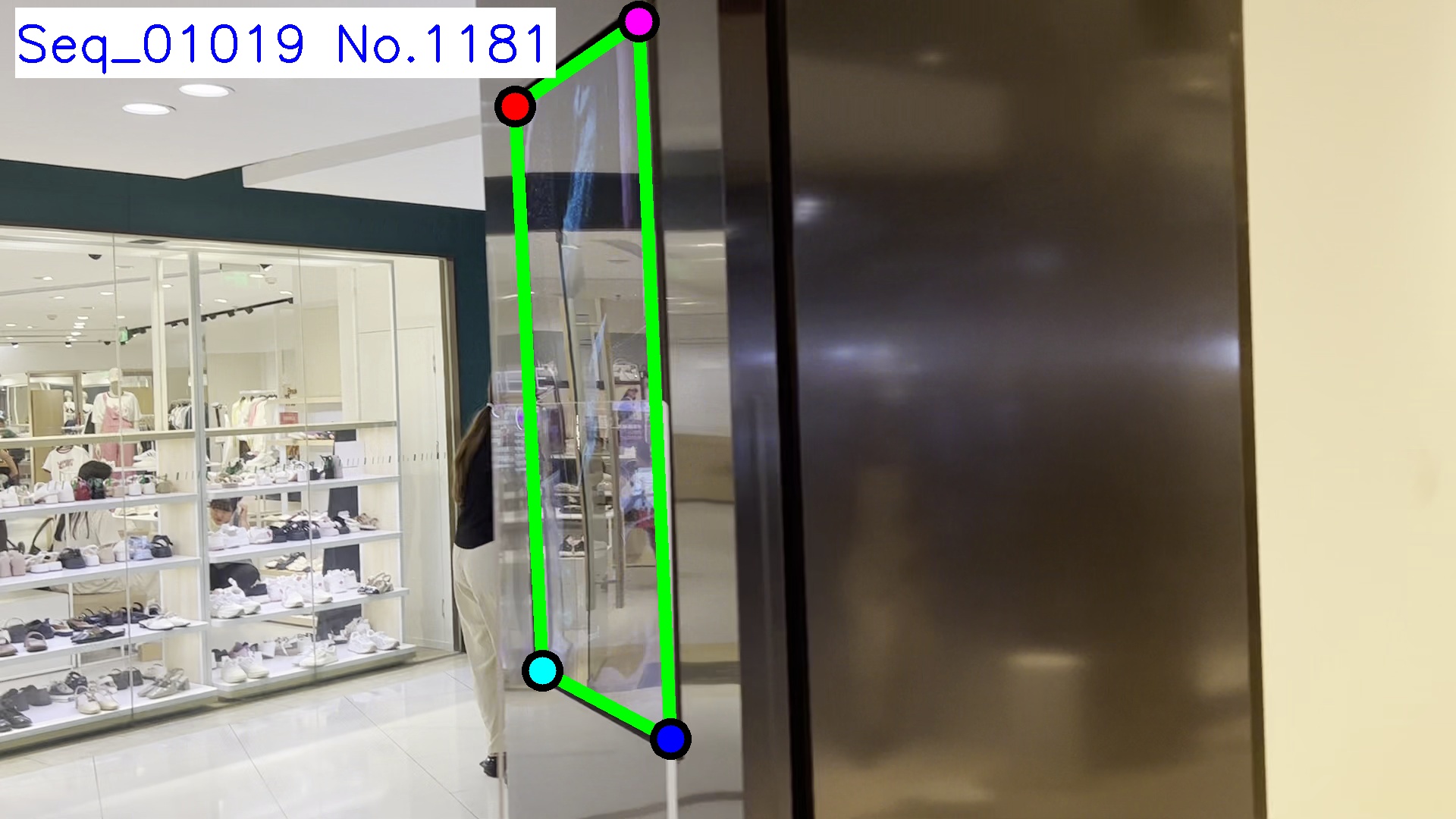} \includegraphics[width=0.158\linewidth]{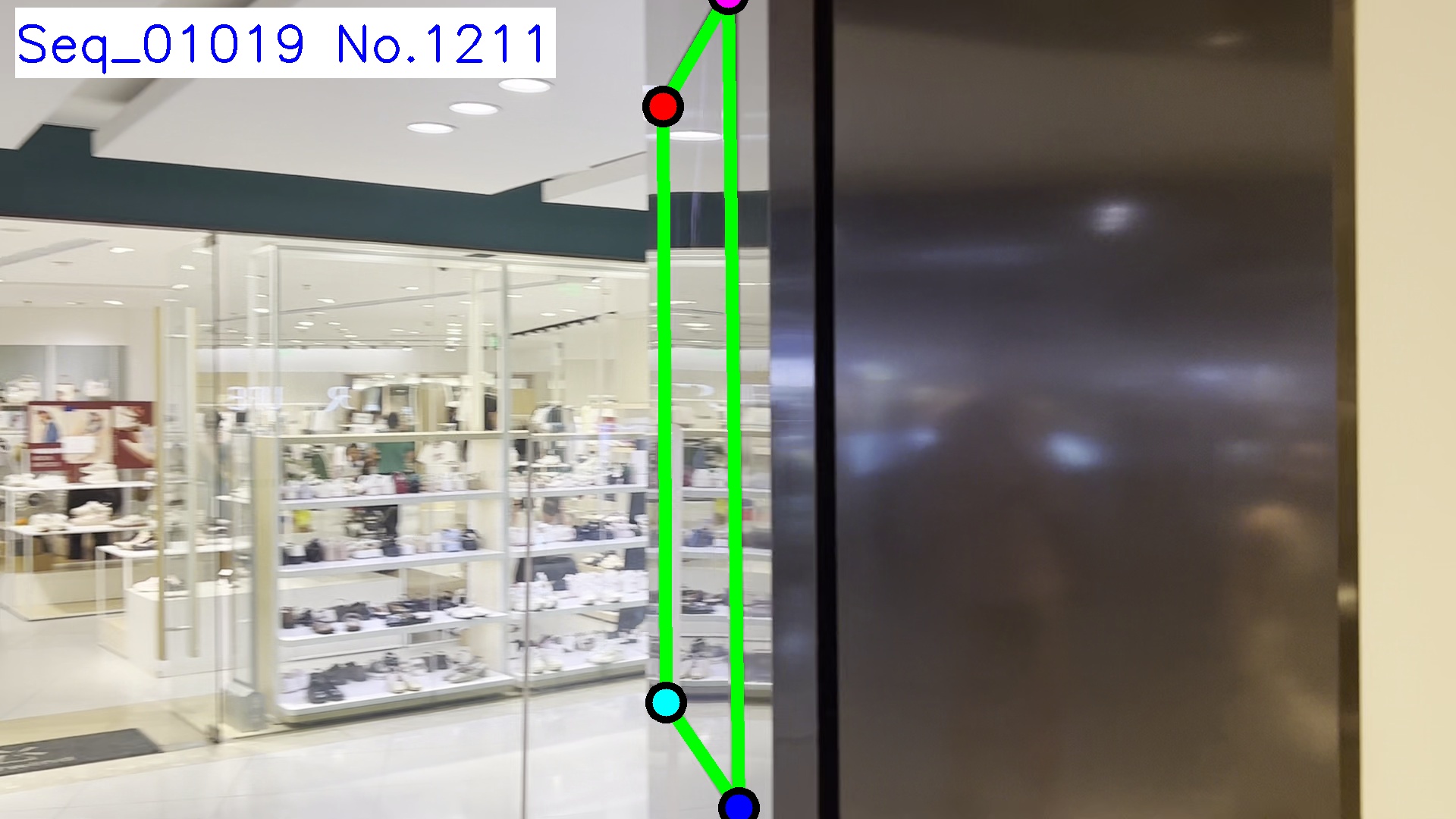} \\
{\small (b) Sequence with sudden appearance change (\eg \ 
electronic screen)} \\
\includegraphics[width=0.158\linewidth]{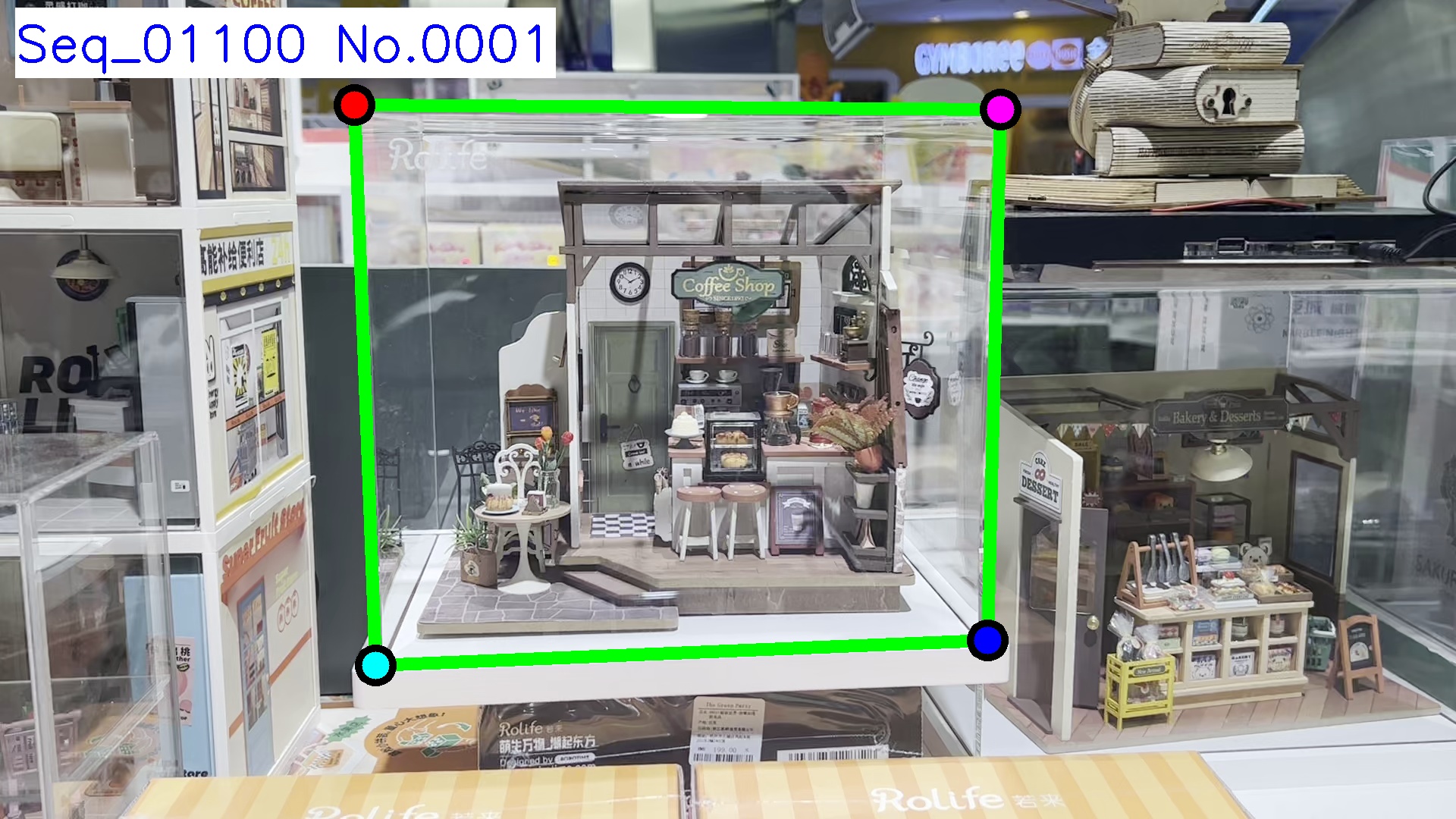} \includegraphics[width=0.158\linewidth]{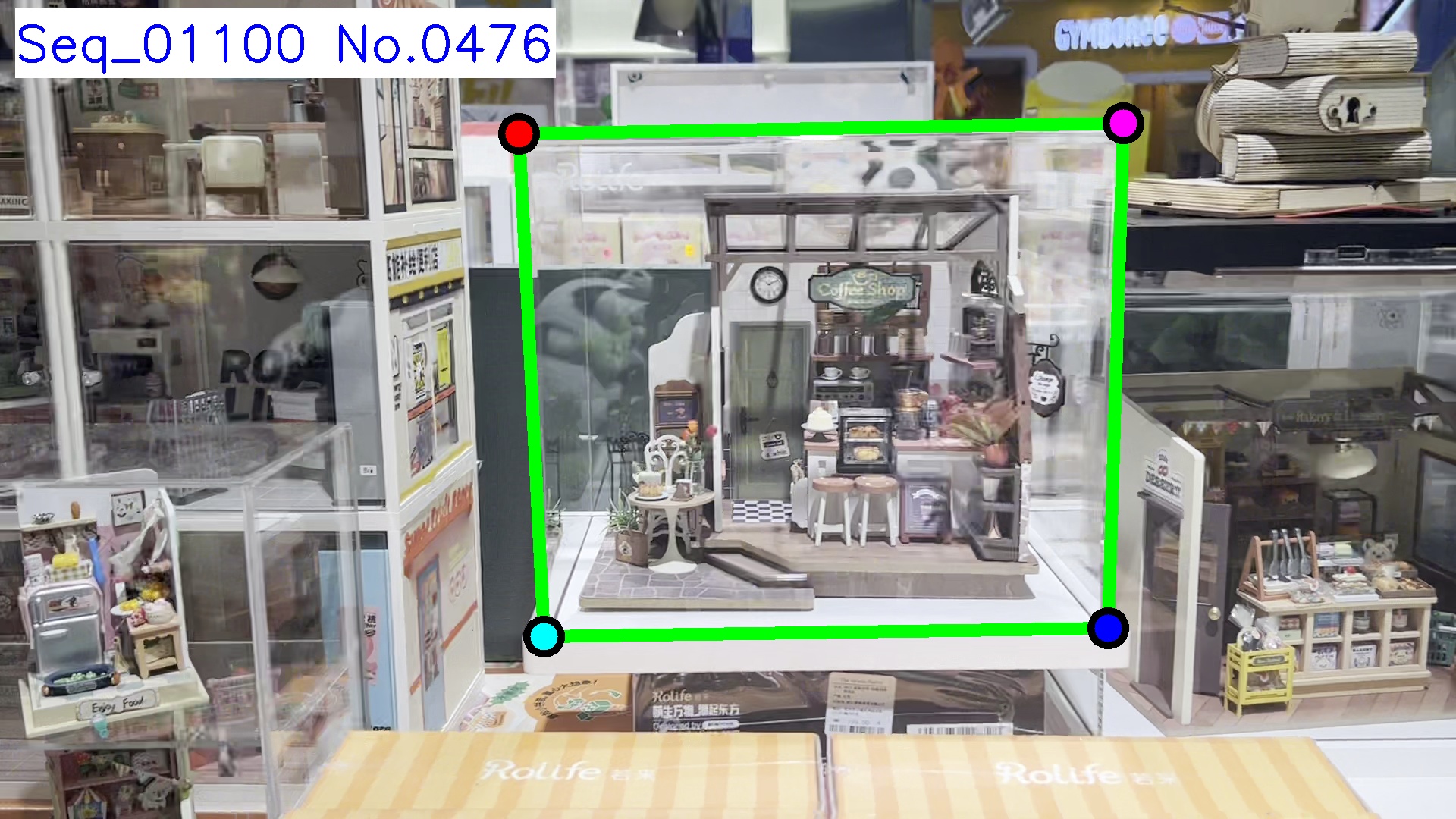} \includegraphics[width=0.158\linewidth]{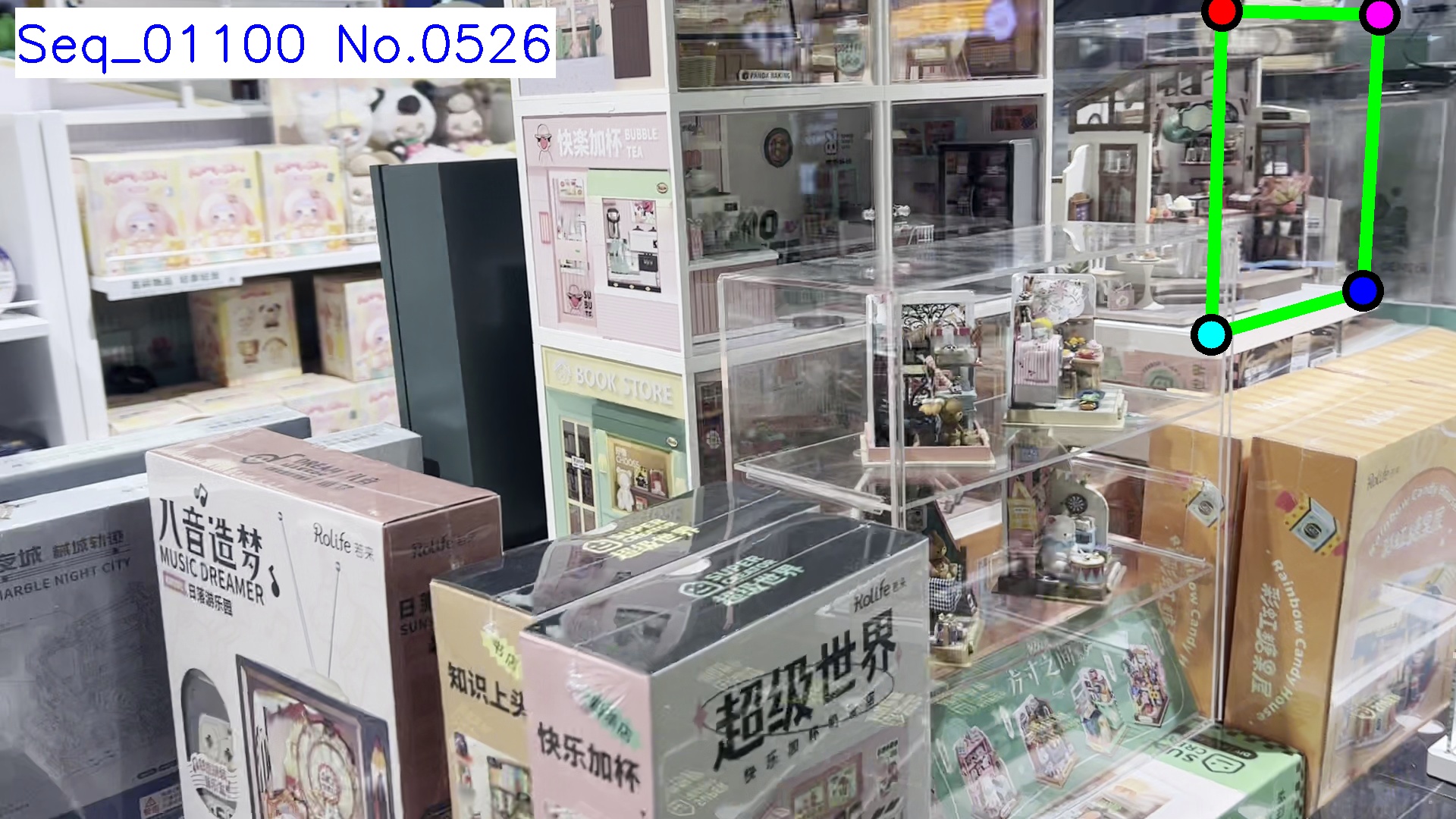} \includegraphics[width=0.158\linewidth]{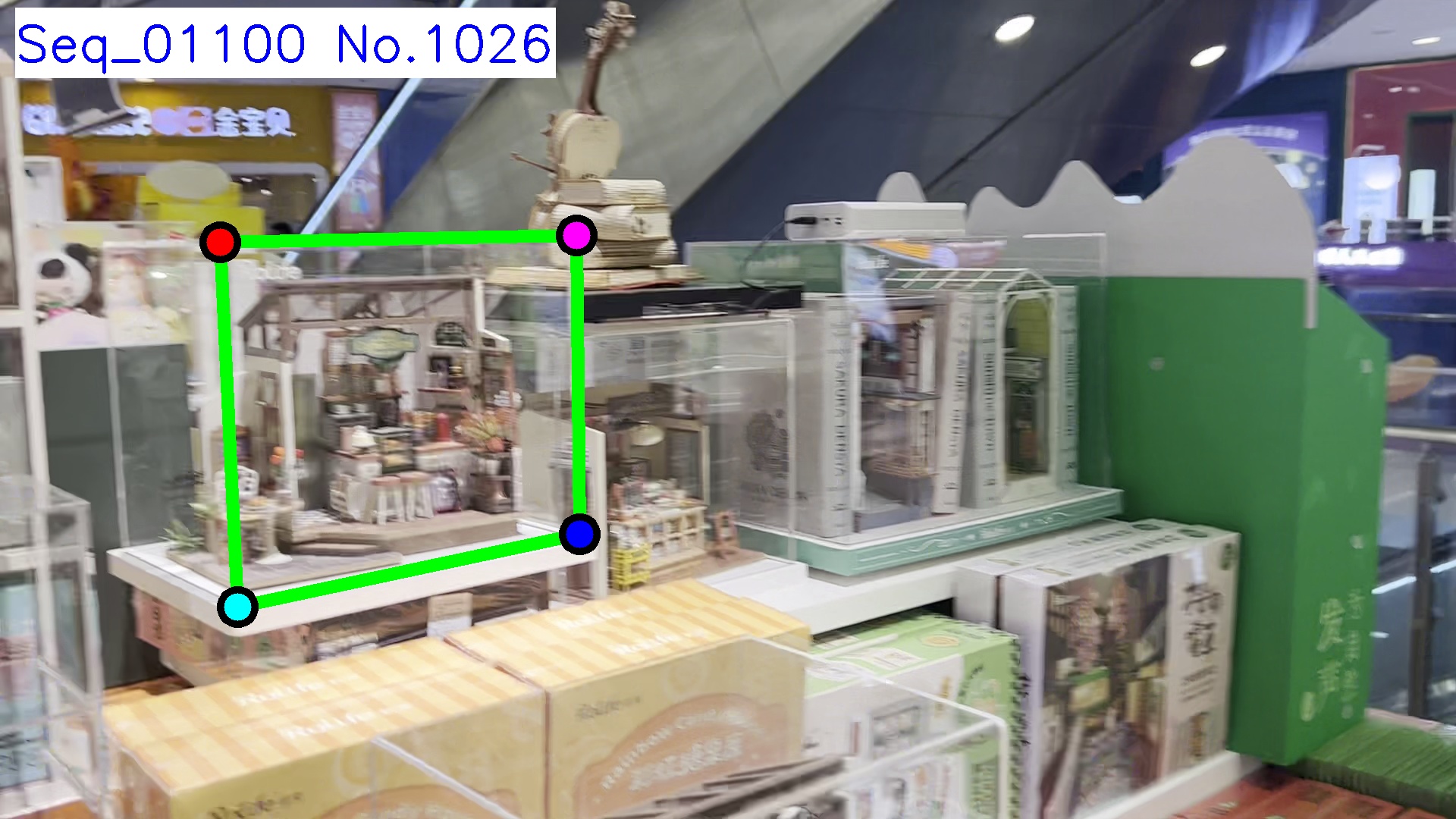}  \includegraphics[width=0.158\linewidth]{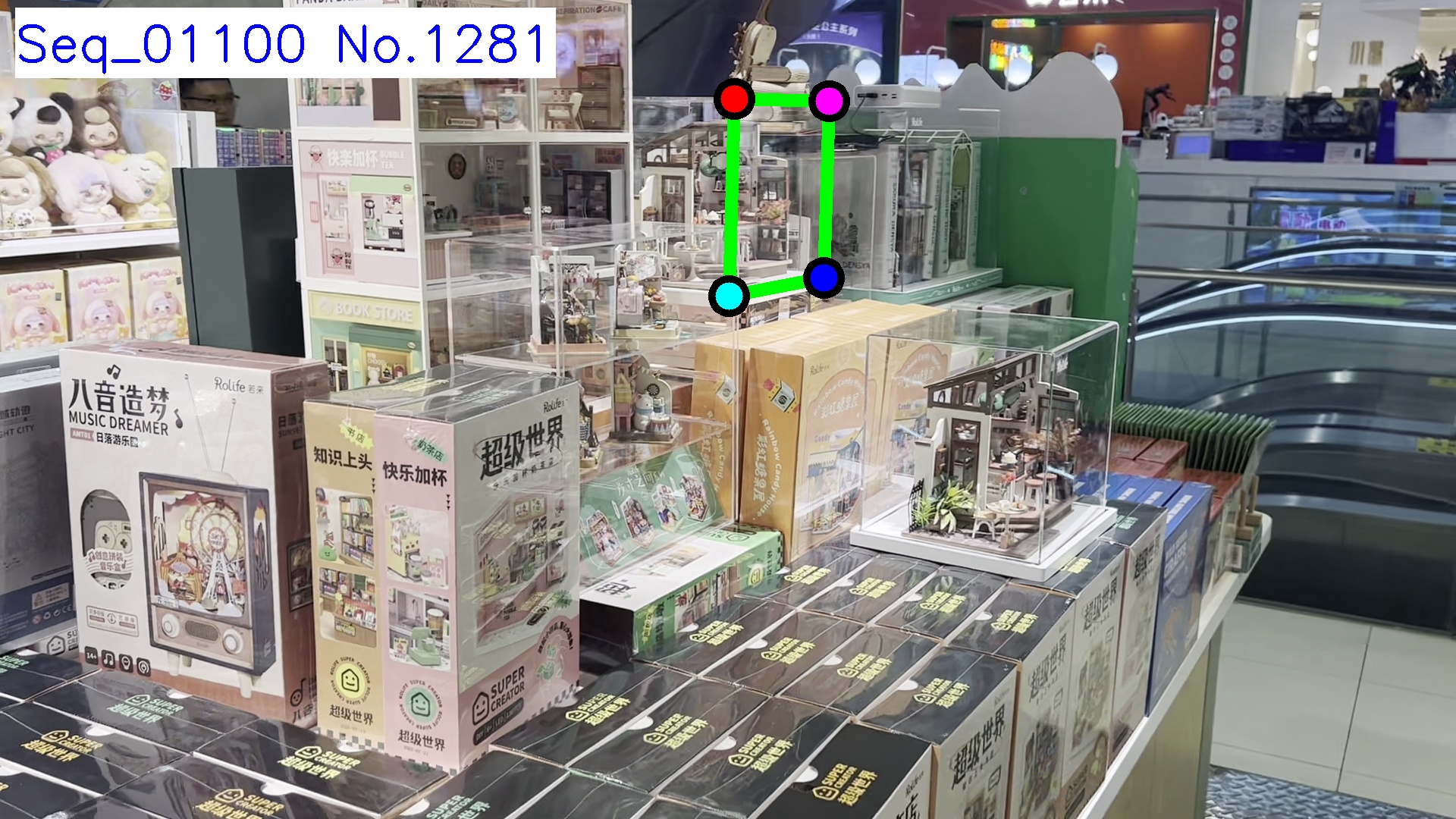} \includegraphics[width=0.158\linewidth]{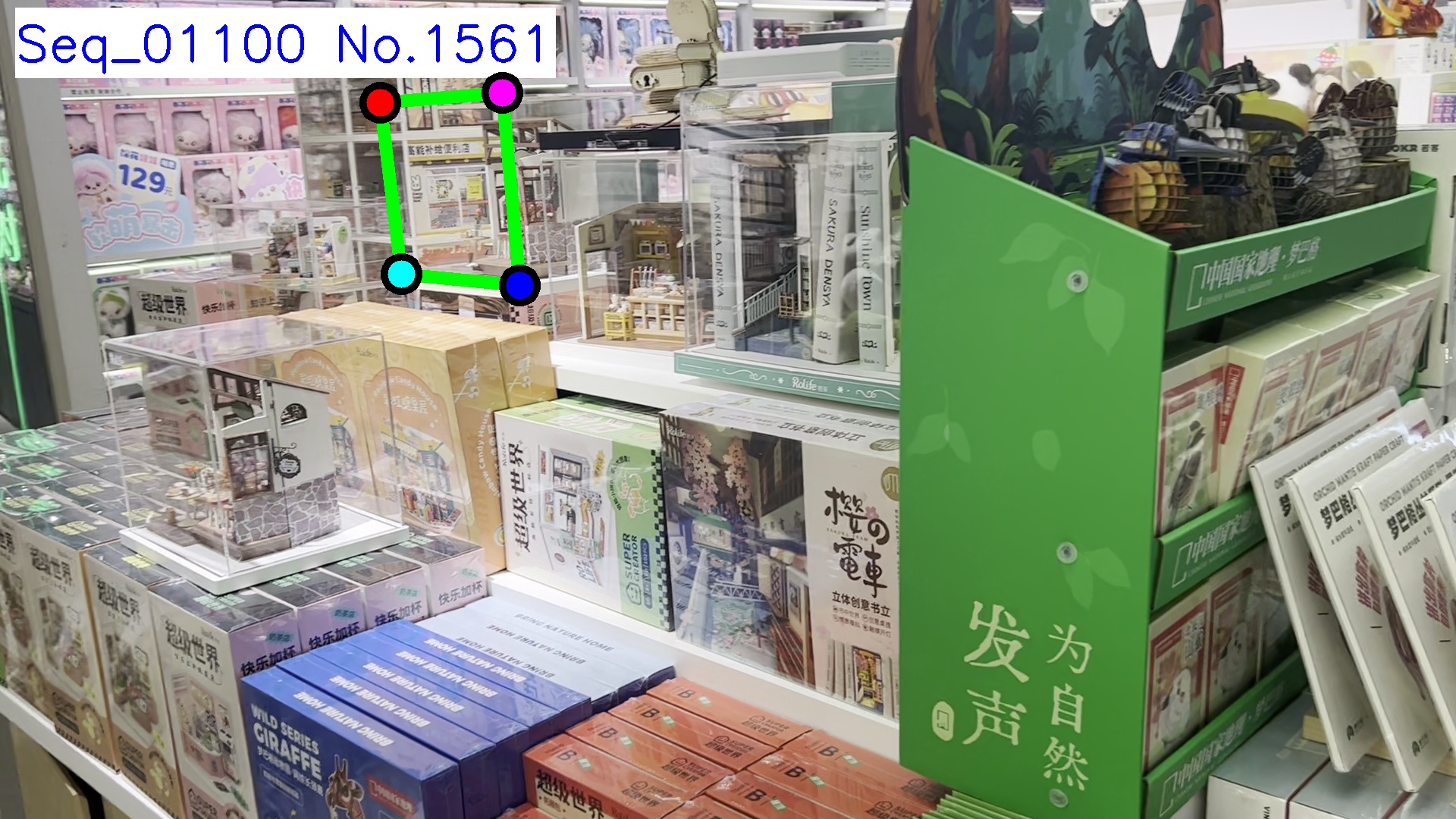} \\
{\small (c) Sequence with transparent planar target} \\
\includegraphics[width=0.158\linewidth]{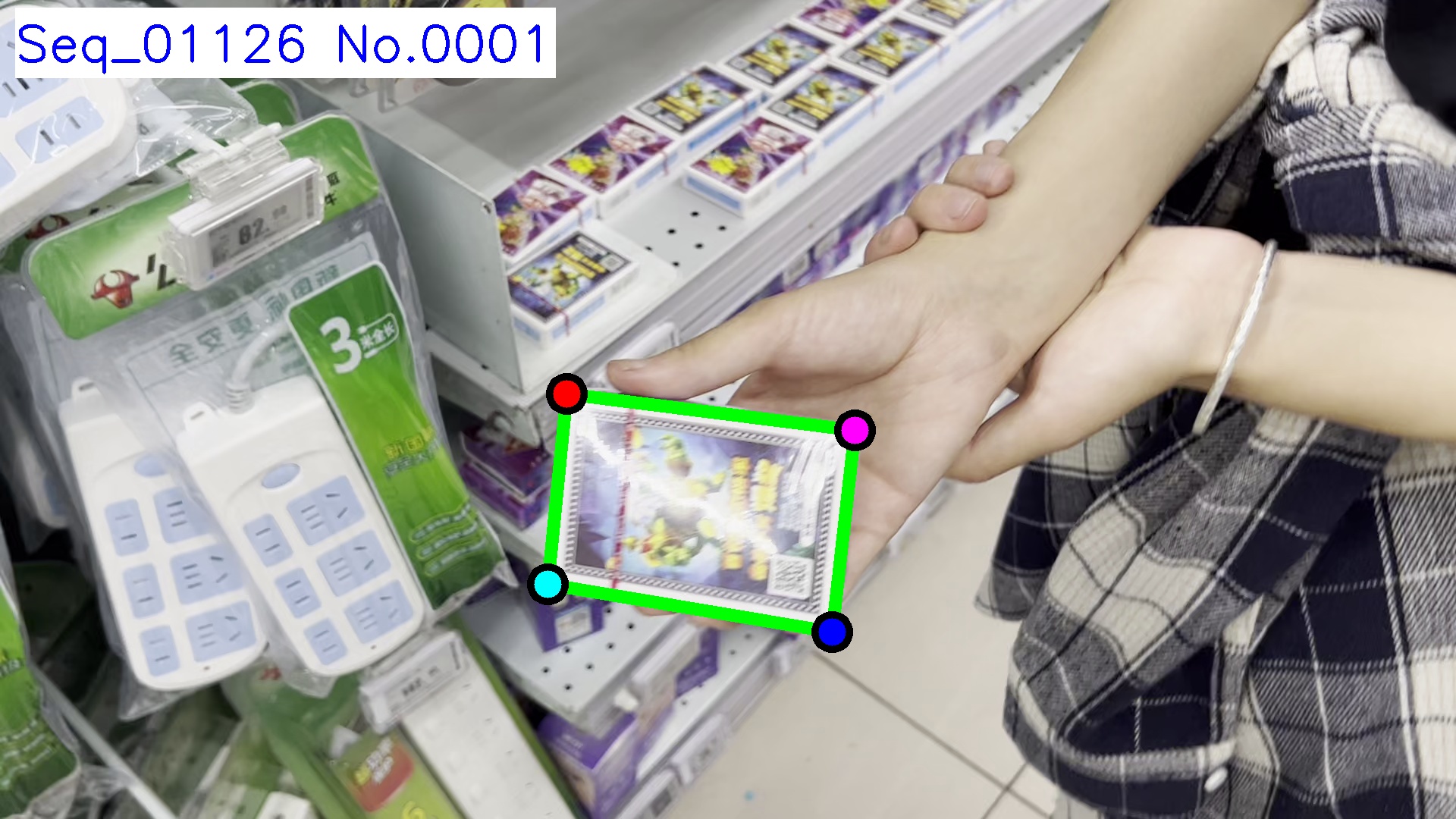}
\includegraphics[width=0.158\linewidth]{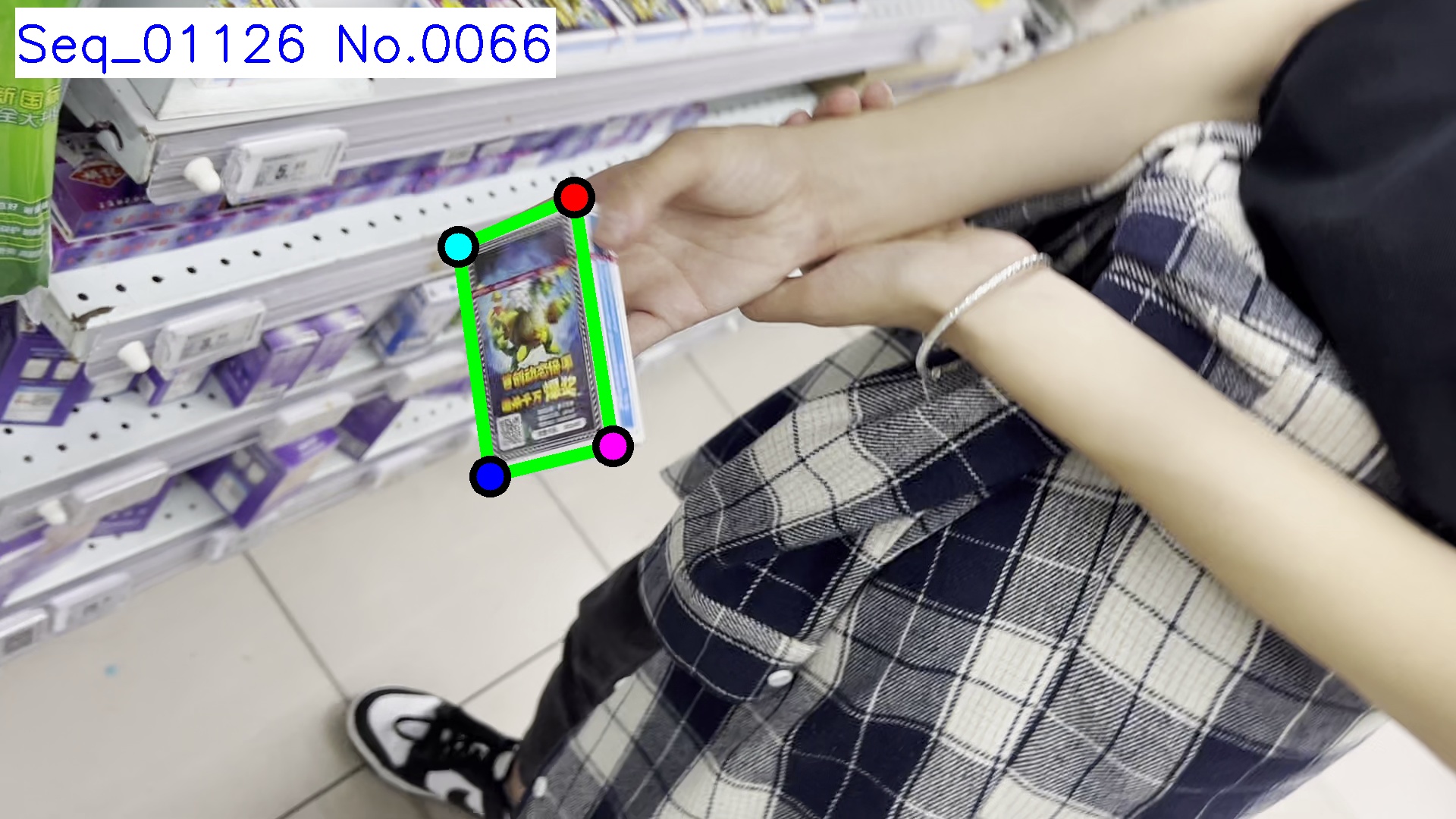}
\includegraphics[width=0.158\linewidth]{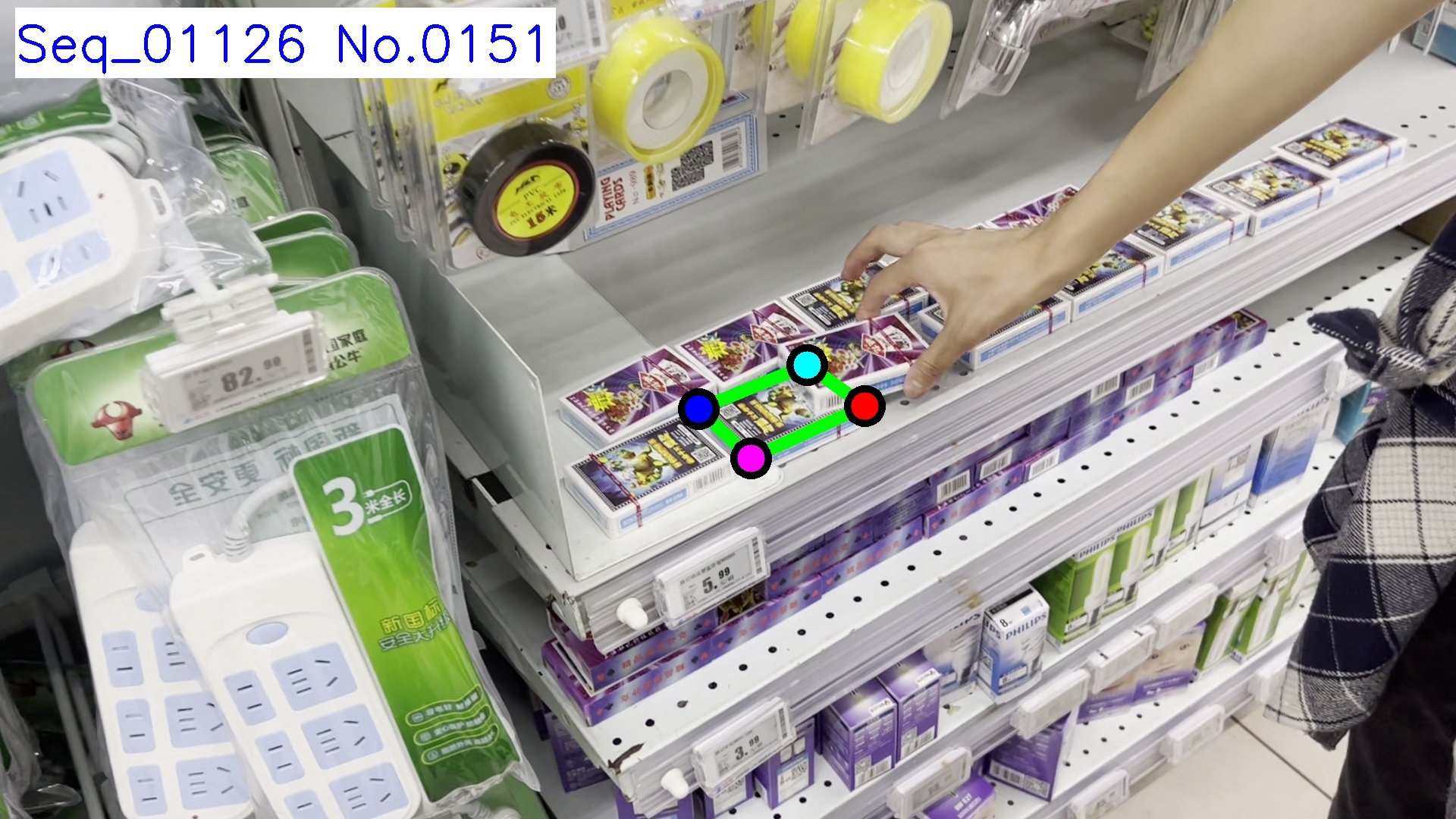}
\includegraphics[width=0.158\linewidth]{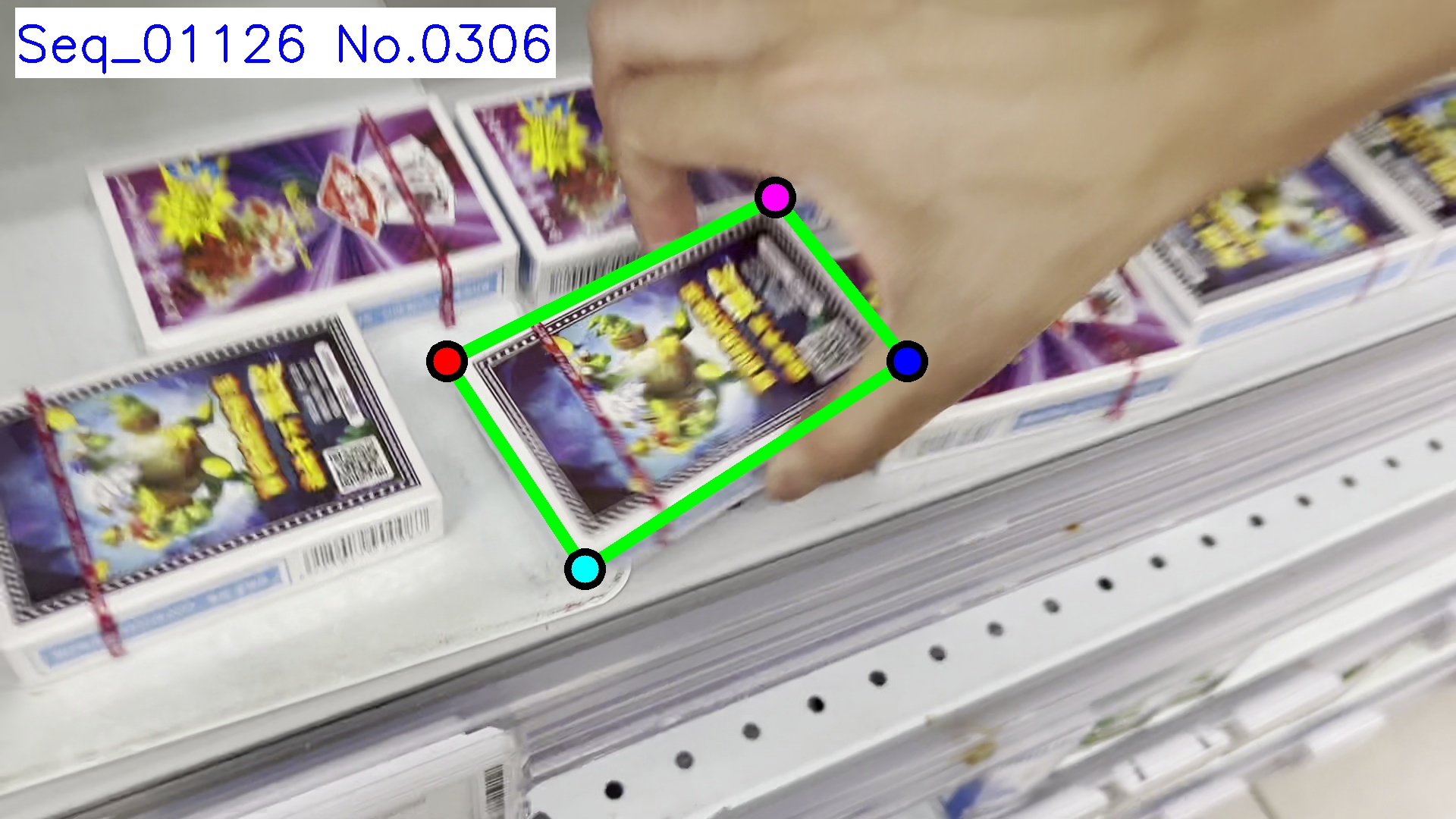}
\includegraphics[width=0.158\linewidth]{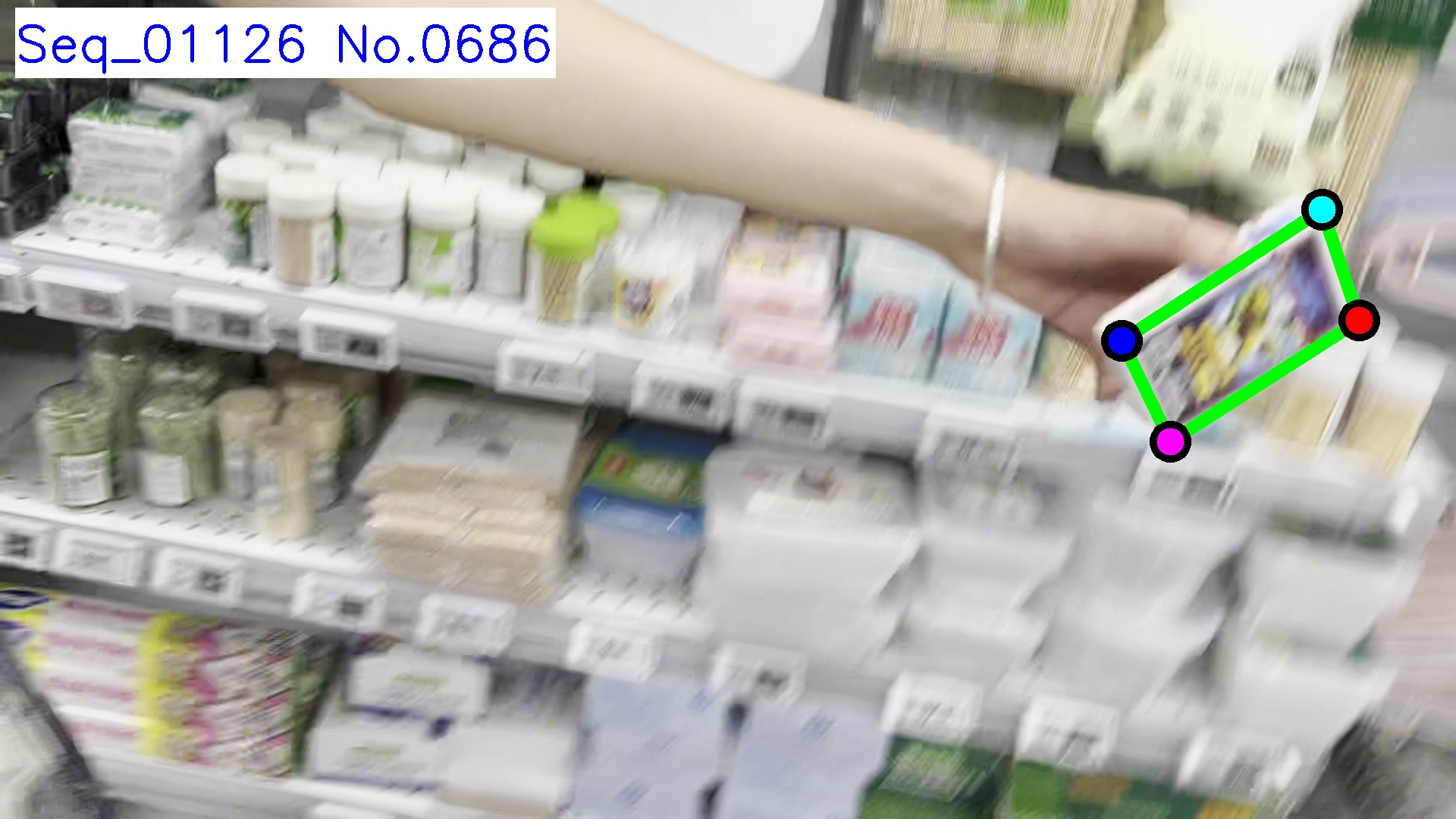}
\includegraphics[width=0.158\linewidth]{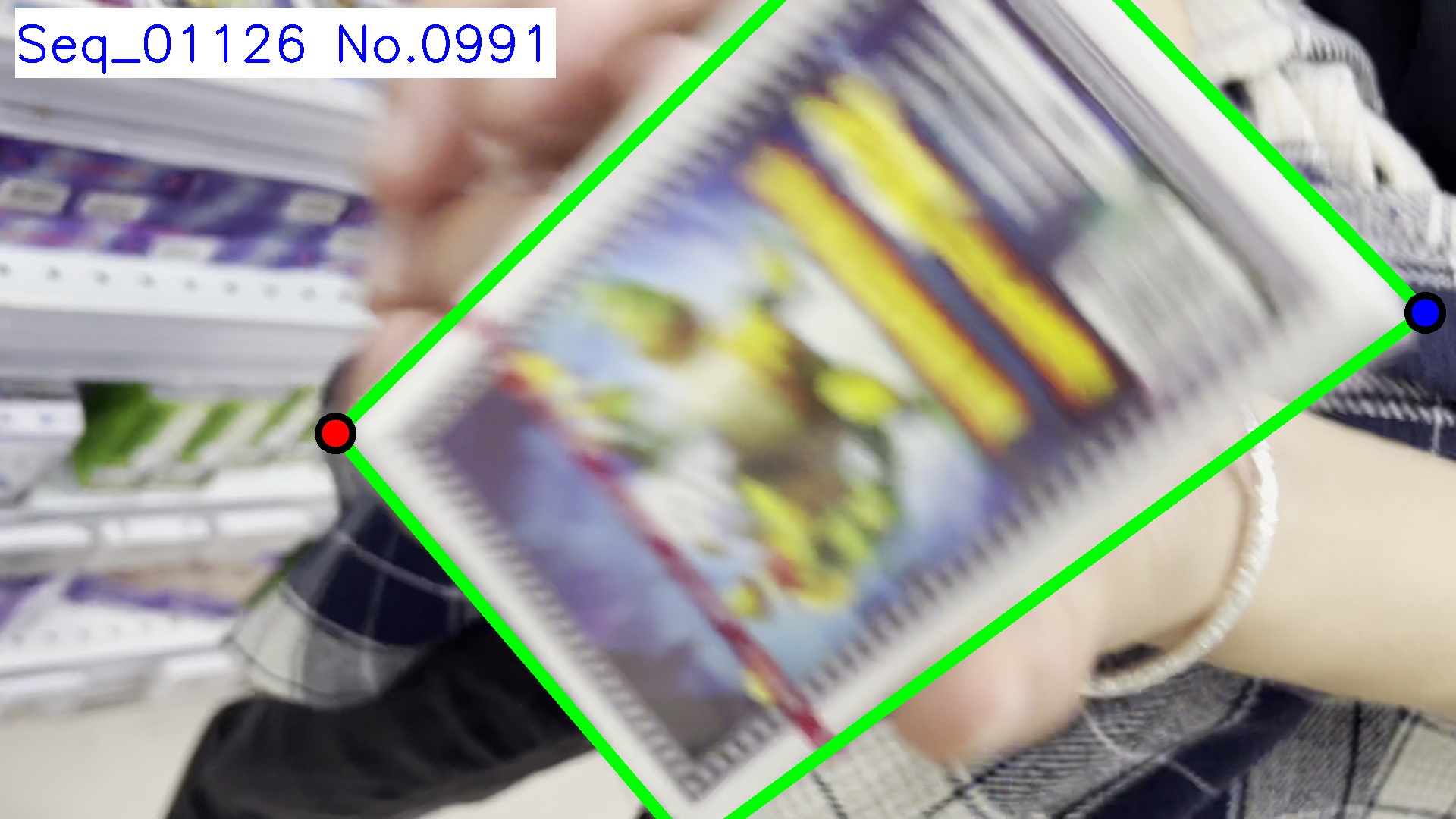} \\
{\small (d) Sequence with large scale variation} \\
\multicolumn{2}{c}{
\includegraphics[width=0.092\linewidth]{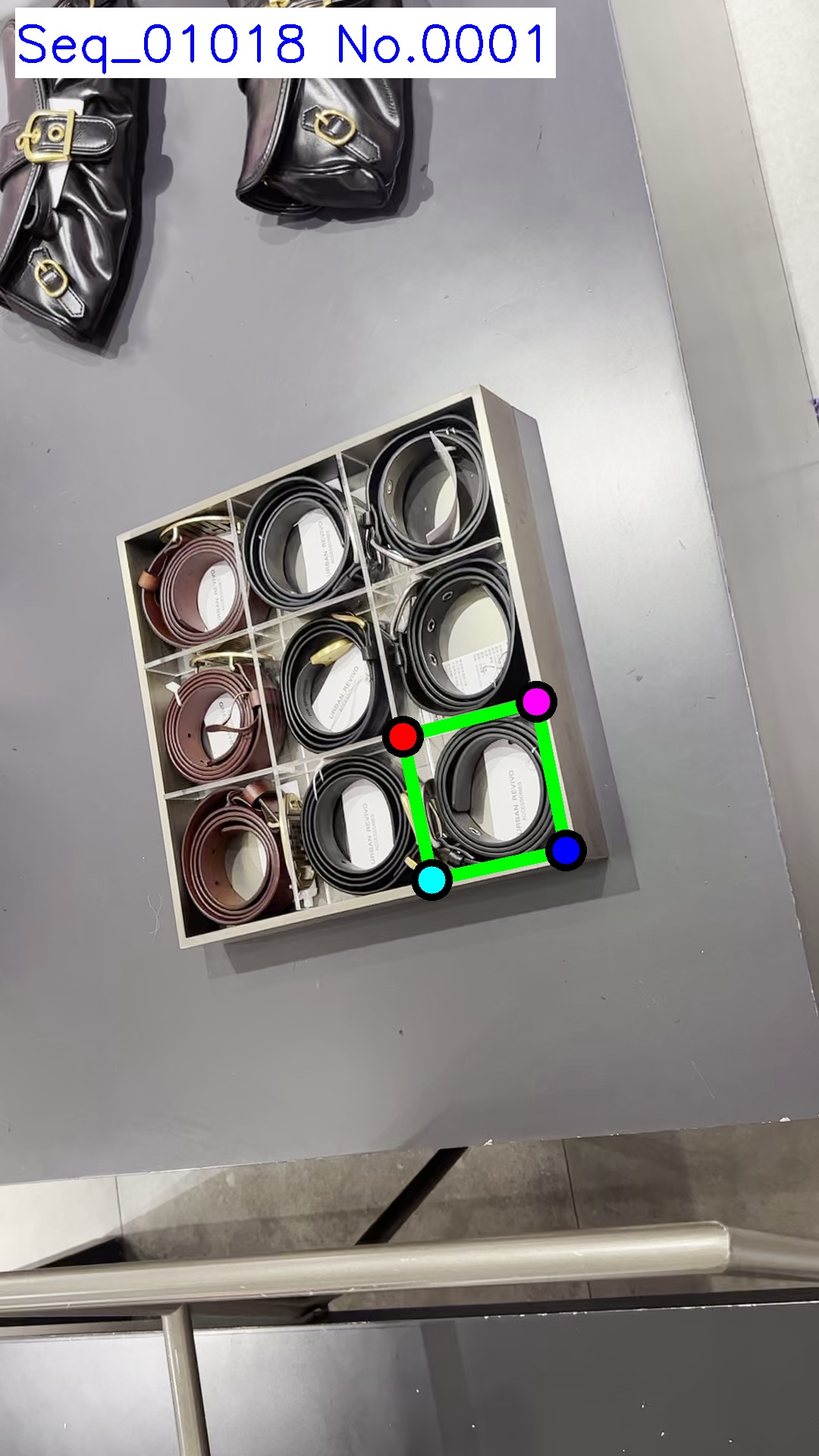}
\includegraphics[width=0.092\linewidth]{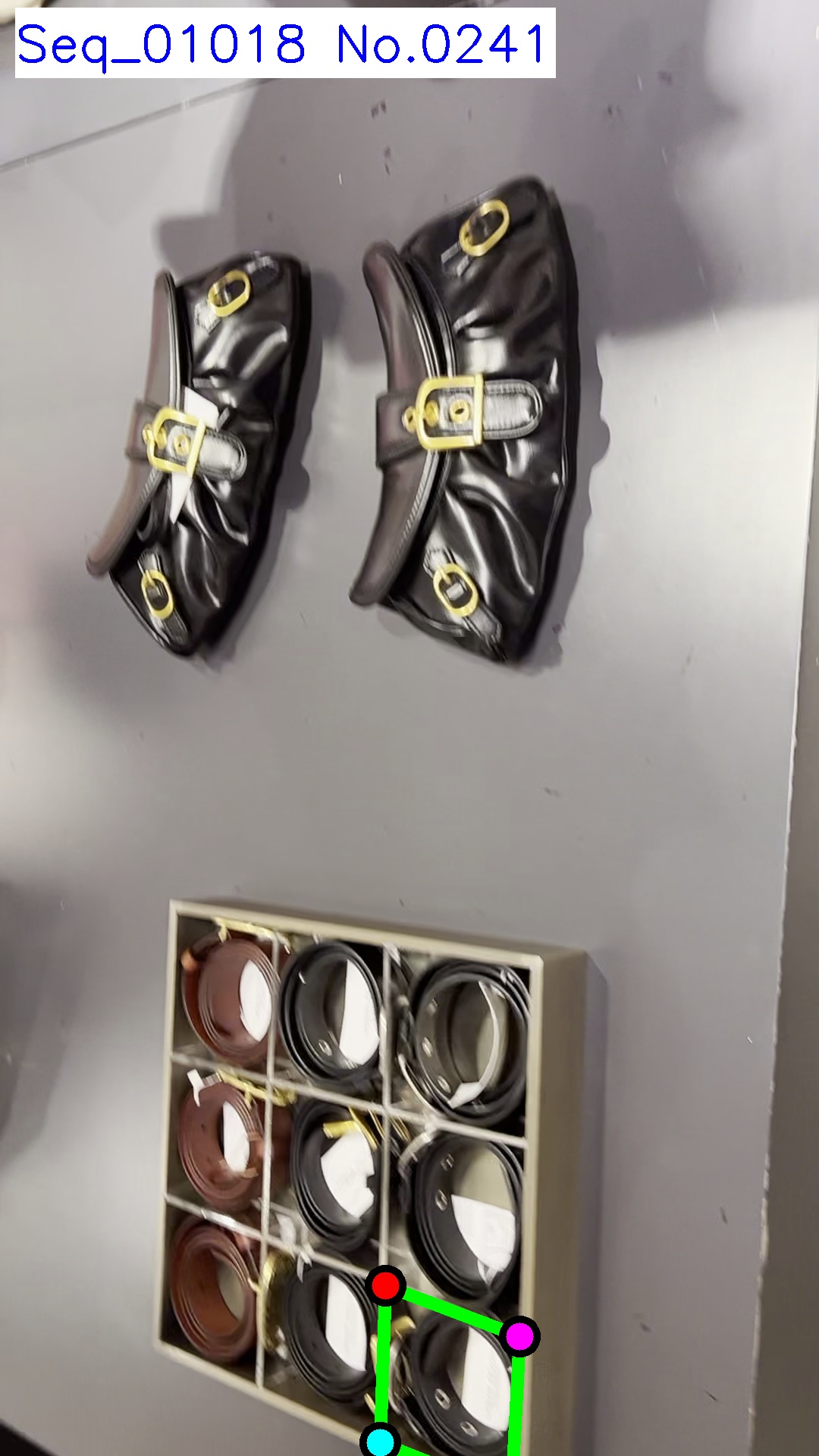}
\includegraphics[width=0.092\linewidth]{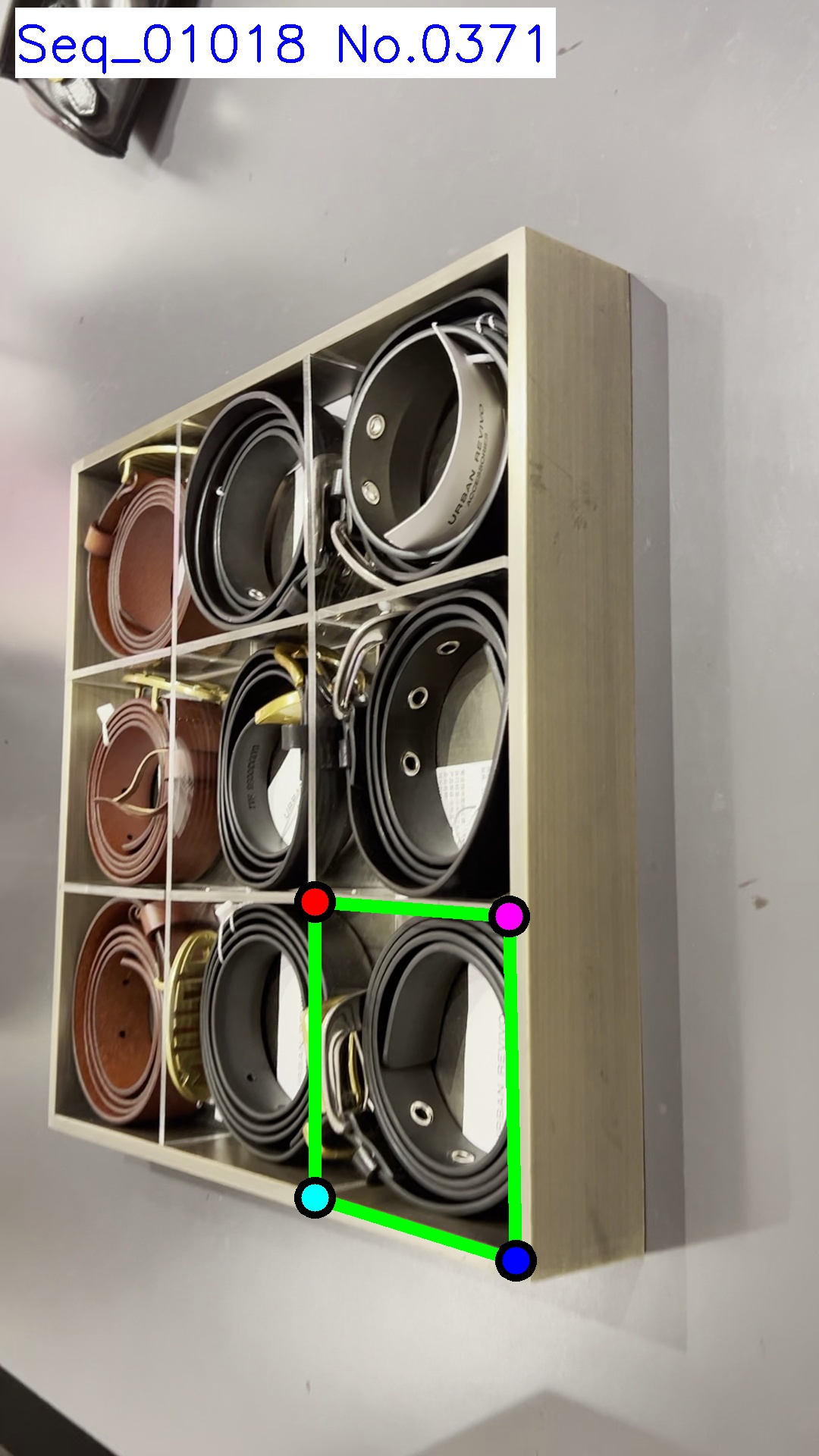}
\includegraphics[width=0.092\linewidth]{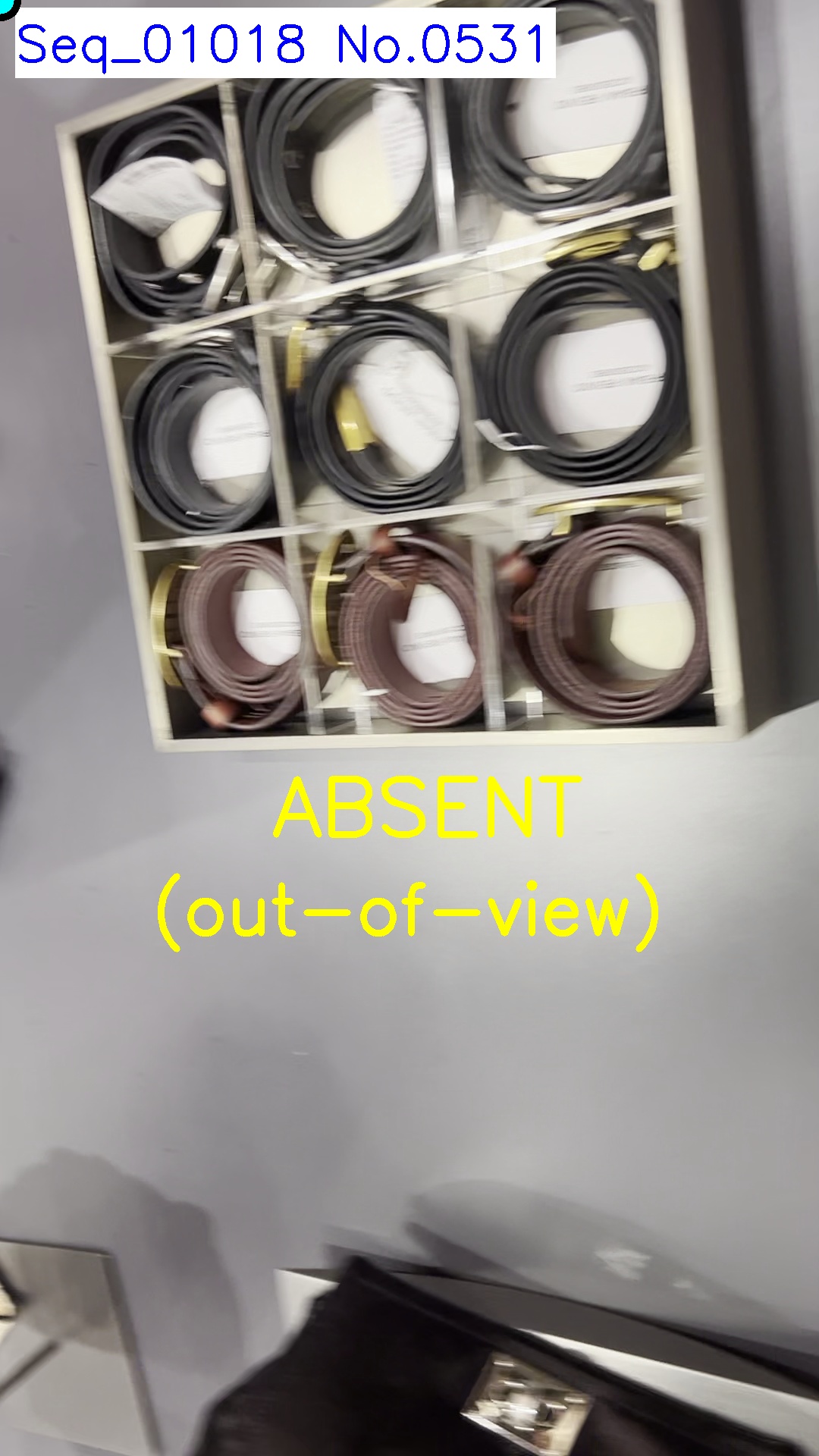}
\includegraphics[width=0.092\linewidth]{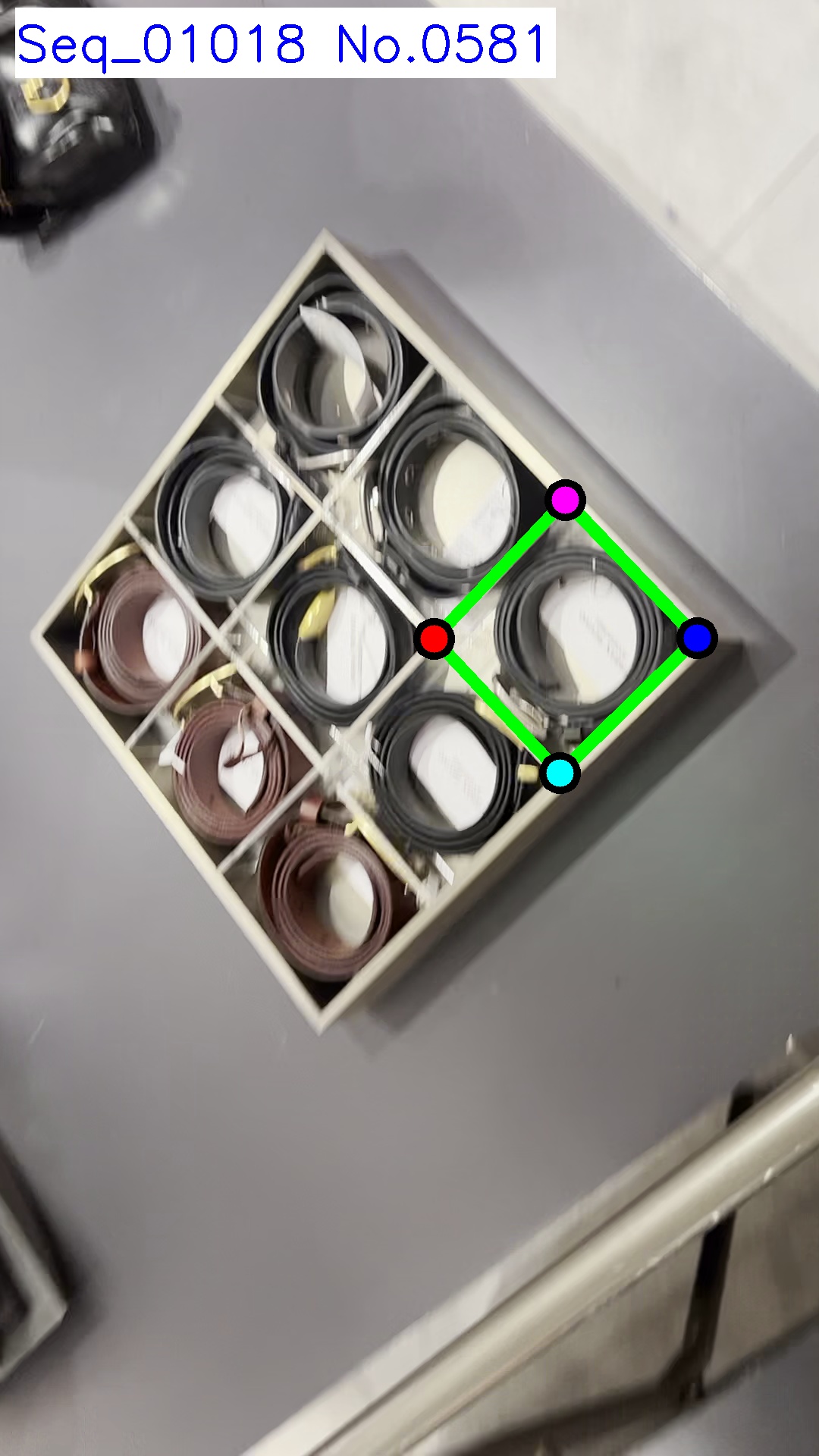}
\includegraphics[width=0.092\linewidth]{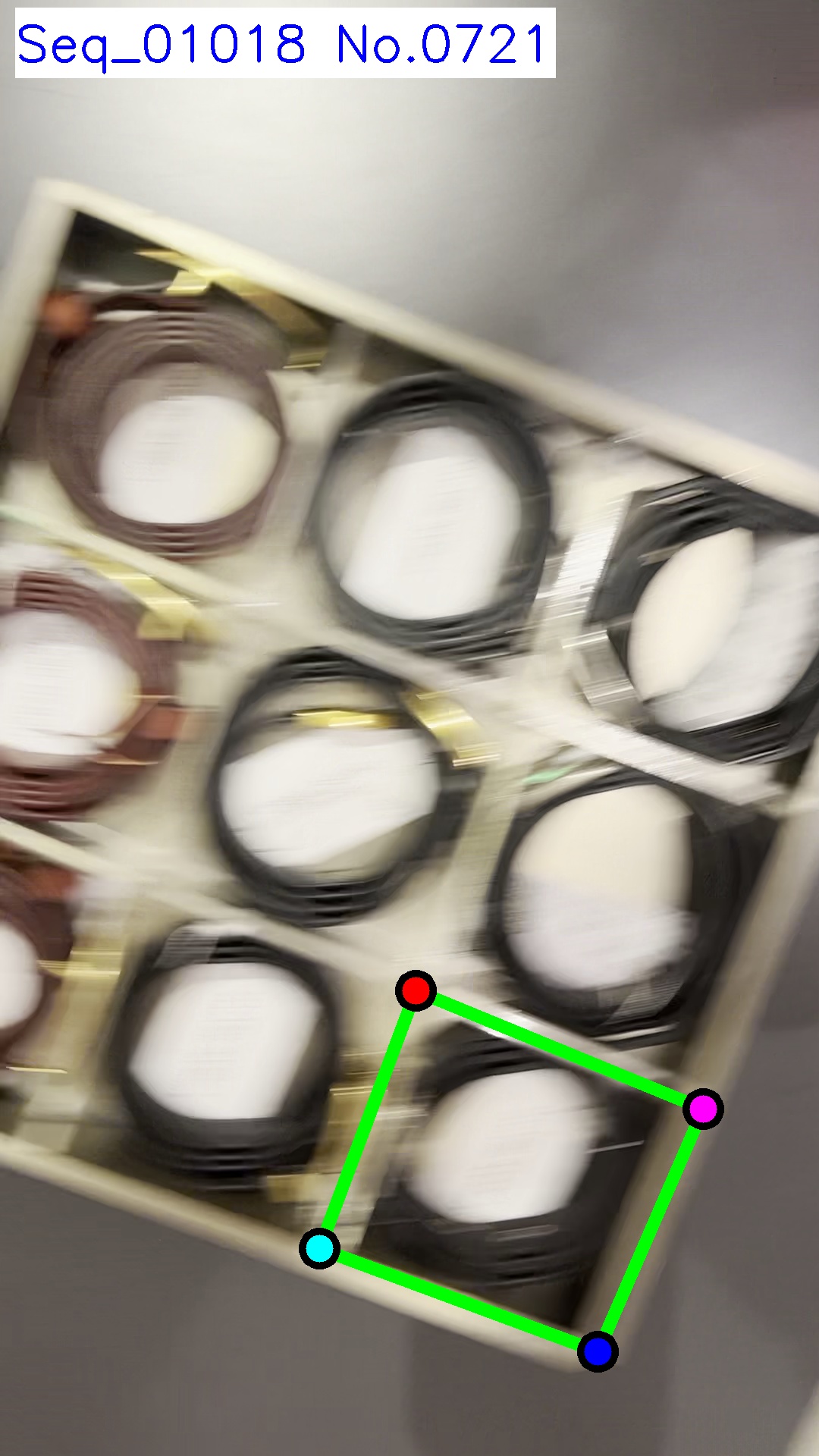}
\includegraphics[width=0.092\linewidth]{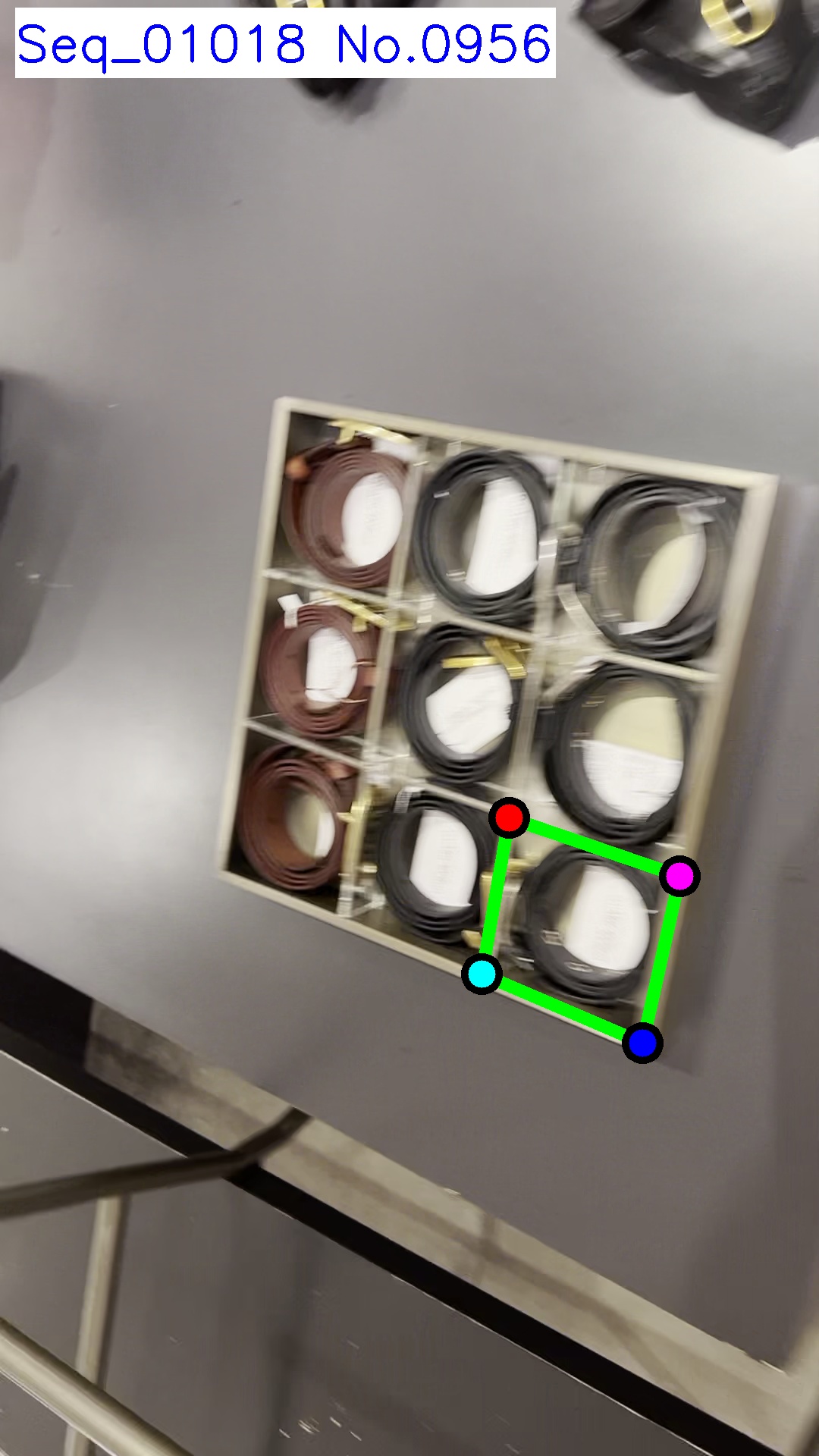}
\includegraphics[width=0.092\linewidth]{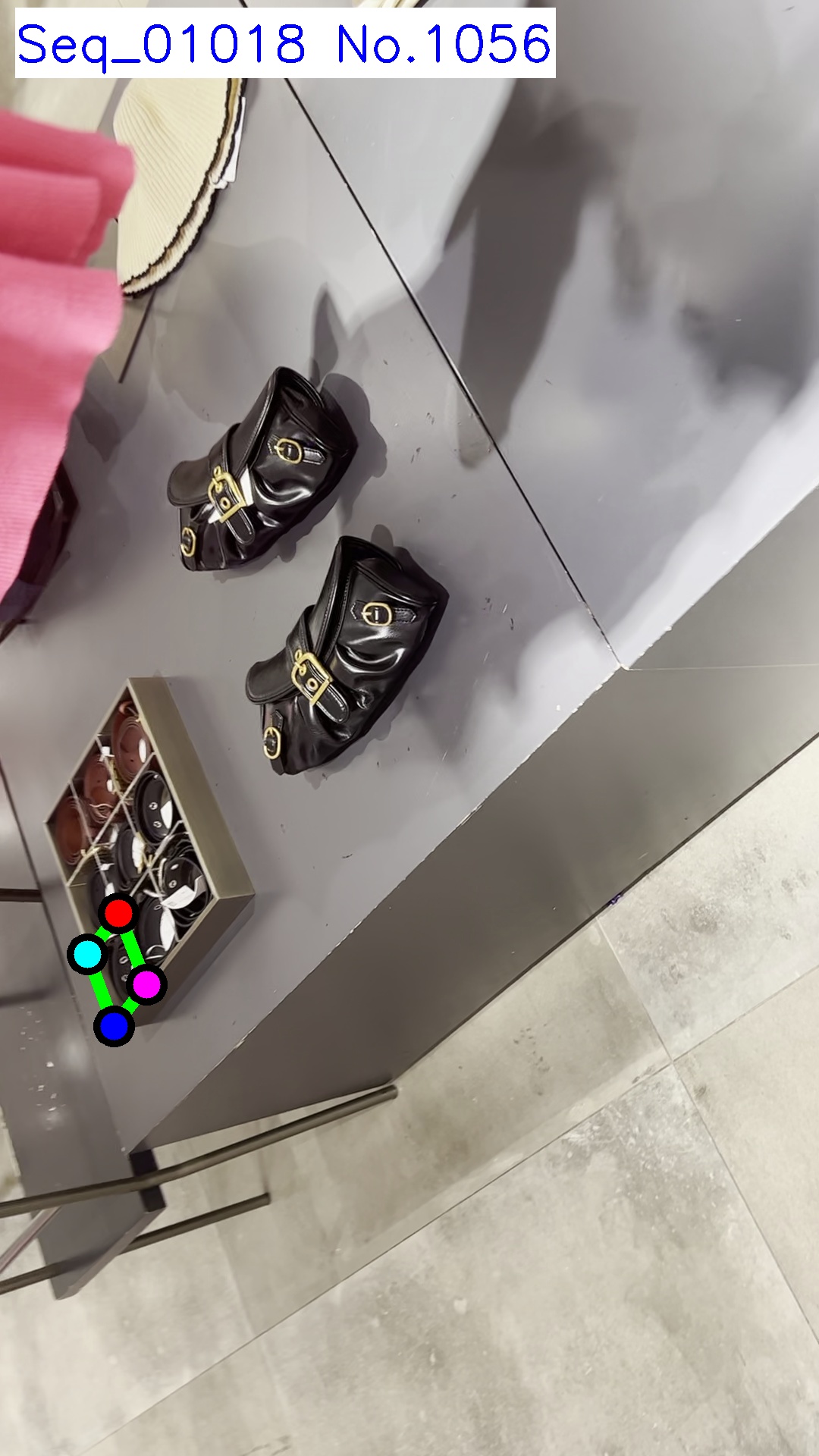}
\includegraphics[width=0.092\linewidth]{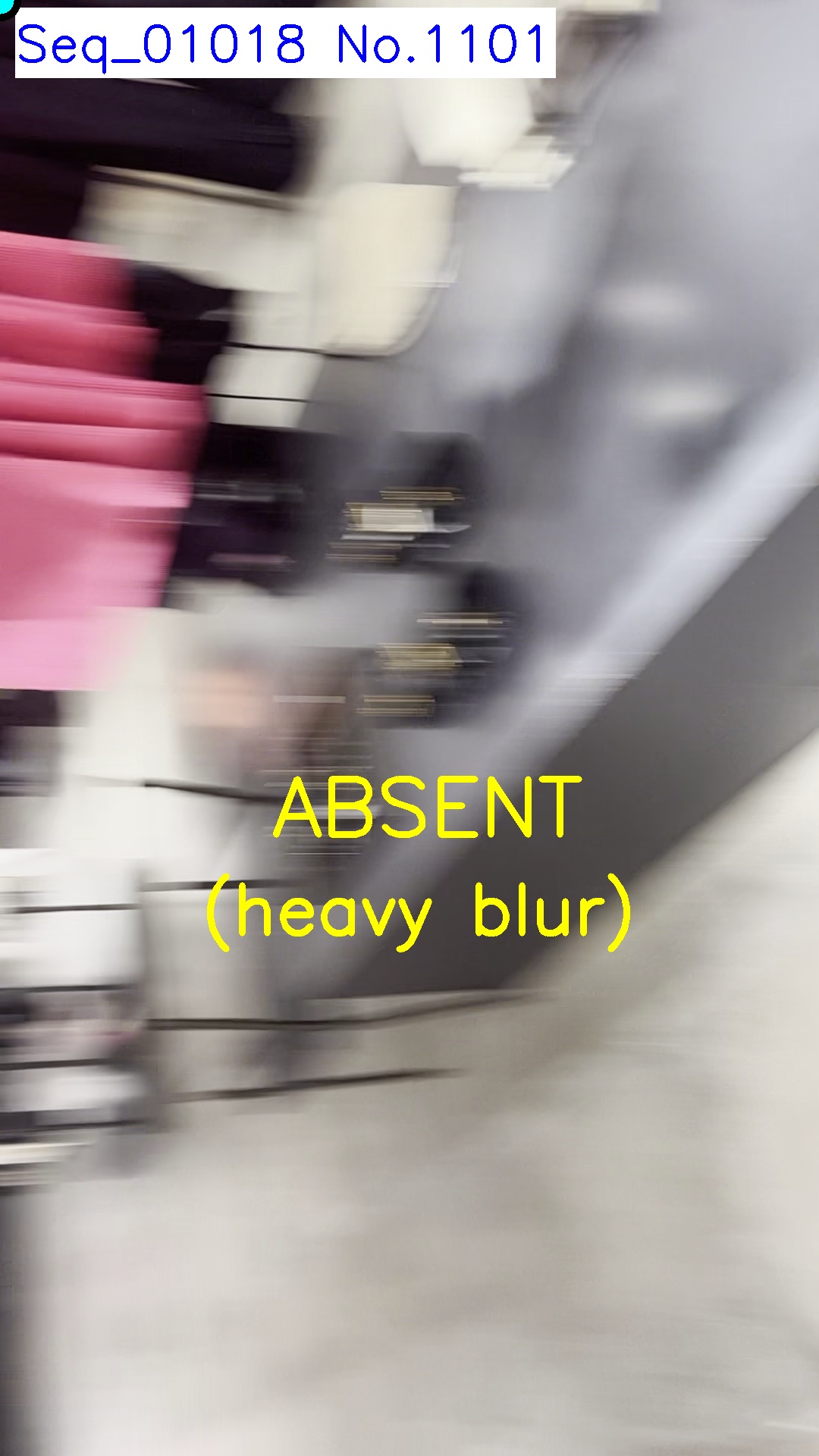}
\includegraphics[width=0.092\linewidth]{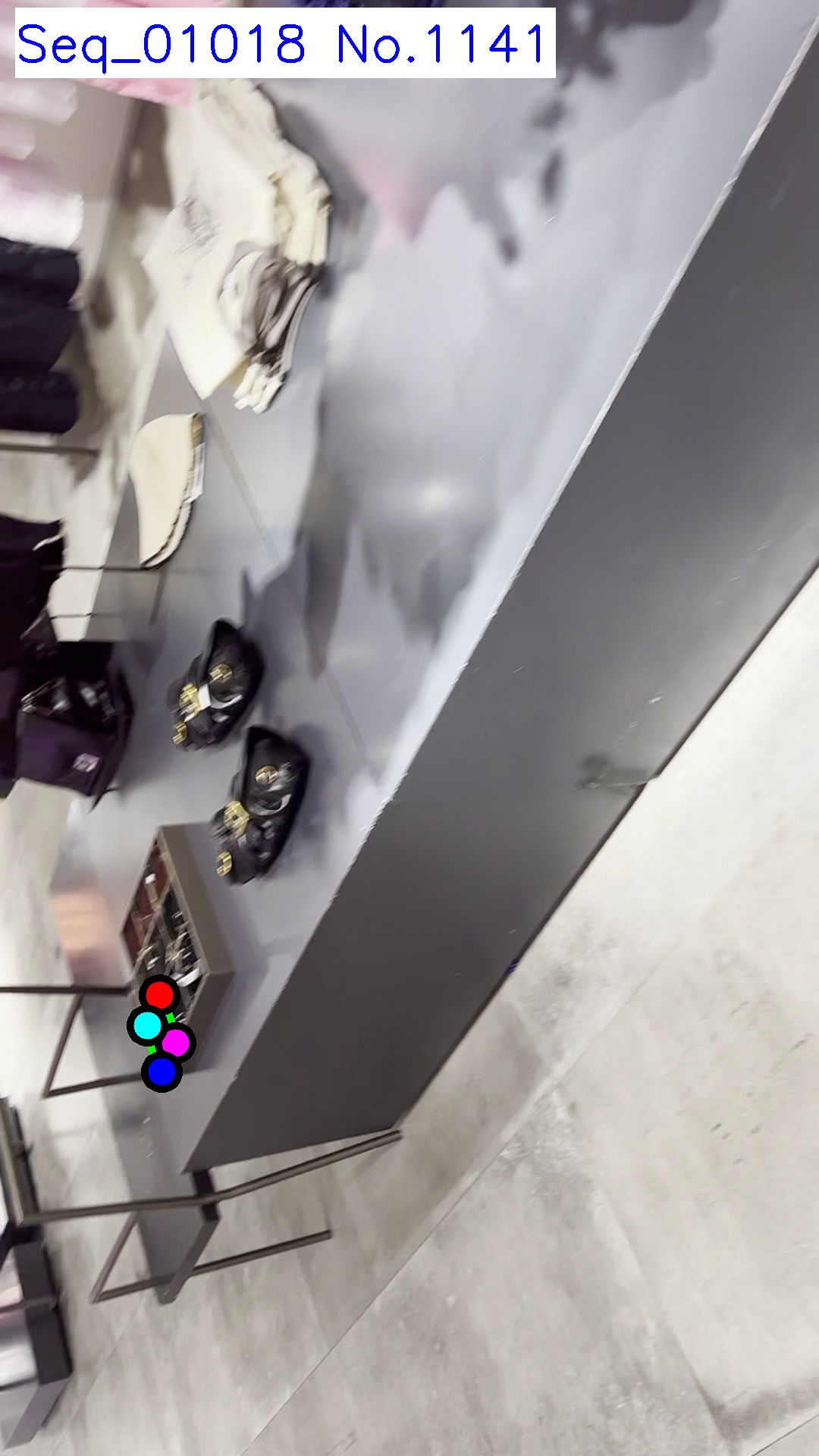}}\\
{\small (e) Sequence different resolution} \\
\end{tabular}
\caption{Examples of annotated sequences in the proposed PlanarTrack. Each video is annotated with four corner points.
}
\label{anno_exa}
\end{figure*}

PlanarTrack is annotated by several well-trained annotators and experts. We manually label each frame to provide a high-quality dense annotation. We employed a customized annotation tool developed in MATLAB, which allowed annotators to mark the four corner points with zoom-in support under challenging conditions. Before annotation, annotators were trained with clear guidelines covering common cases, missing corners, and heavy occlusion or blur. Specifically, we annotate four corner points for the planar target of each frame in the given order if all its four corner points or four edges are clearly visible. When the four corner points and four edges are both hard to recognize due to the occlusion, out-of-view or heavy blur, we will assign an absent flag to this frame.

With the above strategy, we carry out the annotation by the following workflow. Firstly, each sequence is annotated by an annotator. The annotation result is then distributed to two experts for double verification. If the annotation is not unanimously approved by the experts, it will be returned to the original annotator for careful refinement. Such a verification-refinement process will last for multiple rounds until the annotation finally receives unanimous approval in order to ensure the high annotation quality. Fig. \ref{anno_exa} shows some annotation examples of PlanarTrack.

\begin{figure*}[!t]
    \centering
    \includegraphics[width=0.8\linewidth]{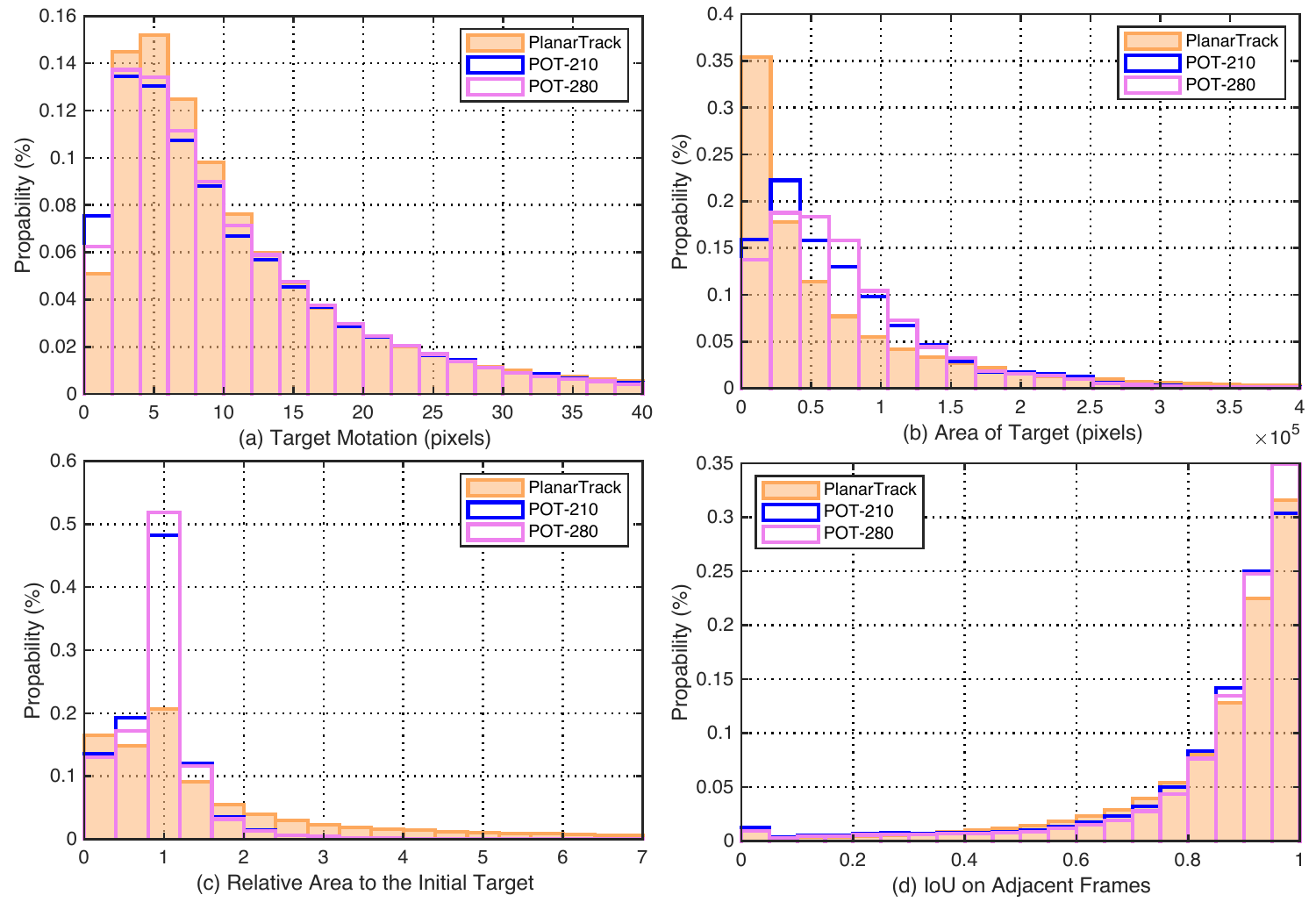}
    \caption{Statistics of planar target motion, size, relative area compared to initial object and IoU of targets in adjacent frames in PlanarTrack and comparison with the recent POT-210/280~\citep{liang2018planar,liang2021planar}. We can see the targets in our dataset have smaller sizes and faster and more challenging motions.}
    \label{stat_comp}
\end{figure*}

In order to better understand our PlanarTrack, we show four representative statistics of the annotations in Fig. \ref{stat_comp}, compared with POT-210/280. Specifically, we present the distributions of target motion, target size (area of target), target scaling (relative area to the initial target) and Intersection over Union (IoU) between targets in adjacent frames. From Fig. \ref{stat_comp}, we find that the planar targets in PlanarTrack have rapid size changes and speed of movement. Compared to POT-210/280 \citep{liang2018planar, liang2021planar}, PlanarTrack has relatively smaller target sizes and faster motions, while most target of POT-210/280 scale around 1 relative to the initial target and only moves a few pixels. Therefore, our PlanarTrack provides new challenges for planar tracking in the wild.

Notice that, since POT-210/280 labels every two frames, we perform linear interpolation on their annotations for statistics comparison.

\subsection{Analysis of Ground Truth Quality}\label{sec:gt_analysis}
Since the ground truth (GT) for each frame in our PlanarTrack dataset is manually annotated, some errors are inevitably introduced. To select appropriate evaluation metric thresholds and prevent researchers from overfitting to GT errors, we conducted an analysis of the GT quality in PlanarTrack.

Specifically, following WOFT~\citep{vserych2023planar}, we randomly selected a small subset from PlanarTrack, consisting of 10,920 frames, which was meticulously annotated by two experts highly familiar with planar object tracking, obtaining a refined GT. Subsequently, we computed the root of the mean square distances between the GT and the refined GT (\ie, the alignment error). Given four GT points $\mathbf{x}_i \in \mathbf{X}$ and four refined GT points $\mathbf{x}^*_i \in \mathbf{X}^*$, the alignment error  $e_{\textrm{AL}}$ can be calculated as
\begin{equation}
    e_{\textrm{AL}}(\mathbf{X},\mathbf{X}^*)=\sqrt{\frac{1}{4} \sum_{i=1}^{4}\left(\mathbf{x}_{i}-\mathbf{x}_{i}^{*}\right)^{2}}.
\end{equation}

\begin{figure}[htb]
    \centering
    \includegraphics[width=0.95\linewidth]{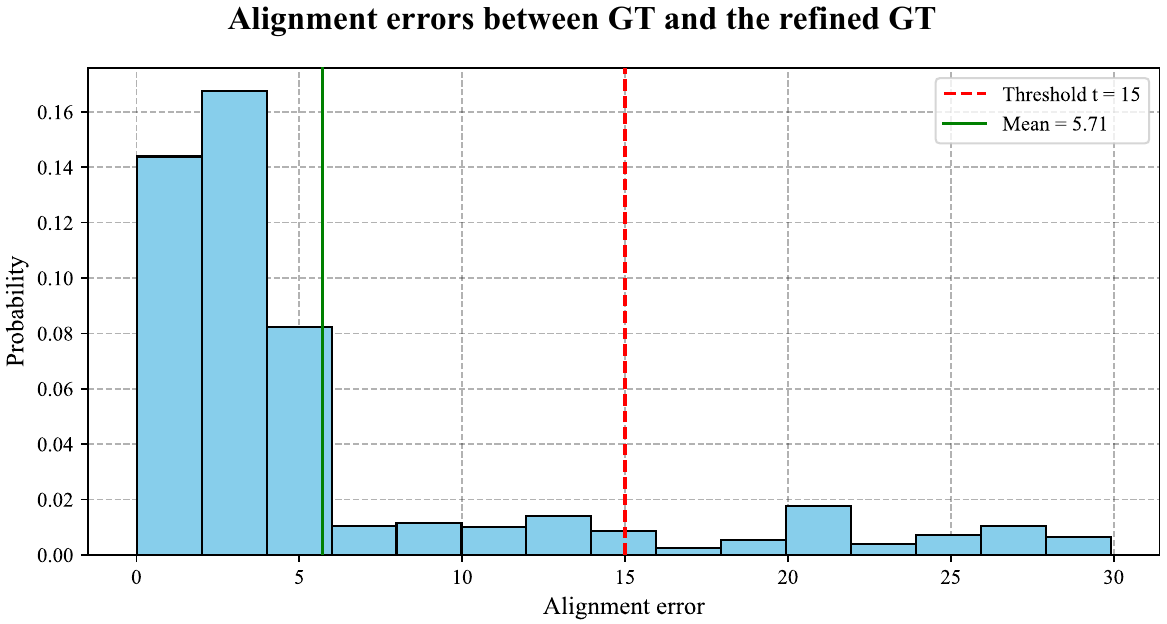}
    \caption{Distribution of alignment error between the GT and the refined GT of PlanarTrack.}
    \label{fig:error_dist}
\end{figure}

The results indicate that the mean alignment error between the GT and the refined GT on the refined-annotated subset of PlanarTrack is \textit{5.71} pixels. Fig.~\ref{fig:error_dist} illustrates the distribution of alignment errors, with \textit{10.71\%} of annotations exhibiting errors exceeding 15 pixels. Please note that, our PlanarTrack includes a greater number of challenging scenarios, such as \textit{heavier blur}, \textit{more extreme illumination changes}, and \textit{faster motion}. These factors make our precise annotation more difficult. Consequently, compared to the GT quality of POT-210 reported in WOFT~\citep{vserych2023planar}, our PlanarTrack exhibits slightly higher errors.

\subsection{Challenging Factors}\label{sec3_4}

Following other tracking benchmarks \citep{liang2018planar, fan2021lasot}, we label each sequence with several challenging factors in PlanarTrack to further analyze planar tracking algorithms in different challenging conditions. Specifically, we define eight challenging factors that widely exist for planar tracking. The challenging factors are listed below:

\vspace{0.3em}
\noindent \textbf{Occlusion (OCC)} Object is occluded by itself or other objects in the background. To increase the difficulty, we also manually occlude the object while moving the camera.

\vspace{0.3em}
\noindent \textbf{Motion Blur (MB)} Motion blur caused by fast camera movement at low frame rates can generate the fuzzy corner points, making it difficult to track a planar object robustly. 

\vspace{0.3em}
\noindent \textbf{Rotation (ROT)} Rotation describes a common situation that an object's direction is changed relative to the camera.

\vspace{0.3em}
\noindent \textbf{Scale Variation (SV)} Scale variation is assigned when the ratio of planar annotation is outside the range $[0.5, 2]$.

\vspace{0.3em}
\noindent \textbf{Perspective Distortion (PD)} Perspective distortion is assigned when the perspective between the object and camera is changed.

\vspace{0.3em}
\noindent \textbf{Out-of-view (OV)} Out-of-view is assigned when part or all of the object leaves the image, which makes some sides or corners of the target invisible.

\vspace{0.3em}
\noindent \textbf{Low Resolution (LR)} Low resolution is assigned when the region of the target in any frame of a sequence is less than 1,000 pixels.

\vspace{0.3em}
\noindent \textbf{Background Clutter (BC)} Background clutter is assigned when the background region looks visually similar to the target, including similar colors, multiple similar targets, etc.

\vspace{0.3em}
\noindent \textbf{Light Interactive Surface (LIS)} Light Interactive Surface is assigned when significant appearance changes of the planar object occur due to light phenomena such as reflection and refraction, \eg, \textit{mirrors} and \textit{transparent plates}. \textit{Screens} are also classified under this category, as the videos displayed on them can cause significant appearance changes.

It is worth mentioning that, some common challenging factors used in generic object tracking are not suitable for planar objects. Thus, we exclude a few of them, such as deformation and illumination change. The vast majority of sequences (1,135 out of 1,150) in PlanarTrack simultaneously contain multiple challenging factors (\ie, recorded in \emph{unconstrained conditions}). Therefore, our PlanarTrack is much more challenging and practical for real applications, compared to POT-210/280. 

\begin{figure}[!t]
    \centering
    \includegraphics[width=0.95\linewidth]{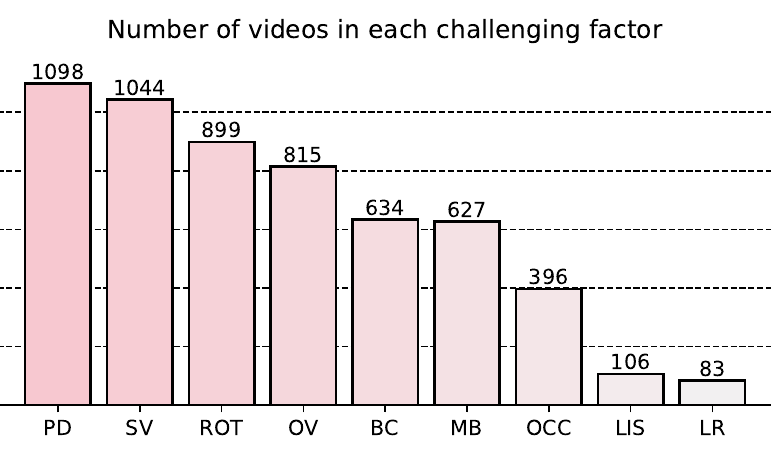}
    \caption{Distribution of sequences on each challenging factor.}
    \label{cf_num}
\end{figure}

The distribution of the above challenging factors on PlanarTrack is presented in Fig. \ref{cf_num}. We notice that perspective distortion is the most common challenging factor in PlanarTrack, which may lead to serious misalignment problems for planar tracking. In addition, scale variation and rotation frequently exist in PlanarTrack. 

\subsection{Dataset Split and Evaluation Metric}\label{sec3_5}

\vspace{0.3em}
\noindent \textbf{Training/Test Set Split} PlanarTrack contains 1,150 sequences. We use 805 sequences for training (PlanarTrack\textsubscript{Tra}) and 345 for evaluation (PlanarTrack\textsubscript{Tst}). We try our best to keep the distributions of training and test sets close to each other. As for the four ultra-long sequences, we put two of them into a training set and the other two into a test set for long-term tracking and evaluation. Table \ref{split_comp} shows a comparison of these two sets. 

\begin{figure}[!t]
    \centering
    \includegraphics[width=0.95\linewidth]{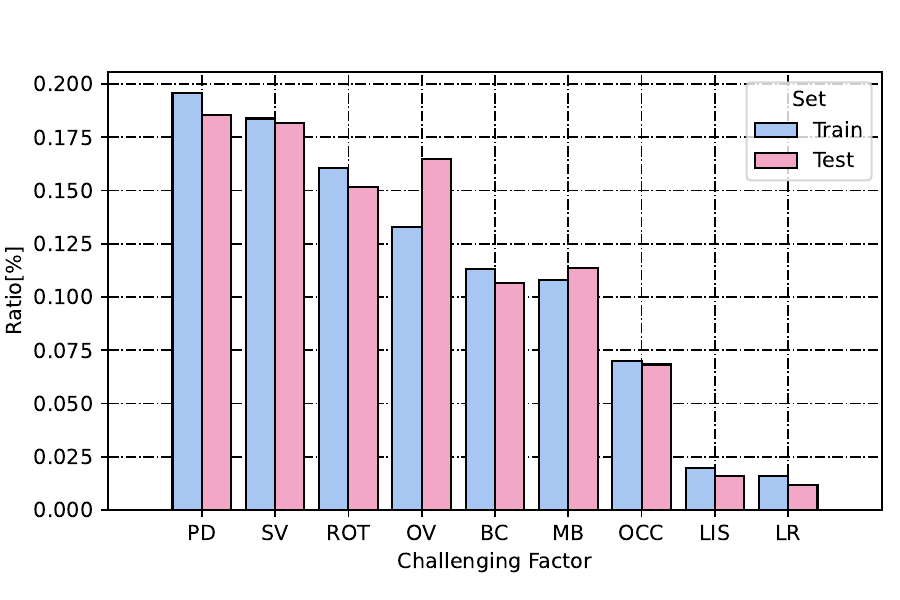}
    \caption{Distribution of challenging factor on training and testing sets.}
    \label{cf_num_split}
\end{figure}

For further comparison between training and test sets of PlanarTrack, we present the ratios of sequences in these two sets on eight different challenging factors in Fig. \ref{cf_num_split}. From Fig \ref{cf_num_split} we can see that, our split makes the training and test sets closing to each other, which ensures the consistency of training/test split in PlanarTrack. Notice that, the number of test sequences is significantly higher than training sequences on OV factor. This is because frequent disappearance may lead to a decrease of training data but make it more challenging for evaluation. Detailed split files will be released on our project website.

\vspace{0.3em}
\noindent \textbf{Evaluation Metric} For the evaluation, we adopt the \emph{precision} (PRE) metric following \citep{liang2021planar}. Please note here, we do \textit{not} utilize the SUC metric as in previous studies for evaluation, because the SUC, that represents the percentage of successful frames in which the error between estimated and real homography is less than or equal to a certain threshold, depends heavily on the position of the target in the image. When the target is located in the bottom-right corner of the image, a very small tracker imprecision can lead to a huge re-projection error. This makes the SUC metric \textit{cannot} access the true accuracy of tracking results.

However, there are some differences between our PRE and that used for generic tracking \citep{wu2013online}. For planar tracking, PRE is defined as the percentage of frames in which the alignment error between corner points of predicted result and groundtruth is within a given threshold. Based on the quality analysis of GT in Sec.~\ref{sec:gt_analysis}, we selected 15 pixels as the primary threshold for the PRE metric. Additionally, since 75.98\% of cases exhibit errors below 5 pixels, we retained the 5px threshold for the PRE metric as used in POT-210. In summary, we adopted 5px and 15px thresholds for the PRE metrics to enable a more comprehensive evaluation, denoted as P@5 and P@15, respectively.

\begin{table}[!tb]
  \centering
  \caption{Comparison of \emph{training} and \emph{test} sets.}
  \setlength{\extrarowheight}{2pt}
  \setlength{\tabcolsep}{1mm}{
  \begin{tabular}{lccccc}
    \specialrule{.1em}{.05em}{.05em} 
    & Videos & \tabincell{c}{Min \\ frames} & \tabincell{c}{Mean \\frames} & \tabincell{c}{Max \\frames} & \tabincell{c}{Total\\ frames} \\
    \hline
    PlanarTrack$_\mathrm{Tra}$ & 805    & 317    & 636    & 3352    & 512K \\
    PlanarTrack$_\mathrm{Tst}$ & 345    & 362    & 641    & 3150    & 221K \\
    \specialrule{.1em}{.05em}{.05em} 
    \end{tabular}}%
  \label{split_comp}%
\end{table}%

\section{Evaluation}\label{sec4}
\subsection{Evaluated Planar Object Tracking Algorithms}\label{sec4_1}

We do several evaluations of planar object trackers on PlanarTrack to demonstrate its reliability and novelty. As there are not many planar object trackers compared to generic tracking (actually, this is the biggest motivation for us to introduce PlanarTrack for promoting research on planar object tracking), we select 10 representative algorithms about planar tracking with accessible source codes. Specifically, these trackers are WOFT \citep{vserych2023planar}, HDN \citep{zhan2022homography}, GIFT \citep{liu2019gift}, LISRD \citep{pautrat2020online}, SIFT \citep{lowe2004distinctive}, Gracker \citep{wang2017gracker}, SOL \citep{hare2012efficient}, SCV \citep{richa2011visual}, ESM \citep{benhimane2004real} and IC \citep{baker2004lucas}. Particularly, WOFT \citep{vserych2023planar} and HDN \citep{zhan2022homography} are two recent planar trackers using deep learning. All other algorithms can be used for homography estimation. We modify them to the corresponding planar object trackers. It's worth mentioning that, we are not able to evaluate generic trackers on PlanarTrack because of the incompatible inputs and results. For this, we construct a new PlanarTrack\textsubscript{BB} for generic tracking evaluation, as described later.

\begin{table}[!t]\footnotesize
\centering
\caption{Summary of evaluated planar trackers. Representation: ``Deep'' for deep-learning-based Method, ``Keypoint'' for Keypoint-based Method, and ``Direct'' for Direct Method.}
\label{tracker_summary}
\setlength{\extrarowheight}{2pt}
\setlength{\tabcolsep}{0.2mm}{
\begin{tabular}{lcccc}
\specialrule{.1em}{.05em}{.05em} 
\multirow{2}{*}{Method} & \multirow{2}{*}{Backbone} & \multicolumn{3}{c}{Representation}  \\ \cmidrule{3-5}
                        &                           & Deep            & Keypoint        & Direct                                   \\ \hline
WOFT~\citep{vserych2023planar}                    & RAFT                      & \cmark &                       &                                    \\
HDN~\citep{zhan2022homography}                     & ResNet-50                 & \cmark &                       &                                    \\
GIFT~\citep{liu2019gift}                    & CNN                       & \cmark &                       &                                 \\
LISRD~\citep{pautrat2020online}                   & VGG16                     & \cmark &                       &                                    \\
SIFT~\citep{lowe2004distinctive}                    & -                         &                       & \cmark &                                    \\
Gracker~\citep{wang2017gracker}                 & -                         &                       & \cmark &                                   \\
SOL~\citep{hare2012efficient}                     & -                         &                       & \cmark &                                    \\
SCV~\citep{richa2011visual}                     & -                         &                       &             & \cmark                         \\
ESM~\citep{benhimane2004real}                     & -                         &                       &           & \cmark                          \\
IC~\citep{baker2004lucas}                      & -                         &                       &              & \cmark                          \\
\specialrule{.1em}{.05em}{.05em} 
\end{tabular}}
\end{table}

\subsection{Evaluation Results}\label{sec4_2}
\subsubsection{Overall Performance}\label{sec4_2_1} Totally, we evaluate 10 representative planar object trackers on PlanarTrack\textsubscript{Tst}, among which WOFT and HDN are utilized without modifications as they are specifically developed for the planar tracking task. For the remaining methods, we modify them so that they can be used for planar object tracking. Their implementations except GIFT and LISRD are borrowed from \citep{liang2018planar}. We adapt GIFT and LISRD to planar object tracking due to some setting problems in \citep{liang2018planar}. Fig. \ref{overall_performance} shows the evaluation results of the above approaches in P@5 and P@15. From Fig. \ref{overall_performance} we can see that, WOFT achieves the best P@5 score of 0.402 and P@15 score of 0.607. GIFT applies transformation-invariant deep visual descriptors for planar object tracking, which demonstrates the second best P@5 score of 0.221 and P@15 score of 0.402. Notice that, all the top four approaches leverage deep neural networks for planar target localization, which shows the great potential of deep-learning-based planar tracking in the future. 

\vspace{0.3em}
\noindent \textbf{Short-term Tracking analysis} Our PlanarTrack consists of 1000 sequences with an average length of 490 frames, which is suitable for short-term tracking. To evaluate the performance of deep-learning-based planar trackers, we perform regular experiments on PlanarTrack\textsubscript{Tst-300}, the test set for short-term tracking. Evaluation results are shown in Table \ref{300_345_comp}. WOFT achieves the highest P@15 score of 0.641, which is obviously better than HDN.

\begin{figure}[htb]
    \centering
    \includegraphics[width=0.95\linewidth]{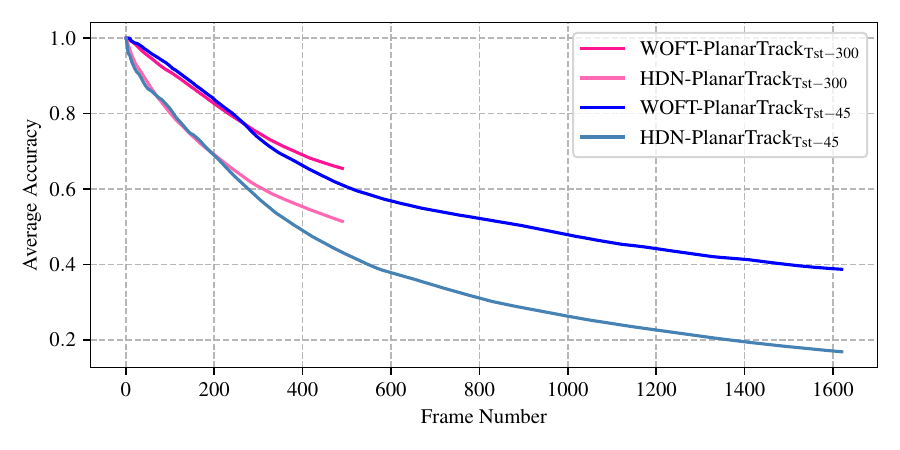}
    \caption{Accuracy changes of two planar trackers WOFT and HDN with respect to frame number.}
    \label{fig:longterm-analysis}
\end{figure}

\vspace{0.3em}
\noindent \textbf{Long-term Tracking analysis} To analyze the performance of the top four methods in long-term planar object tracking, we demonstrate the tracking results on PlanarTrack\textsubscript{Tst-300}, PlanarTrack\textsubscript{Tst-45} and PlanarTrack\textsubscript{Tst-345} in Table \ref{300_345_comp}. Notice that, PlanarTrack\textsubscript{Tst-45} is the test set consisting entirely of long sequences, while PlanarTrack\textsubscript{Tst-345} is the test set of the whole PlanarTrack. From Table \ref{300_345_comp}, we can observe that both WOFT and HDN show performance degradation while HDN has the most significant decline in the long-term tracking scenario. Additionally, we plot the accuracy of these tow planar trackers as a function of frame number, as shown in Fig.~\ref{fig:longterm-analysis}. From Fig.~\ref{fig:longterm-analysis}, it can be observed that, the accuracy trends for the same tracker in the short-term intervals of PlanarTrack\textsubscript{Tst-300} and PlanarTrack\textsubscript{Tst-45} are relatively similar. However, during long-term tracking on PlanarTrack\textsubscript{Tst-45}, the accuracy consistently declines, suggesting that current trackers struggle to maintain target capture over extended periods. Several factors may contribute to this issue. For example, frequent disappearances and reappearances of the target over time can cause significant spatial shifts relative to the last successfully tracked frame, which is particularly detrimental to trackers relying on displacement prediction. Additionally, repeated appearance changes of the target over a long duration may exceed the trackers’ ability to manage long-term associations.

This highlights the need for a dedicated platform dedicated for long-term planar object tracking, which could drive the development of advanced long-term tracking algorithms.

\begin{table}[!ht]\footnotesize
\centering
    \caption{Comparison and analysis of two planar trackers in short-term tracking and long-term tracking. }
    \label{300_345_comp}
    \setlength{\extrarowheight}{2pt}
    \setlength{\tabcolsep}{3.8mm}{
        \begin{tabular}{lccc}
        \specialrule{.1em}{.05em}{.05em} 
        \multicolumn{2}{l}{}                  & WOFT  & HDN   \\ \hline
        \multirow{2}{*}{PlanarTrack\textsubscript{Tst-300}} & P@5 & 0.433 & 0.263 \\
                                        & P@15 & 0.641 & 0.499 \\
        \multirow{2}{*}{PlanarTrack\textsubscript{Tst-45}}  & P@5 & 0.253 & 0.085 \\
                                        & P@15 & 0.379 & 0.164 \\
        \multirow{2}{*}{PlanarTrack\textsubscript{Tst-345}} & P@5 & 0.402 & 0.211 \\
                                        & P@15 & 0.607 & 0.455 \\
        \specialrule{.1em}{.05em}{.05em} 
        \end{tabular}}
\end{table}

\begin{figure}[!t]
    \centering
    \includegraphics[width=0.95\linewidth]{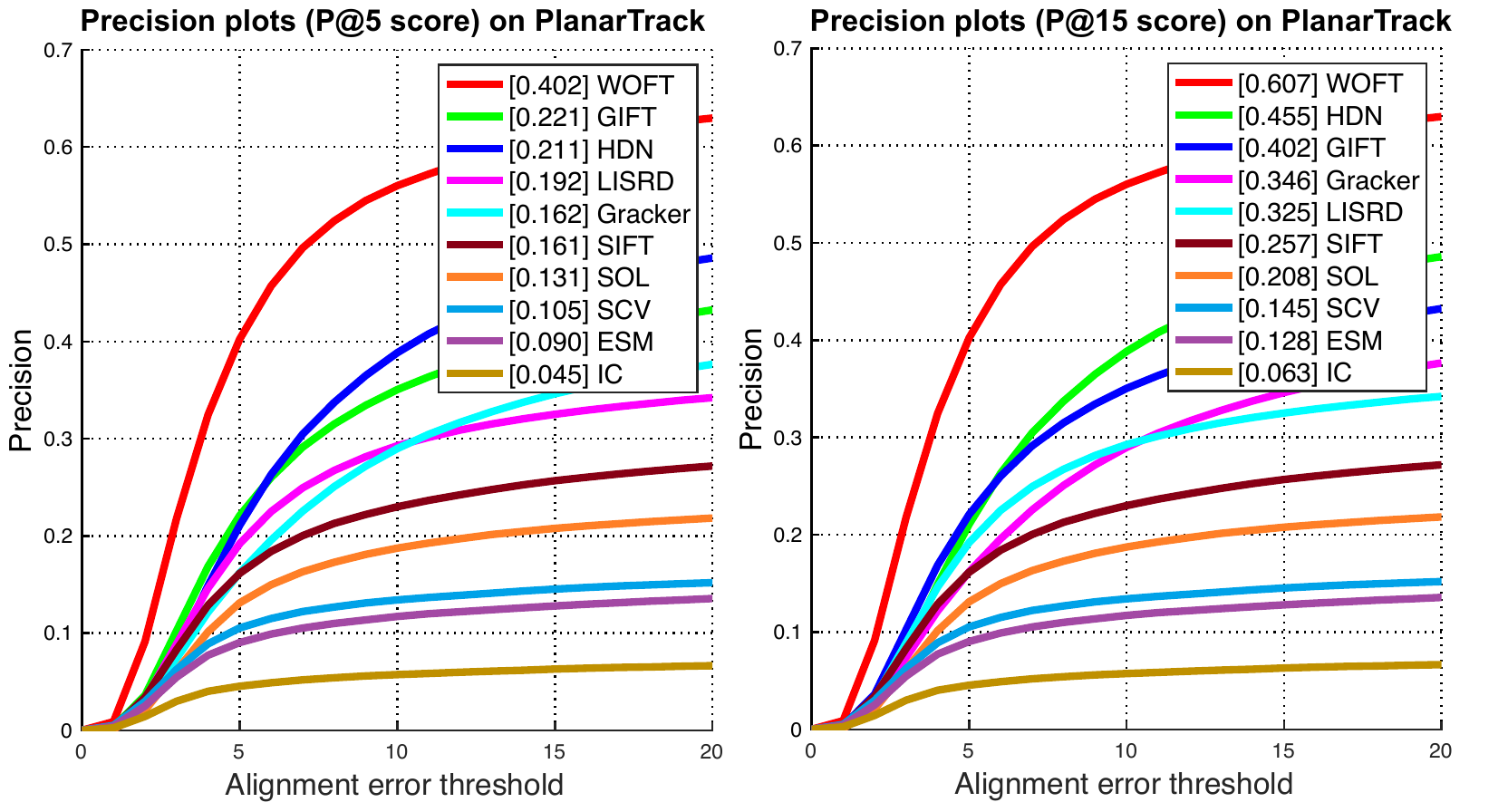}
    \caption{Precision plots of all planar trackers on PlanarTrack\textsubscript{Tst} using P@5 score and P@15 score, respectively.}
    \label{overall_performance}
\end{figure}

\subsubsection{Challenging Factor-based Evaluation}\label{sec4_2_2}

\begin{figure}[!t]
    \centering
    \includegraphics[width=0.95\linewidth]{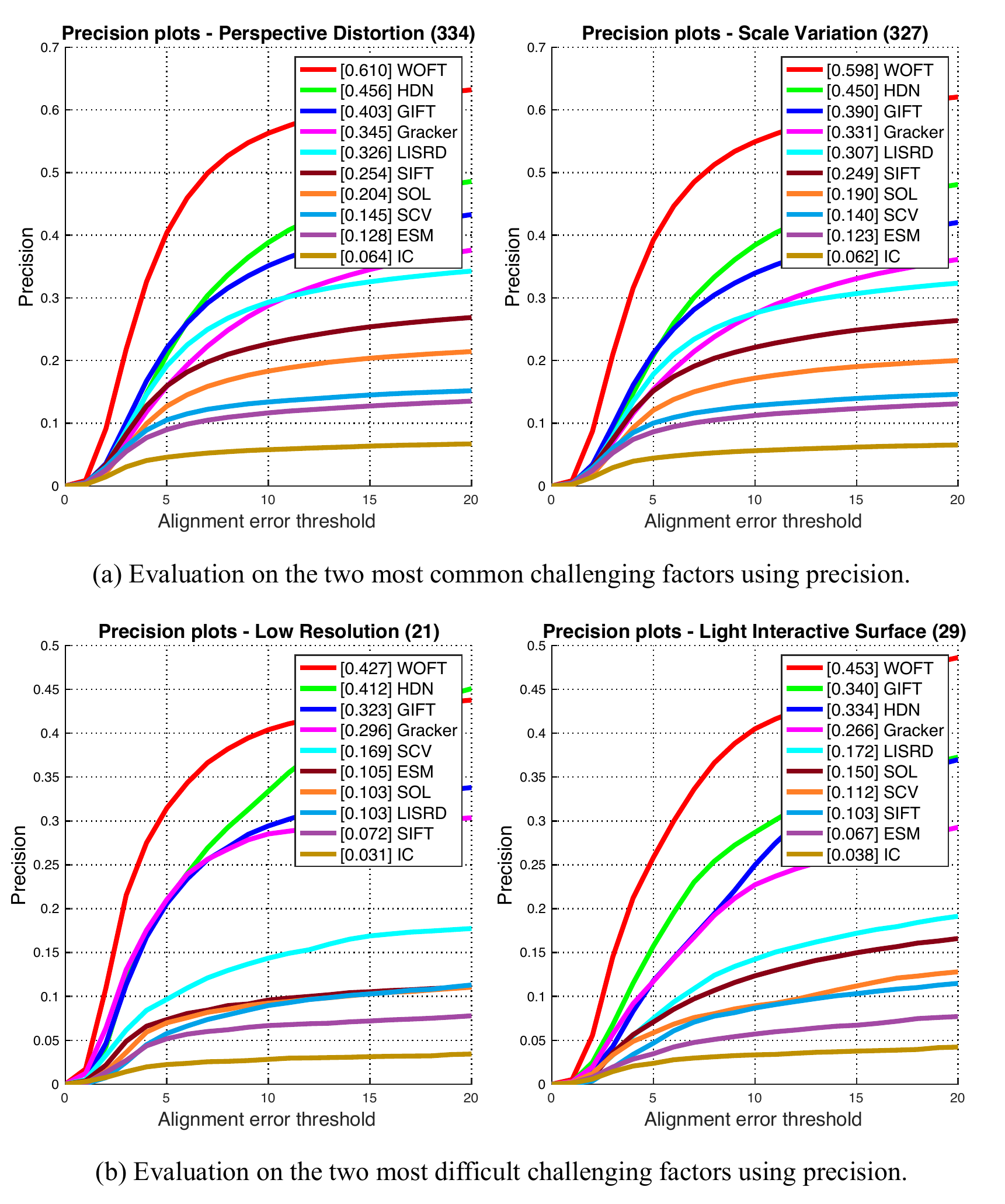}
    \caption{Precision plots of trackers on the two most common challenging factors including \emph{perspective distortion} and \emph{scale variation} and on the two most difficult challenging factors including \emph{low resolution} and \emph{light interactive surface} using P@15. }
    \label{com_diff_performance}
\end{figure}

For better analysis of different planar trackers, we further evaluate the above trackers on the eight challenging factors. Fig. \ref{com_diff_performance} displays the tracking results on the two most common challenging factors (\emph{perspective distortion} (PD) and \emph{scale variation} (SV)) and on the two most difficult challenging factors (\emph{low resolution} (LR) and \emph{light interactive surface} (LIS)). From Fig. \ref{com_diff_performance} we can see that, WOFT achieves the best performance on both the commonest and most difficult scenarios. Specifically, WOFT achieves the best P@15 scores of 0.610, 0.598, 0.427 and 0.453 on PD, SV, LR and LIS, which again shows the importance of temporal information for planar tracking. Besides, the tracking performances severely decrease on LR and LIV. A reasonable explanation is that these two challenges may be harmful to the feature extraction of points or targets, leading to tracking drifts or failures. From our perspective, research should be devoted to improvements in these two situations. 

\begin{figure*}[!t]
    \centering
    \includegraphics[width=0.95\linewidth]{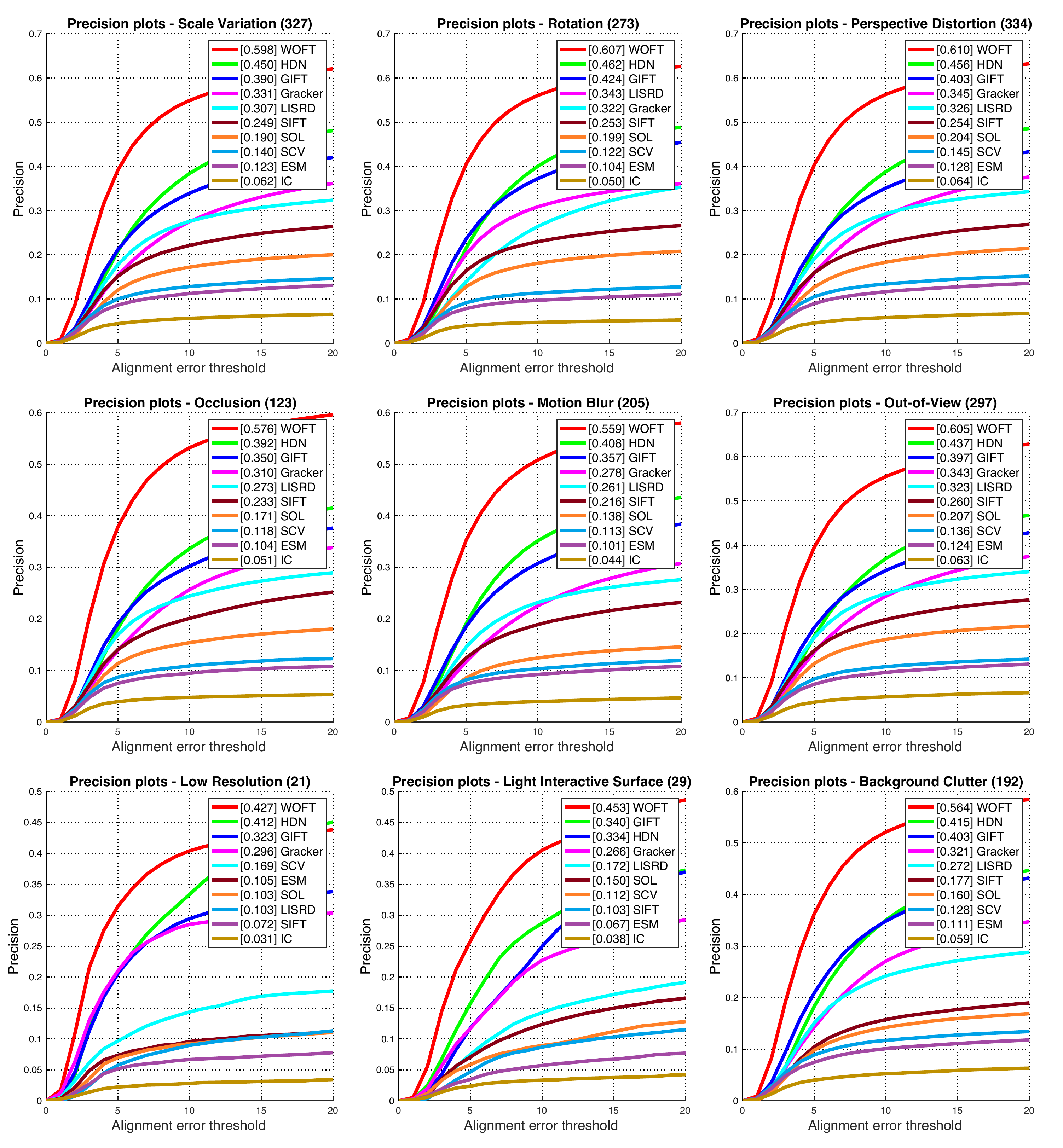}
    \caption{Precision plots of trackers on each challenging factor using P@15 score. Best viewed in color}
    \label{challenging_performance}
\end{figure*}

Fig. \ref{challenging_performance} shows the whole results on all 9 challenging factors with P@15 score. From Fig. \ref{challenging_performance} we observe that WOFT achieves the best performance on all 9 challenging factors with P@15 scores. HDN obtains the second best results on 8 out of 9 factors with P@15 score. Among the four deep-learning-based tracking methods, WOFT is far ahead of the rest three approaches due to the introduction of temporal information. An interesting observation is that LISRD performs extremely poorly on LR. A potential reason is that the small target information is buried in background when extract features by its CNN-based backbone.

\subsubsection{Qualitative Evaluation}\label{sec4_2_3}

\begin{figure*}[!t]
		\centering
		\begin{tabular}{c@{\hspace{1.8mm}}c}
			%\begin{tabular}{@{\hspace{.0mm}}c@{\hspace{1.75mm}} @{\hspace{.0mm}}c@{\hspace{.0mm}} @{\hspace{.0mm}}c@{\hspace{.0mm}}}
   
        \includegraphics[width=0.19\linewidth]{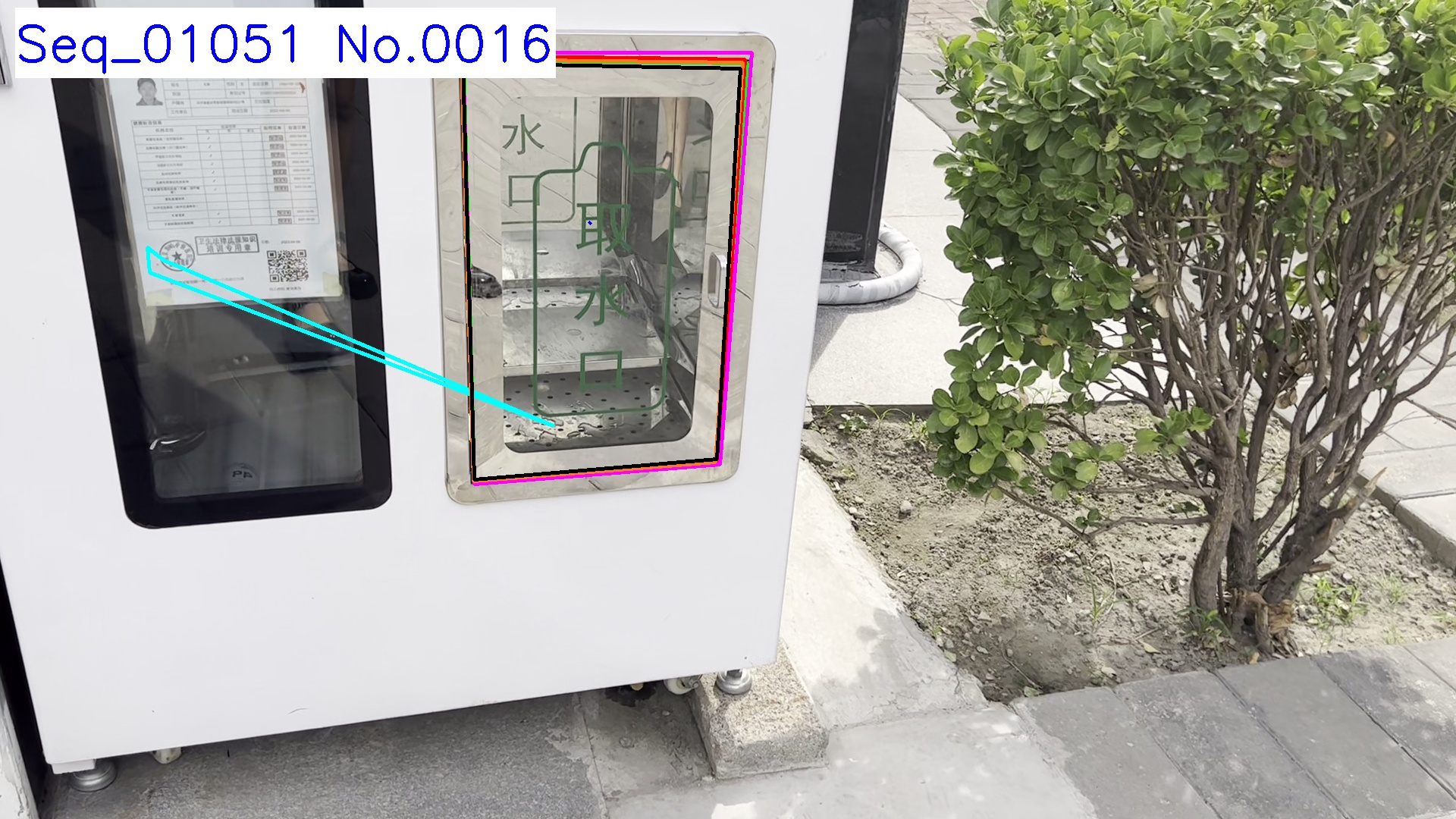} \includegraphics[width=0.19\linewidth]{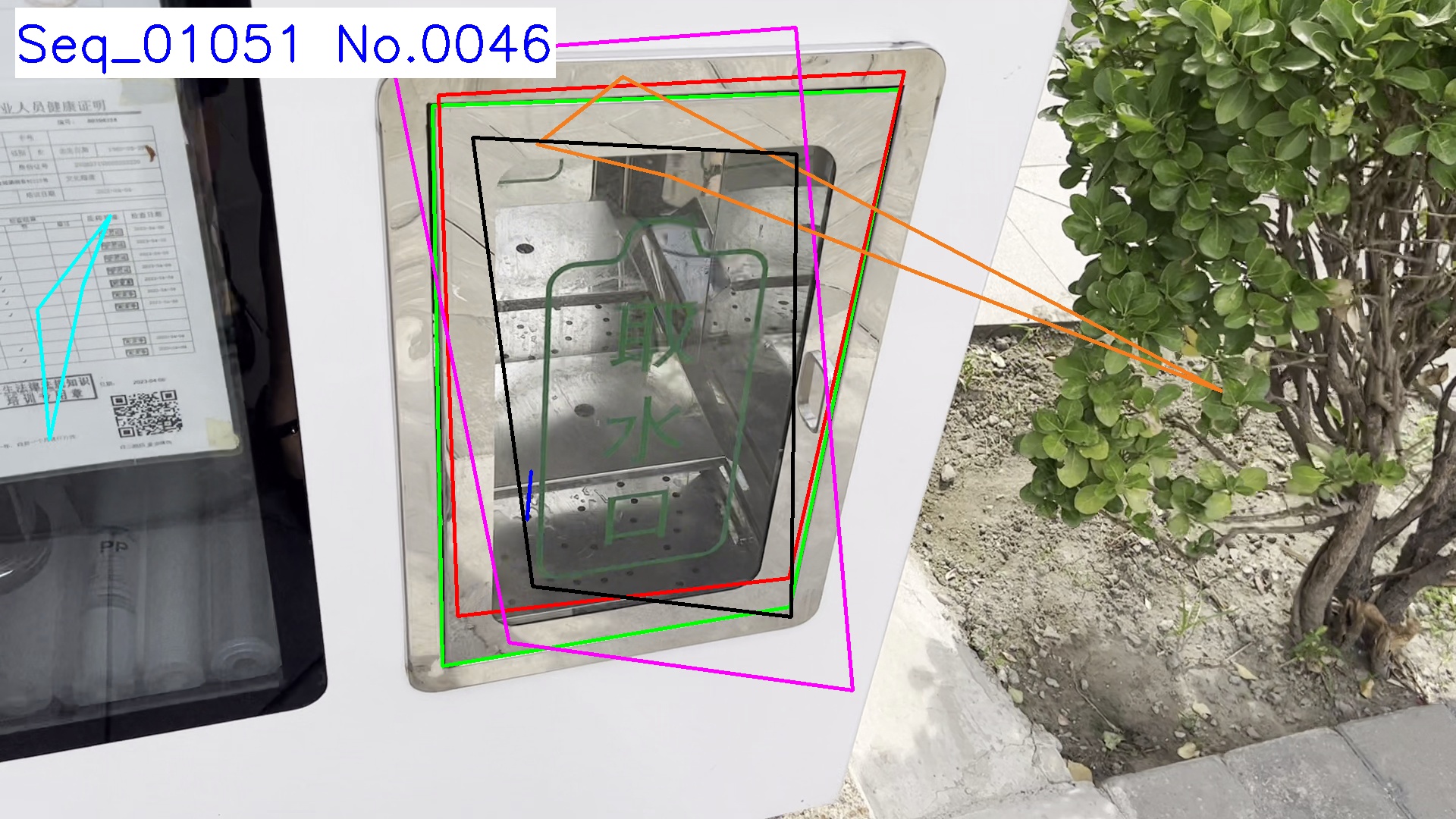} \includegraphics[width=0.19\linewidth]{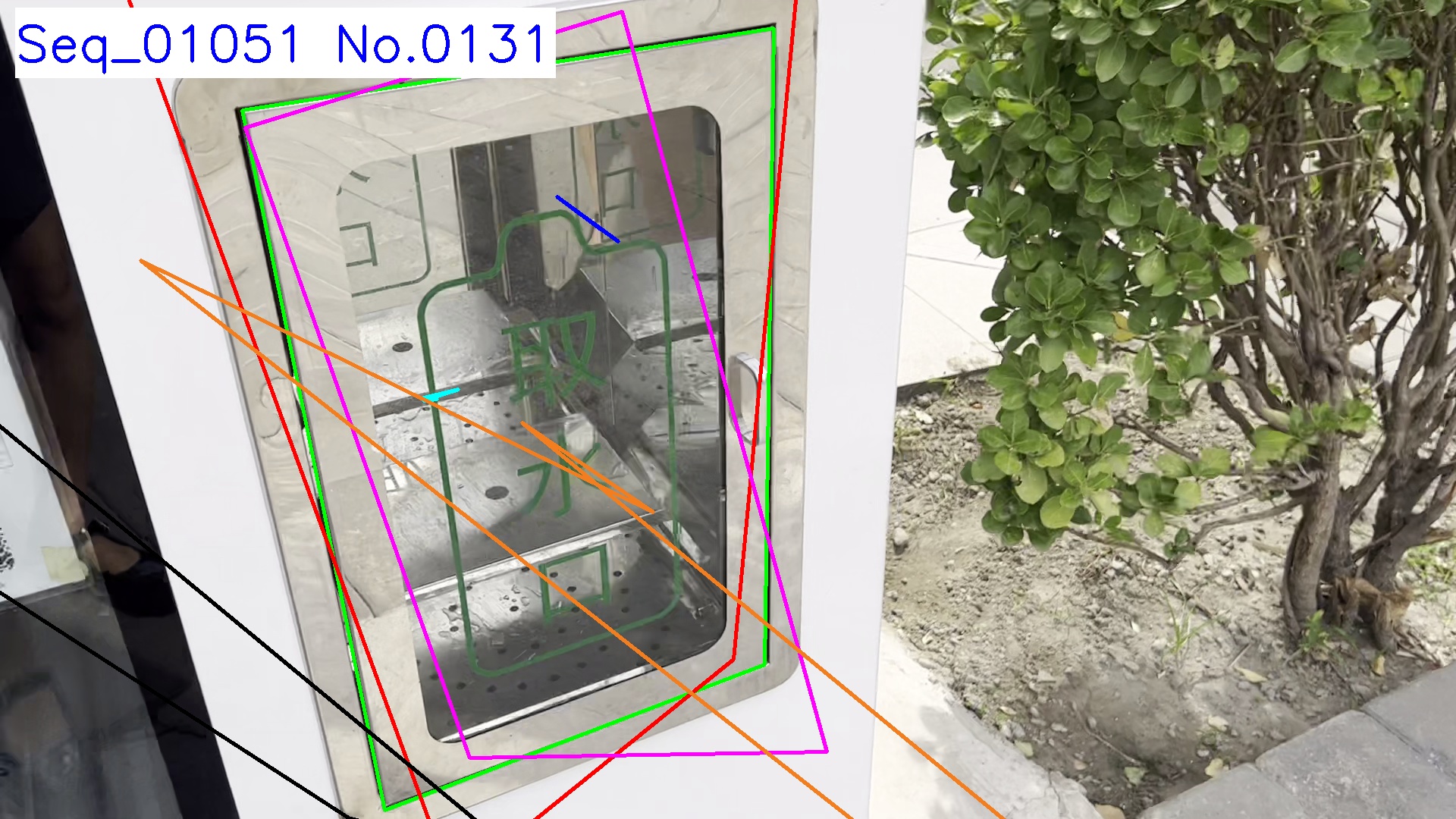} \includegraphics[width=0.19\linewidth]{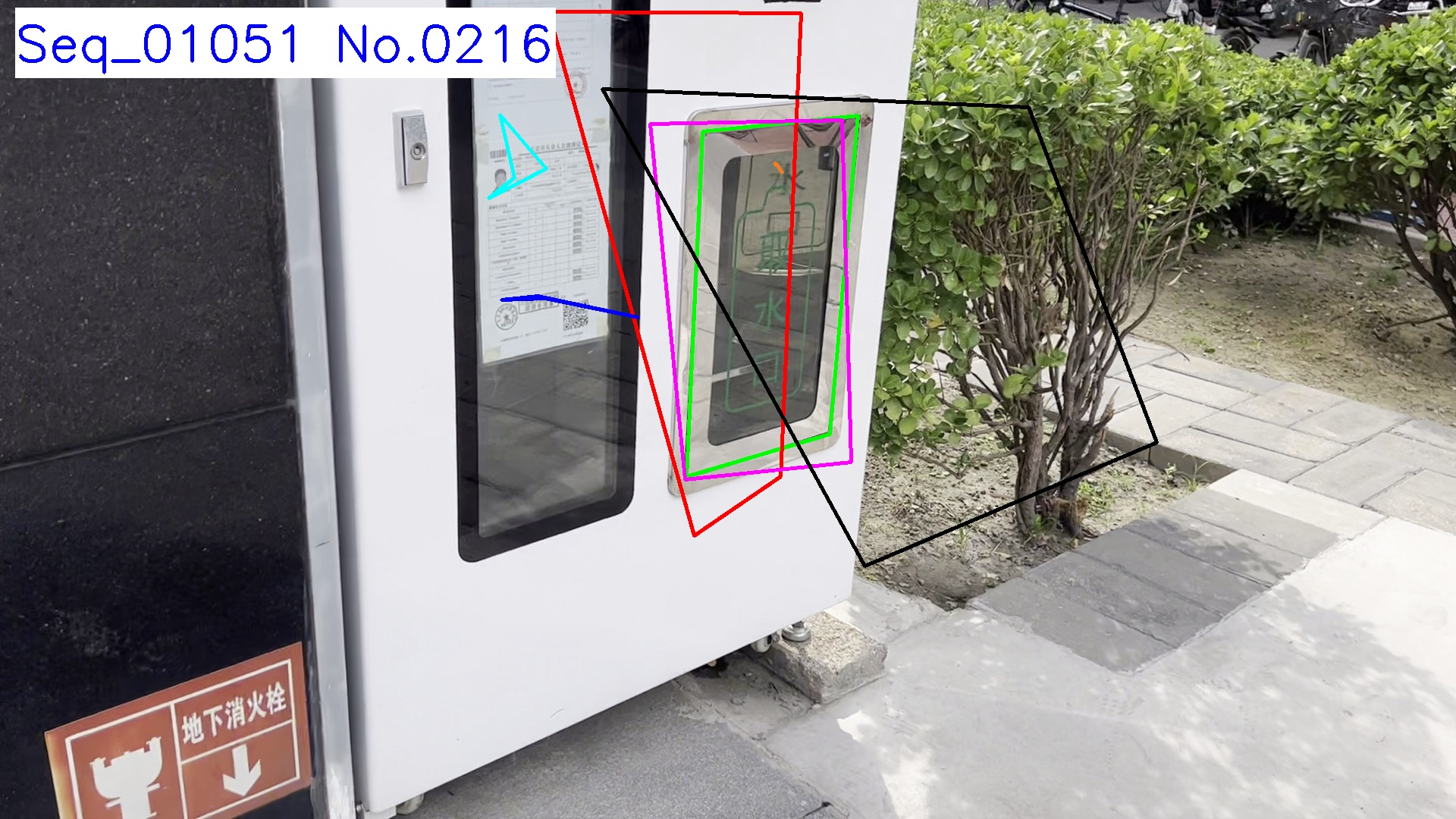} \includegraphics[width=0.19\linewidth]{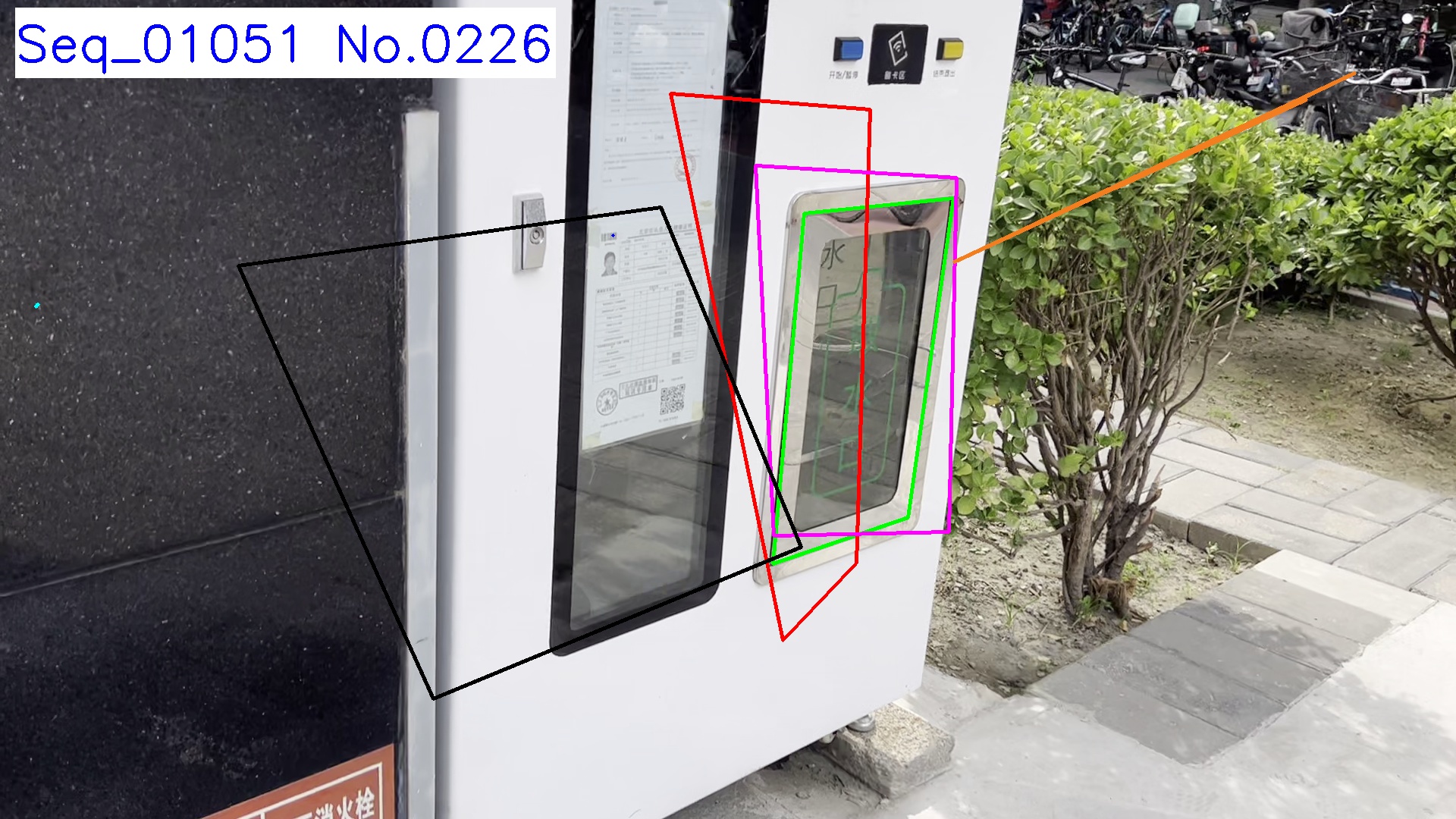} \\
        {\small (a) Sequence with BC and SV.} \\
        
        \includegraphics[width=0.19\linewidth]{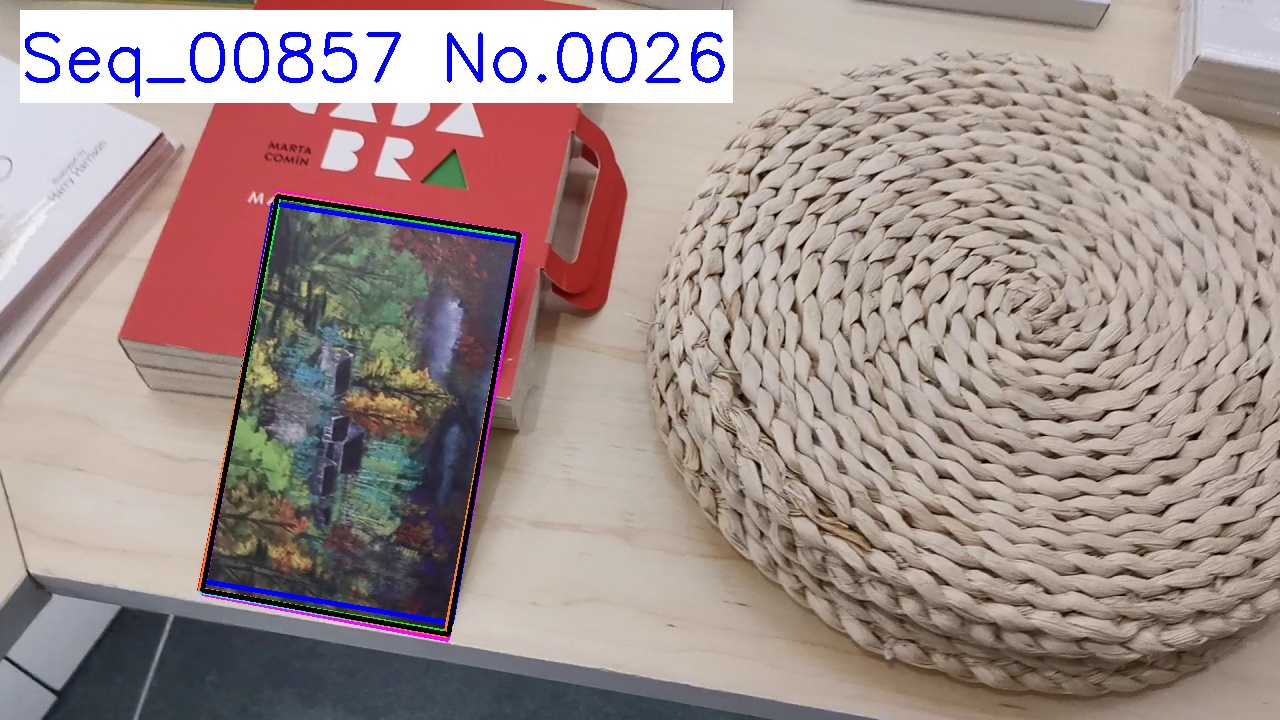} \includegraphics[width=0.19\linewidth]{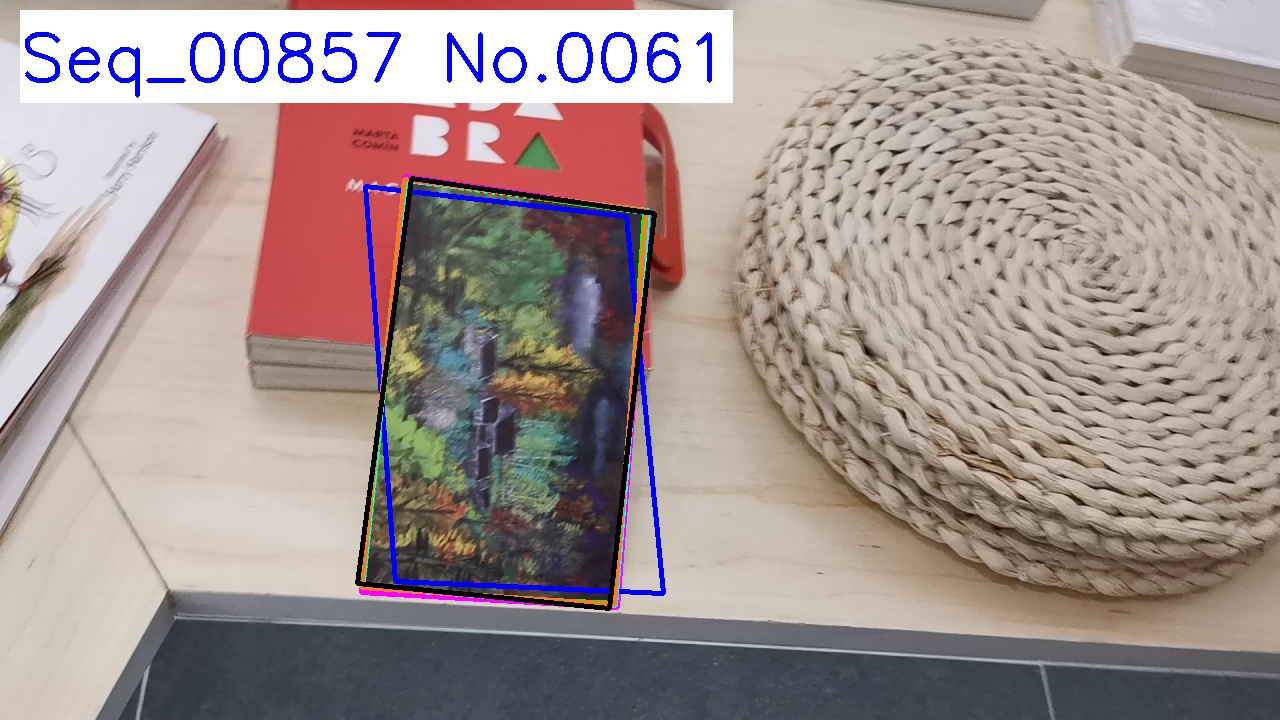} \includegraphics[width=0.19\linewidth]{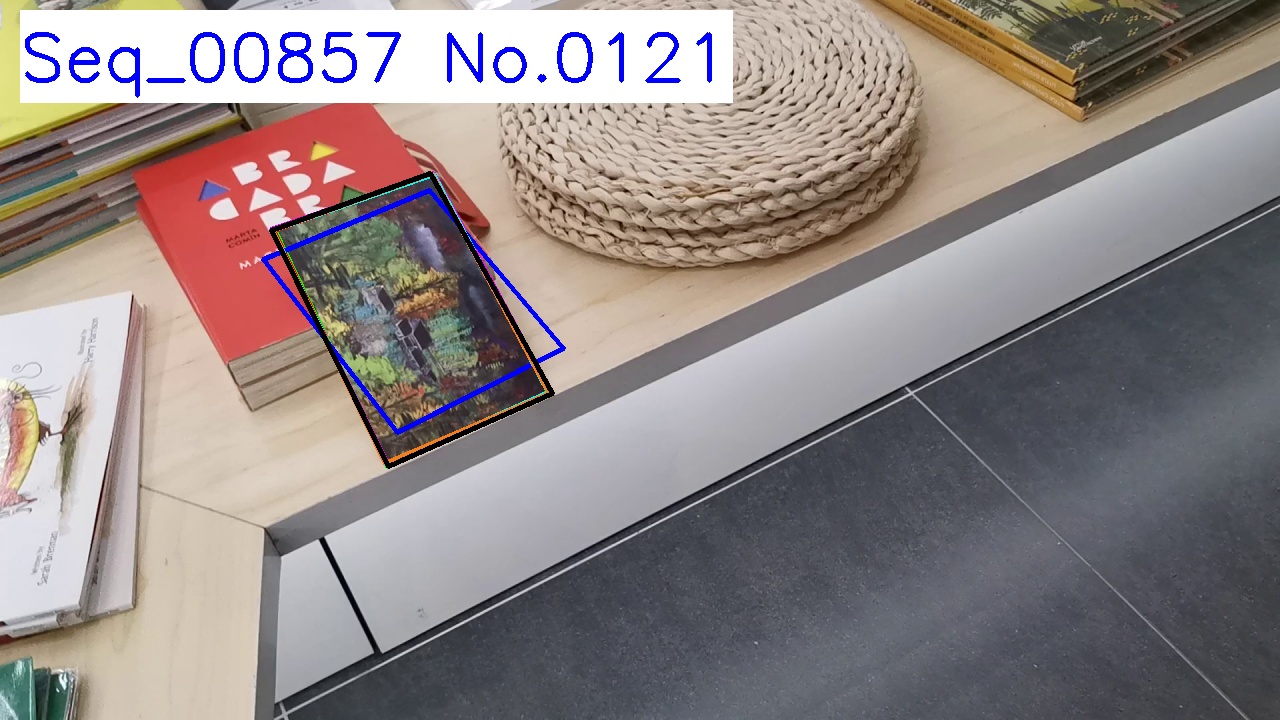} \includegraphics[width=0.19\linewidth]{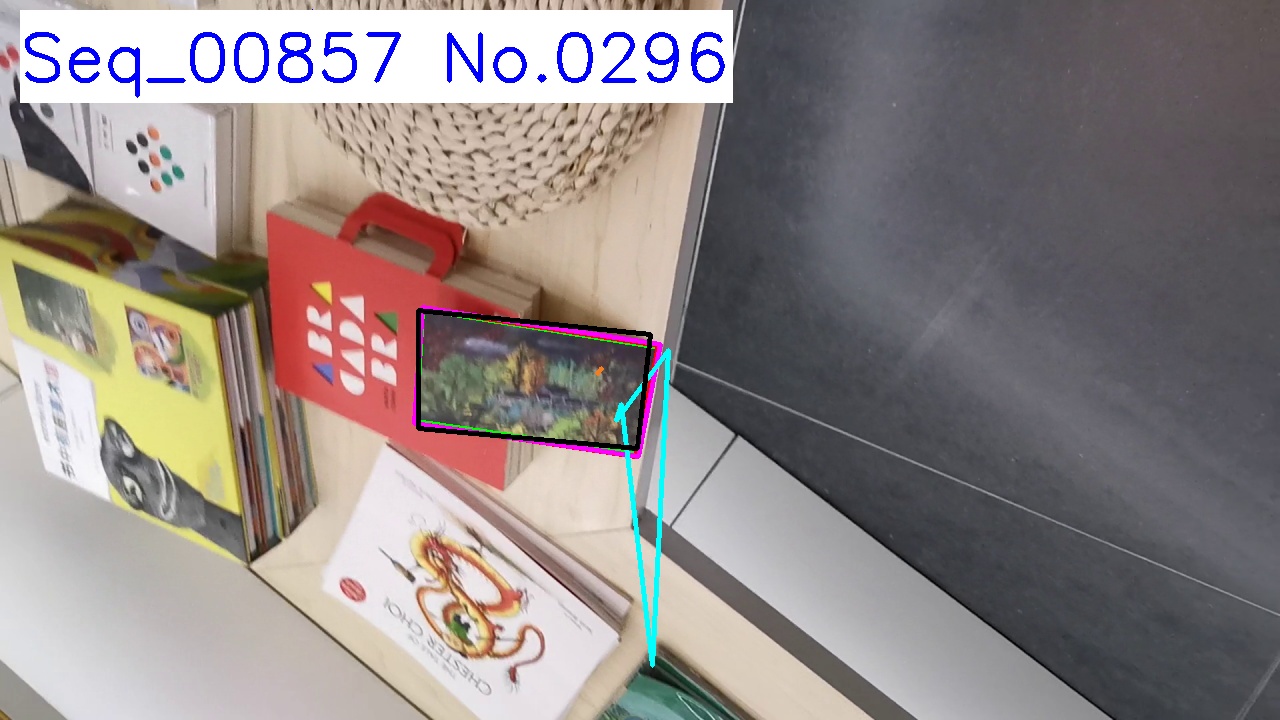} \includegraphics[width=0.19\linewidth]{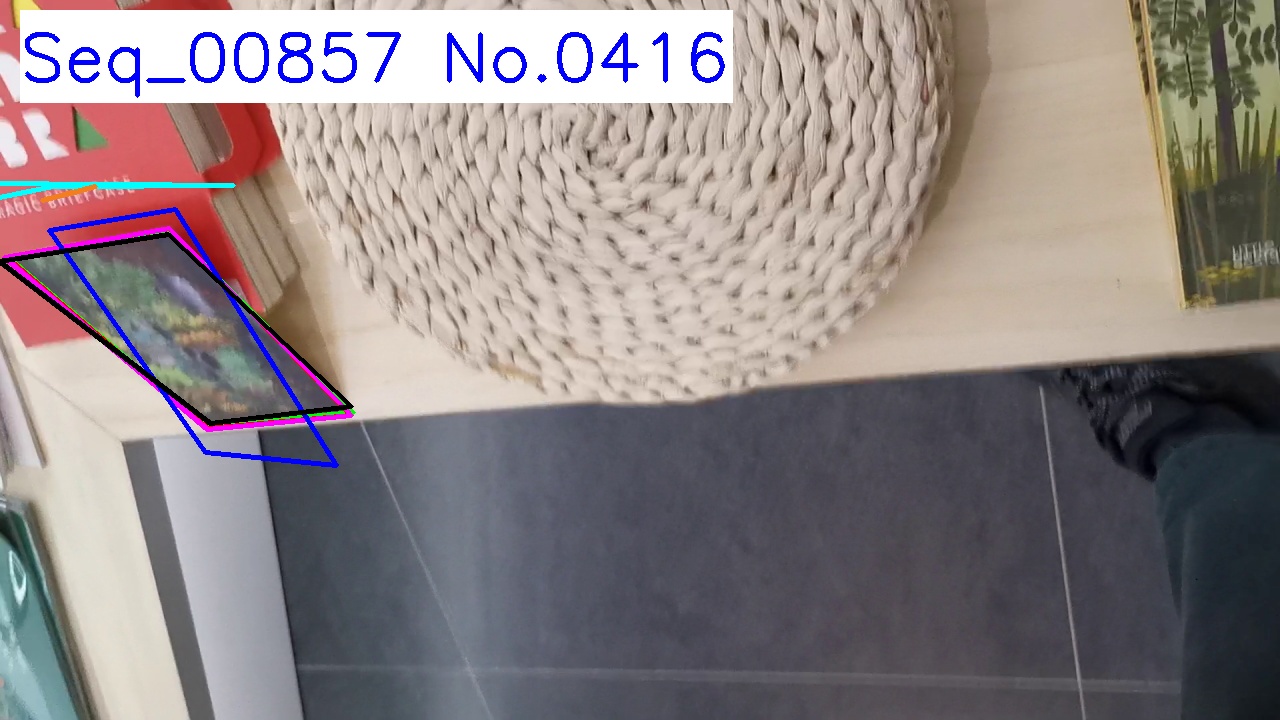} \\
        {\small (b) Sequence with SV and PD.} \\
        
        \includegraphics[width=0.19\linewidth]{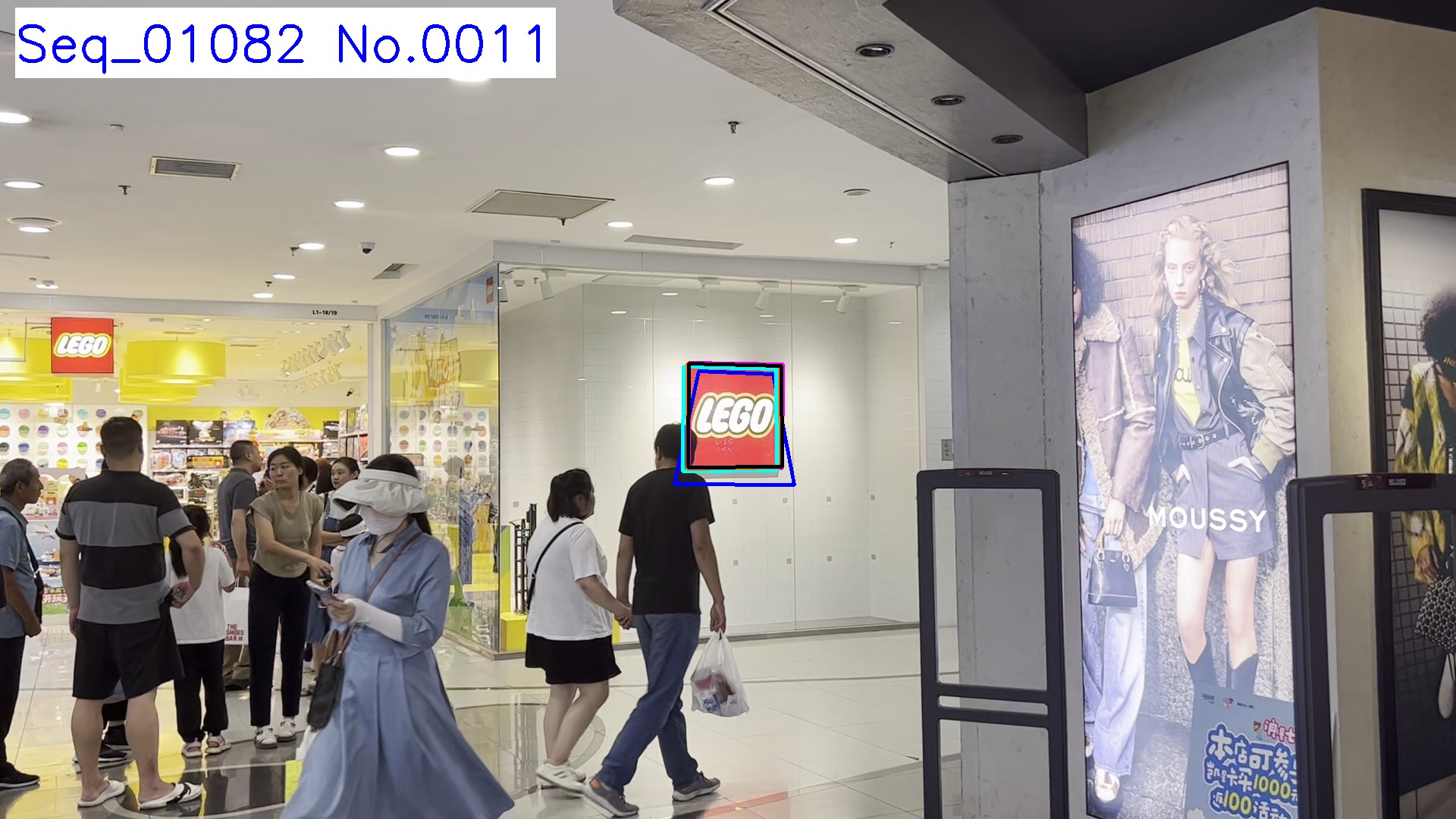} \includegraphics[width=0.19\linewidth]{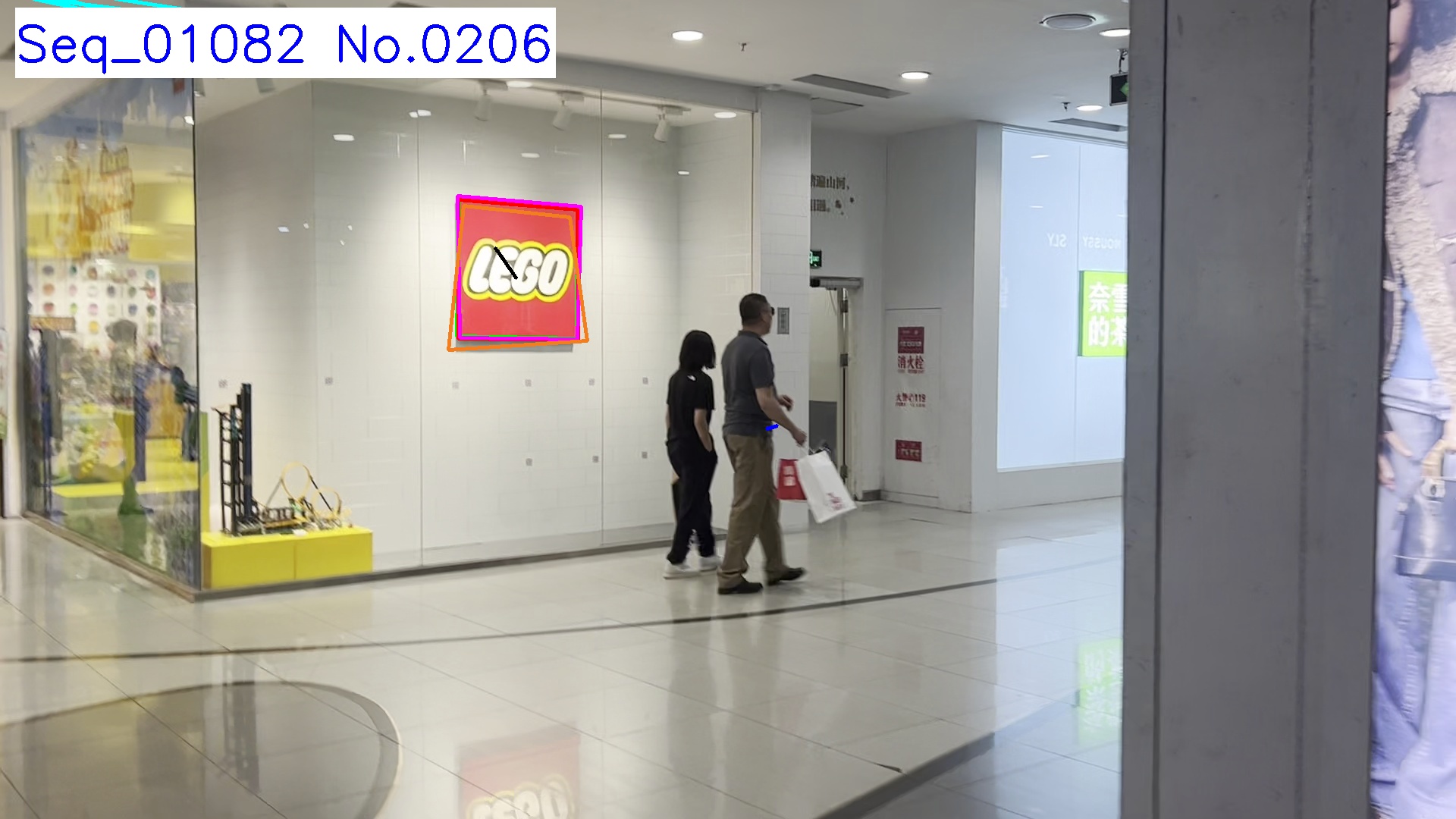} \includegraphics[width=0.19\linewidth]{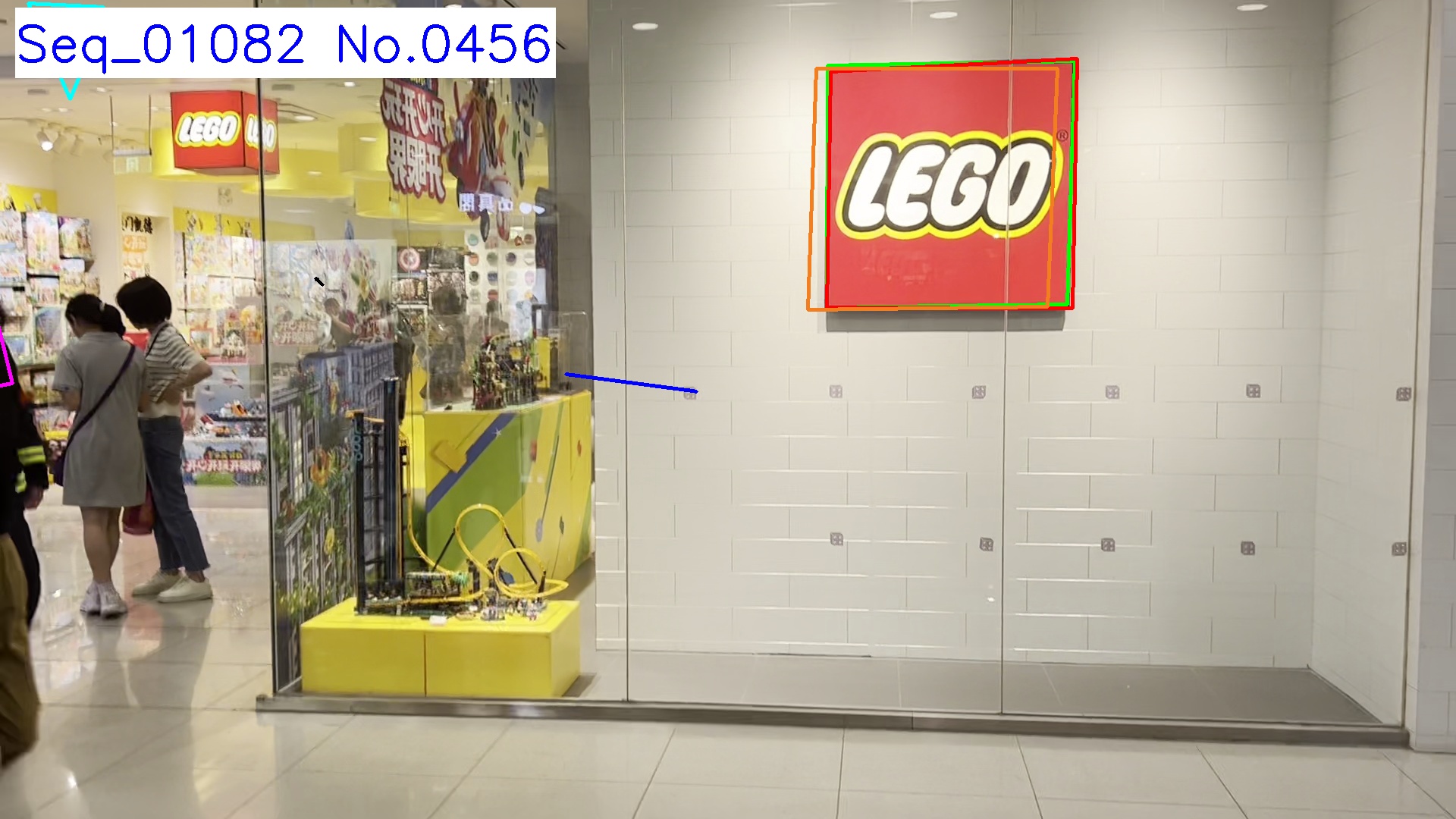}  \includegraphics[width=0.19\linewidth]{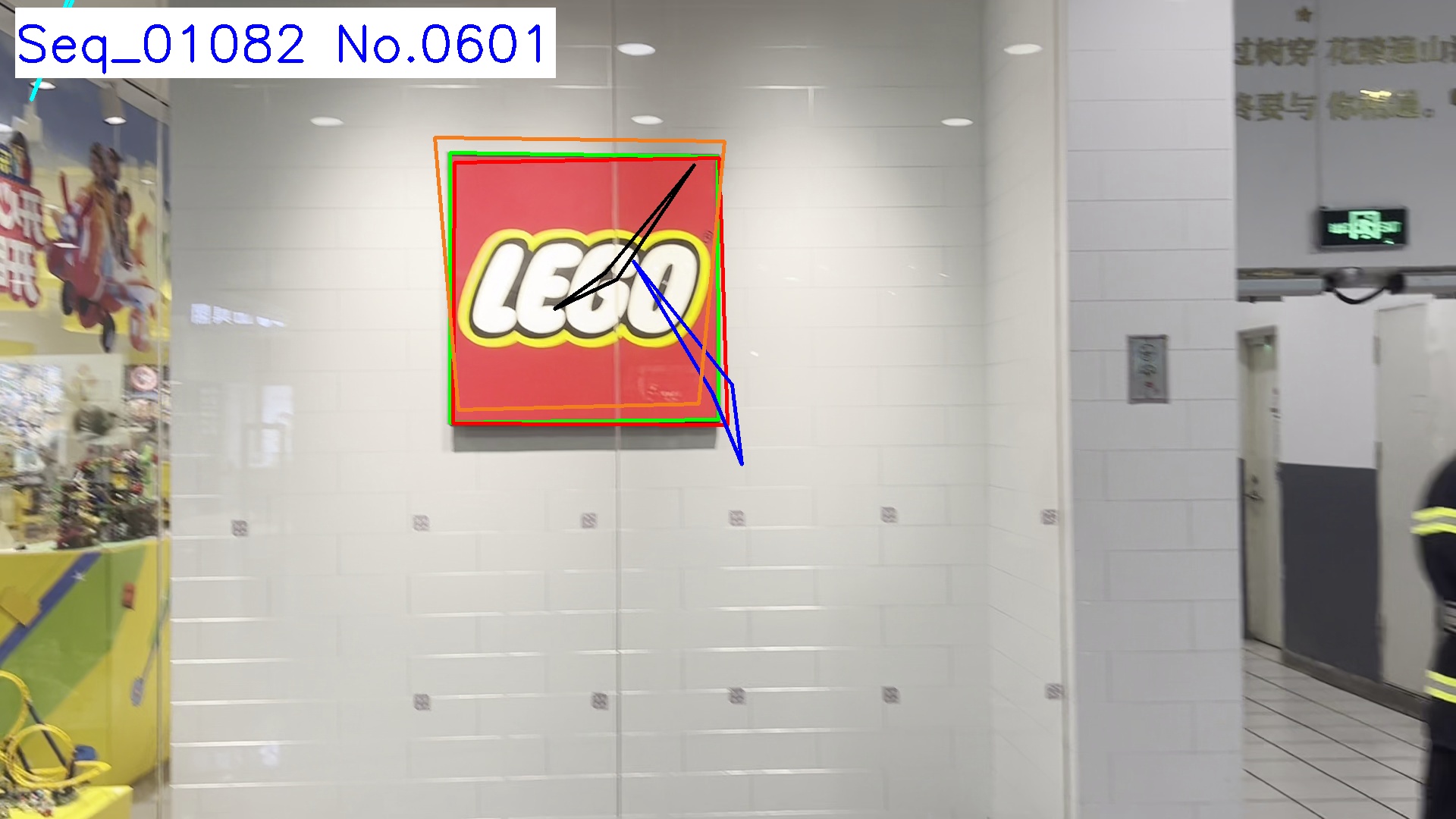} \includegraphics[width=0.19\linewidth]{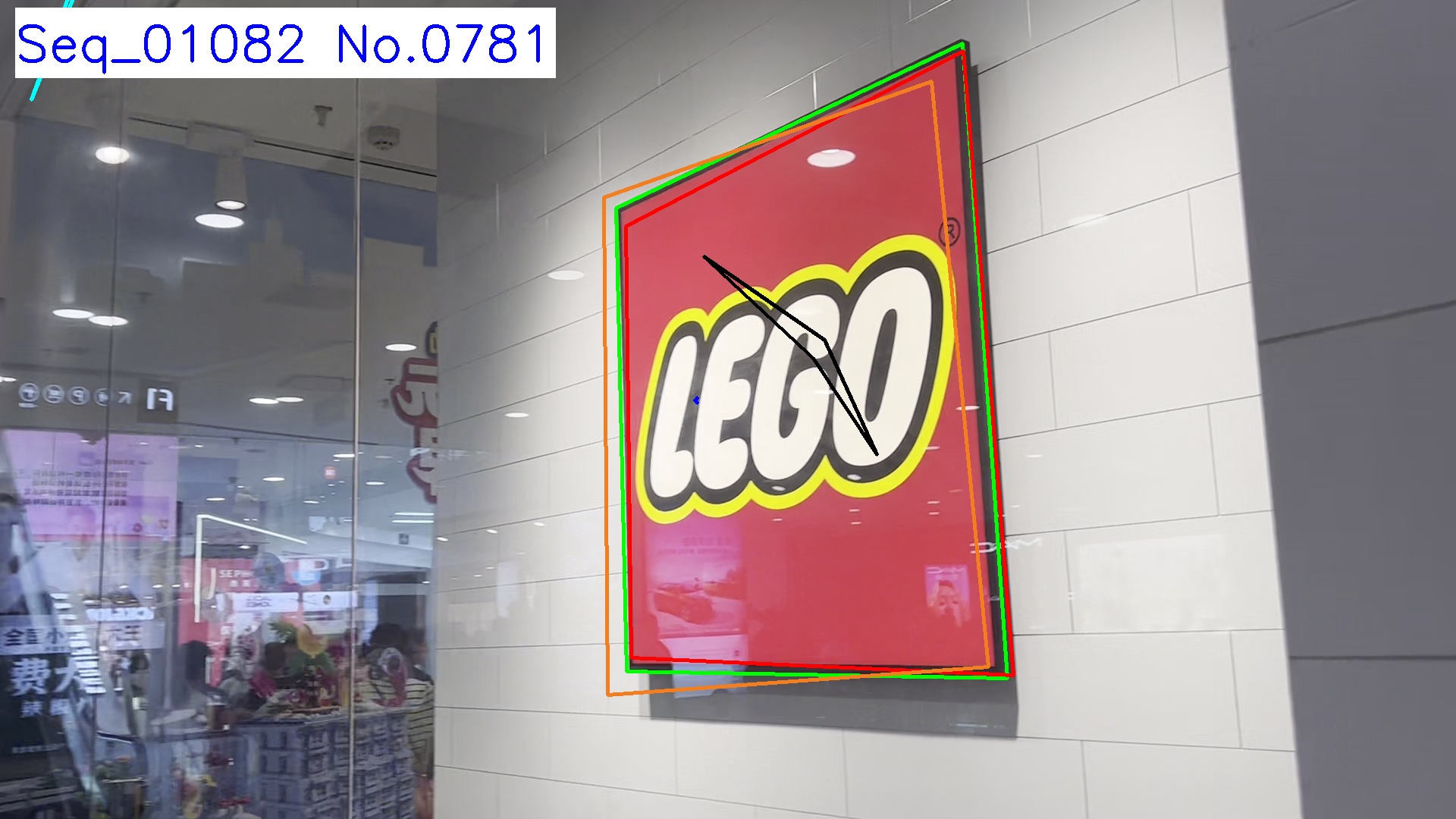} \\
        {\small (c) Sequence with BC and PD.} \\
        
        \includegraphics[width=0.19\linewidth]{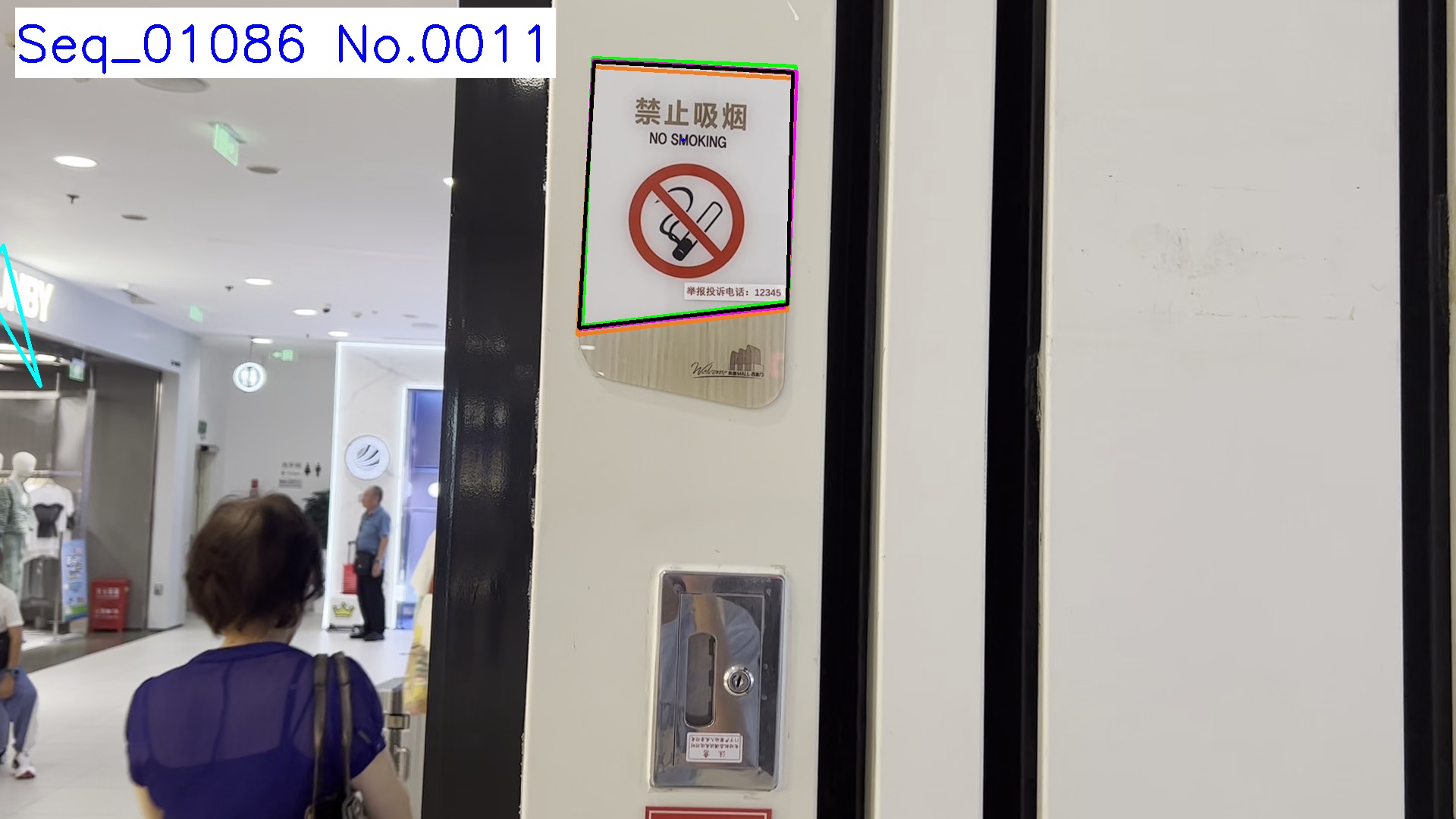} \includegraphics[width=0.19\linewidth]{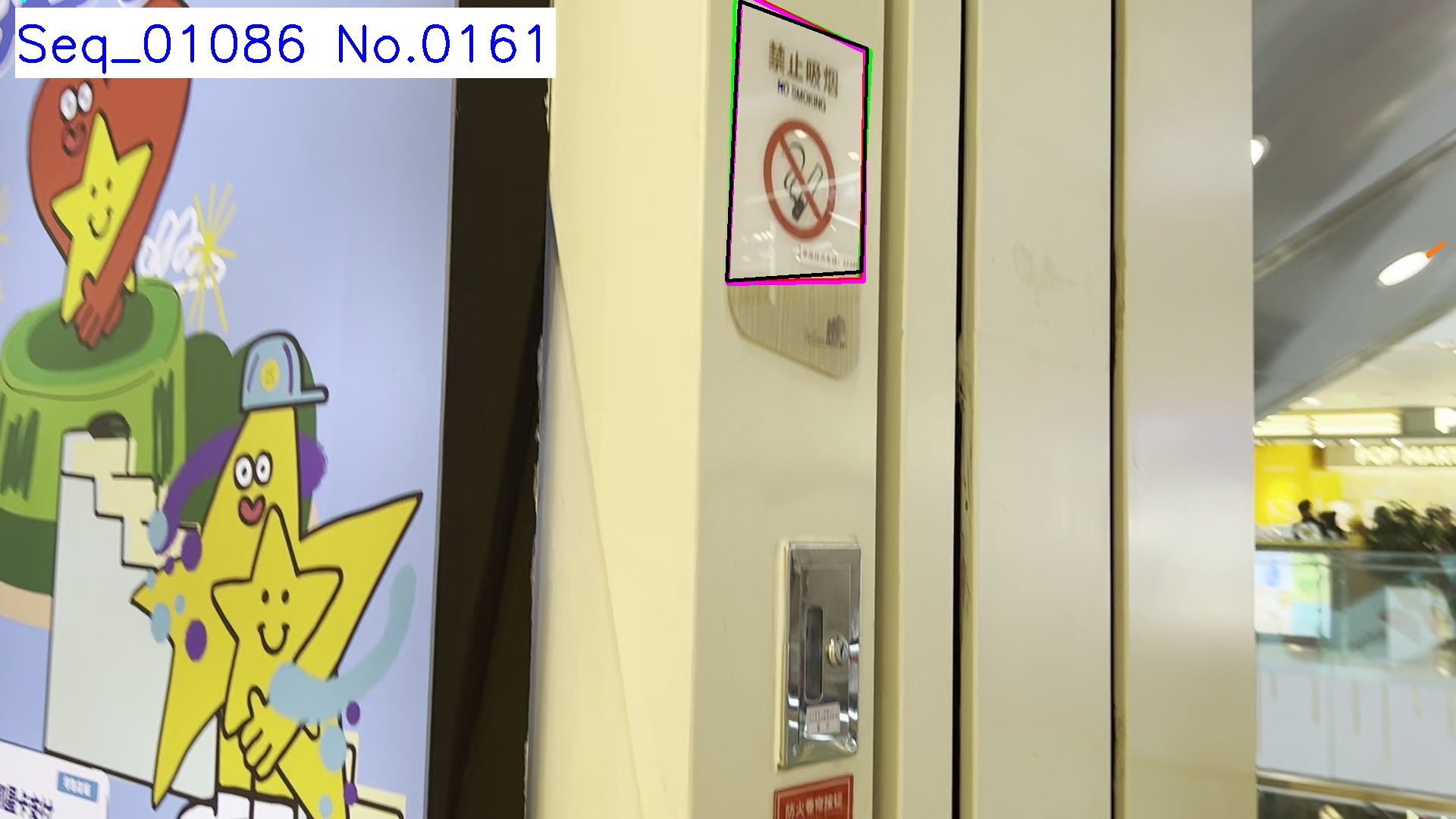} \includegraphics[width=0.19\linewidth]{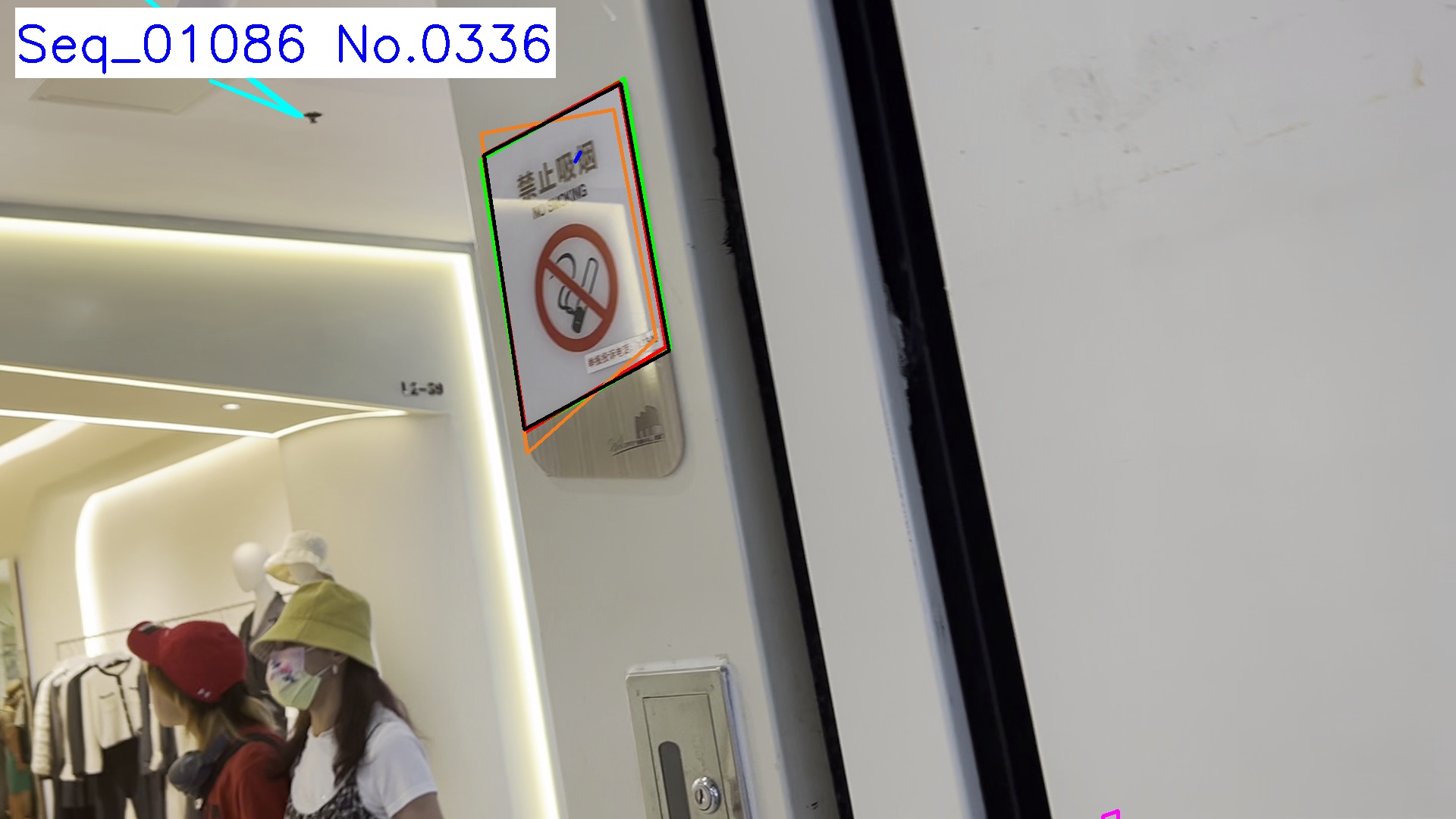} \includegraphics[width=0.19\linewidth]{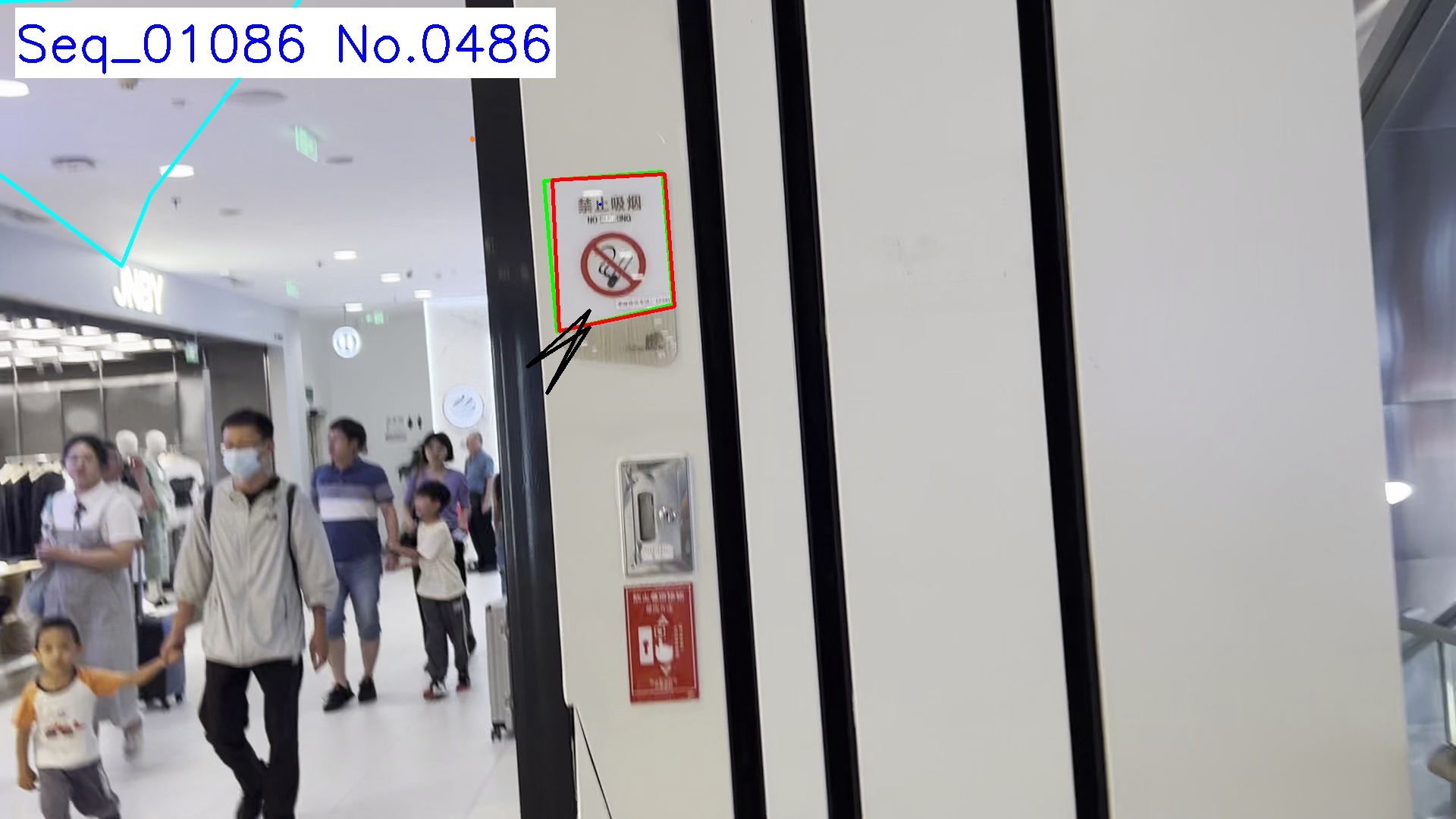}  \includegraphics[width=0.19\linewidth]{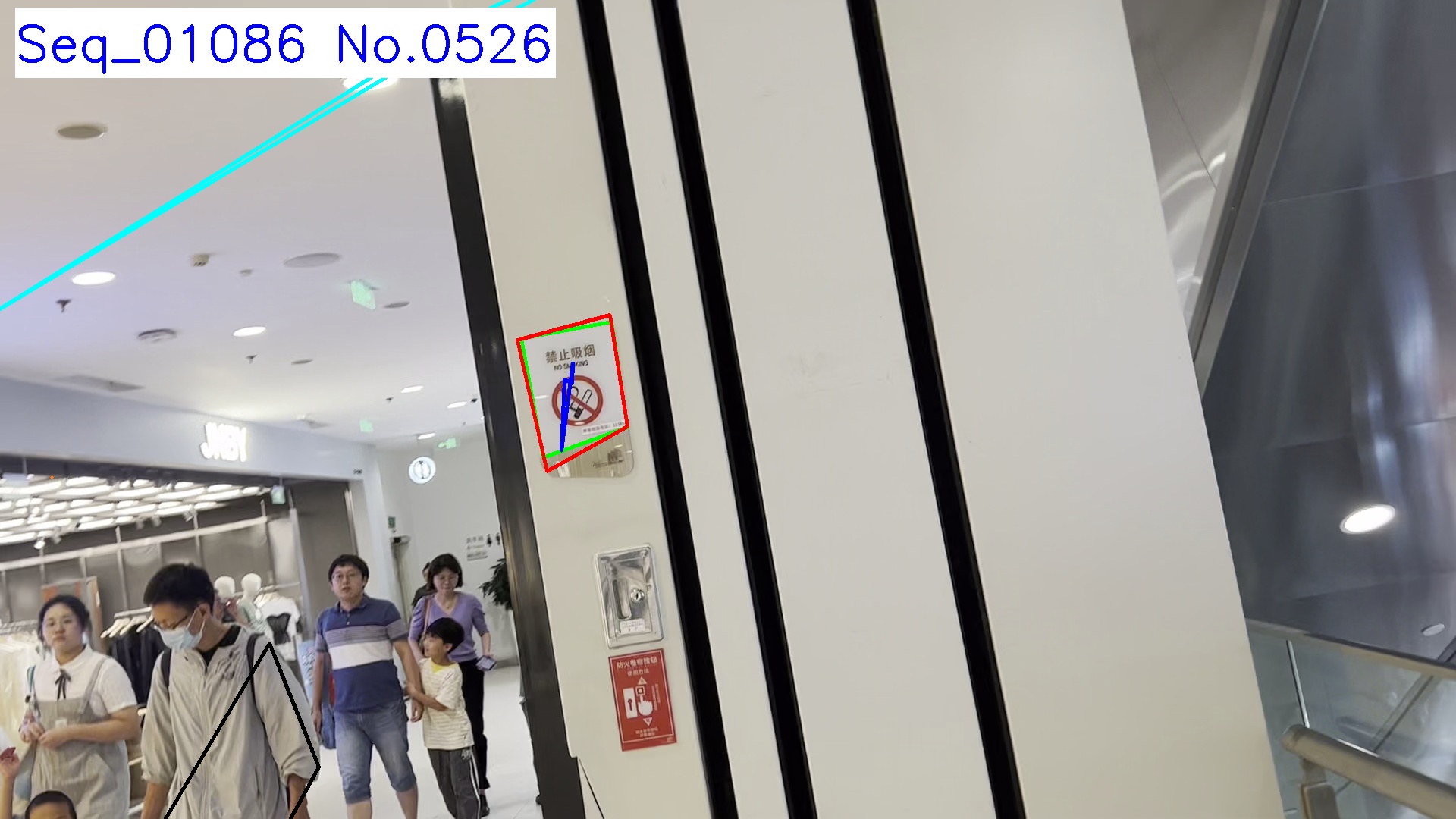} \\
        {\small (d) Sequence with BC and ROT.} \\
        
        \includegraphics[width=0.19\linewidth]{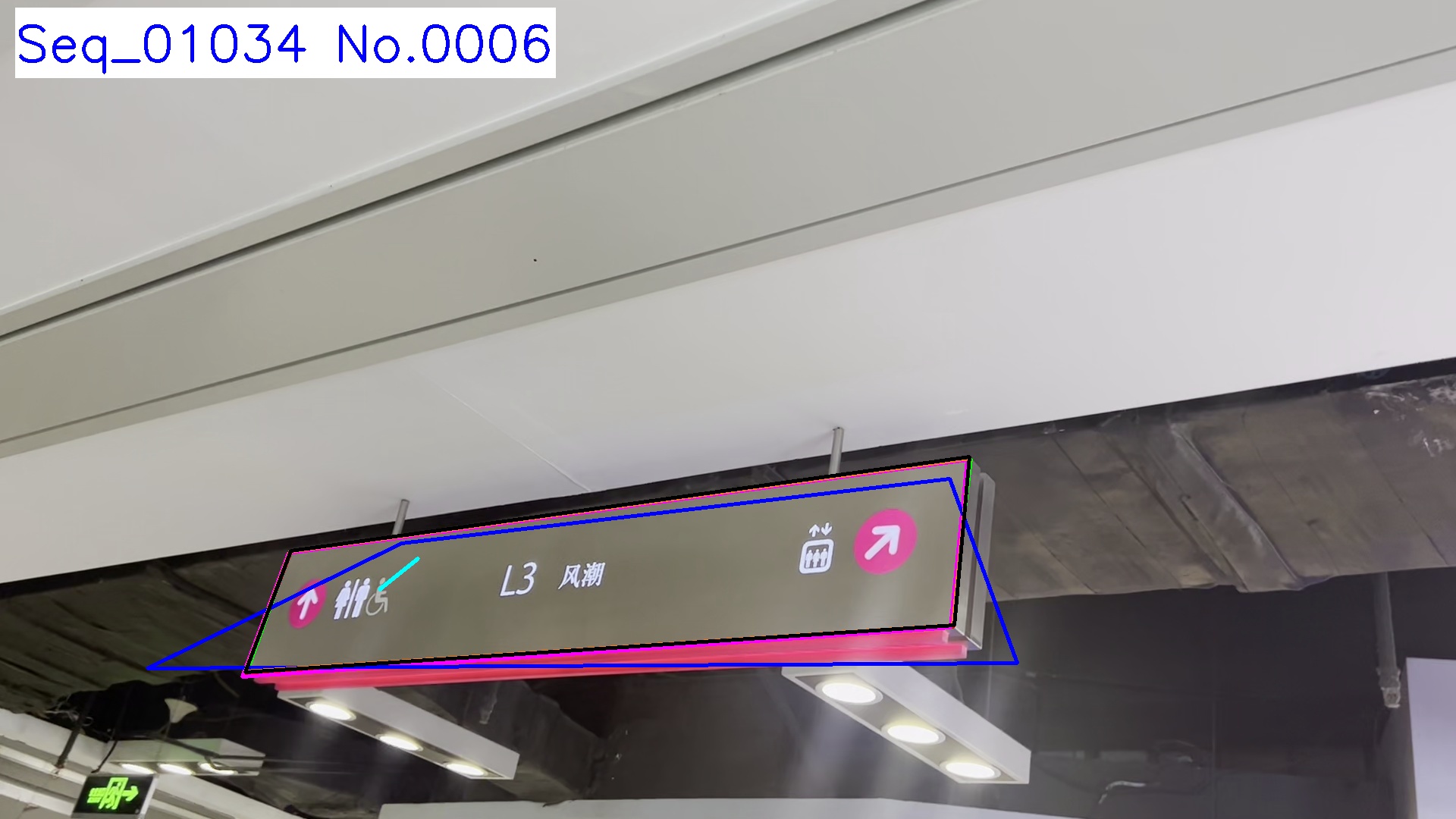} \includegraphics[width=0.19\linewidth]{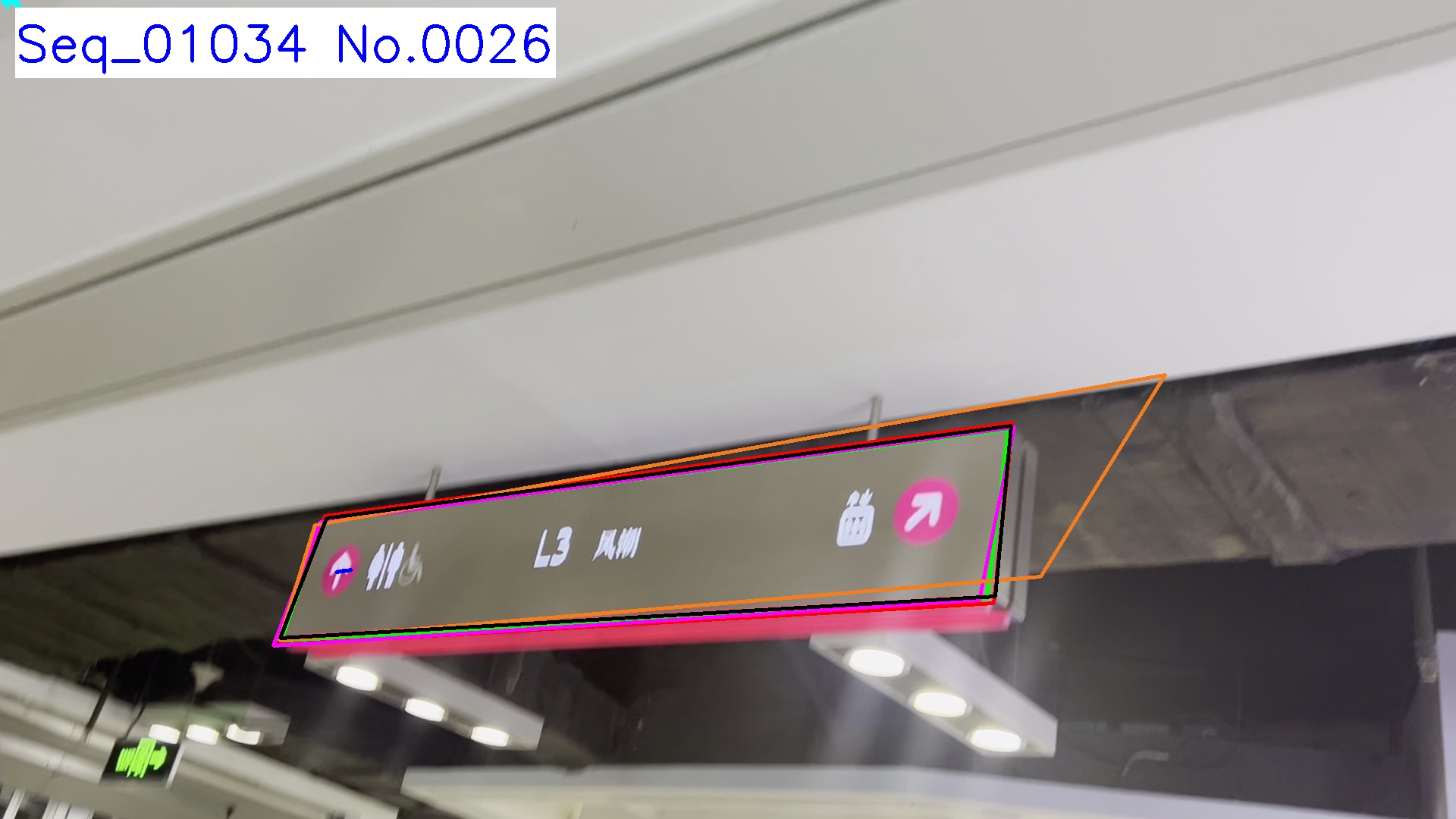} \includegraphics[width=0.19\linewidth]{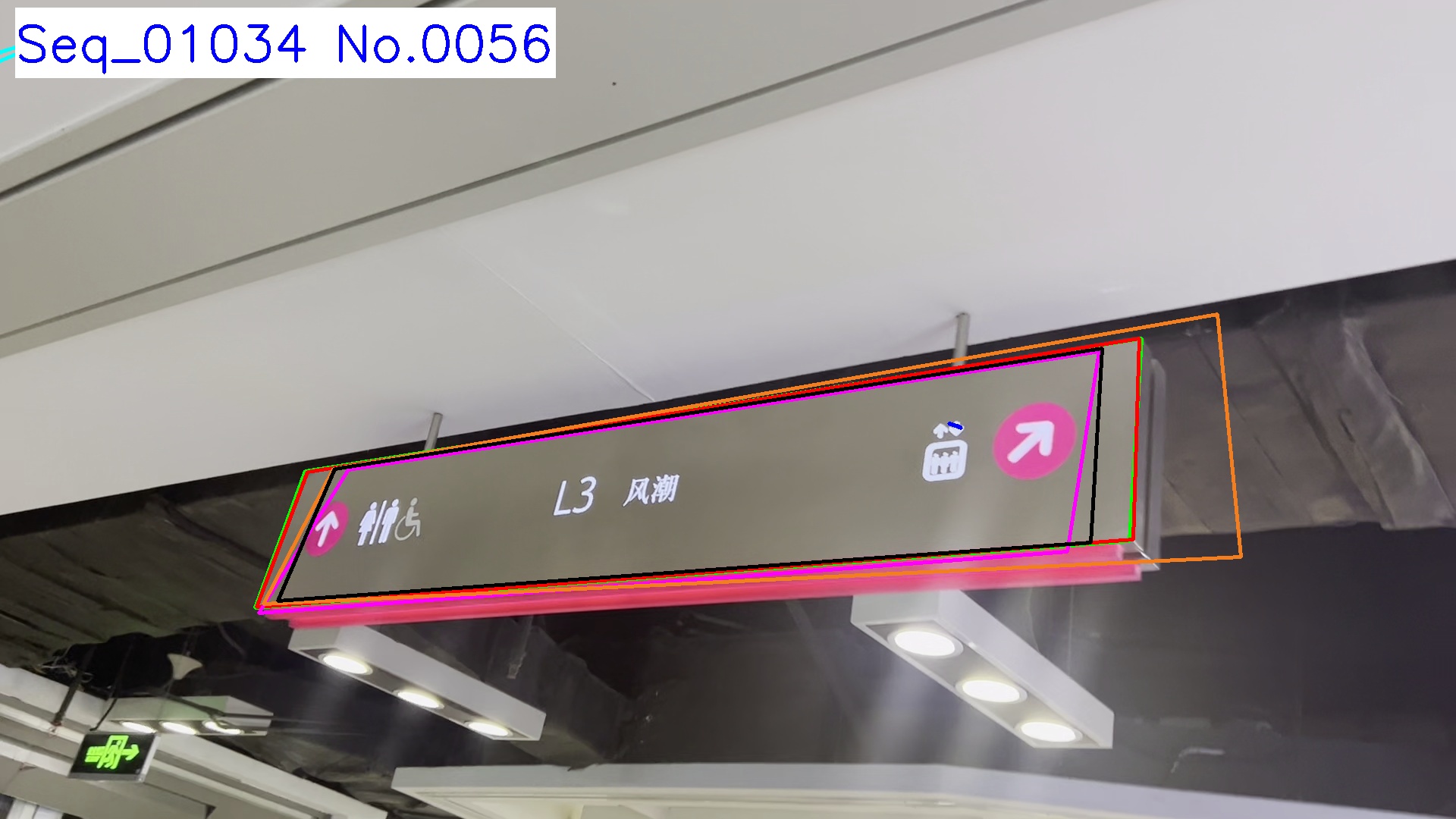} \includegraphics[width=0.19\linewidth]{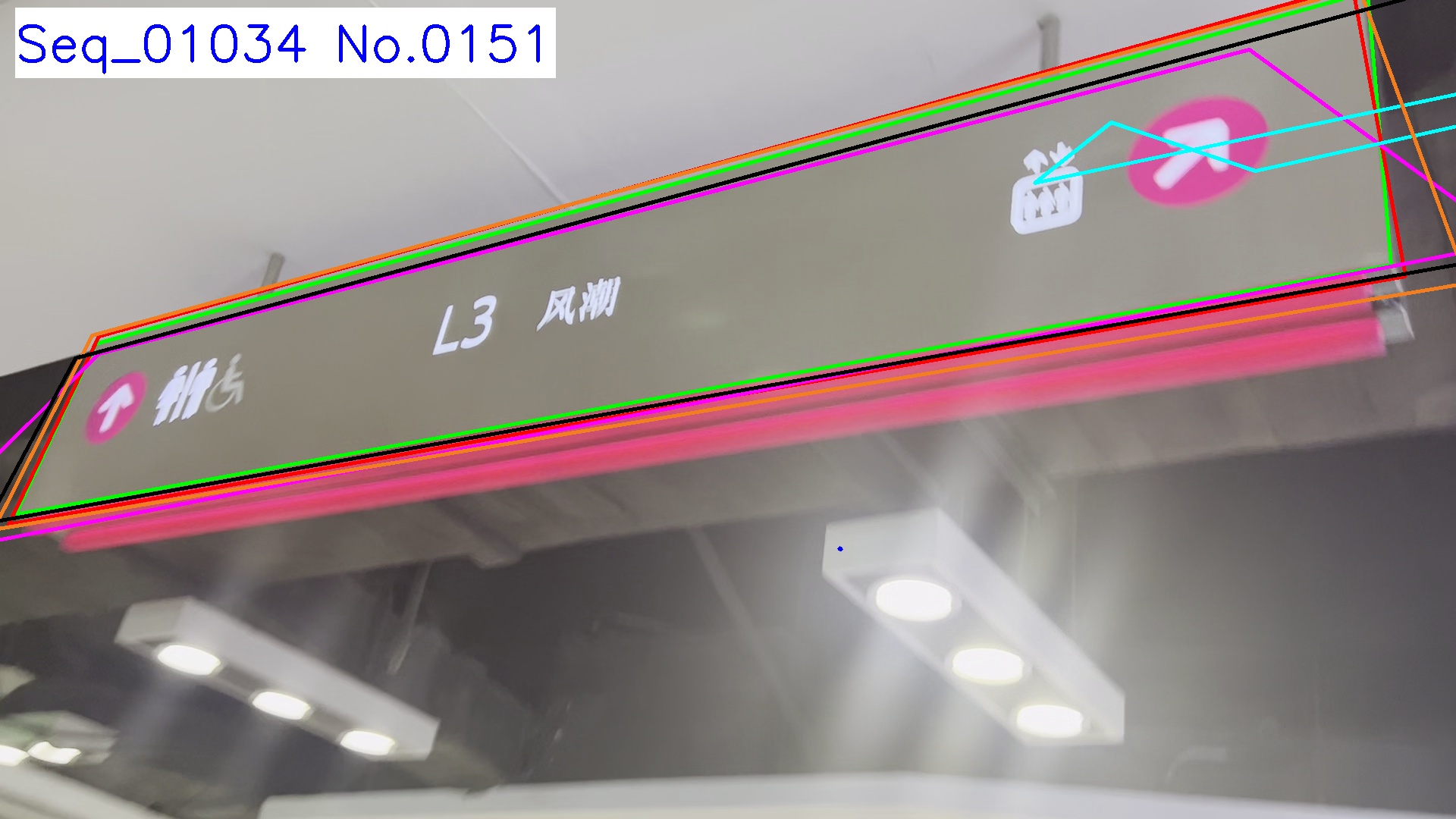}  \includegraphics[width=0.19\linewidth]{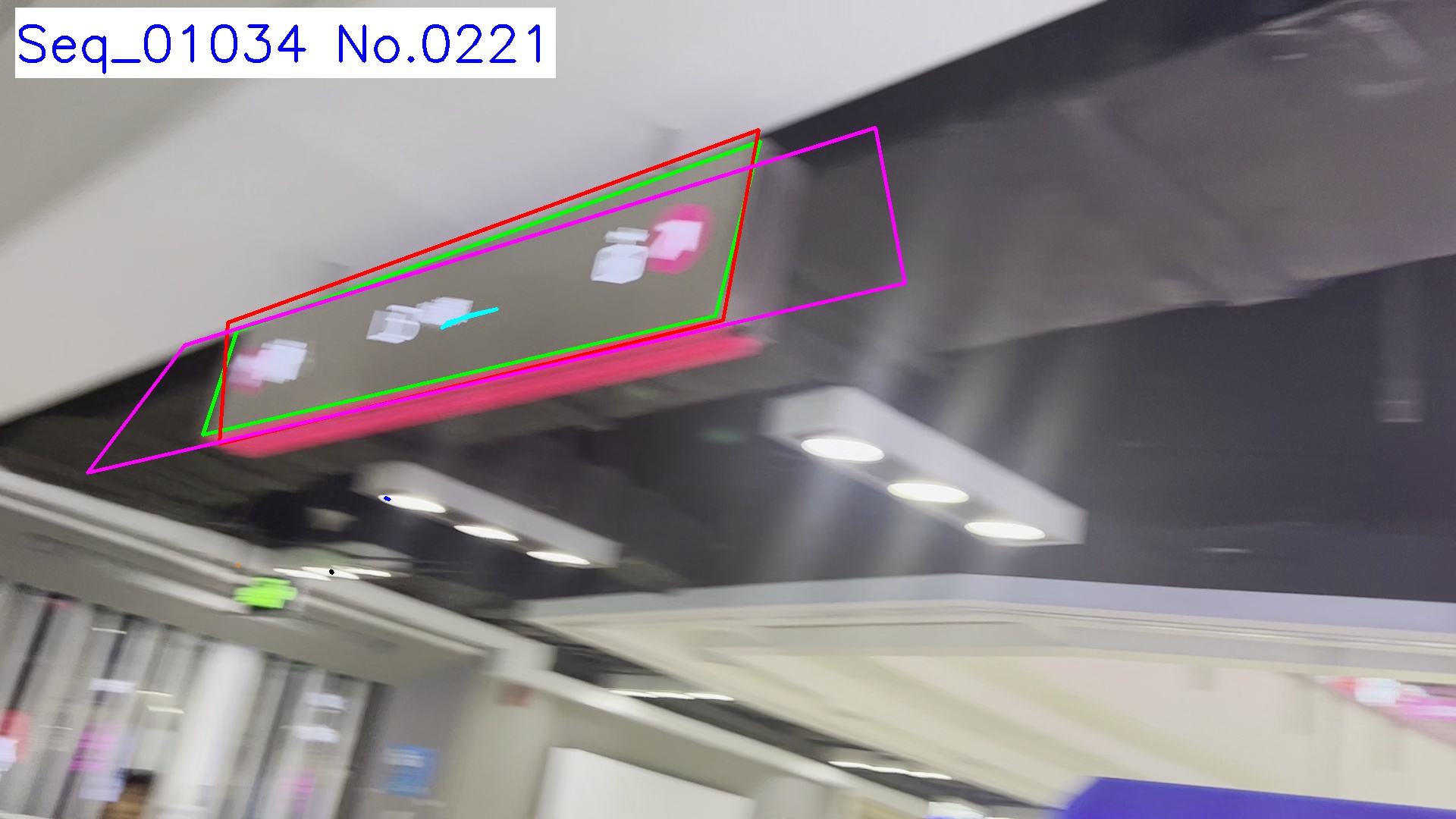} \\
        {\small (e) Sequence with MB and OV.} \\
        
        \includegraphics[width=0.19\linewidth]{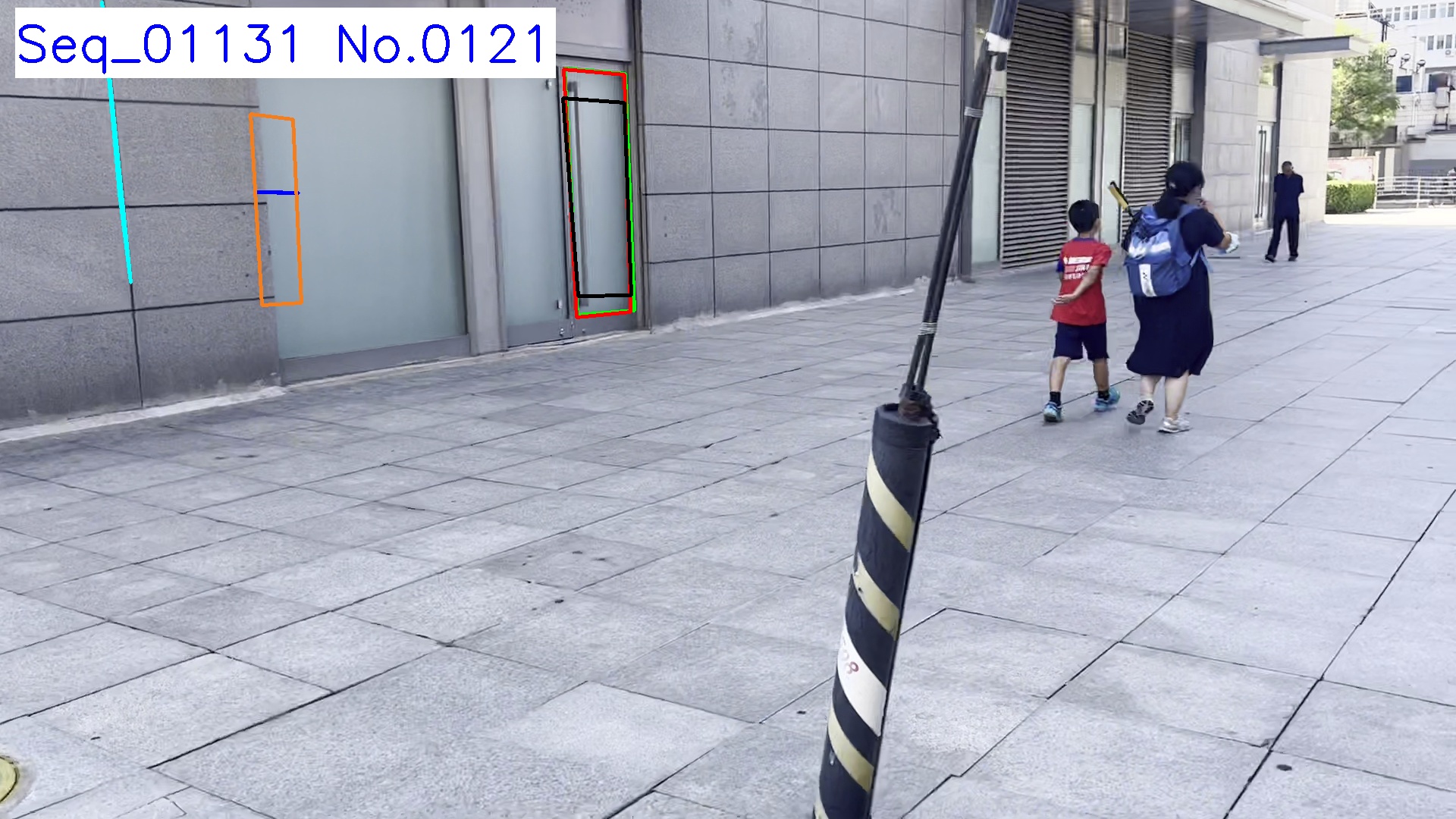} \includegraphics[width=0.19\linewidth]{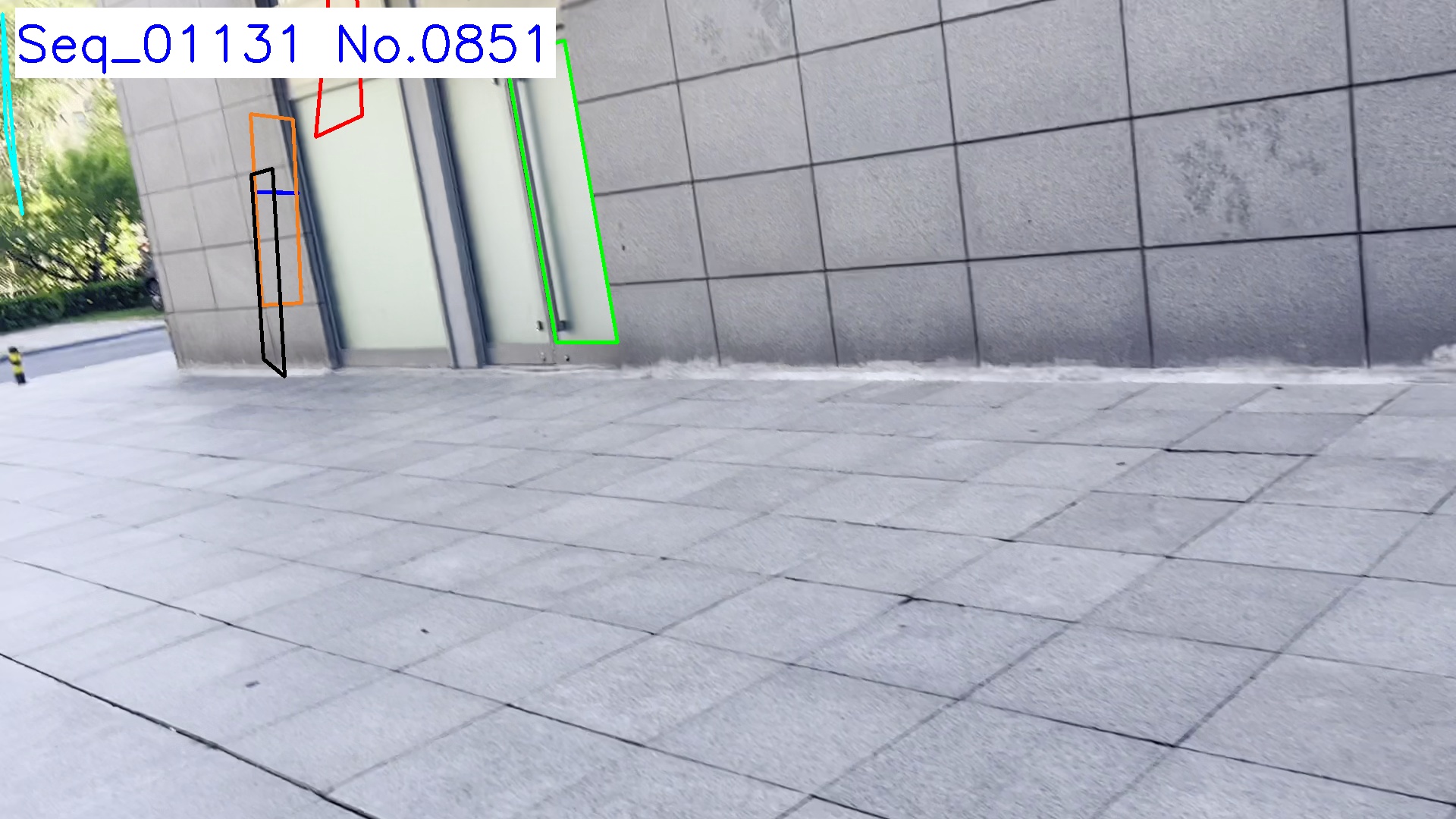} \includegraphics[width=0.19\linewidth]{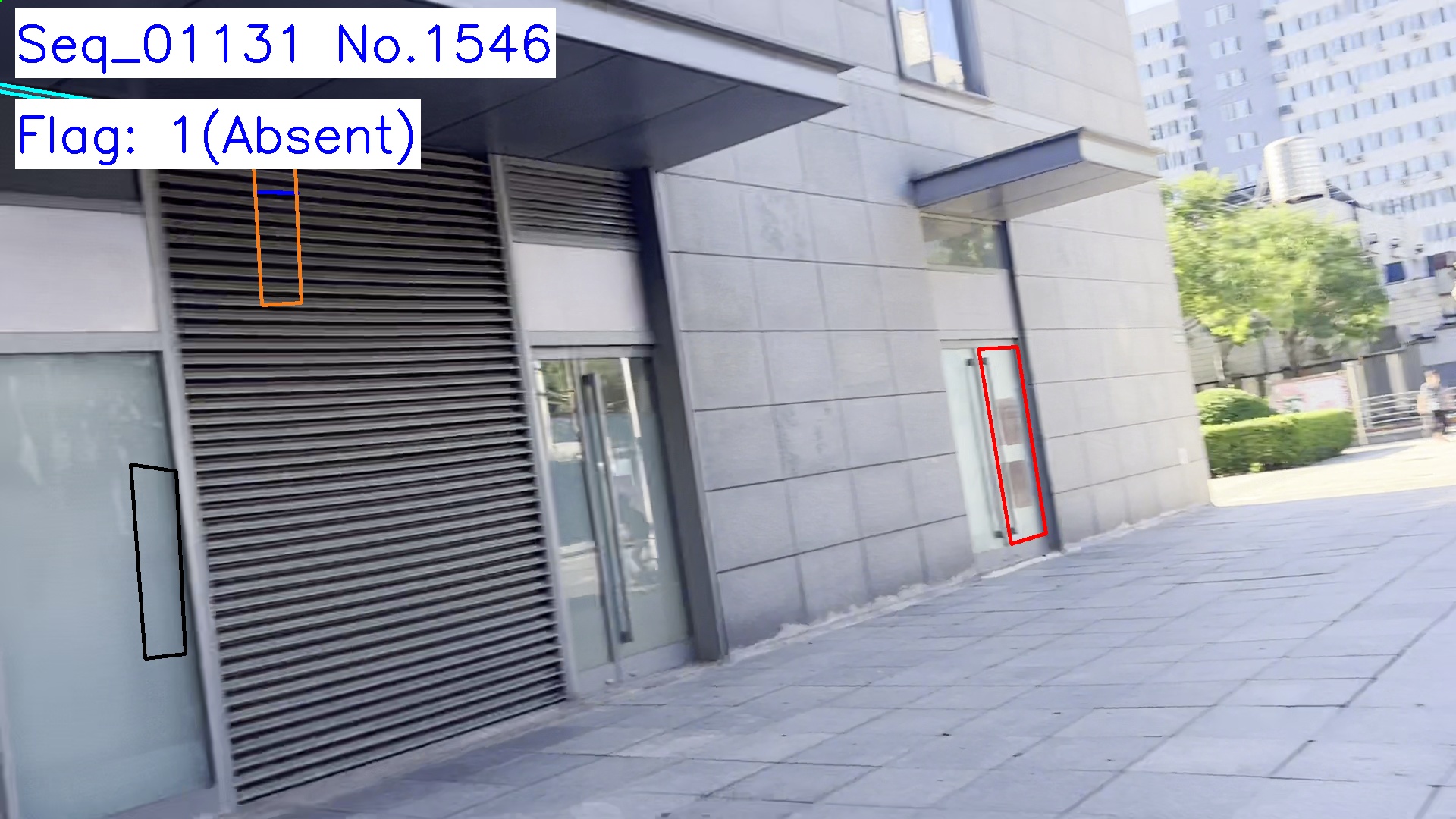} \includegraphics[width=0.19\linewidth]{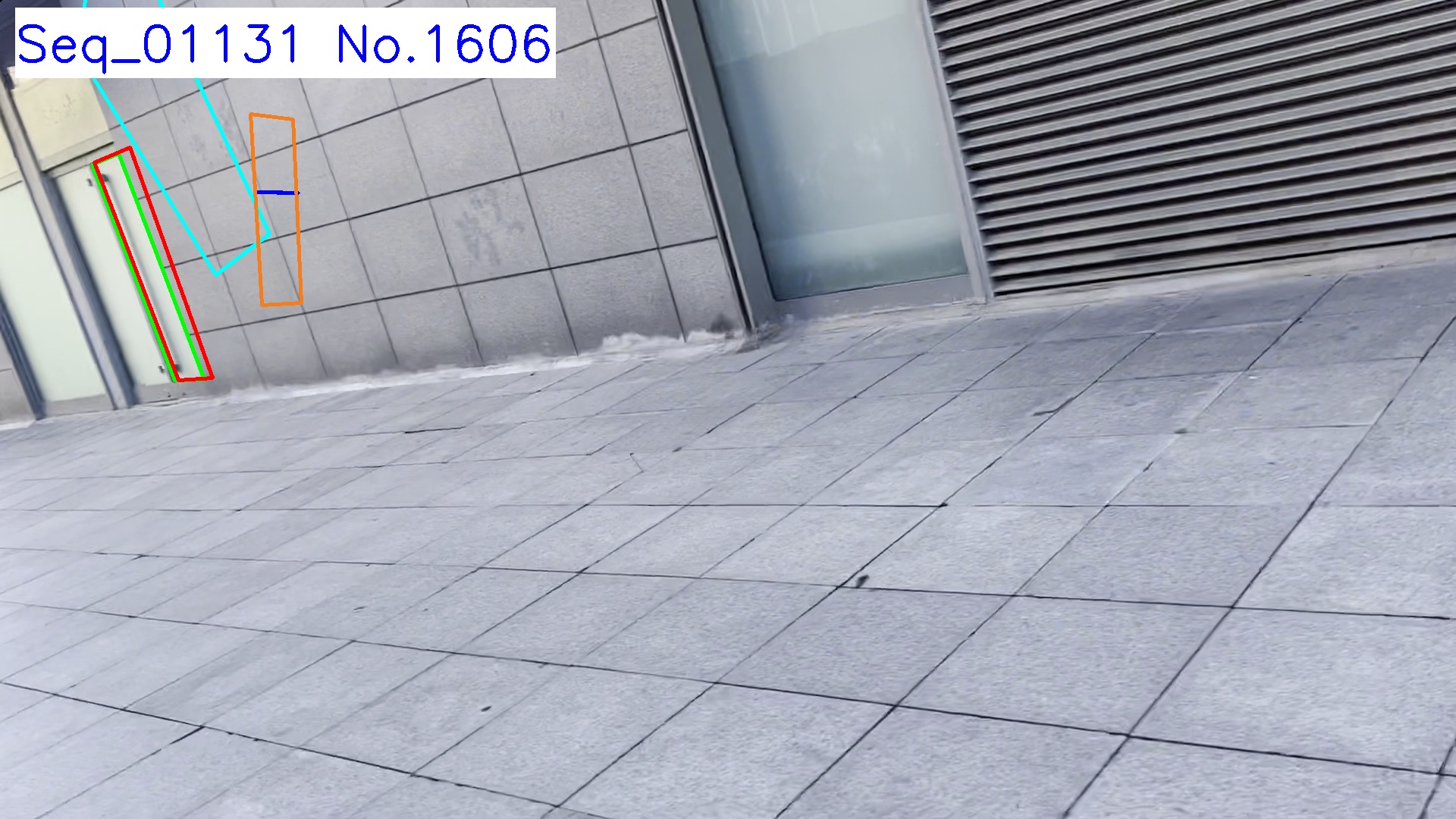}  \includegraphics[width=0.19\linewidth]{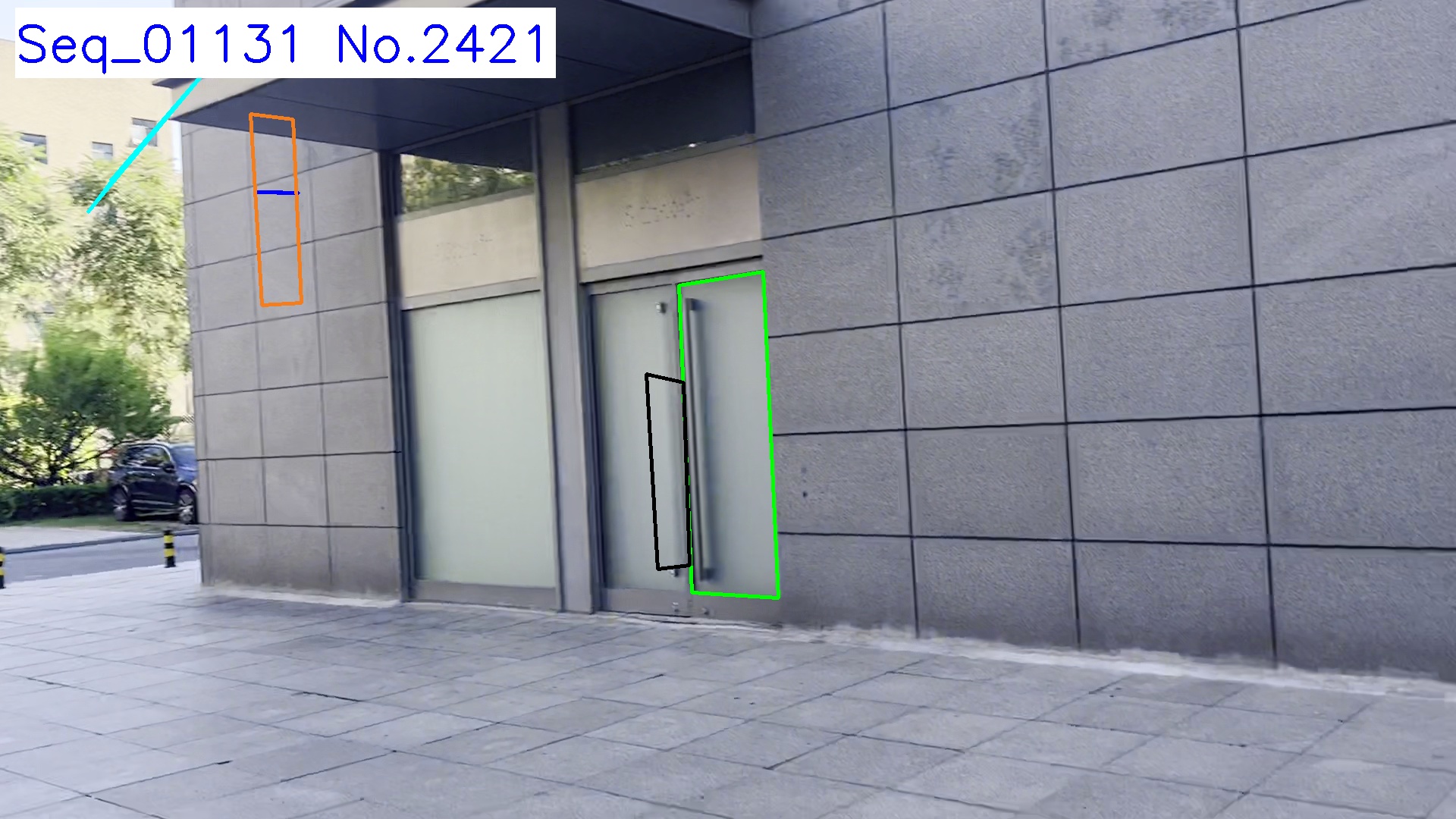} \\
        {\small (f) Ultra-long sequence specially for long-term tracking} \\

        \includegraphics[width=0.19\linewidth]{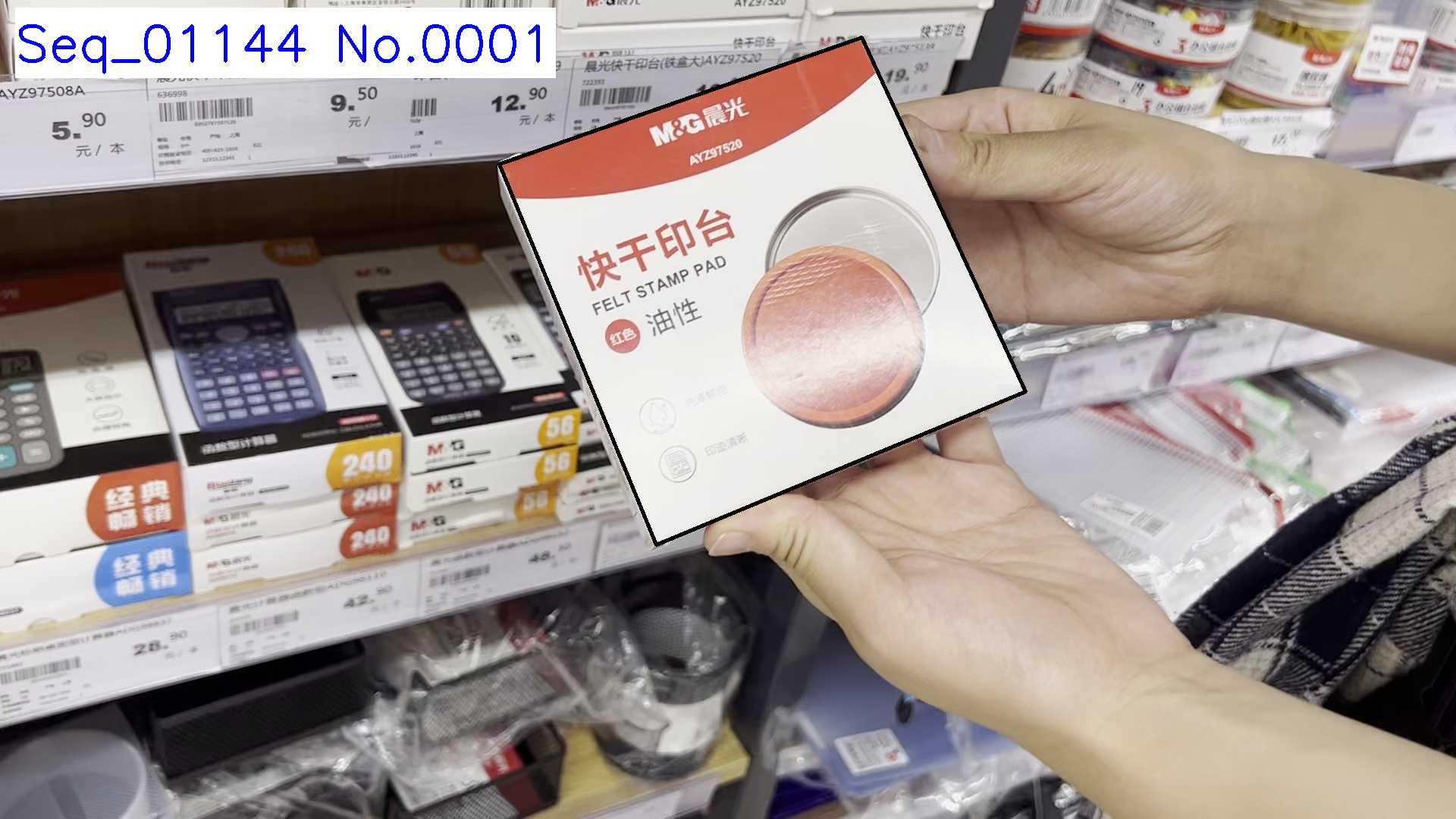} \includegraphics[width=0.19\linewidth]{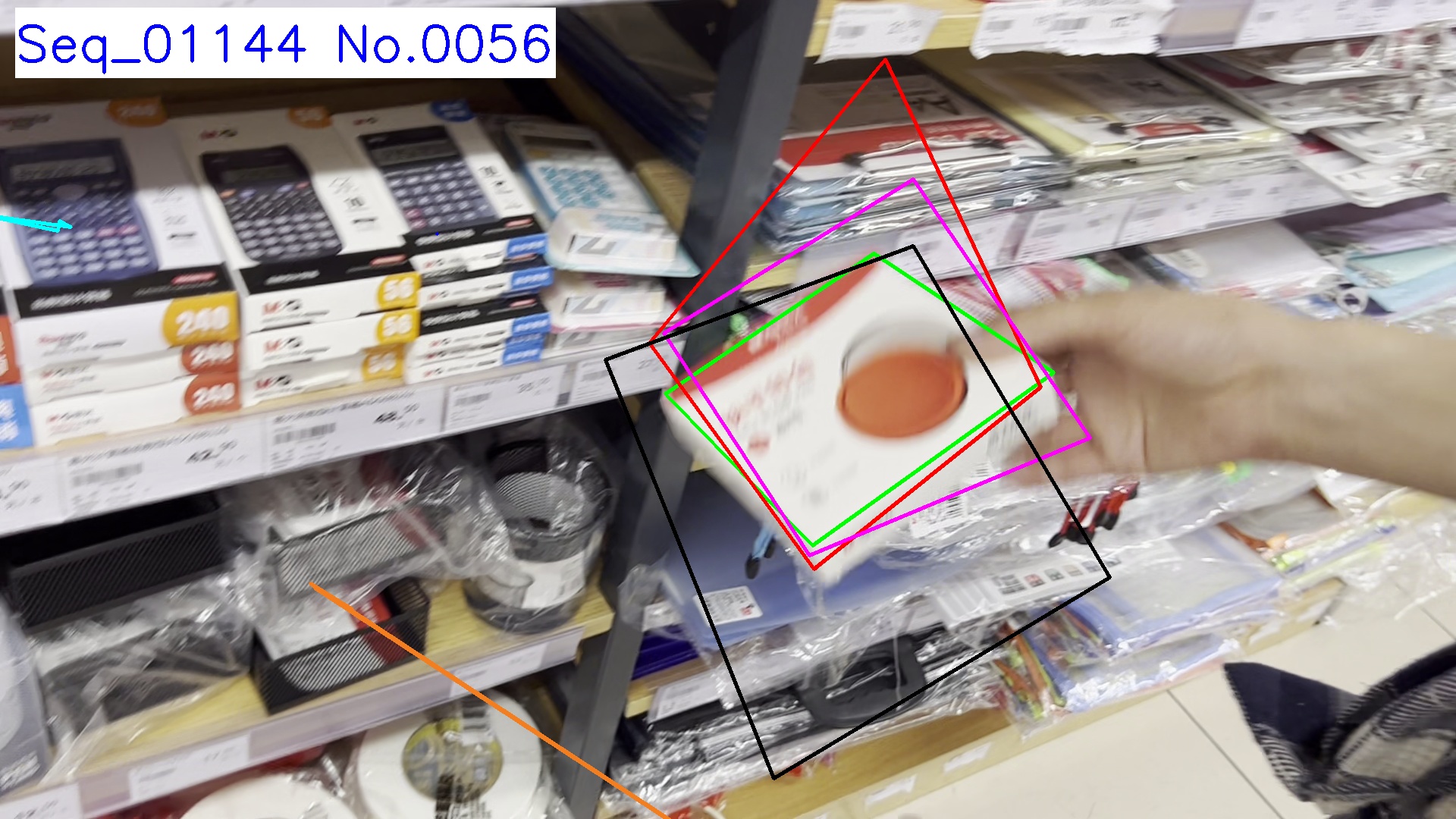} \includegraphics[width=0.19\linewidth]{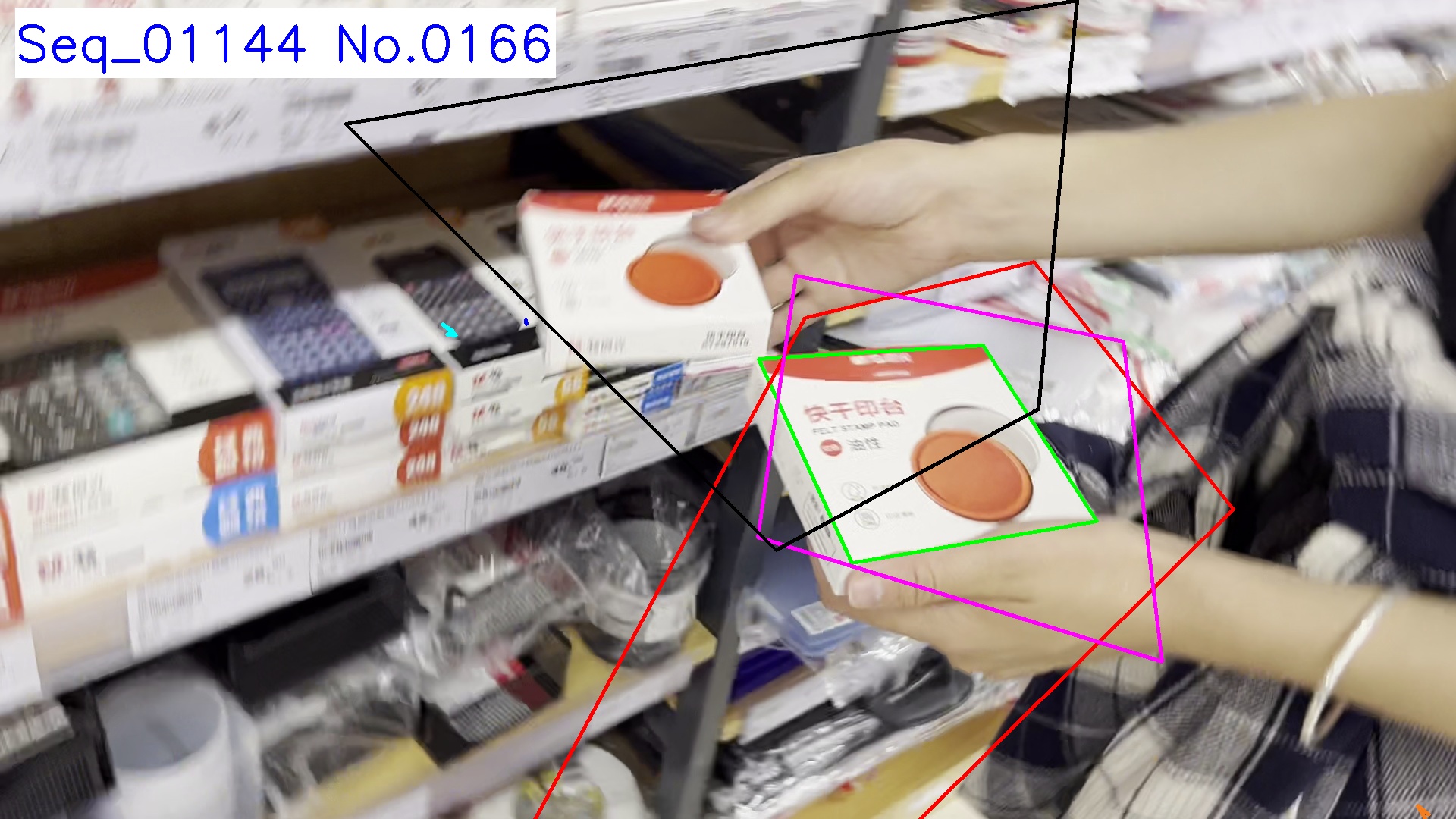} \includegraphics[width=0.19\linewidth]{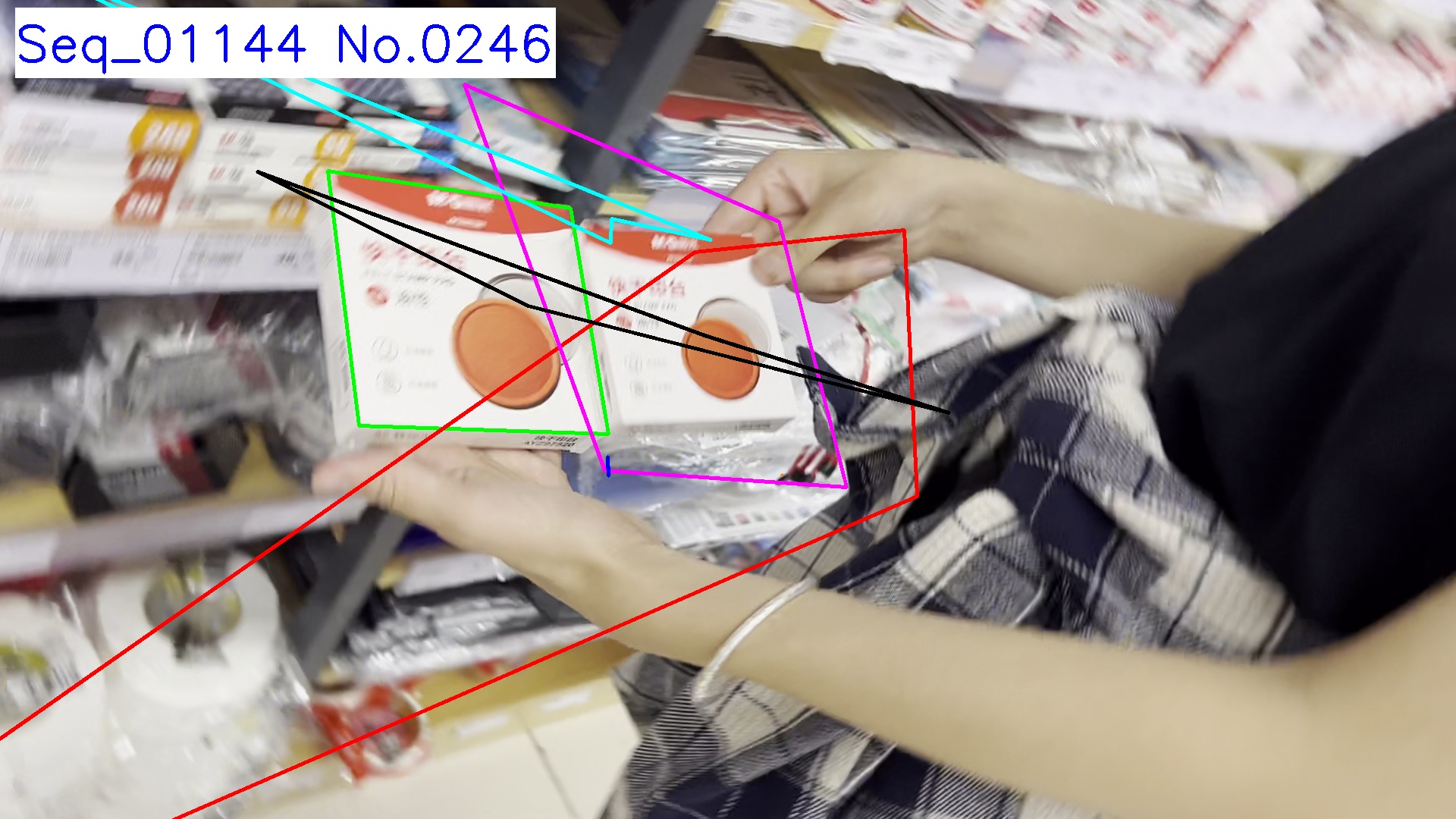} \includegraphics[width=0.19\linewidth]{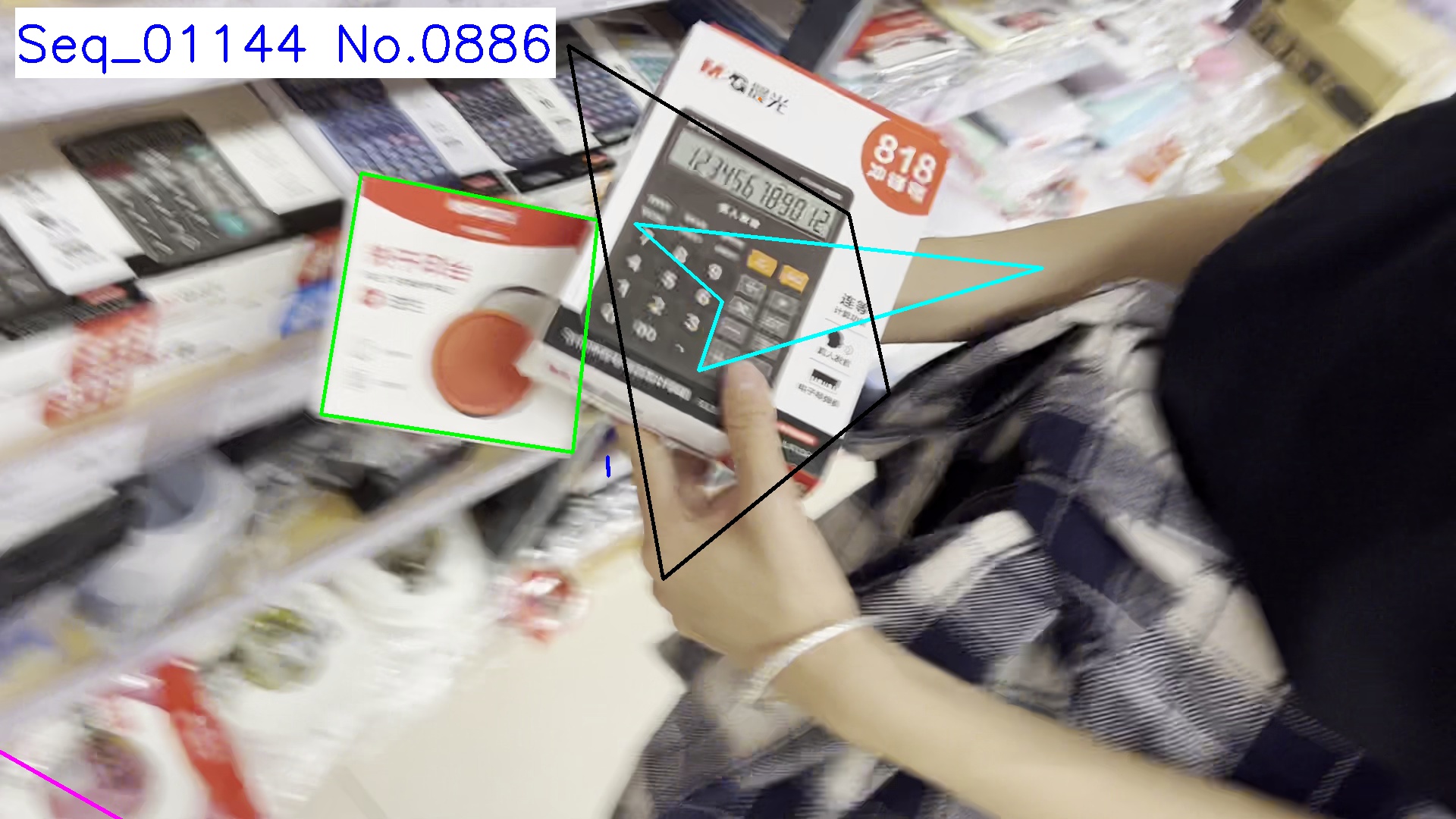} \\
        {\small (g) Sequence with multiple challenging factors, including OCC, MB, ROT, SV, PD, OV, LR.} \\
        
        \includegraphics[width=0.19\linewidth]{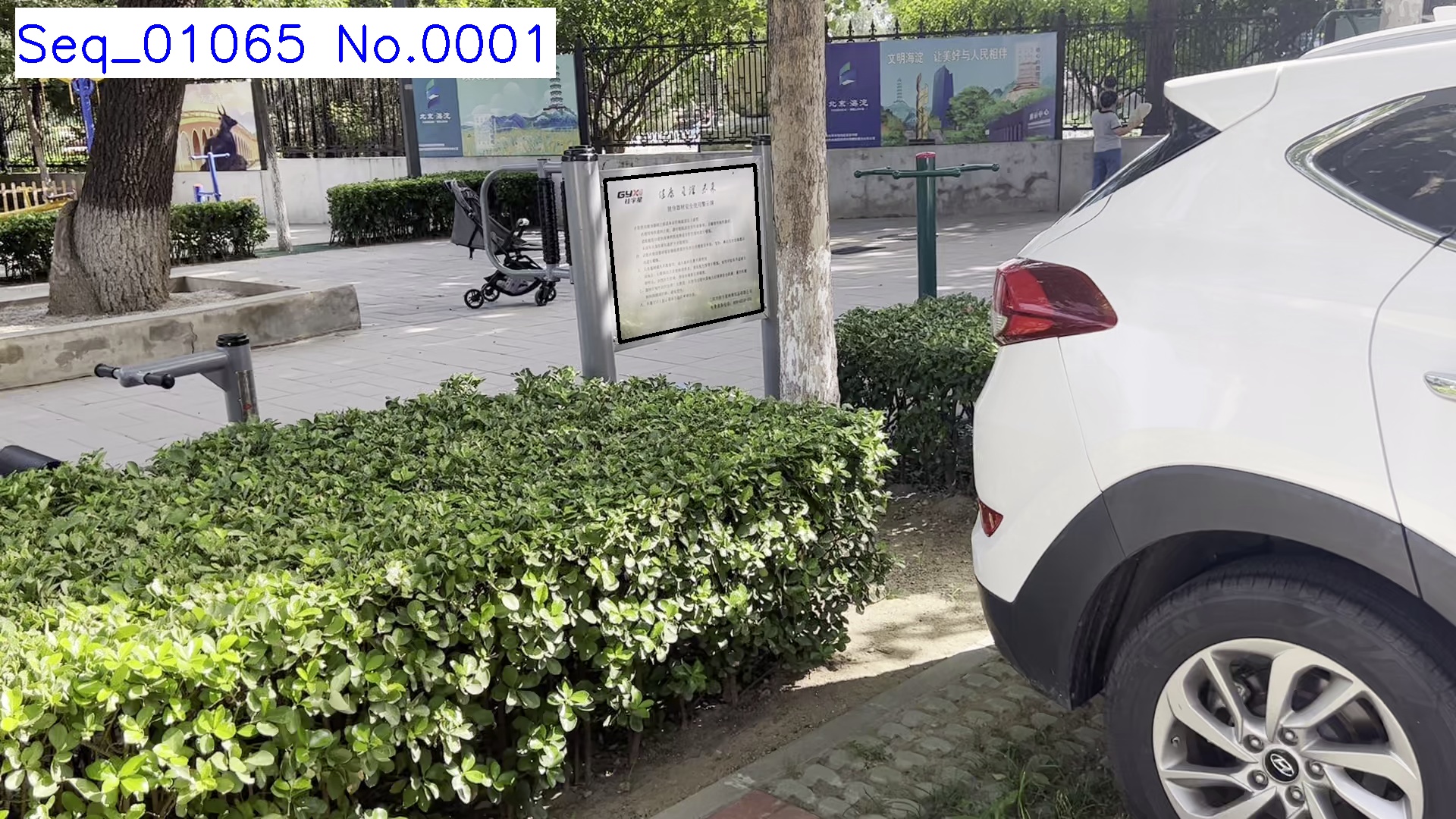} \includegraphics[width=0.19\linewidth]{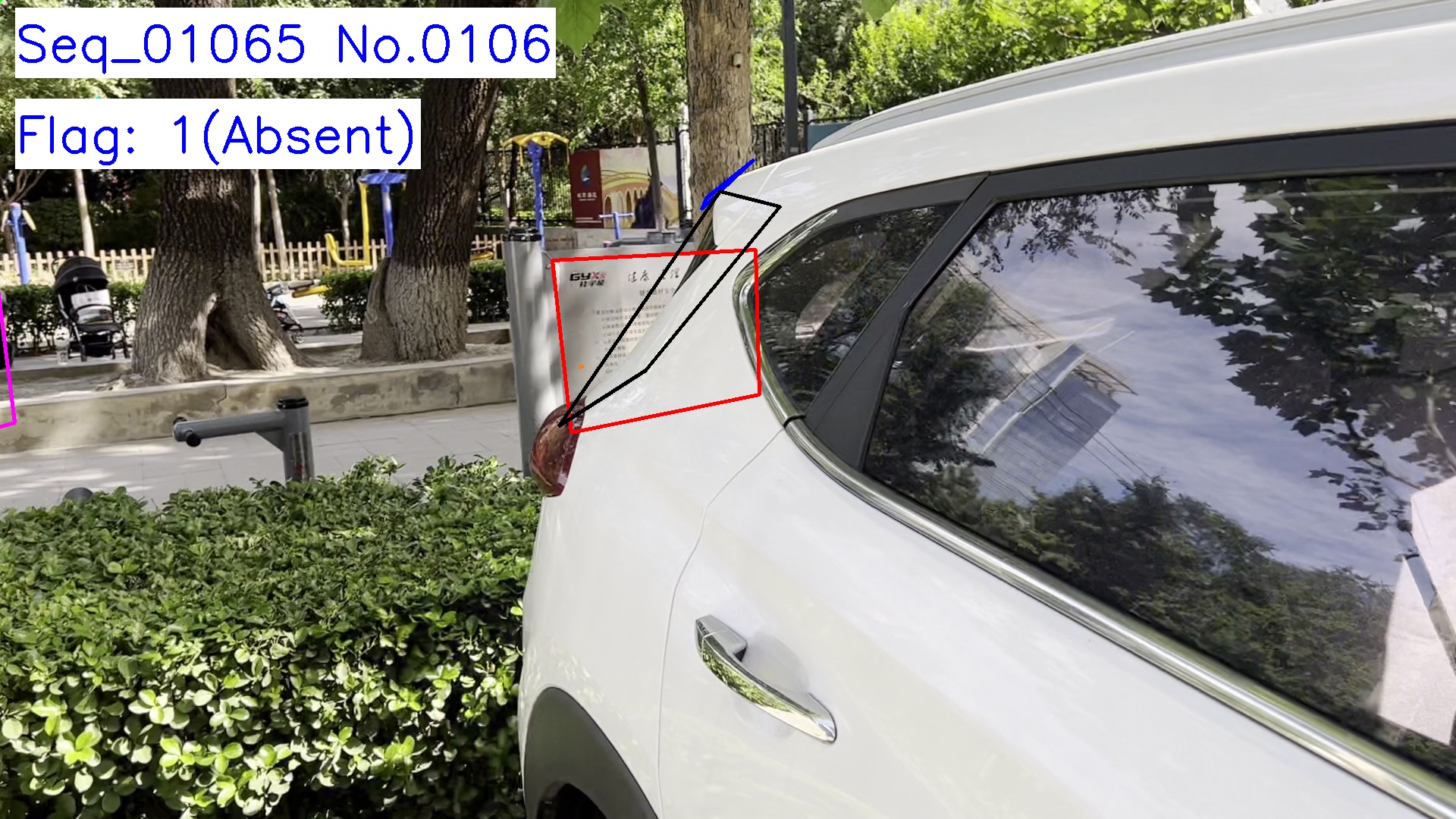} \includegraphics[width=0.19\linewidth]{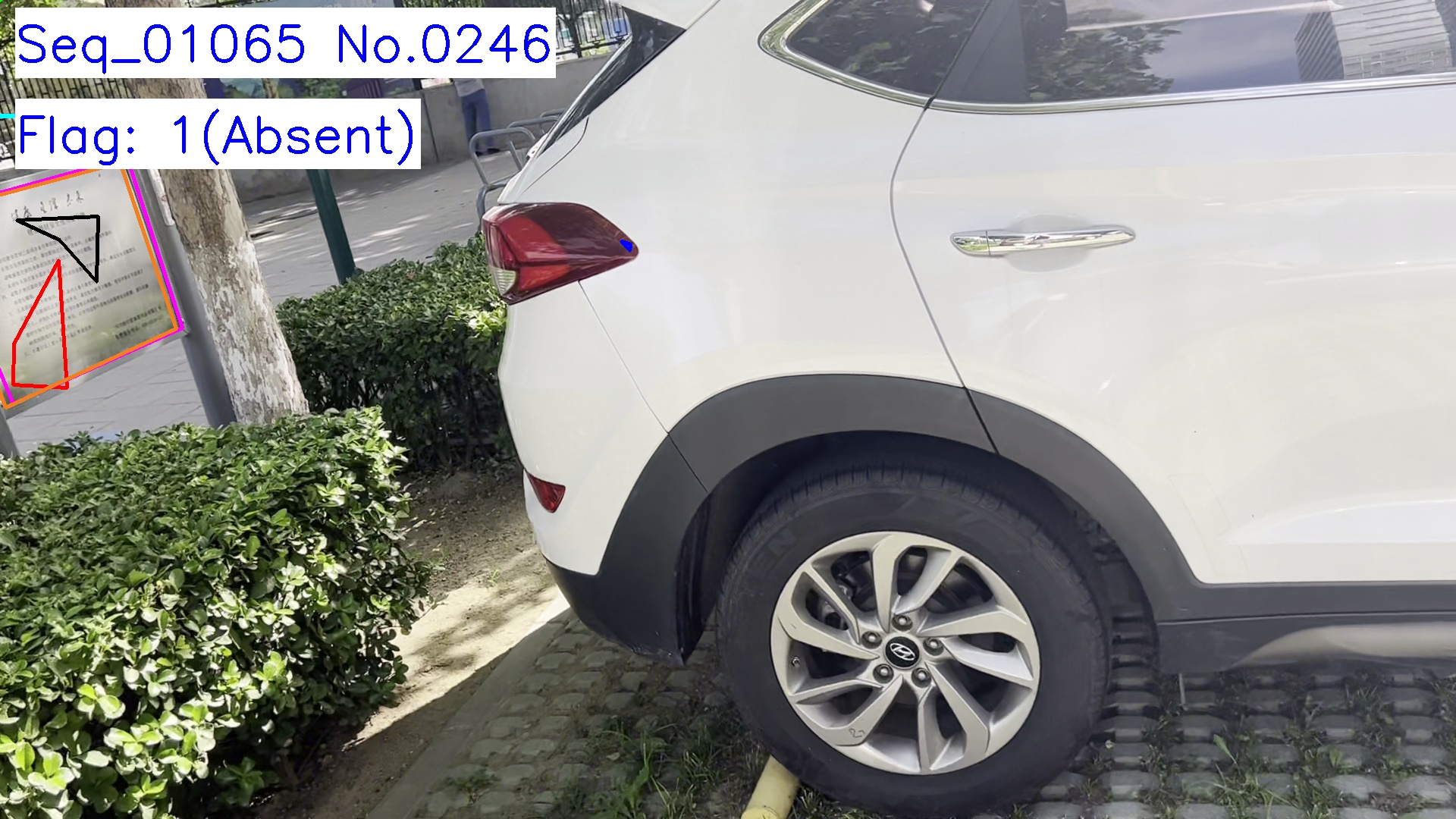} \includegraphics[width=0.19\linewidth]{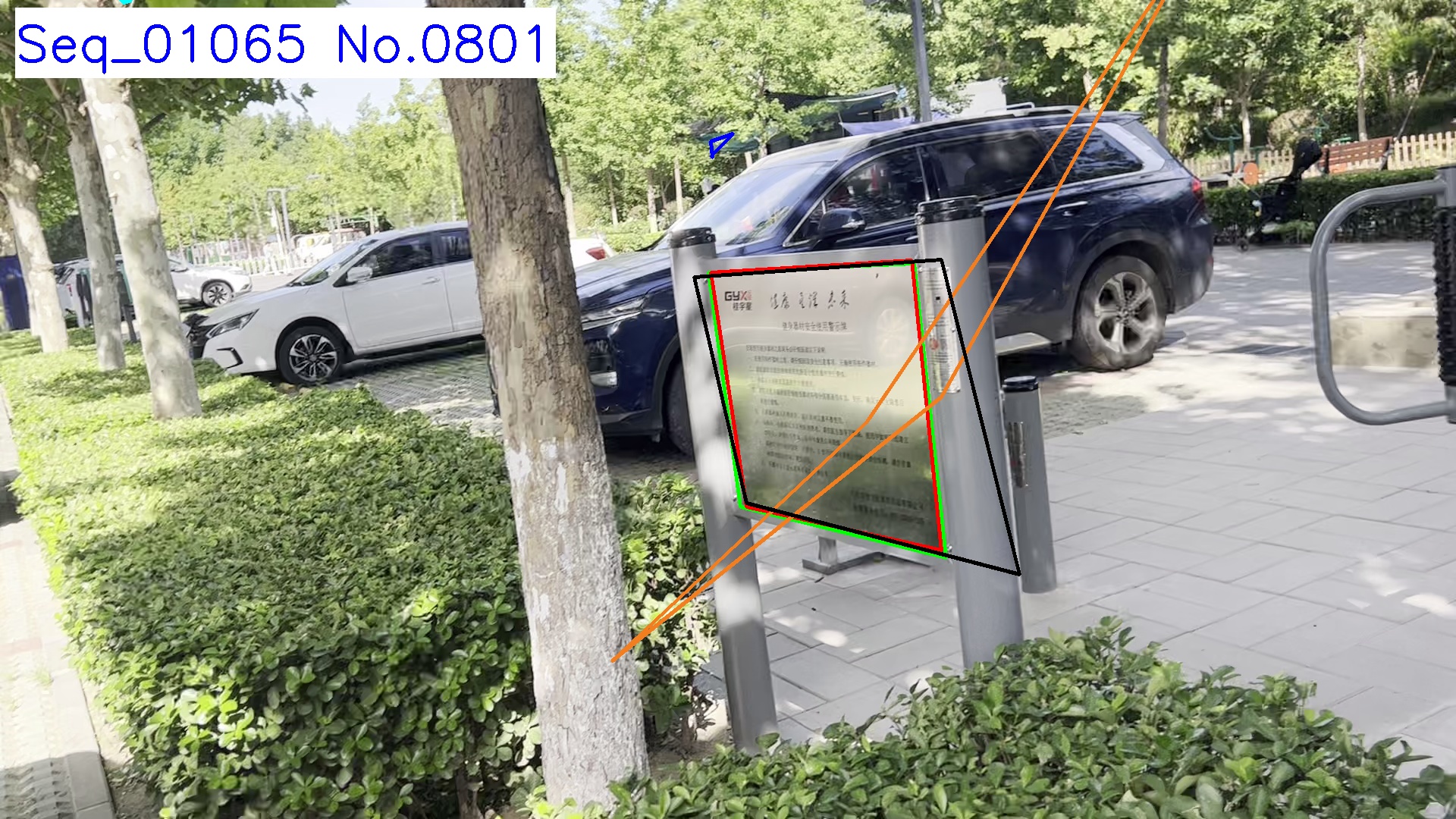} \includegraphics[width=0.19\linewidth]{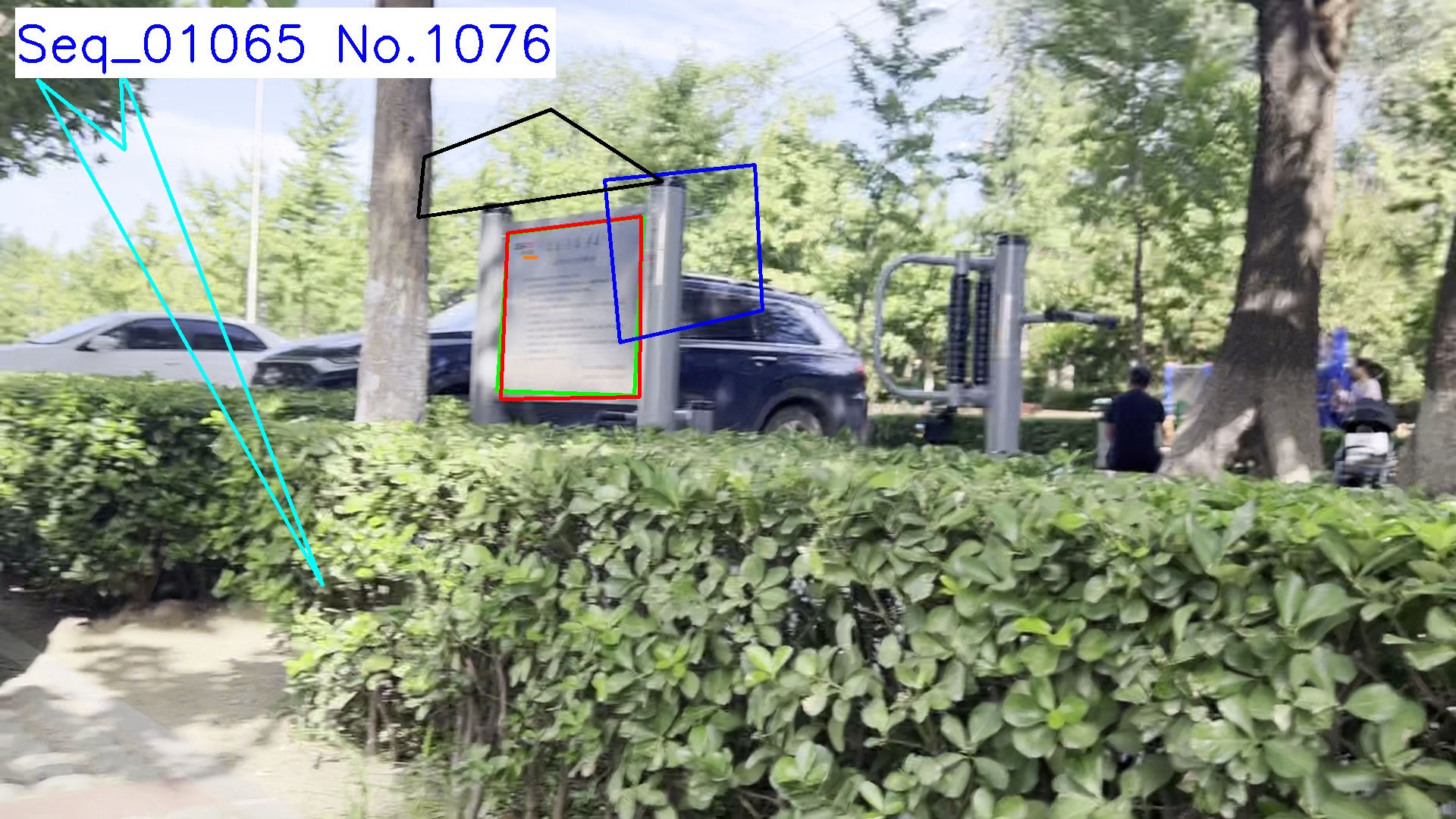} \\
        {\small (h) Sequence with multiple challenging factors, including OCC, MB, ROT, SV, PD, OV.} \\
        
        \includegraphics[width=0.19\linewidth]{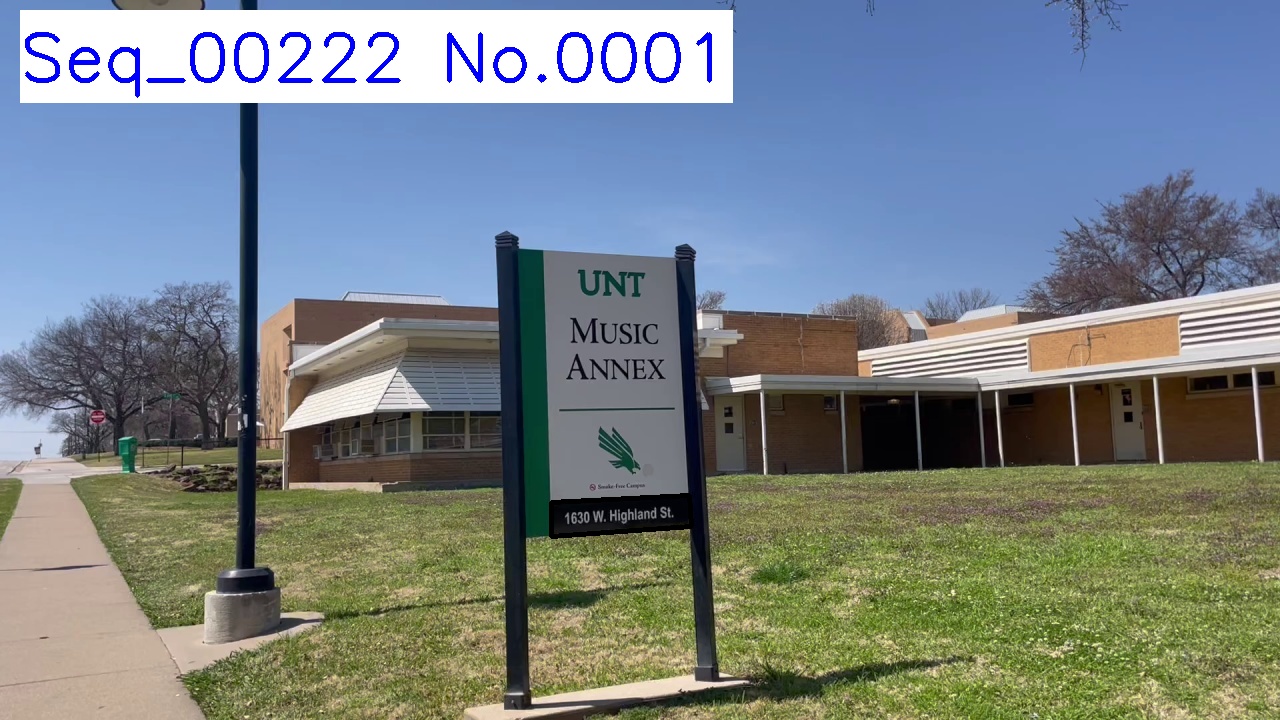} \includegraphics[width=0.19\linewidth]{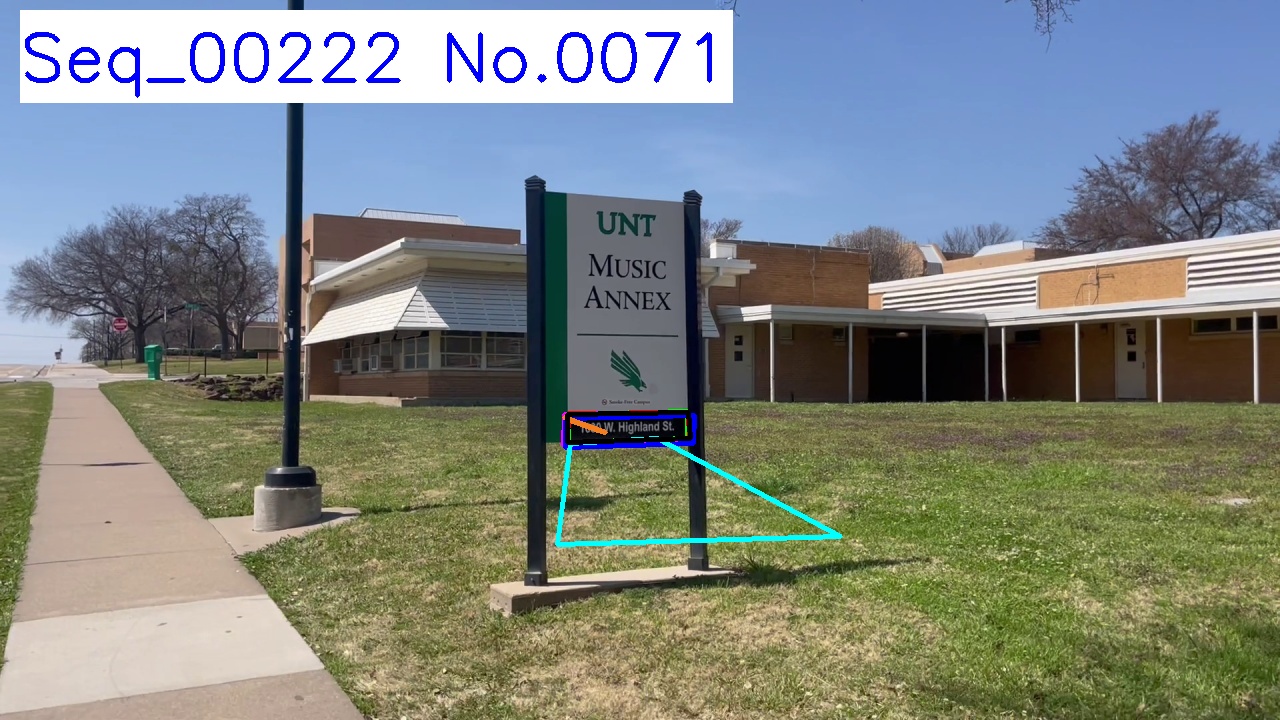} \includegraphics[width=0.19\linewidth]{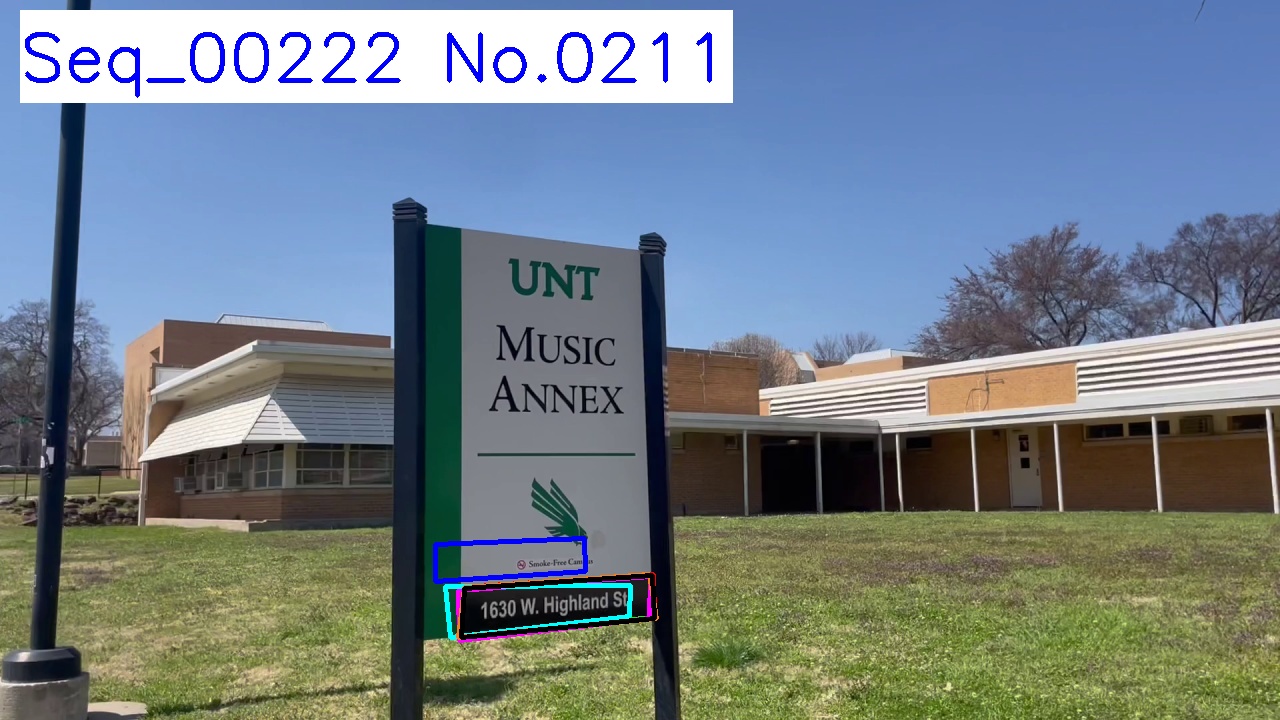}  \includegraphics[width=0.19\linewidth]{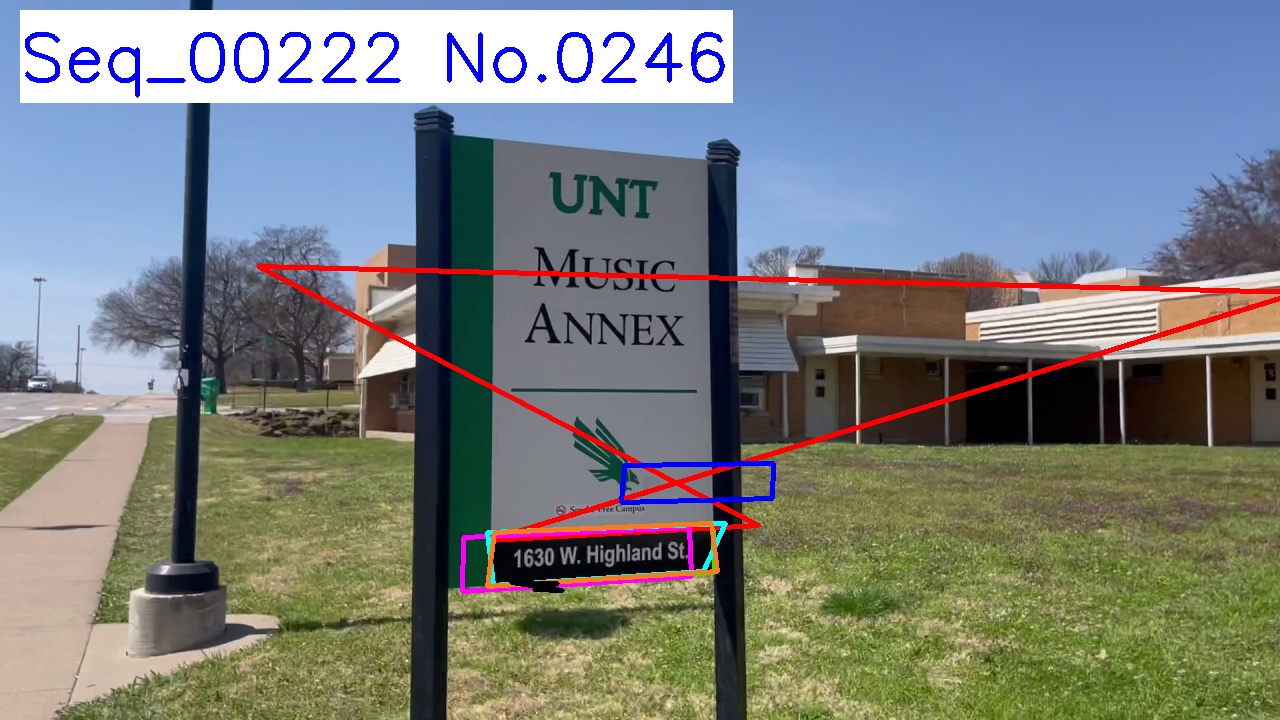} \includegraphics[width=0.19\linewidth]{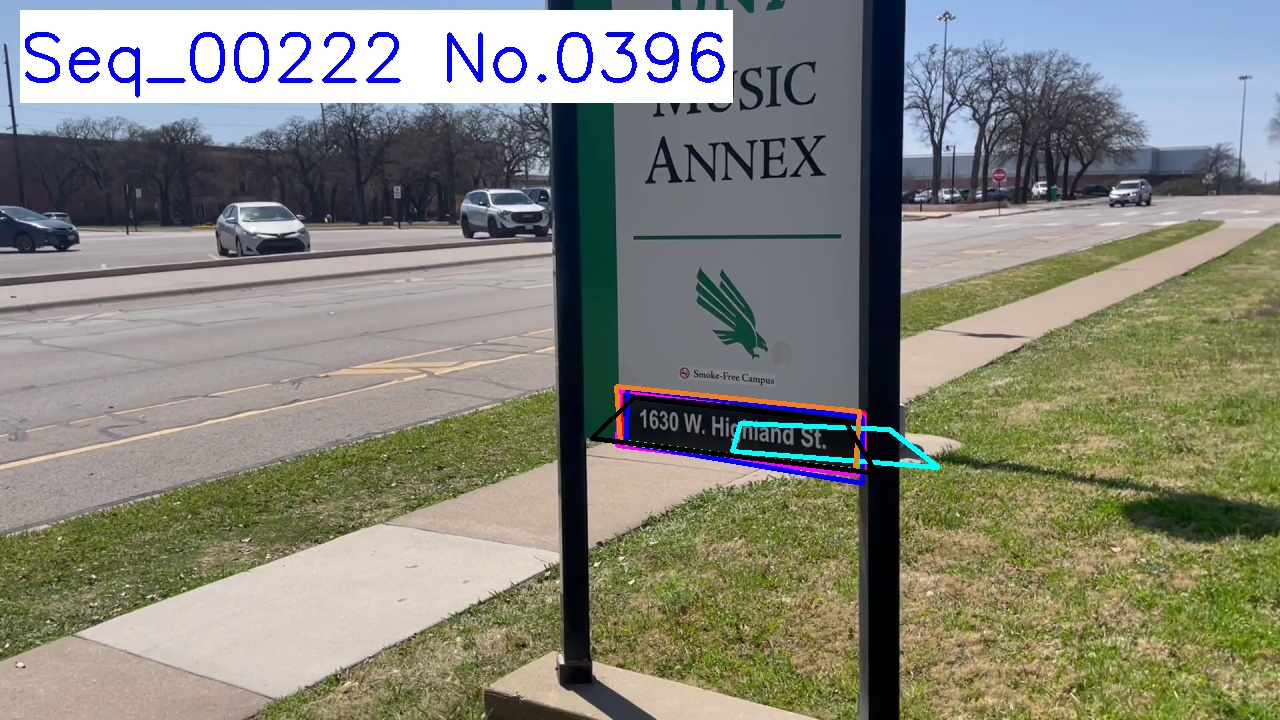} \\
        {\small (i) Sequence with multiple challenging factors, including ROT, SV, PD, OV.} \\
        
        \includegraphics[width=0.8\linewidth]{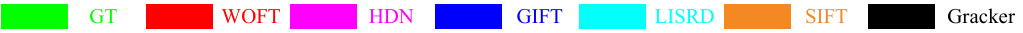}
\end{tabular}
\caption{Qualitative results of six trackers with the highest precision scores on different sequences. We observe that these planar trackers drift to the background region or even lose the target object due to different challenging factors in the videos such as background clutter, scale variation, perspective distortion, motion blur, rotation, out-of-view and low resolution.
}
\label{qualitative_result}
\end{figure*}

To better understand the above planar trackers, we demonstrate sampled tracking results of them in different challenging factors such as \emph{background clutter}, \emph{scale variation}, \emph{perspective distortion}, \emph{motion blur}, \emph{rotation}, \emph{out-of-view}, \emph{low resolution} and \emph{ultra-long-term tracking} in Fig. \ref{qualitative_result}. From Fig. \ref{qualitative_result} we observe that, although some trackers can deal with certain challenging factors, they may drift to the background region or fail to localize the planar target when multiple challenging factors occur simultaneously. For Fig. \ref{qualitative_result}-(a), trackers except WOFT can only roughly localize the target with large alignment error because of the varying reflection and large scale variation. A possible solution to handle this issue is to use some temporal information with the last and current frames (like optical flow in WOFT). We also evaluate the trackers on our proposed \emph{ultra-long} sequences (see Fig. \ref{qualitative_result}-(f)). WOFT can localize the planar target in most frames benefit from its motion clues. However, it may misidentify when there are similar targets (Fig. \ref{qualitative_result}-(g)).

\subsection{Comparison with POT-210}\label{sec4_3}

POT-210 \citep{liang2018planar} is currently one of the most popular benchmarks for planar object tracking. However, there remain some issues that limit the development of deep-learning-based planar object tracking algorithms. Firstly, most videos of POT-210 contain mainly one challenging factor and very few (\ie \ 30 in POT-210 and 40 in POT-280) are involved in unconstrained conditions. This could not faithfully reflect the difficulties and complexities in reality for evaluation. Besides, the lack of planar target diversity also limits its usage. In addition, the biggest drawback is that POT-210 only contains 53K annotated frames(70K in POT-280), which is far from enough for training and fair evaluation. To address these issues, we first construct PlanarTrack with 1,150 sequences and totally 733K frames, making it a large-scale benchmark for planar object tracking. For each sequence, we freely capture a unique target for diversity with multiple challenging factors. Therefore, our PlanarTrack is more challenging and realistic in practical applications.

To verify the above, we compare existing planar trackers on POT-210 and PlanarTrack\textsubscript{Tst}. Please note that, among the ten selected trackers, only four trackers are deep-based (\ie, WOFT, HDN, GIFT and LISRD) that require training before inference, as shown in Tab.~\ref{tracker_summary}. The remaining six trackers are training-free and can directly track planar objects. Therefore, in Tab.~\ref{POT_comp}, we evaluate the performance of the six training-free trackers by directly performing inference on POT-210, $\mathrm{POT\-210_{UC}}$, and $\mathrm{PlanarTrack_{Tst}}$. For the four deep-based trackers, we first train them on POT-210 and then perform inference on POT-210, $\mathrm{POT\-210_{UC}}$, and $\mathrm{PlanarTrack_{Tst}}$ to obtain the evaluation results.

Table \ref{POT_comp} shows the tracking results. From Table \ref{POT_comp} we observe that, WOFT achieves the best P@5 score of 0.805 and 0.768 on POT-210 and $\mathrm{POT\-210_{UC}}$. However, when used for tracking planar targets on PlanarTrack\textsubscript{Tst}, its performance is significantly degenerated. GIFT with the second best performance also absolutely declines from POT-210 to PlanarTrack\textsubscript{Tst}. Other trackers are declined more or less on PlanarTrack\textsubscript{Tst}.

\begin{table*}[!t]\footnotesize
  \centering
  \caption{Comparison of PlanarTrack\textsubscript{Tst} to POT-210~\citep{liang2018planar} and its subset POT-210$_\text{UC}$ in unconstrained condition using P@5 score. We also compare the P@5 score and P@15 score on our PlanarTrack\textsubscript{Tst}.}
  \setlength{\extrarowheight}{3pt}
    \begin{tabular}{lccccc}
    \specialrule{.1em}{.05em}{.05em}  
    \multirow{2}{*}{\textbf{Method}} & \textbf{POT-210} & \textbf{POT-210$_\text{UC}$} & \multicolumn{2}{c}{\textbf{PlanarTrack\textsubscript{Tst}}} \\
    & \textbf{P@5} & \textbf{P@5} & \textbf{P@5} & \textbf{P@15} \\
    \hline
    WOFT~\citep{vserych2023planar} & 0.805 & 0.768 & \cellcolor{gray!15}0.402 & \cellcolor{gray!15}0.607 \\
    HDN~\citep{zhan2022homography} & 0.612 & 0.567 & \cellcolor{gray!15}0.211 & \cellcolor{gray!15}0.455 \\
    GIFT~\citep{liu2019gift} & 0.553 & 0.528 & \cellcolor{gray!15}0.221 & \cellcolor{gray!15}0.402 \\
    LISRD~\citep{pautrat2020online} & 0.617 & 0.581 & \cellcolor{gray!15}0.192 & \cellcolor{gray!15}0.325 \\
    SIFT~\citep{lowe2004distinctive} & 0.692 & 0.578 & \cellcolor{gray!15}0.161 & \cellcolor{gray!15}0.257 \\
    Gracker~\citep{wang2017gracker} & 0.392 & 0.185 & \cellcolor{gray!15}0.162 & \cellcolor{gray!15}0.346 \\
    SOL~\citep{hare2012efficient} & 0.417 & 0.289 & \cellcolor{gray!15}0.131 & \cellcolor{gray!15}0.208 \\
    SCV~\citep{richa2011visual} & 0.228 & 0.105 & \cellcolor{gray!15}0.105 & \cellcolor{gray!15}0.145 \\
    ESM~\citep{benhimane2004real} & 0.204 & 0.100 & \cellcolor{gray!15}0.090 & \cellcolor{gray!15}0.128 \\
    IC~\citep{baker2004lucas} & 0.121 & 0.053 & \cellcolor{gray!15}0.045 & \cellcolor{gray!15}0.063 \\
    \specialrule{.1em}{.05em}{.05em} 
    \end{tabular}%
  % }
  \label{POT_comp}%
\end{table*}%

In addition to POT-210, we further compare POT-210\textsubscript{UC}, a small subset of POT-210 with all videos captured in unconstrained conditions, with PlanarTrack\textsubscript{Tst} in Table \ref{POT_comp}, as they are both have multiple challenging factors in a sequence. As in Table \ref{POT_comp}, tracking performances on POT-210\textsubscript{UC} are significantly worse than those on POT-210, which means that POT-210\textsubscript{UC} is more challenging than POT-210. Compared to POT-210\textsubscript{UC}, all trackers achieve the worst P@5 score on PlanarTrack\textsubscript{Tst}, which implies that our PlanarTrack is challenging. The best tracker WOFT on POT-210\textsubscript{UC} shows P@5 score of 0.768, while it degrades to 0.402 on PlanarTrack\textsubscript{Tst} with an absolute drop of 36.6\%.

From the above comparisons and analysis, we clearly see that POT-210 is a little simple for existing deep-learning-based planar trackers, which limits the development of planar object tracking algorithms. By contrast, our PlanarTrack is more challenging, complicated and large enough for planar object tracking. There is still a big room for improving tracking performance on PlanarTrack.

\subsection{Retraining on PlanarTrack}\label{sec4_4}

Deep-based algorithms often face the challenge of data hungry, where increasing the dataset size can significantly enhance generalization performance. As one of our central aspirations is to provide a large-scale platform for promoting the development of deep-learning-based planar trackers, we conduct retraining experiments on PlanarTrack. Please note that, among the four deep-based algorithms, GIFT and LISRD are not end-to-end trackers and are \textit{not} well-suited for retraining. WOFT released code but did \textit{not} provide a training script. As a result, we performed the retraining experiments solely on HDN. Specifically, we retrain the recent HDN using PlanarTrack\textsubscript{Tra}, instead of the synthetic data. While retraining, all the parameters and settings are kept the same as in the original method. After retraining, we demonstrate the results of HDN on POT-210 and PlanarTrack\textsubscript{Tst} in Table \ref{retraining}. From Table \ref{retraining}, we observe consistent performance gains on the two benchmarks. In other words, leveraging enough task-specific data in training can obviously improve the tracking performance. In specific, after retraining and testing on POT-210 by a fixed training/test split, the P@5 scores on POT-210 are increased from 0.612 to 0.637, with an absolute improvement of 2.5\%. On PlanarTrack\textsubscript{Tst}, the P@5 and P@15 scores have a more significant rise of 7.0\%/6.5\%, from 0.211/0.455 to 0.281/0.520. These improvements show that a large-scale training set is effective and necessary for improving planar object tracking performance. 

\begin{table}[!t]\footnotesize
  \centering
  \caption{Retraining of HDN~\citep{zhan2022homography} using PlanarTrack$_\textbf{Tra}$.}
  \label{retraining}
  \setlength{\extrarowheight}{3pt}
  \setlength{\tabcolsep}{0.6mm}{
    \begin{tabular}{lccc}
    \specialrule{.1em}{.05em}{.05em}  
          &       & \begin{tabular}[c]{@{}c@{}}Original\\ HDN\end{tabular} & \tabincell{c}{Retrained\\HDN} \\
    \hline
    \multirow{1}[0]{*}{POT-210~\citep{liang2018planar}} & P@5   & 0.612 & 0.637 (\textcolor[RGB]{220,20,60}{+2.5\%}) \\
    \hline
    \multirow{2}[0]{*}{PlanarTrack\textsubscript{Tst}} & P@5   & 0.211 & 0.281 (\textcolor[RGB]{220,20,60}{+7.0\%}) \\
          & P@15   & 0.455 & 0.520 (\textcolor[RGB]{220,20,60}{+6.5\%}) \\
    \specialrule{.1em}{.05em}{.05em} 
    \end{tabular}}
\end{table}%

\section{PlanarTrack \textsubscript{BB} and experiments}\label{sec5}

A certified generic tracker should be able to locate the targets robustly without prior knowledge of their categories. Planar objects (\eg \ posters, screen, board) are very common things in our daily life. Surprisingly, there is little study on localization of planar targets with \textit{generic visual trackers}at large scale, even in the existing large-scale generic tracking benchmarks (\eg \ \citep{fan2021lasot, huang2019got, muller2018trackingnet}). 

In order to figure out the capacities of these generic trackers in tracking planar targets, we further develop a new benchmark named PlanarTrack\textsubscript{BB} based on our PlanarTrack. To be specific, PlanarTrack\textsubscript{BB} shares the same images and training/test split as PlanarTrack. The only difference between PlanarTrack\textsubscript{BB} and PlanarTrack is that we convert annotations from four annotated corner points to an axis-aligned bounding box in PlanarTrack\textsubscript{BB}, especially used for large-scale evaluation of generic trackers. Specifically, we calculate the axis-aligned bounding box based on the four annotated corner points and adjust it to ensure it completely fits within the image boundaries. Notice that, in PlanarTrack\textsubscript{BB} we actually represent the coordinates of the axis-aligned bounding box in XYWH format (\ie \ $[x_\text{min},y_\text{min},width,height]$) like LaSOT \citep{fan2019lasot} and GOT-10k \citep{huang2019got}. The difference and some examples of PlanarTrack and PlanarTrack\textsubscript{BB} are demonstrated in Fig. \ref{gen_planar_compare}.

To further understand PlanarTrack\textsubscript{BB}, we select 15 recent state-of-the-art generic trackers for evaluation. All the trackers are transformer-based, including SeqTrack \citep{chen2023seqtrack}, ROMTrack \citep{cai2023robust}, DropTrack \citep{wu2023dropmae}, MixFormerV2 \citep{cui2024mixformerv2}, MixFormer \citep{cui2022mixformer}, OStrack \citep{ye2022joint}, SwinTrack \citep{lin2022swintrack}, ARTrack \citep{wei2023autoregressive}, TransInMo \citep{guo2022learning}, STARK \citep{yan2021learning}, AiATrack \citep{gao2022aiatrack}, TransT \citep{chen2021transformer}, SimTrack \citep{chen2022backbone}, ToMP \citep{mayer2022transforming}, TrDiMP \citep{wang2021transformer}. We employ the best version of each generic tracker for evaluation except SimTrack and ARTrack. Sim-L/14 performs best but only Sim-B/16 is released in Simtrack, while ARTrack-L\textsubscript{384} achieves the best performance but only ARTrack-B\textsubscript{384} is given. For metrics, we use the success score for bounding box-based tracking \citep{wu2013online}, named SUC\textsubscript{BB}. 

\begin{table*}[!ht]\footnotesize
  \centering
  \caption{Evaluation of generic trackers on PlanarTrack$_\text{BB}$ and comparison with other popular generic benchmarks using SUC$_\text{BB}$.}
    \label{bb_results}
    \setlength{\extrarowheight}{3pt}
    \begin{tabular}{lcc>{\columncolor{gray!15}}c}
    \specialrule{.1em}{.05em}{.05em} 
          & \tabincell{c}{\textbf{TrackingNet}\\\citep{muller2018trackingnet}} & \tabincell{c}{\textbf{LaSOT}\\\citep{fan2019lasot}} &  \tabincell{c}{\textbf{PlanarTrack\textsubscript{BB}}\\(ours)} \\
    \hline
    SeqTrack~\citep{chen2023seqtrack} & 0.855 & 0.725 & 0.670 \\
    ROMTrack~\citep{cai2023robust} & 0.841 & 0.714 & 0.667 \\
    DropTrack~\citep{wu2023dropmae} & 0.841 & 0.718 & 0.665 \\
    MixFormerV2~\citep{cui2024mixformerv2} & 0.834 & 0.706 & 0.648 \\
    MixFormer~\citep{cui2022mixformer} & 0.839 & 0.701 & 0.647 \\
    OStrack~\citep{ye2022joint} & 0.839 & 0.711 & 0.642 \\
    SwinTrack~\citep{lin2022swintrack} & 0.840 & 0.713 & 0.638 \\
    ARTrack~\citep{wei2023autoregressive} & 0.856 & 0.731 & 0.633 \\
    TransInMo~\citep{guo2022learning} & 0.817 & 0.657 & 0.620 \\
    STARK~\citep{yan2021learning} & 0.820 & 0.671 & 0.615 \\
    AiATrack~\citep{gao2022aiatrack} & 0.827 & 0.690 & 0.613 \\
    TransT~\citep{chen2021transformer} & 0.814 & 0.649 & 0.603 \\
    SimTrack~\citep{chen2022backbone} & 0.834 & 0.705 & 0.601 \\
    ToMP~\citep{mayer2022transforming}  & 0.815 & 0.685 & 0.597 \\
    TrDiMP~\citep{wang2021transformer} & 0.784 & 0.639 & 0.589 \\
    \specialrule{.1em}{.05em}{.05em} 
    \end{tabular}
    % }
\end{table*}

Table \ref{bb_results} shows the evaluation results of the above generic trackers and comparisons with existing large-scale generic tracking benchmarks including LaSOT \citep{fan2019lasot} and TrackingNet \citep{muller2018trackingnet}. Due to the different evaluation metrics, we do not compare our PlanarTrack\textsubscript{BB} with GOT-10k \citep{huang2019got}. From Table \ref{bb_results} we observe that, although existing generic trackers can achieve remarkable performance on LaSOT and TrackingNet, they are significantly degraded when handling planar-like targets on PlanarTrack\textsubscript{BB}. For instance, the best generic tracker SeqTrack obtains 0.855/0.725 SUC scores on LaSOT/TrackingNet, but obviously declines to 0.670 on PlanarTrack\textsubscript{BB}, with an absolute drop of 18.5\%/5.5\%. The second best ROMTrack is also decreased from 0.841/0.714 to 0.667. This may indicate that more attention should be paid to improve such planar trackers, though they are rigid.

For in-depth analysis of generic tracking performances on PlanarTrack\textsubscript{BB}, we further demonstrate the evaluation results of the above generic trackers in Fig. \ref{figbb} by using a modified LaSOT \citep{fan2019lasot} evaluation toolkit. Under One Pass Evaluation (OPE) protocol, we utilize bounding box-based precision and success plots as in generic tracking \citep{wu2013online} for assessment. From Fig. \ref{figbb} we can see that, the top two generic trackers SeqTrack and ROMTrack achieve 0.684/0.670 and 0.674/0.667 relatively on PlanarTrack\textsubscript{BB}.

\newcommand{\MMM}{0.315}
\newcommand{\MM}{0.234}
\begin{figure}[!t]
		\centering
		\begin{tabular}{c}
\includegraphics[width=\MMM\linewidth]{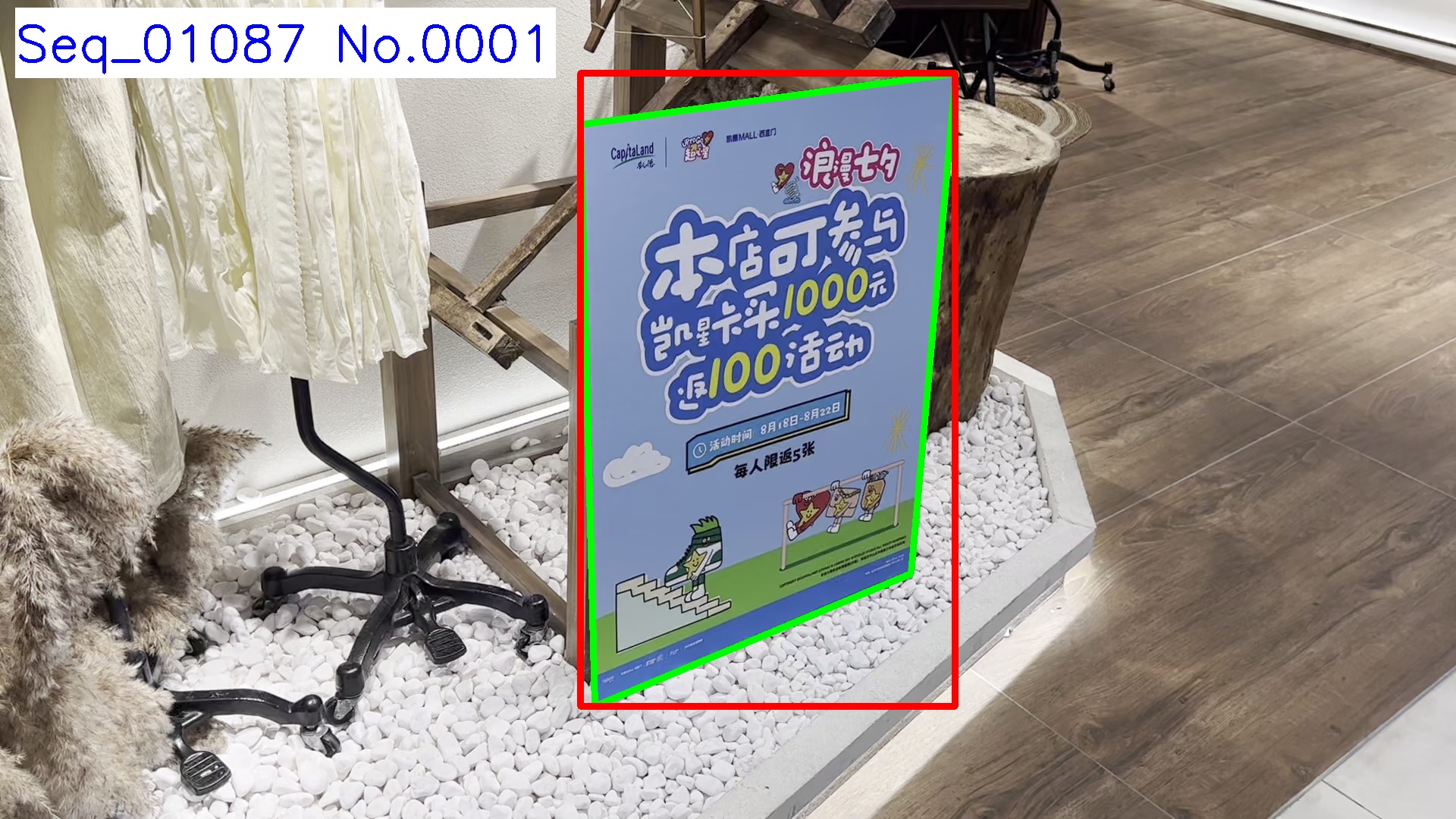} \includegraphics[width=\MMM\linewidth]{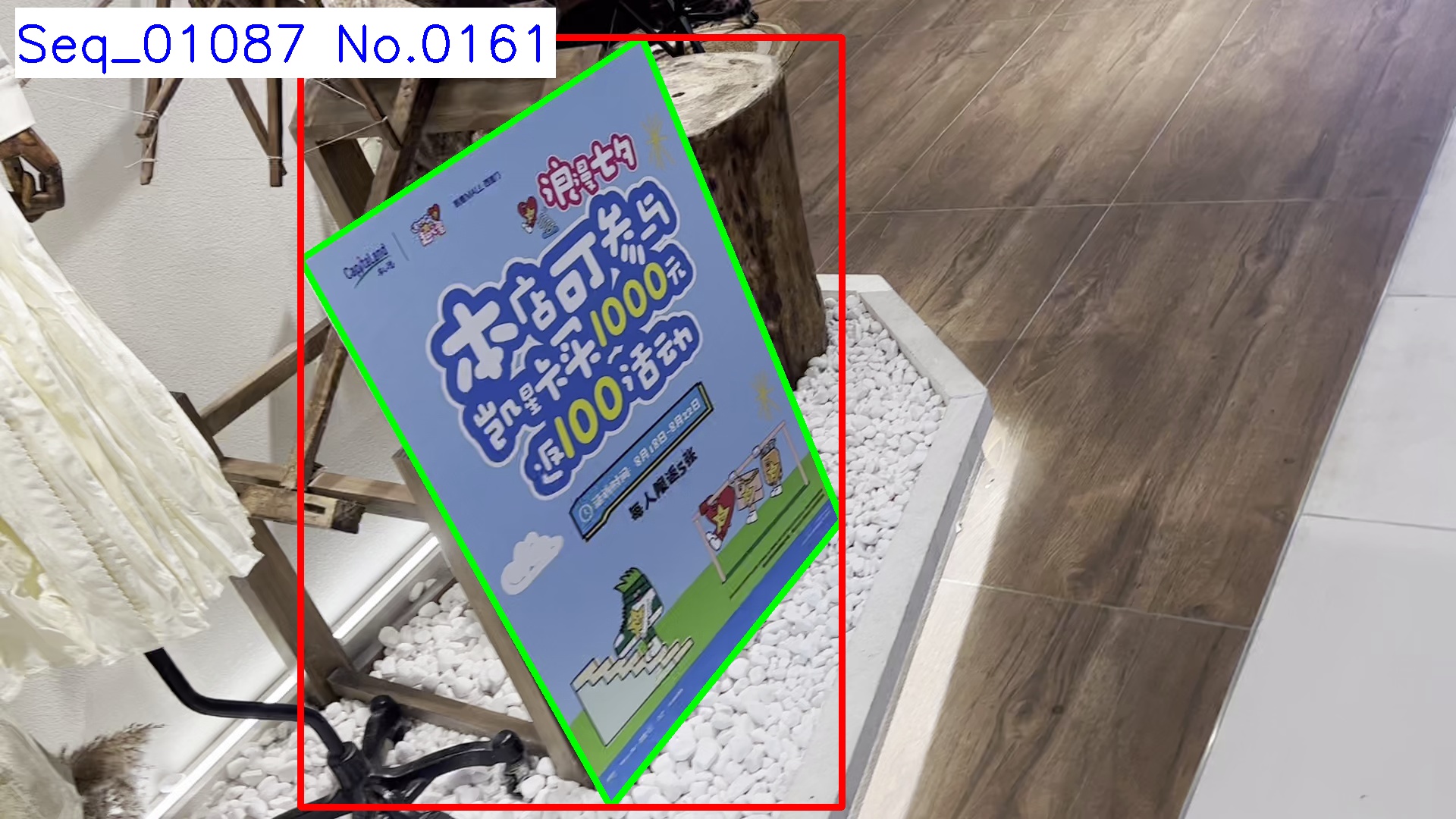}
\includegraphics[width=\MMM\linewidth]{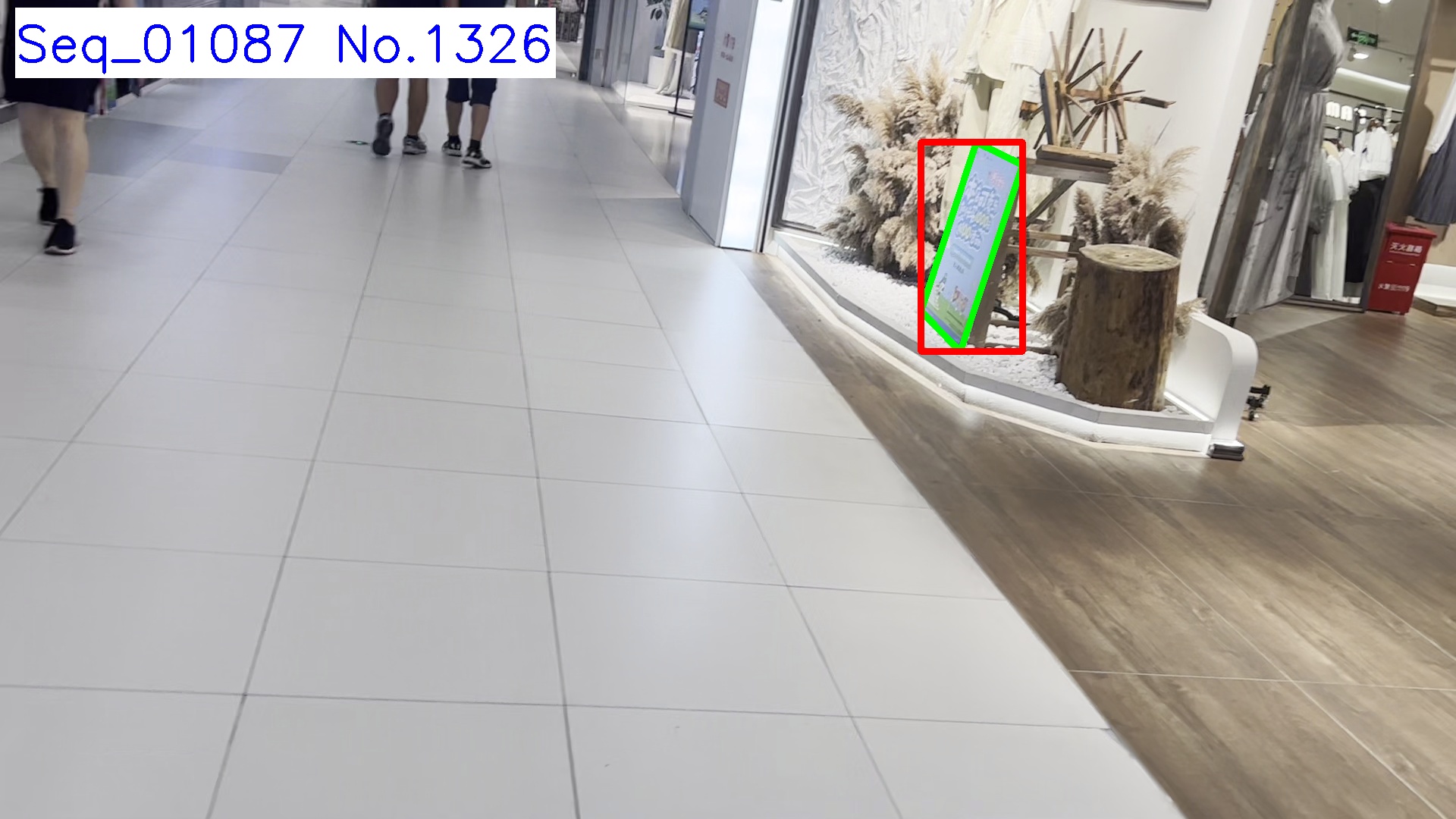}\\
\includegraphics[width=\MMM\linewidth]{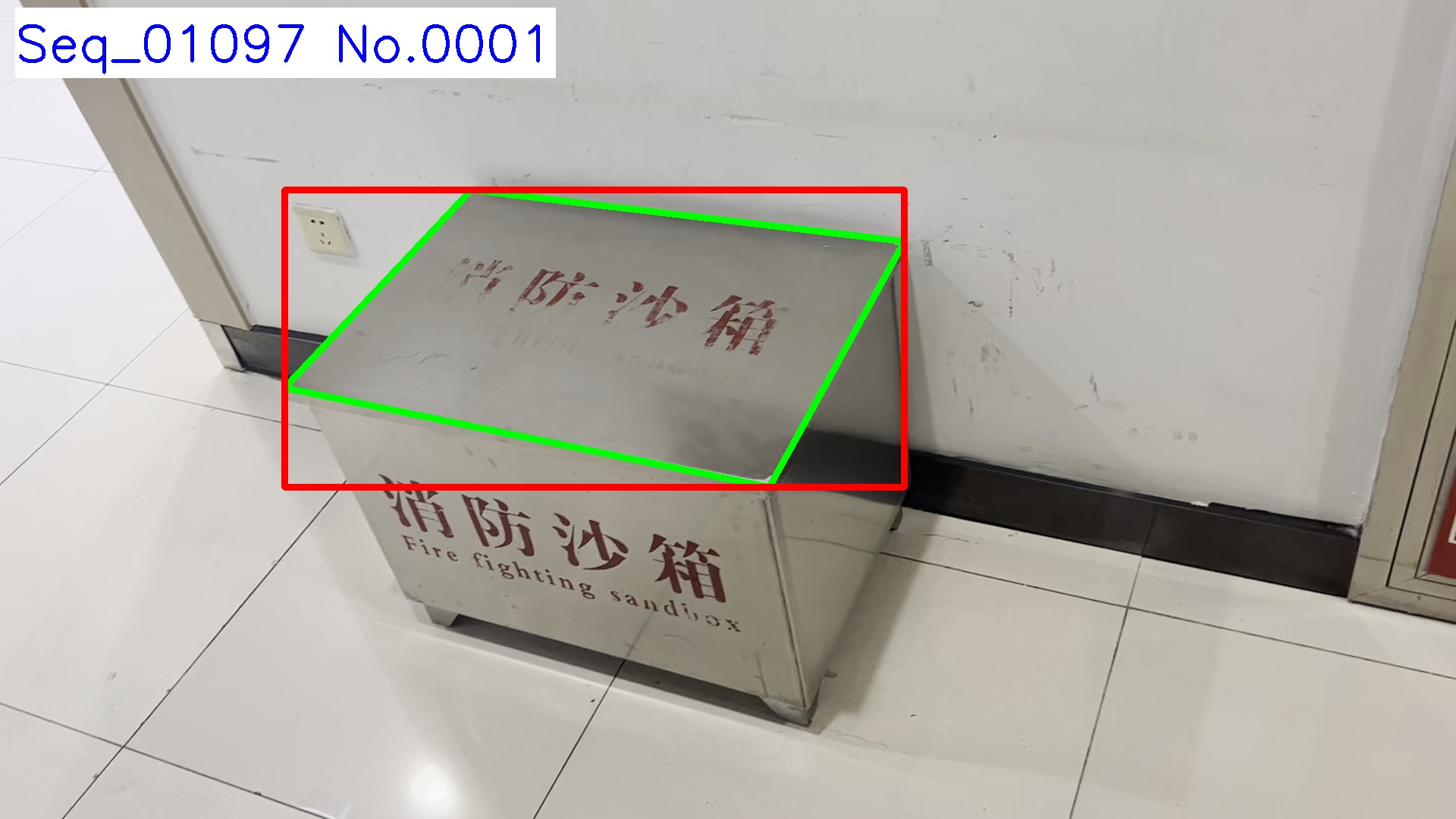} \includegraphics[width=\MMM\linewidth]{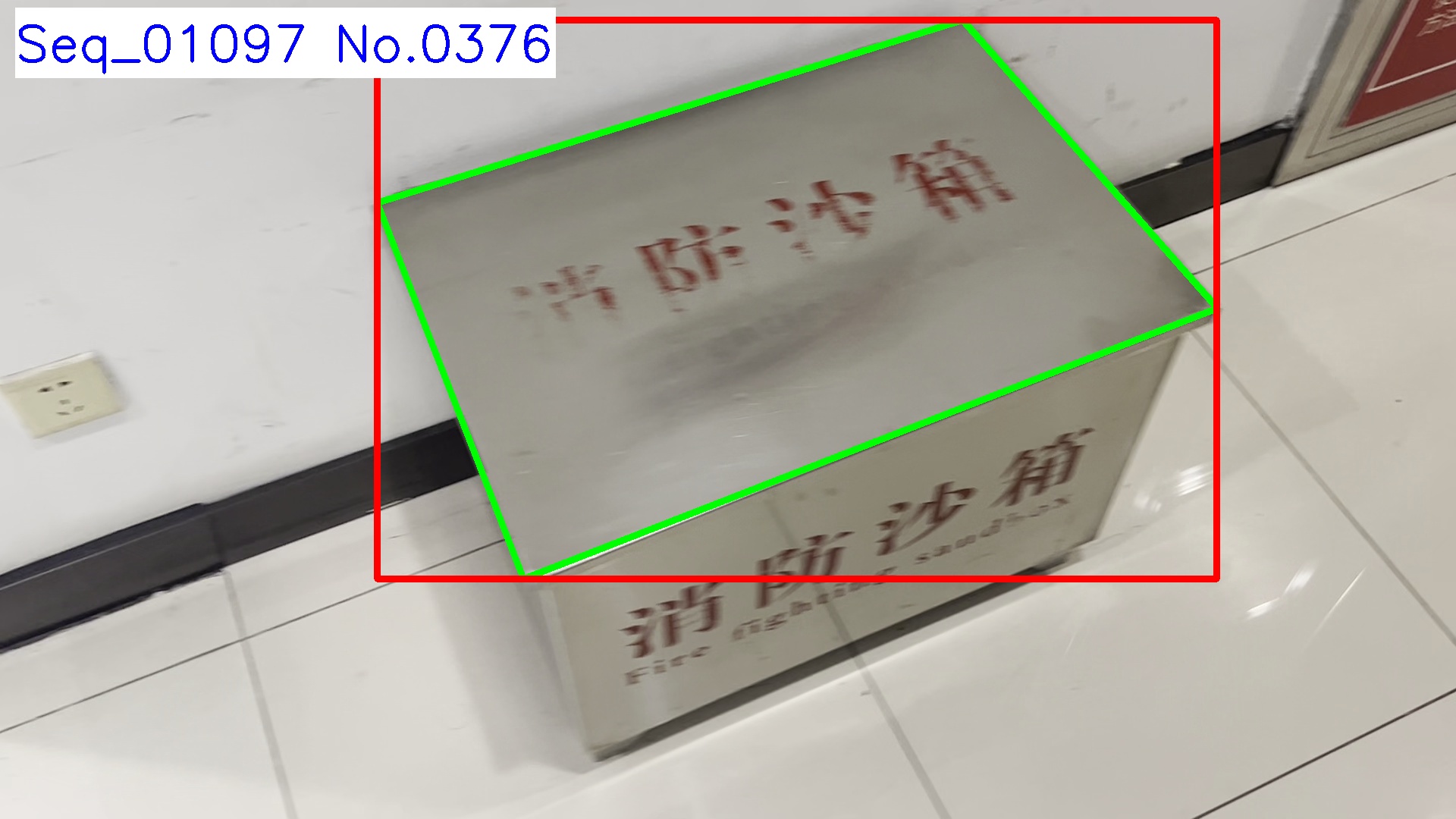}
\includegraphics[width=\MMM\linewidth]{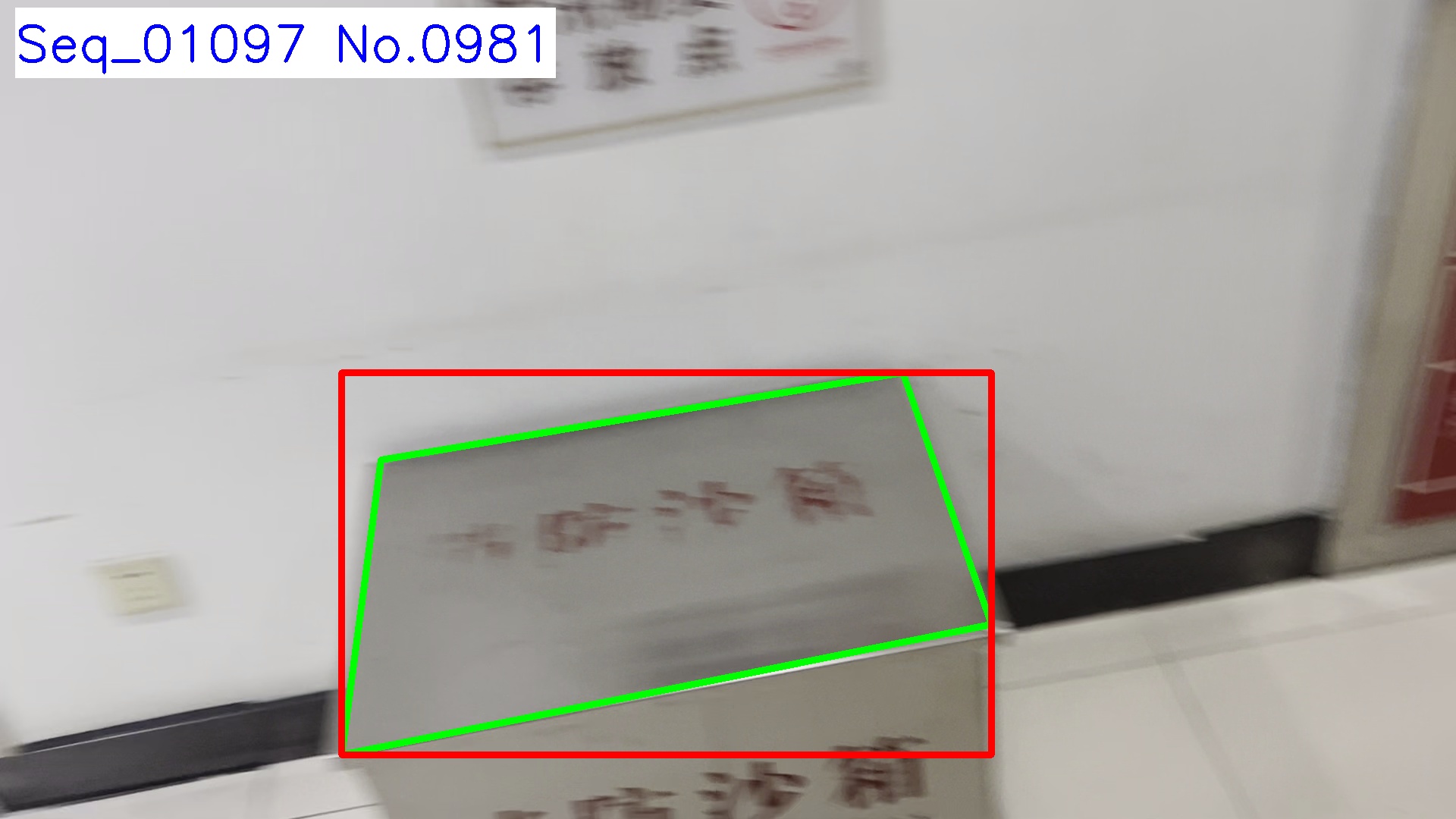}\\
\includegraphics[width=\MM\linewidth]{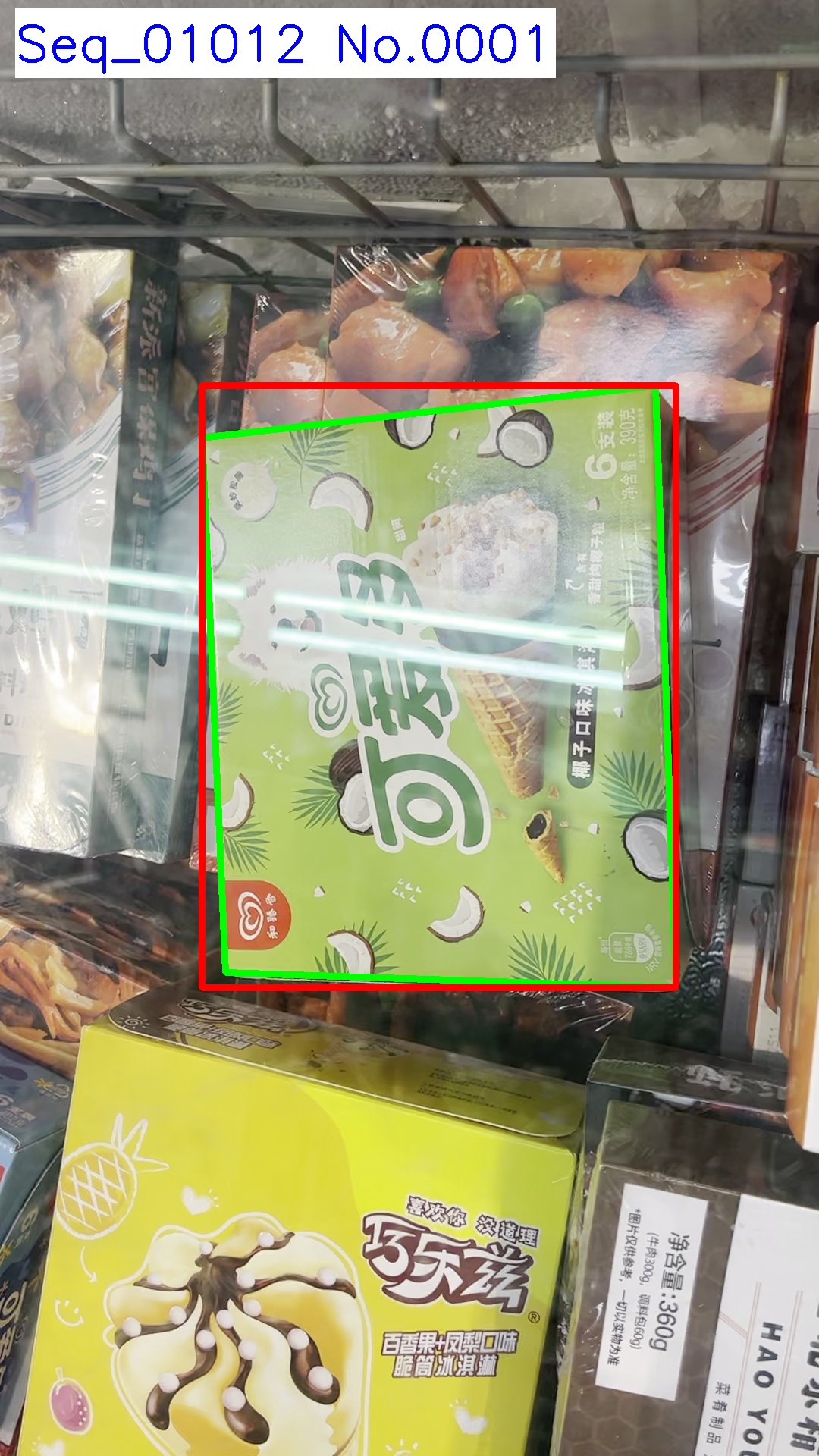} \includegraphics[width=\MM\linewidth]{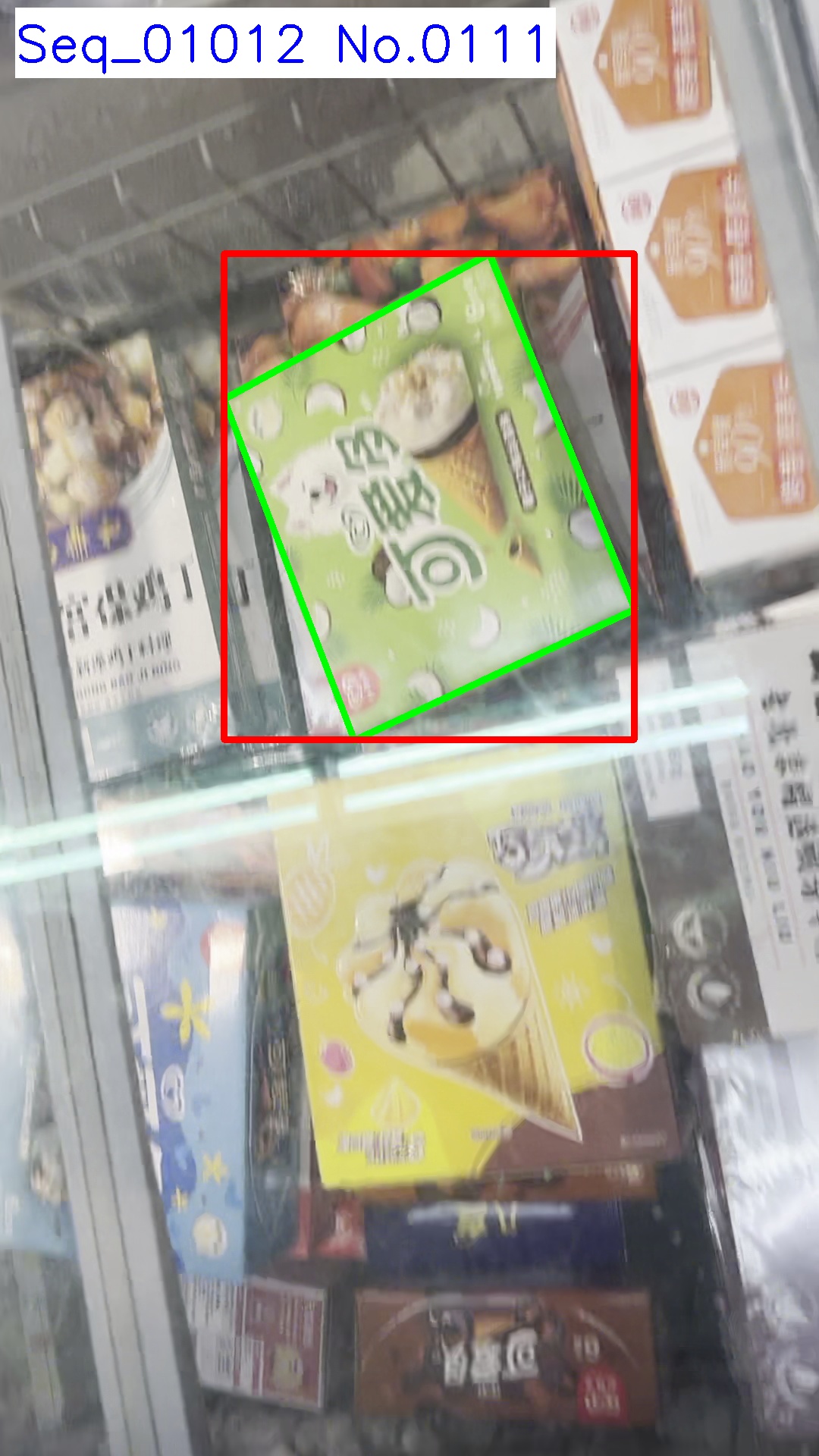} 
\includegraphics[width=\MM\linewidth]{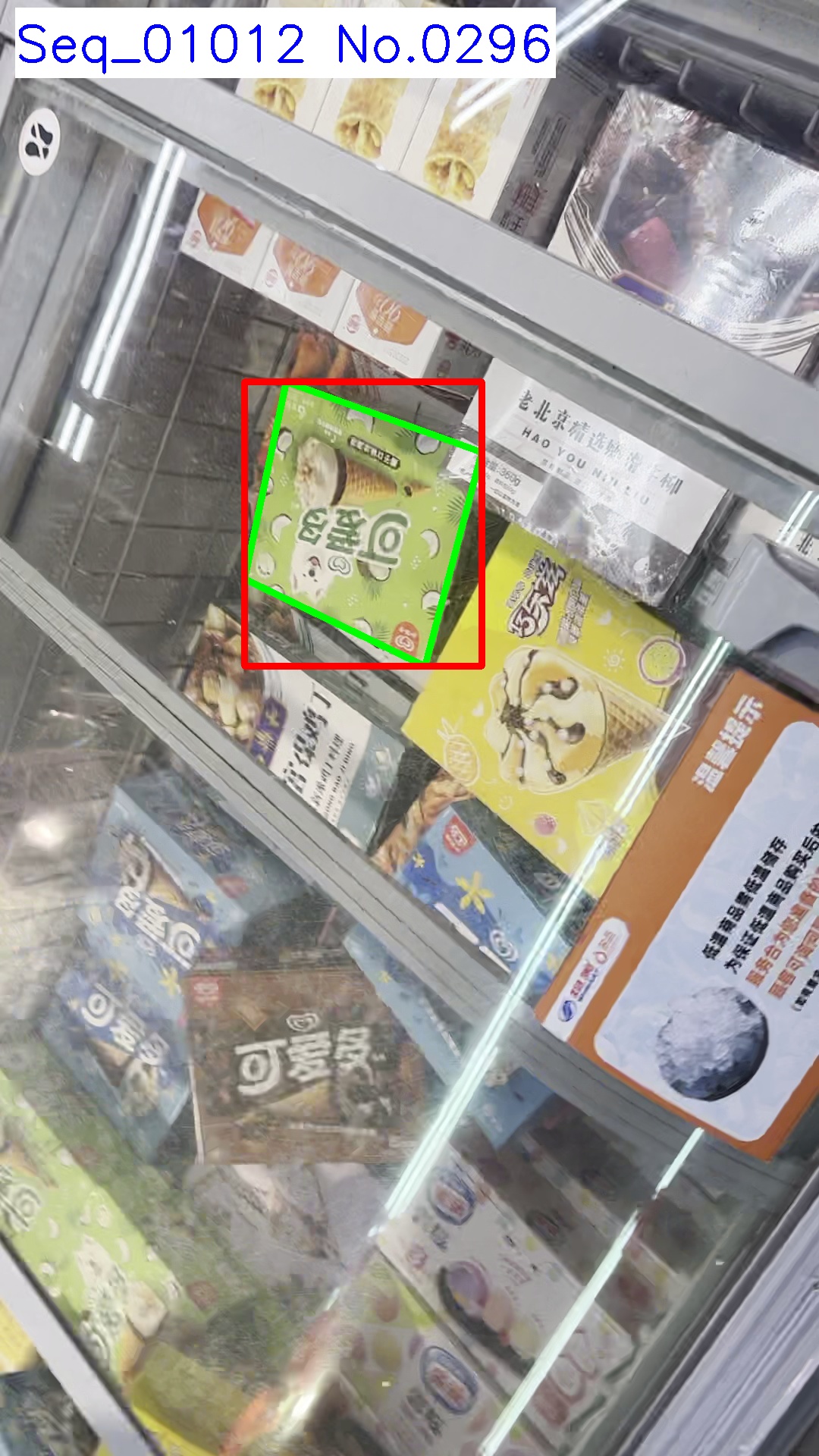} \includegraphics[width=\MM\linewidth]{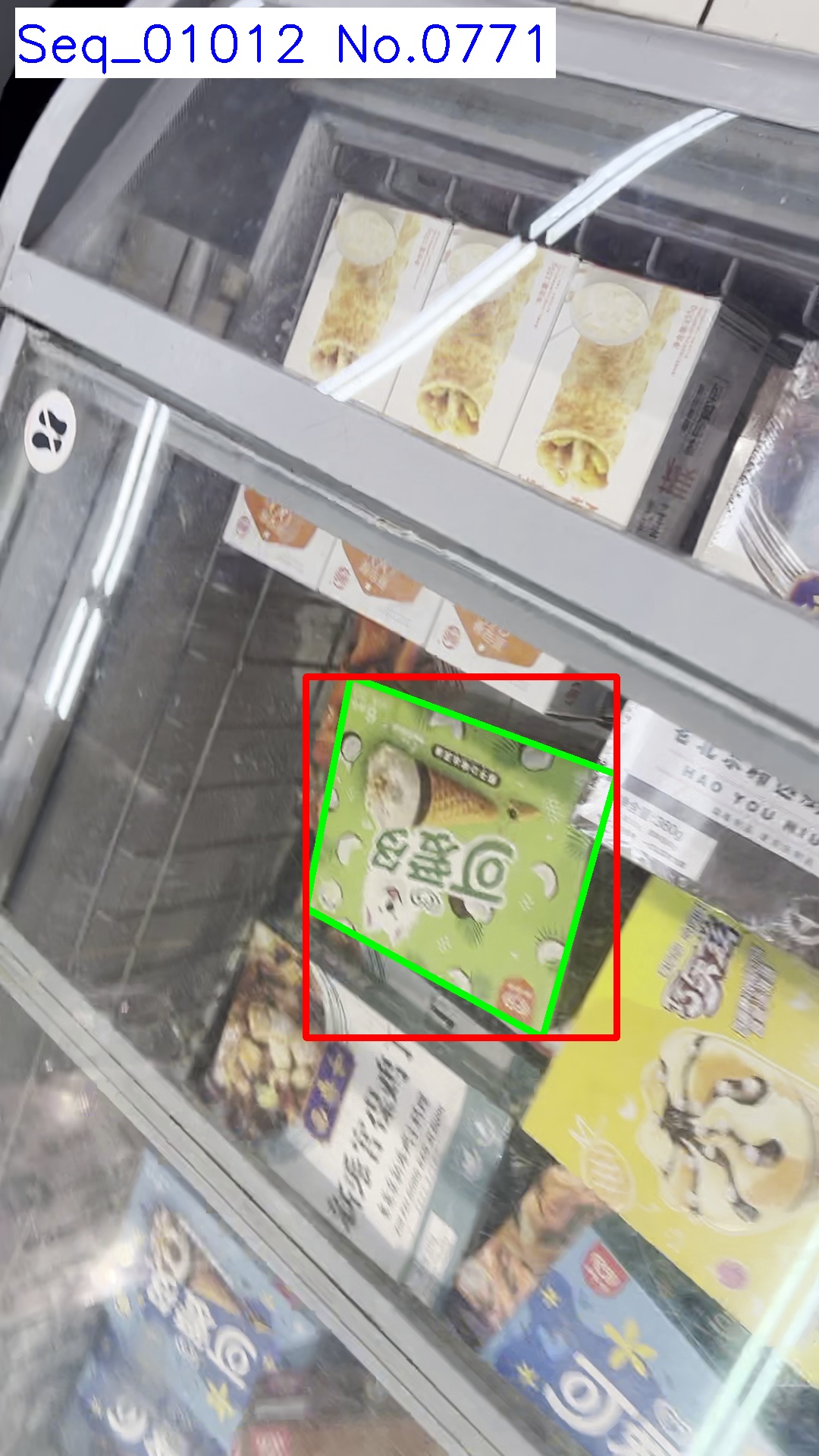} \\
\includegraphics[width=0.8\linewidth]{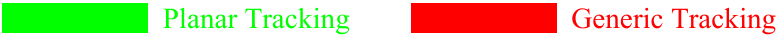}\\
\end{tabular}
\caption{Examples from PlanarTrack$_\text{BB}$. The targets are annotated by white axis-align bounding boxes for genetic visual tracking. Best viewed in color.
}
\label{gen_planar_compare}
\end{figure}
 
\begin{figure}[!t]
    \centering
    \includegraphics[width=\linewidth]{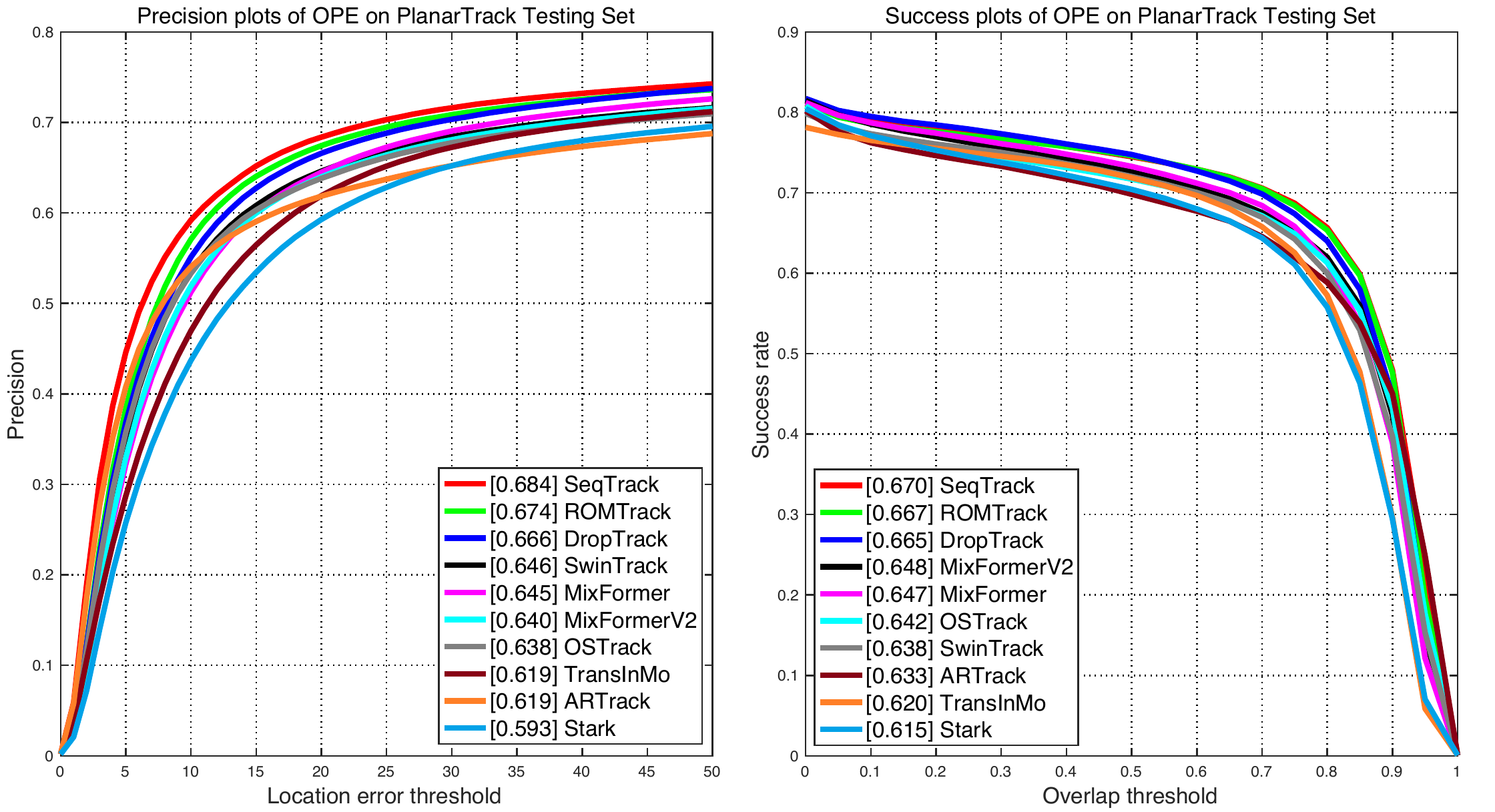}
    \caption{Performance of evaluated generic visual trackers on PlanarTrack\textsubscript{BB} using bounding box-based precision and success plots. To facilitate clearer analysis, we exclusively present the top 10 trackers. Best viewed in color.}
    \label{figbb}
\end{figure}

\section{Conclusion}\label{sec6}

In this paper, we introduced a brand new benchmark named PlanarTrack. PlanarTrack consists of 1,150 videos recorded in unconstrained conditions from realistic scenarios, and has more than 733K annotated image frames in total. High-quality dense annotations are provided and great diversity of targets is ensured in PlanarTrack. To the best of our knowledge, PlanarTrack is the \emph{first} challenging large-scale dataset dedicated to planar object tracking. To further understand existing approaches and provide a comparison for further research, we perform experiments by evaluating ten recent planar trackers and carry out a detailed analysis of PlanarTrack. By releasing PlanarTrack, we sincerely hope that we can offer the community a dedicated platform for research and applications of planar tracking. In addition, we provide PlanarTrack\textsubscript{BB}, a by-product dataset based on PlanarTrack, for studying generic trackers on tracking planar-like target objects. Evaluation results indicate that there is still huge room for future improvement on PlanarTrack and PlanarTrack\textsubscript{BB}. For future research, we see several promising directions: (i) robust feature learning under low resolution and light-interactive surfaces, (ii) better temporal modeling for long-term tracking, (iii) integration of multi-modal cues such as depth or inertial data, and (iv) effective re-detection strategies for disappeared objects.

\vspace{0.3em}
\noindent
{\bf Acknowledgement.} We sincerely thank volunteers for their help in constructing PlanarTrack.

% \appendix
% \section{My Appendix}
% Appendix sections are coded under \verb+\appendix+.

% \verb+\printcredits+ command is used after appendix sections to list 
% author credit taxonomy contribution roles tagged using \verb+\credit+ 
% in frontmatter.


\bibliographystyle{cas-model2-names}

\bibliography{cas-refs}


% Author biography without author photo.
% Author biography. Author biography. Author biography.
% Author biography. Author biography. Author biography.
% Author biography. Author biography. Author biography.
% Author biography. Author biography. Author biography.
% Author biography. Author biography. Author biography.
% Author biography. Author biography. Author biography.
% Author biography. Author biography. Author biography.
% Author biography. Author biography. Author biography.
% Author biography. Author biography. Author biography.

% \bio{figs/pic1}
% Author biography with author photo.
% Author biography. Author biography. Author biography.
% Author biography. Author biography. Author biography.
% Author biography. Author biography. Author biography.
% Author biography. Author biography. Author biography.
% Author biography. Author biography. Author biography.
% Author biography. Author biography. Author biography.
% Author biography. Author biography. Author biography.
% Author biography. Author biography. Author biography.
% Author biography. Author biography. Author biography.

% \bio{figs/pic1}
% Author biography with author photo.
% Author biography. Author biography. Author biography.
% Author biography. Author biography. Author biography.
% Author biography. Author biography. Author biography.
% Author biography. Author biography. Author biography.

\end{document}